\documentclass[opre, nonblindrev]{informs3_hide}

\usepackage{algcompatible}
\usepackage{algorithm}
\usepackage{float}
\floatstyle{ruled}
\newfloat{algorithm}{htbp}{lop}
\floatname{algorithm}{Algorithm}
\usepackage{booktabs}
\usepackage{diagbox}

\usepackage{appendix}
\usepackage{soul}

\OneAndAHalfSpacedXI 

\newcommand{\cD}{\mbox{$\mathcal{D}$}}

\newcommand{\bI}{\mathbbm{1}}
\newcommand{\bN}{\mathbb{N}}
\newcommand{\bR}{\mathbb{R}}
\newcommand{\bE}{\mathbb{E}}
\newcommand{\bp}{\bm{p}}

\newcommand{\bq}{\bm{q}}
\newcommand{\bC}{\bm{C}}
\newcommand{\bmR}{\bm{R}}
\newcommand{\bx}{\bm{x}}

\newcommand{\cZ}{\mathcal{Z}}
\newcommand{\radius}{\mathcal{r}}
\newcommand{\mathbbm}[1]{\text{\usefont{U}{bbm}{m}{n}#1}} 

\newcommand{\bQ}{{\bm{Q}}}

\newcommand{\Prob}{\mathbb{P}}

\newcommand{\Ber}{\mathrm{Ber}}
\newcommand{\Rev}{\mathsf{Rev}}
\newcommand{\DSL}{\mathsf{\Lambda}}
\newcommand{\cDSL}{\mathcal{X}}
\newcommand{\Bmin}{B_\mathsf{min}}
\newcommand{\DLP}{\mathsf{J}^{\mathsf{DLP}}}
\newcommand{\DLPG}{\mathsf{J}^{\mathsf{DLP-G}}}
\newcommand{\DLPGO}{\mathsf{J}^{\mathsf{DLP-G-0}}}
\newcommand{\pmax}{p_{\mathsf{max}}}
\newcommand{\pmin}{p_{\mathsf{min}}}
\newcommand{\amax}{a_{\mathsf{max}}}
\newcommand{\rmax}{R_{\mathsf{max}}}
\newcommand{\cmax}{C_{\mathsf{max}}}
\newcommand{\cI}{\mathcal{I}}
\newcommand{\bmu}{\bm{\mu}}
\newcommand{\Imu}{\mathcal{I}_{\bmu}}
\newcommand{\gap}{\eta}

\newcommand{\nrm}{\textsf{NRM}\xspace}
\newcommand{\stp}{\textsf{SP}\xspace}
\newcommand{\bnrm}{\textsf{BNRM}\xspace}
\newcommand{\bwk}{\textsf{BwK}\xspace}
\newcommand{\mab}{\textsf{MAB}\xspace}
\newcommand{\lsbnrm}{\textsf{BNRM-LS}\xspace}
\newcommand{\lsbwk}{\textsf{BwK-LS}\xspace}
\newcommand{\mainalgo}{\textsf{LS-2SLP}\xspace}
\newcommand{\lsete}{\textsf{LS-ETE}\xspace}

\usepackage{natbib,multirow,multicol,xspace,enumitem,subcaption,caption}
\bibpunct[, ]{(}{)}{,}{a}{}{,}%
\def\bibfont{\small}%
\usepackage{bm}
\usepackage[normalem]{ulem}
\usepackage{xcolor}
\definecolor{DarkBlue}{rgb}{0.0,0.0,0.55}
\usepackage[backref = false, bookmarks, colorlinks = true, plainpages = false, citecolor = DarkBlue , urlcolor = DarkBlue, filecolor = DarkBlue, linkcolor = DarkBlue, hypertexnames=false]{hyperref}

{\theoremstyle{TH}
\newtheorem{statement}{Statement}}

\TheoremsNumberedThrough     
\ECRepeatTheorems

\EquationsNumberedThrough    

\MANUSCRIPTNO{}

\begin{document}


\RUNAUTHOR{Simchi-Levi, Xu, and Zhao}
\RUNTITLE{BNRM and BwK Under Limited Switches}

\TITLE{\Large{Blind Network Revenue Management and Bandits with Knapsacks Under Limited Switches}}


\ARTICLEAUTHORS{
\AUTHOR{David Simchi-Levi}
\AFF{Department of Civil and Environmental Engineering, Operations Research Center, and Institute for Data, Systems, and Society, Massachusetts Institute of Technology, Cambridge, MA 02139,
\EMAIL{dslevi@mit.edu}}
\AUTHOR{Yunzong Xu}
\AFF{Department of Industrial and Enterprise Systems Engineering, Grainger College of Engineering, University of Illinois, Urbana-Champaign, IL 61801,
\EMAIL{xyz@illinois.edu}}
\AUTHOR{Jinglong Zhao}
\AFF{Questrom School of Business, Boston University, Boston, MA, 02215,
\EMAIL{jinglong@bu.edu}}
}

\ABSTRACT{
This paper studies the impact of limited switches on resource-constrained dynamic pricing with demand learning. We focus on the classical price-based blind network revenue management problem and extend our results to the bandits with knapsacks problem. In both settings, a decision maker faces stochastic and distributionally unknown demand, and must allocate finite initial inventory across multiple resources over time. In addition to standard resource constraints, we impose a switching constraint that limits the number of action changes over the time horizon. We establish matching upper and lower bounds on the optimal regret and develop computationally efficient limited-switch algorithms that achieve it. We show that the optimal regret rate is fully characterized by a piecewise-constant function of the switching budget, which further depends on the number of resource constraints. Our results highlight the fundamental role of resource constraints in shaping the statistical complexity of online learning under limited switches. Extensive simulations demonstrate that our algorithms maintain strong cumulative reward performance while significantly reducing the number of switches.
}


\maketitle

\section{Introduction} \label{sec:Introduction}

In this paper, we consider the classical price-based blind network revenue management (\bnrm) problem (\citealt{besbes2012blind}) and its extension to the bandits with knapsacks (\bwk) problem  (\citealt{badanidiyuru2013bandits}).
In the \bnrm problem, a firm is endowed with some finite inventory of multiple resources to sell over a finite time horizon.
The starting inventory is unreplenishable and exogenously given.
The firm can control its sales through sequential decisions on the prices offered. 
The firm's objective is to maximize its expected cumulative revenue.

We consider the setup in which demand is stochastic, independent and time-homogeneous.
Yet the distributional information is unknown to the firm, and has to be sequentially learned over the selling horizon.
In such a setup, any optimal policy must instantaneously adjust its prices and switch between the prices in real time \citep{badanidiyuru2013bandits, besbes2012blind, ferreira2018online}.

However, not all firms have the infrastructure to query the realized demand in real time, to adjust their decisions instantaneously, or to switch between the prices as freely as possible.
Changing the posted prices is costly for many firms \citep{bray2022menu, levy1998price, stamatopoulos2020effects, zbaracki2004managerial}, and frequent price changes may confuse the customers \citep{jorgensen2003retail}.
A common practice for many firms is to restrict the number of price changes to be within a budgeted number \citep{chen2015real, chen2020data, cheung2017dynamic, netessine2006dynamic, perakis2023dynamic, simchi2023phase}.

Motivated by this challenge, we analyze the impact of limited switches on the classical blind network revenue management (\bnrm) problem.
We incorporate an additional constraint of limited switching budget into the classical \bnrm model and formulate a new problem: blind network revenue management under limited switches (\lsbnrm). 
For the \lsbnrm problem, we establish tight upper and lower bounds on the optimal regret, and design limited-switch algorithms that achieve the optimal regret rate. 
Our results characterize the optimal regret rate as a function of the switching budget, which further depends on the number of resources.

Additionally, we consider the more general bandits with knapsacks (\bwk) problem \citep{badanidiyuru2013bandits, slivkins2014online}.
The \bwk problem generalizes the \bnrm problem in the sense that rewards (revenue) and costs (consumption of resources) can have any arbitrary relationship, that is, they are not necessarily connected through demand variables.
Similar to how we formulate the blind network revenue management under
limited switches (\lsbnrm) problem, we formulate the bandits with knapsacks under limited switches (\lsbwk) problem, which extends the classical bandits with knapsacks (\bwk) problem; see Sections~\ref{sec:BwK}~--~\ref{sec:BwKLS} in the Online Appendix.

\subsection{Contributions}

\begin{figure}[!tb]
\centering
\includegraphics[width=0.6\textwidth,trim={4 15.5 12 56},clip]{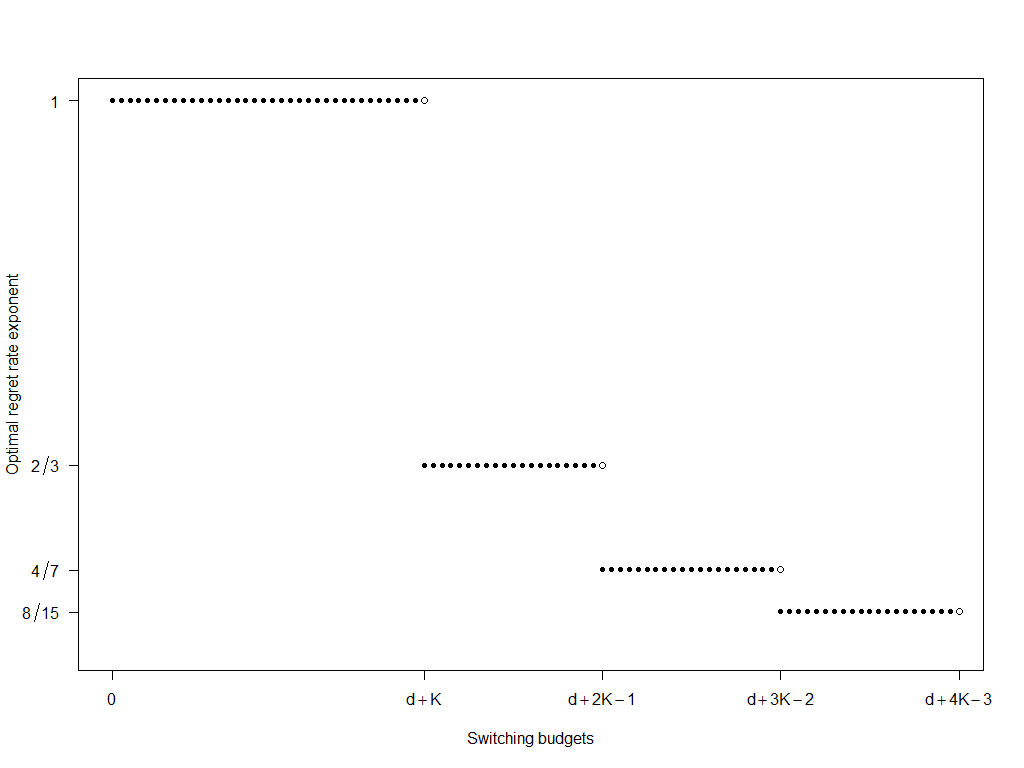}
\vspace{-5pt}
\caption{Optimal regret rate exponent $\lim_{T\rightarrow\infty}\log R^*(T)/\log T$ as a function of switching budget $s$ in the \lsbnrm problem. Here $R^*(T)$ stands for the optimal (i.e., minimax) regret.}
\label{fig:BNRMRegret}
\end{figure}

This paper appears to be one of the first papers to study online learning problems with both resource and switching constraints. 
We fully characterize the statistical complexity of the \lsbnrm problem (as well as the generalization to the \lsbwk problem in the Online Appendix), and provide optimal algorithms to solve the \lsbnrm problem.
We show matching upper and lower bounds on the optimal regret, and design novel limited-switch algorithms to achieve such regret. 
We adopt Landau notations and use $O,\Omega,\Theta$ to hide constant factors, and use $\widetilde{O},\widetilde{\Omega},\widetilde{\Theta}$ to hide both constant and logarithmic factors.
For any positive real number $x \in \bR_+$, we use $\lfloor x \rfloor$ for the largest integer that is smaller or equal to $x$; for any non-positive real number $x \in \bR \setminus \bR_+$, we write $\lfloor x \rfloor = 0$.
Using the above notations, our main results can be summarized as follows. 

First, we provide a computationally efficient limited-switch algorithm and show that the regret is upper bounded by $\widetilde{O}\Big(T^{\frac{1}{2-2^{-\nu(s,d)}}}\Big)$, where $\nu(s,d)=\left\lfloor\frac{s-d-1}{K-1}\right\rfloor$.
Here $s$ stands for the switching budget, $d$ the number of resources, and $K$ the number of price vectors in the \lsbnrm setup (the number of arms in the \lsbwk setup).
Second, we provide matching lower bounds (i.e., impossibility results) on the optimal regret. Specifically,  
for any algorithm with switching budget $s$, 
we construct a class of \bnrm instances such that the algorithm must incur $\widetilde{\Omega} \Big(T^{\frac{1}{2-2^{-\nu(s,d)}}}\Big)$ expected revenue loss on one of these instances.
Combining the above upper and lower bounds, we show that the optimal regret is on the order of $\widetilde{\Theta} \Big(T^{\frac{1}{2-2^{-\nu(s,d)}}}\Big)$. 
Our results show that the optimal regret rate is completely characterized by a piece-wise constant function of the switching budget $s$, which further depends on the number of resources $d$. See Figure~\ref{fig:BNRMRegret} for an illustration.

Our results have two implications.
First, a total number of $\Theta(\log\log T)$ switching budget is necessary and sufficient to achieve the optimal $\widetilde{\Theta}(\sqrt{T})$ regret for the classical \bnrm problem (and \bwk problem as well). 
Compared with existing optimal algorithms for \bnrm that would incur an $\Omega(T)$ switching cost in the worst case, our algorithm achieves a doubly exponential (and best possible) improvement in the switching cost.

Second, our results reveal a separation on the optimal regret between the resource-constrained problem (\bnrm) and the resource-unconstrained problem (\mab).
Under the standard regime, prior literature has shown that both \mab and \bnrm have the same optimal regret rate $\widetilde{\Theta}(\sqrt{T})$.
Our paper shows that when there is a switching budget, the resource-constrained problem (\bnrm) exhibits a larger regret rate than the resource-unconstrained problem (\mab).
We explicitly characterize how the optimal regret rate depends on the number of resource constraints. 
If we fix all the other problem primitives unchanged and only add in one more resource constraint, then the regret bound is going to be larger, illustrating that resource constraints indeed increase the regret --- they make the problem ``harder.''

In addition, we conduct experiments to examine the performance of our algorithms on a numerical setup that is widely used in the literature\footnote{The setup is a \bnrm setup without any switching constraint. We note that when there is a switching constraint, most existing algorithms would incur $\Omega(T)$ regret, and our algorithms should obviously be the best choice.}. 
Compared with benchmark algorithms from the literature (\citealt{besbes2012blind,badanidiyuru2013bandits,ferreira2018online}), our proposed algorithms achieve promising performance with clear advantages on the number of incurred switches. 
Our numerical results also provide practical suggestions to firms on how to design their switching budgets.


\subsection{Challenges and Approaches} \label{sec:Techniques}

\subsubsection{Algorithmic techniques.} 
It is the co-existence of both resource and switching constraints that make our problems  particularly challenging. 
Indeed, when there is no resource constraint, the topic of switching cost has been well-studied in the online learning literature (for both limited and unlimited switching budgets setups); see \cite{agrawal1988asymptotically,agrawal1990multi,cesa2013online,chen2019parametric,chen2020data,cheung2017dynamic,dong2020multinomial,guha2009multi,perakis2023dynamic,simchi2023phase} for various models and results. 
In particular, prior literature has established an ``epoch-based elimination + uniform exploration'' strategy (see, e.g., \citealt{perchet2016batched,gao2019batched}) which enables researchers to design optimal algorithms for \mab with switching constraints (\citealt{simchi2023phase}). 
The ``epoch-based elimination + uniform exploration'' strategy divides the time horizon into multiple {pre-determined} epochs; at each epoch, the algorithm constructs confidence bounds for (the expected reward of) each action,
eliminates actions that are obviously sub-optimal, and then uniformly explores each remaining action for an equal number of periods.

However, the above strategy could easily fail in \bnrm, as two core ideas of the above strategy, namely ``eliminating individual actions'' and ``uniform exploration'', could result in substantial (i.e., linear) regret in \bnrm. 
First, eliminating individual actions is a bad idea in \bnrm. In \bnrm, due to resource constraints, an action's ``earning power'' depends not only on its own reward and resource consumption pattern but also on the reward and resource consumption patterns of other actions that could be effectively combined with it. For example, two actions that each heavily exhaust resources 1 and 2 respectively may be considered ``bad'' if used individually. 
However, combining the two actions proportionally could be optimal and generate significantly higher total revenue than any single action. Therefore, any effective algorithm must decide on \emph{action combinations} rather than individual actions; see Appendix~\ref{sec:failure} for a concrete example of the failure of eliminating individual actions. Given that the space of action combinations is exponentially large, this poses unique statistical and computational challenges for our problem.



Second, unlike in \mab, uniform exploration over actions is not a good idea in \bnrm. Due to the existence of resource constraints, it could be necessary for any effective algorithm to combine different actions in a non-uniform manner in the long term --- the optimal proportion of each action could depend on the characteristics of all actions and has to be learned over time.
The idea of ``choosing actions uniformly'' from \citet{simchi2023phase} could result in substantial (i.e., linear) regret in \bnrm; see Appendix \ref{sec:failure} for a concrete example. 

To address the two challenges described above, we develop a novel ``two-stage linear programming'' (\textsf{2SLP}) approach in Section~\ref{sec:BlindNRMUB}, which provides guidance on how to conduct efficient elimination and exploration in \bnrm.
This approach has two features. 
First, the elimination and exploration is conduced over \textit{action combinations} rather than individual actions.
Second, the complicated optimization tasks involved in computing the elimination and exploration strategies are reduced to solving $(K+1)$ simple linear programs in two stages. 
Since the \textsf{2SLP} approach is computationally efficient, easy to modify, and provides a powerful generalization of the well-known \textsf{successive elimination} principle from \mab to \bnrm, we believe this approach is of independent interest and will be useful in efficiently solving more complex resource-constrained online learning problems.

It is worth mentioning that, without developing \textsf{2SLP} as a new algorithmic principle, it is not apparent whether one can directly modify existing \bnrm and \bwk algorithms to obtain efficient limited-switch algorithms. 
Note that \citet{badanidiyuru2013bandits} and \citet{immorlica2019adversarial} design computationally efficient algorithms for \bwk using adversarial online learning subroutines, which does not seem to work for our purpose as adversarial online learning is shown to require frequent switches \citep{dekel2014bandits,altschuler2018online}.
Note also that by incorporating the delayed update techniques \citep{auer2002finite} into the UCB-type algorithms \citep{agrawal2014bandits}, one may design a modified UCB-type algorithm that achieves $\widetilde{\Theta}(\sqrt{T})$ regret using $O(\log T)$ switches. 
This guarantee is exponentially worse than our guarantee, as our algorithm achieves $\widetilde{\Theta}(\sqrt{T})$ regret using only $O(\log\log T)$ switches.

\subsubsection{Lower bound techniques.}
\label{sec:lower-tech}


Our lower bound proof builds on the ``tracking the cover time'' argument of \cite{simchi2023phase}, who establish regret lower bounds for \mab with a single switching constraint by tracking a series of ordered stopping times and constructing hard \mab instances based on (algorithm-dependent) realizations of the stopping times. 
Extending the argument of \cite{simchi2023phase} from their resource-unconstrained setting to the resource-constrained setting of \lsbnrm is non-trivial, due to the following two reasons.
First, the argument of \cite{simchi2023phase} critically utilizes the fact that the regret is measured against a single fixed action, but in \lsbnrm the regret is measured against a complex dynamic policy (which itself requires switches).
Second, the analysis of \cite{simchi2023phase} is not sensitive to the number of resource constraints $d$ at all, but in order to match the upper bound of \lsbnrm, we need to establish a strengthened lower bound that gradually increases with $d$. 
We address the above two challenges by developing an LP-based analysis framework to construct hard \bnrm instances with specially-designed resource constraints and demand structures, and measuring several revenue gaps based on clean event analysis of the demand realization process.

It is worth mentioning that no prior work in \bnrm and \bwk has tried to construct hard instances that involve multiple ($>1$) resource constraints\footnote{One reason is that using zero or one resource constraint is already enough for their purposes.}. 
Moreover, prior lower bound constructions that involve a single resource constraint (\citealt{badanidiyuru2013bandits,sankararaman2020advances}) are all in the \bwk setup instead of the \bnrm setup.
Since \bwk is more general than \bnrm, constructing lower bound examples for \bnrm is more challenging than for \bwk.
All prior constructions break the specific reward-cost structure of \bnrm, thus failing to provide lower bounds for \bnrm. 
Compared with prior work, our lower bound instance construction is considerably more complicated, as we have to deal with $d$ resource constraints and we do not want to break the \bnrm structure.

\subsubsection*{Roadmap.}

In Section~\ref{sec:Considered} we start with the classical discrete price \bnrm model and then introduce its limited switching budget variant \lsbnrm.
In Section~\ref{sec:NRM} we introduce the deterministic linear program, and present the results in the (simple) distributionally-known case.
While the techniques and results are standard, they build intuitions for our main results.
In Section~\ref{sec:BlindNRM} we introduce our main results in the distributionally-unknown case.
We prove matching upper and lower bounds on the optimal regret and provide optimal and efficient algorithms that achieve the optimal regret.
In Section~\ref{sec:simulation} we conduct an extensive simulation study.
In Section~\ref{sec:LinearDemands} we extend our results to the continuous price setting by assuming that the demand is linear.
We conclude the paper in Section~\ref{sec:Conclusions}.
All the generalizations to the \lsbwk problem are deferred to Sections~\ref{sec:BwK}--~\ref{sec:BwKLS} in the Appendix.

\section{Problem Formulation}
\label{sec:Considered}

We introduce the blind network revenue management (\bnrm) model, and defer the bandits with knapsacks (\bwk) model to Section~\ref{sec:BwK} in the Online Appendix.
The \bnrm problem is a learning version of the classical price-based network revenue management (\nrm) problem. 
The \nrm problem is a distributionally-known stochastic control problem which originates from the airline industry \citep{gallego1997multiproduct, talluri1998analysis}, and has been extensively studied in the revenue management literature with diverse applications \citep{adelman2007dynamic, jasin2014reoptimization, ma2020approximation, topaloglu2009using}. 
The \bnrm problem extends the classical \nrm problem by assuming that the demand distribution is unknown and has to be sequentially learned over time.

The \nrm problem and the \bnrm problem have two distinct setups: a discrete price setup and a continuous price setup.
In Sections~\ref{sec:Considered}~--~\ref{sec:simulation}, we focus on the discrete price setup.
In Section~\ref{sec:LinearDemands} we extend to the continuous price setup when demand is linear.
We refer to \citet{perakis2023dynamic, miao2021network} for more discussions of the continuous price setup when demand is general.

\subsubsection*{\bnrm setup.}
Let $\bN$, $\bR$ and $\bR_+$ be the set of positive integers, real numbers and non-negative real numbers, respectively.
For any $n \in \bN$, define $[n] = \{1,2,...,n\}$.
Let there be a discrete, finite time horizon with $T$ periods.
Time starts from period $1$ and ends in period $T$.
Let there be $n$ different products generated by $d$ different resources.
Each resource is endowed with finite initial inventory $B_i$, $\forall i \in [d]$, and $\Bmin = \min_{i \in [d]} B_i$.
Let $A = (a_{ij})_{i \in [d], j \in [n]}$ be the consumption matrix.
Each entry $a_{ij} \in \bR_+$ stands for the amount of inventory $i \in [d]$ consumed, if one unit of product $j \in [n]$ is sold. 
Let $A_i$ denote the $i$-th row of $A$.
Let $\amax = \max_{i,j} a_{ij}$ be a bounded constant.

In each period $t$, a decision maker can post prices for the $n$ products by selecting a price vector from a finite set of $K$ price vectors $P:=\{\bp_1, ..., \bp_K\}$, which we denote using $\bm{z}_t \in \{\bp_1, ..., \bp_K\}$.
A price vector is $\bp_k = (p_{1,k}, ..., p_{n,k})$, and $p_{j,k}\in[0,\pmax]$ is the price for product $j$ under $\bp_k$.
This captures situations where the prices of all the products must choose from a menu of price points that have been pre-determined by market standards, e.g., a common menu of prices that end in \$9.99: \$59.99, \$69.99, \$79.99, \$89.99. 
In online fashion retailing, the number of feasible price vectors $K$ is usually a small number relative to the number of products $n$.
Price ordering constraints, such as the product with basic features must be cheaper than the premium product, significantly reduce the number of price vectors to choose from \citep{ferreira2016analytics}.
In the above example of 4 prices in a menu, if there are 3 products that must satisfy a price ordering, there are only 4 feasible price vectors to choose from, which is significantly smaller than $4^3 = 64$, the number of feasible price vectors without such price ordering constraints.

Given price vector $\bp_k$, the demand for each product $j \in [n]$ is an unknown but bounded random variable\footnote{All our results can be straightforwardly extended to the more general sub-Gaussian random variables.}, $Q_{j,k} := Q_j(\bp_k) \in [0,1]$, which has to be sequentially learned over time.
Let $q_{j,k} := \bE[Q_{j,k}]$ denote the unknown mean demand for product $j$ under price $\bp_k$, and we collect $\bQ=(Q_{j,k})_{j\in[n],k\in[K]}$, $\bm{q}=(q_{j,k})_{j\in [n],k\in[K]}$. 
For each unit of demand generated for product $j \in [n]$ under price vector $\bp_k$, the decision maker generates $p_{j,k}$ revenue by depleting $a_{ij}$ units of each inventory $i \in [d]$.
If no demand is generated, all the remaining inventory is carried over to the next period.
The selling process stops immediately when the total cumulative demand of any resource exceeds its initial inventory; see Section~\ref{sec:StoppingCriterion} for discussions of alternative stopping rules.
We use $\mathcal{I} = (T,\bm{B},K,d,n,P,A; \bQ)$ to stand for a \bnrm problem instance.


The objective of the decision maker is to maximize the expected total cumulative revenue over $T$ periods. 
The performance is measured by \textit{regret}, which is defined as the worst-case expected revenue loss compared with a clairvoyant decision maker who is endowed with infinite switching budget and knows the true demand distributions but not the realizations. 
The revenue maximization problem is equivalent to a regret minimization problem.

\subsubsection*{Regime for regret analysis.}

We derive non-asymptotic bounds on the regret of policies in terms of the number of time periods $T$. 
For all of our results (except Theorem~\ref{thm:NRMLB}, which uses a linear scaling regime), we adopt the following regret analysis regime: there exists an arbitrary constant $\underline{b}>0$, such that $\Bmin \geq \underline{b} T$.
In other words, we do not assume any specific form of dependence between $T$ and $\bm{B}$.
We only require that inventory is not too scarce compared to the time horizon.
This regime generalizes the standard linear scaling regime in the network revenue management literature; see, e.g.,
\citet{besbes2012blind, bumpensanti2020re, chen2019nonparametric, chen2019network, ferreira2018online, gallego1997multiproduct, jasin2014reoptimization, liu2008choice, sun2020near}.

Following the literature, we assume $\pmax,\amax$ are all constants that do not depend on $T$ or $\bm{B}$. 
The other parameters $K$, $d$, and $n$ do not depend on $T$ or $\bm{B}$, either.
Yet we write out our regret bounds' exact dependence on  $K$, $d$ and $n$ in our theorems for better managerial insights. 
Obtaining regret upper and lower bounds that are tight on the orders of $K$, $d$ and $n$ is an interesting future direction.
For ease of presentation, we also assume that $d<\min\{n,K\}-1$. 
Note that this assumption is only for the purpose of avoiding repeatedly using the notation $\min\{d,n-1,K-1\}$; all our results readily extend to the general case without assuming $d<\min\{n,K\}-1$ by replacing $d$ with $\min\{d,n-1,K-1\}$ and replacing $\nu(s,d)$ with $\left\lfloor\frac{s-\min\{d,n-1,K-1\}-\mathbbm{1}\{\min\{d,n-1\}<K-1\}}{K-1}\right\rfloor$, where $\mathbbm{1} \{ \cdot \}$ stands for an indicator function that takes value 1 if the condition is met..

\subsubsection*{New constraint to \bnrm.}
We model the business constraint of limited price changes as a hard constraint, and define the blind network revenue management under limited switches (\lsbnrm) problem as the \bnrm problem with an extra limited switches constraint.
Specifically, on top of the initial resource capacities, the decision maker is initially endowed with a fixed number of switching budget $s$, to change the price vector from one to another.
When two consecutive price vectors are different, i.e., $\bm{z}_t \ne \bm{z}_{t+1}$, one unit of switching budget is consumed.
For example, if there is only one product, a sequence of prices $(\$79.99, \$69.99, \$79.99, \$69.99)$ uses two distinct prices, and makes three price changes. 
When there is no switching budget remaining, the decision maker cannot change the price vector anymore, and has to keep using the last price vector used.
There are other ways to model the business constraint of limited switches, but all are beyond the scope of this paper.
See Section~\ref{sec:SwitchingBudget} for more discussions.
We can view the \bnrm problem as the \lsbnrm problem under an infinite switching budget.
Since a limited switching budget restricts the family of feasible policies, any feasible policy for the \lsbnrm problem is also a feasible policy for the \bnrm problem.

\subsubsection*{The impact of limited switches constraint.}
The \bnrm problem is extensively studied in the literature, with multiple algorithms developed, e.g., the explore-then-exploit algorithm from \citet{besbes2012blind}, the Balanced Exploration algorithm from \citet{badanidiyuru2013bandits}, the primal-dual algorithms from \citet{badanidiyuru2013bandits} and \citet{immorlica2019adversarial}, the UCB-type algorithm from \citet{agrawal2014bandits}, and Thompson Sampling algorithm from \citet{ferreira2018online}. 
Under the standard linear scaling regime where $T$ and $\Bmin$ are in the same order, it has been shown that the optimal regret rate of the \bnrm problem is $\widetilde{\Theta}(\sqrt{T})$, which is the same as the optimal regret rate of the classical \mab problem, which has no resource constraints.
Existing results characterize a relatively complete picture of the statistical complexity and algorithmic principles for stochastic online learning problems with resource constraints.

The limited switches constraint, however, has not been explored in the literature. 
Notably, all existing near-optimal algorithms for \bnrm require frequently switching between actions --- they all incur $\Omega(T)$ switching cost over $T$ periods. 
The only exception is the explore-then-exploit algorithm of \cite{besbes2012blind}, which controls its number of price changes within $K+d$, but unfortunately suffers from $\Omega(T^{2/3})$ sub-optimal regret rate.

\section{Warm-up: Network Revenue Management under Limited Switches} \label{sec:NRM}

Before we proceed to consider the learning problems, we study the distributionally known case to build better intuitions.
In such case, the distributions of $\bQ$ are known to the decision maker, and the learning problem reduces to a stochastic control problem. 
In this section, since the distributions are known, the \textit{regret} of any policy (with limited switching budget) refers to the expected revenue loss from the optimal policy endowed with unlimited switching budget.\footnote{Unlike the learning problems where the regret is defined by taking a worst case over $\bQ$, the regret considered in this section is instance-dependent because $\bQ$ are already given.}  
Our techniques and results in this section are standard, yet serve as a foundation of Section~\ref{sec:BlindNRM}.

For any set $X$, let $\Delta(X)$ be the set of all probability distributions over $X$.
For any problem instance $\mathcal{I} = (T,\bm{B},K,d,n,P,A;\bm{Q})$, we adopt the general notation $\pi: \bR^d \times [s] \times [T] \to \Delta([K])$ to denote any policy with the full information about stochastic distributions, which suggests a (possibly randomized) price vector to use given the remaining inventory, remaining switching budget, and the remaining periods.
For any $s \in \bN$, let $\Pi[s]$ be the set of policies that changes prices for no more than $s$ times.
For any $s,s' \in \bN$ such that $s \leq s'$, we know that $\Pi[s] \subseteq \Pi[s']$.
Let $\Pi[\infty]$ be the set of all admissible policies.
Let $\Rev(\pi)$ be the expected revenue that policy $\pi$ generates.
Let $\pi^*[s]\in\argmax_{\pi\in\Pi[s]}\Rev(\pi)$ be one of the optimal dynamic policies with switching budget $s$, and $\pi^*[\infty]$ be one of the optimal dynamic policies with an infinite switching budget, i.e., without a switching constraint.

\subsection{The Deterministic Linear Programs} \label{sec:NRMOverview}


For any problem instance $\mathcal{I} = (T,\bm{B},K,d,n,P,A;\bm{Q})$, the literature have extensively studied the following deterministic linear program (DLP) in the \nrm setup. See \cite{gallego1997multiproduct, cooper2002asymptotic, maglaras2006dynamic, liu2008choice}.
\begin{align}
\DLP = \max_{(x_1,\dots,x_K)} \sum_{k\in[K]}\sum_{j\in[n]}p_{j,k} \ q_{j,k} \ x_k &&& \label{eqn:obj} \\
\text{s.t.} \ \sum_{k\in[K]}\sum_{j\in[n]} a_{ij} \ q_{j,k} \ x_k &\leq B_i && \forall\ i \in [d] \label{eqn:constraint:inventory} \\
\sum_{k\in [K]}x_k &\leq T && \label{eqn:constraint:time} \\
x_k & \geq 0 && \forall\ k\in[K] \label{eqn:constraint:NonNeg}
\end{align}
It is well known that in the \nrm setup, the above DLP serves as an upper bound on the expected revenue of any policy, even an optimal policy with an infinite switching budget.
It is well-known that the gap between the expected revenue obtained by the optimal policy and the DLP upper bound is bounded by $O(\sqrt{T})$, i.e., $\Rev(\pi^*[\infty])=\DLP-O(\sqrt{T})$.

Let the set of optimal solutions to the DLP be $X^* = \arg\max_{\bm{x} \in \bR^K} \{ \eqref{eqn:obj} \left| \eqref{eqn:constraint:inventory}, \eqref{eqn:constraint:time}, \eqref{eqn:constraint:NonNeg} \text{ are satisfied} \right.\}$.
For any vector $\bm{x} \in \bR^K$, let $\left\| \bm{x} \right\|_0 = \sum_{k \in [K]} \bI\{x_k \ne 0\}$ be the $L_0$ norm of $\bm{x}$, i.e. the number of non-zero elements in vector $\bm{x}$.
Let $\DSL = \min \{ \left\| \bx \right\|_0 \left| \bx \in X^* \right.\}$ be the least number of non-zero variables of any optimal solution. 
Let $\cDSL = \arg\min \{ \left\| \bx \right\|_0 \left| \bx \in X^* \right.\}$ be the set of such solutions.
For any $\bx^* \in \cDSL$, let $\cZ(\bx^*) = \{ k \in [K] \left| \bx^* \ne 0 \right.\} \subseteq [K]$ be the subset of dimensions that are non-zero in $\bx^*$.
Note that $\DSL$ is an instance-dependent quantity such that $\DSL \leq d+1$, where $d+1$ is the number of all constraints ($d$ resource constraints and $1$ time constraint) in the linear program.
When DLP is non-degenerate, then equality holds $\DSL = d+1$.

In Sections~\ref{sec:NRMLB} and~\ref{sec:NRMUB} we show that for any problem instance, the instance-dependent quantity $\DSL-1$ is a critical switching budget.
If the switching budget is greater or equal to $\DSL-1$, the regret is on the order of $\widetilde{O}(\sqrt{T})$; if the switching budget is less than $\DSL-1$, the regret is on the order of $\Theta(T)$.


\subsection{Lower Bounds} \label{sec:NRMLB}
In this section we show that when the switching budget is less than $\DSL-1$ (at most $\DSL-2$), then a linear regret rate is inevitable.
Recall that $\Pi[\DSL-2]$ stands for the family of admissible policies that make at most $\DSL-2$ changes.

\begin{theorem}
\label{thm:NRMLB}
Let $\bm{b}=\bm{B}/T$ (i.e., $b_1= B_1/T ,\dots,b_d= B_d/T$) be any arbitrary vector of constants.
For any \nrm instance $\mathcal{I} = (T,\bm{B},K,d,n,P,A;\bm{Q})$ with $\bm{B} = \bm{b} T$, $d \geq 0$, $K > d+1$, $n \geq 1$, there is an associated $\DSL$ number (defined in Section \ref{sec:NRMOverview}), such that any policy $\pi \in \Pi[\DSL-2]$ earns an expected revenue:
\begin{align*}
\Rev(\pi) \leq \DLP - c\cdot T,
\end{align*}
where $c>0$ is some distribution-dependent constant that possibly depends on $\bm{b}$ but does not depend on $T$.
\end{theorem}

As a direct implication of Theorem~\ref{thm:NRMLB}, we combine the inequality in Theorem~\ref{thm:NRMLB} with the known fact that $\Rev(\pi^*[\infty]) \ge \DLP - O(\sqrt{T})$ and have $\Rev(\pi) \leq \Rev(\pi^*[\infty]) - \Omega(T)$.
That is, the regret scales linearly with $(T,\bm{B})$ when other parameters are fixed.
Note that, only Theorem~\ref{thm:NRMLB} requires this linear scaling regime; all the other theorems in this paper are described under a more general asymptotic regime (see discussions in Section~\ref{sec:Considered}).

The lower bound established in Theorem~\ref{thm:NRMLB} holds for any $\bm{Q}$. 
Such a result is much stronger than the worst-case type lower bounds (which takes the worst case over $\bm{Q}$) that are widely considered in the revenue management literature.

We outline three key steps here and defer the details of our proof to Appendix~\ref{app:NRMLB}.
We first identify a clean event, such that the realized demands are close to the expected demands that the LP suggests.
This clean event happens with high probability $(1 - \frac{2}{T^3})$.
Second, conditioning on such an event, the maximum amount of revenue we generate is no more than $O(\sqrt{T})$ compared to what the LP suggests;
and the minimum amount of inventory demanded is no less than $O(\sqrt{T})$ compared to what the LP suggests, resulting in no more than $O(\sqrt{T})$ of realized revenue.
In the third step, we show that the regret from insufficient price changes scales on the order of $\Omega(T)$, which dominates the $O(\sqrt{T})$ amount revenue due to randomness.
Such clean event analysis, originating from the online learning literature to prove upper bounds \citep{badanidiyuru2013bandits, slivkins2019introduction, lattimore2020bandit}, was recently used in \citet{arlotto2019uniformly} to prove lower bounds.


\subsection{Upper Bounds} \label{sec:NRMUB}
In this section we show that when the switching budget is greater or equal to $\DSL-1$, then the regret is $\widetilde{O}(\sqrt{T})$.
Such a sub-linear guarantee is achieved by tweaking the well-known static control policy in the network revenue management literature \citep{gallego1997multiproduct, cooper2002asymptotic, maglaras2006dynamic, liu2008choice, ahn2019certainty}.
We tweak the static control policy, so that with high probability the selling horizon never stops earlier than the last period $T$.
See Algorithm~\ref{alg:NRMUB} below.
This is achieved by selecting the value of $\gamma$ in the first step of Algorithm~\ref{alg:NRMUB}.
Similar ideas have been used in \citet{hajiaghayi2007automated, ma2021dynamic, balseiro2019dynamic} to prove asymptotic results in different setups.


\begin{algorithm}[htb]
\caption{Tweaked LP Policy for the \nrm Problem}
\label{alg:NRMUB}
\leftline{{\bf Input:} $\mathcal{I} = (T,\bm{B},K,d,n,P,A;\bm{Q})$.}
\leftline{{\bf Policy:}}
\begin{algorithmic}[1]
\STATE{Define $\gamma = \max\big\{1 - 2 \frac{\amax}{\Bmin}\sqrt{n T\log{T}},0\big\}$.}
\STATE{Solve the DLP as defined by \eqref{eqn:obj}, \eqref{eqn:constraint:inventory}, \eqref{eqn:constraint:time}, and \eqref{eqn:constraint:NonNeg}. Find an optimal solution with the least number of non-zero variables, $\bx^* \in \cDSL$. We assume that $x^*_k, \forall k \in [K]$ are integers, because rounding issues incur a regret of at most $( d \cdot \max_k \bp_k^\top \cdot \bq_k )$, which is negligible compared with $\sqrt{T}$.}
\STATE{Arbitrarily choose any permutation $\sigma: [\DSL] \to \cZ(\bx^*)$ from all $(\DSL)!$ possibilities.}
\STATE{Execute: Set the price vector to be $\bp_{\sigma(1)}$ for the first $\gamma \cdot x^*_{\sigma(1)}$ periods, then $\bp_{\sigma(2)}$ for the next $\gamma \cdot x^*_{\sigma(2)}$ periods, ..., and finally $\bp_{\sigma(\DSL)}$ for the last $T - \gamma \cdot \sum_{l=1}^{\DSL-1}x^*_{\sigma(l)}$ periods.}
\end{algorithmic}
\end{algorithm}

We explain the third step permutation.
Suppose $\cZ(\bx^*) = \{1,3,4\}$.
In this case, $\DSL = 3$ and there are $6$ permutations.
There are $6$ possible policies as suggested in Algorithm~\ref{alg:NRMUB}.
While some of these policies may have better performance than others, they all achieve $\widetilde{O}(\sqrt{T})$ regret.

\begin{theorem}
\label{thm:NRMUB}
Let $\underline{b}>0$ be an arbitrary constant.
For any \nrm instance $\mathcal{I} = (T,\bm{B},K,d,n,P,A;\bm{Q})$ with $T\ge1, d \ge 0, K>d+1$ and $\Bmin/T\ge\underline{b}$, any policy $\pi$ as defined in Algorithm~\ref{alg:NRMUB} satisfies $\pi\in\Pi[\DSL-1]$ and earns an expected revenue:
\begin{align*}
\Rev(\pi) & \geq \DLP-\max\{c/\underline{b}, c'd\} \sqrt{n^3} \sqrt{T\log T} \\
& \geq \Rev(\pi^*[\infty])-\max\{c/\underline{b}, c'd\} \sqrt{n^3} \sqrt{T\log T}
\end{align*}
where $c,c'>0$ are some absolute constants completely determined by $\amax,\pmax$.
\end{theorem}

As we will see in Section~\ref{sec:BlindNRM}, since the loss from an unknown distribution is on the order of $\widetilde{O}(\sqrt{T})$, the $\widetilde{O}(\sqrt{T})$ regret from Algorithm~\ref{alg:NRMUB} suffices to serve as a sub-routine in the last epoch of the main algorithm. 
Even though there are many advanced techniques that improve the $\widetilde{O}(\sqrt{T})$ result, they are beyond the scope of this paper.



\section{Blind Network Revenue Management under Limited Switches} 
\label{sec:BlindNRM}

In this section, we study the \lsbnrm problem, introduce an efficient algorithm, and provide matching upper and lower bounds of the optimal regret.
We start with some definitions.

\subsubsection*{Learning policies and clairvoyant policies.}
In this section, we distinguish between a \bnrm \textit{instance} $\mathcal{I}=(T,\bm{B}, K,d,n,P,A;\bQ)$ and a \bnrm \textit{problem} $\mathcal{P}=(T,\bm{B}, K,d,n,P,A)$ based on whether the underlying demand distributions $\bQ$ are specified or not. Consider a \bnrm problem $\mathcal{P}=(T,\bm{B}, K,d,n,P,A)$.   
Let $\phi$ denote {any} non-anticipating learning policy; specifically, $\phi$ consists of a sequence of (possibly randomized) decision rules $(\phi^t)_{t\in[T]}$, where each $\phi^t$ establishes a probability kernel mapping from the space of historical actions and observations in periods $1,\dots,t-1$ to the space of actions at period $t$. 
For any $s \in \bN$, let $\Phi[s]$ be the set of learning policies that change price vectors for no more than $s$ times almost surely under all possible demand distributions $\bQ$.
For any $s,s' \in \bN$ such that $s \leq s'$, $\Phi[s] \subseteq \Phi[s']$.
Let $\Phi[\infty]$ be the set of all admissible learning policies. 
Let $\Rev_\bQ(\phi)$ be the expected revenue that a learning policy $\phi$ generates under demand distribution $\bQ$.

As we have defined in Section~\ref{sec:Considered}, $\pi$ refers to a \textit{clairvoyant} policy endowed with full distributional information about the distributions $\bQ$.
For any $s \in \bN$, let ${\Pi_\bQ}[s]$ be the set of clairvoyant policies that change price vectors for no more than $s$ times under the true distributions $\bQ$. For any $s,s' \in \bN$ such that $s \leq s'$, $\Pi_{\bQ}[s] \subseteq \Pi_{\bQ}[s']$.
Let $\Pi_\bQ[\infty]$ be the set of all admissible clairvoyant policies.
Let $\Rev_\bQ(\pi)$ be the expected revenue that a clairvoyant policy $\pi\in\Pi_\bQ$ generates under distributions $\bQ$.
Let $\pi_\bQ^*[s] \in \arg\sup_{\pi\in\Pi_\bQ[s]}\Rev(\pi)$ be one optimal clairvoyant policy with switching budget $s$, and $\pi_\bQ^*$ be one of the optimal dynamic policies with an infinite switching budget (i.e., without a switching constraint).

\subsubsection*{Performance metrics.}
The performance of an $s$-switch learning policy $\phi\in\Phi[s]$ is measured against the performance of the optimal $s$-switch clairvoyant policy $\pi_\bQ^*[s]$.
Specifically, for any \bnrm problem $\mathcal{P}$ and switching budget $s$, we define the \textit{$s$-switch regret} of a learning policy $\phi\in\Phi[s]$ as the worst-case difference between the expected revenue of the optimal $s$-switch clairvoyant policy  $\pi_\bQ^*[s]$ and the expected revenue of policy $\phi$: 
\begin{align*}
R_s^\phi(T)& =\sup_{\bQ}\left\{\Rev_{\bQ}(\pi_\bQ^*[s])-\Rev_{\bQ}(\phi)\right\}.
\end{align*}
We also measure the performance of policy $\phi$ against the performance of the optimal unlimited-switch clairvoyant policy $\pi_\bQ^*$. 
Specifically, we define the \textit{overall regret} of a learning policy $\phi\in\Phi[s]$ as the worst-case difference between the expected revenue of clairvoyant policy  $\pi_\bQ^*$ without a switching constraint and the expected revenue of the policy $\phi$:
\begin{align*}
{R}^\phi(T) & =\sup_{\bQ}\left\{\Rev_{\bQ}(\pi_\bQ^*)-\Rev_{\bQ}(\phi)\right\}.
\end{align*}
Intuitively, the $s$-switch regret $R_s^\phi(T)$ measures the ``informational revenue loss'' due to not knowing the demand distributions, while the overall regret $R^\phi(T)$ measures the ``overall revenue loss'' due to not knowing the demand distributions and not being able to switch freely. Clearly, the overall regret $R^\phi(T)$ is always larger than the $s$-switch regret $R_s^\phi(T)$. 
Interestingly, as we will show later, for all $s$, ${R}^\phi(T)$ and ${R}_s^\phi(T)$ are always on the same order in terms of dependence on $T$.

\subsection{Upper Bounds} \label{sec:BlindNRMUB}

We propose a computationally efficient algorithm that provides an upper bound on both the $s$-switch regret and the overall regret. Our algorithm, called \textit{Limited-Switch Learning via Two-Stage Linear Programming} (\textsf{LS-2SLP}), is described in Algorithm~\ref{alg:bsse}.
In Algorithm~\ref{alg:bsse} and onwards, for any vector $\bm{x} \in \bR^K$ and any $k \in [K]$, let $(\bm{x})_k$ be the $k^{\text{th}}$ element of vector $\bm{x}$.




The design of our algorithm builds on the \textsf{SS-SE} algorithm proposed in \cite{simchi2023phase}, the \textsf{Balanced Exploration} algorithm proposed in \cite{badanidiyuru2013bandits}, and the \textsf{Tweaked LP} policy defined in Algorithm \ref{alg:NRMUB}. 
To address the fundamental challenges inherent in our problems (as illustrated in Section~\ref{sec:Techniques}), we go beyond the above algorithms and develop novel ingredients for efficient exploration and exploitation under both resource and switching constraints. 
We provide more details below.

\begin{algorithm}[htbp]
\caption{Limited-Switch Learning via Two-Stage Linear Programming (\textsf{LS-2SLP})}
\label{alg:bsse}
\leftline{{\bf Input:} Problem parameters $(T,\bm{B},K,d,n,P,A)$; switching budget $s$.}
\leftline{{\bf Initialization:} Calculate $\nu(s,d)=\left\lfloor\frac{s-d-1}{K-1}\right\rfloor$. Define $t_0=0$ and}
\begin{equation}\label{eq:points}
t_l=\left\lfloor K^{1-\frac{2-2^{-(l-1)}}{2-2^{-\nu(s,d)}}}T^{\frac{2-2^{-(l-1)}}{2-2^{-\nu(s,d)}}}\right\rfloor,~~\forall l=1,\dots,\nu(s,d)+1.
\end{equation}
{Set $\gamma=\max\bigg\{1- 17\frac{\amax\sqrt{n(d+1)\log{[(d+1)KT]}}\log T}{\Bmin}{t_1},0\bigg\}$.} \\
\leftline{{\bf Notation:} {Let $T_l$ denote the ending period of epoch $l$ (which will be determined by the algorithm).}}
{Let $\bm{z}_t \in \{\bp_1,\dots,\bp_K\}$ denote the algorithm's selected price vector at period $t$. Let $\bm{z}_0 \in \{\bp_1,\dots,\bp_K\}$ be a random price vector.}\\
{{\bf Policy:}}
\begin{algorithmic}[1]
\FOR{epoch $l=1,\dots,\nu(s,d)$}
\IF{$l=1$}
\STATE{Set $T_0=L^{\mathsf{rew}}_k(0)=L^{\mathsf{cost}}_{i,k}(0)=0$ and $U^{\mathsf{rew}}_k(0)=U^{\mathsf{cost}}_{i,k}(0)=\infty$ for all $i\in[d],k\in[K]$.}
\ELSE
\STATE{Let $n_{k}(T_{l-1})$ be the total number of periods that price vector $\bp_k$ is chosen up to period $T_{l-1}$, and $\bar{q}_{j,k}(T_{l-1})$ be the empirical mean demand of product $j$ sold at price vector $\bp_k$ up to period $T_{l-1}$. Calculate $\radius_k(T_{l-1})=\sqrt{\frac{\log\left[(d+1)KT\right]}{n_k(T_{l-1})}}$ and 
\[
\begin{cases}U^{\mathsf{rew}}_{k}(T_{l-1})=\min\left\{\sum_{j\in[n]}p_{j,k}\bar{q}_{j,k}(T_{l-1})+||\bp_k||_2\radius_k(T_{l-1}),U^{\mathsf{rew}}_k(T_{l-2})\right\}
,\\L^{\mathsf{rew}}_{k}(T_{l-1})=\max\left\{\sum_{j\in[n]}p_{j,k}\bar{q}_{j,k}(T_{l-1})-||\bp_k||_2\radius_k(T_{l-1}),L_k^{\mathsf{rew}}(T_{l-2})\right\},\end{cases}~~\forall k\in[K],
\]
\[
\begin{cases}U^{\mathsf{cost}}_{i,k}(T_{l-1})=\min\left\{\sum_{j\in[n]}a_{ij}\bar{q}_{j,k}{(T_{l-1})}+||A_i||_2\radius_k(T_{l-1}),U^{\mathsf{cost}}_{i,k}(T_{l-2})\right\}
,\\L^{\mathsf{cost}}_{i,k}(T_{l-1})=\max\left\{\sum_{j\in[n]}a_{ij}\bar{q}_{j,k}{(T_{l-1})}-||A_i||_2\radius_k(T_{l-1}),L_{i,k}^{\mathsf{cost}}(T_{l-2})\right\},\end{cases}~~\forall i\in[d],\forall k\in[K].
\]}
\ENDIF
\STATE{Solve the first-stage pessimistic LP:
\begin{align*}
\mathsf{J}^{\mathsf{PES}}_{l} = \max_{(x_1,\dots,x_K)} \sum_{k\in[K]}L^{\mathsf{rew}}_k(T_{l-1})  x_k &&  \\
\text{s.t.} \ \sum_{k\in[K]}U_{i,k}^{\mathsf{cost}}(T_{l-1})  x_k &\leq B_i & \forall i \in [d] \\
\sum_{k\in [K]}x_k &\leq T &  \\
x_k & \geq 0 & \forall k\in[K] 
\end{align*}}
\algstore{myalg}
\end{algorithmic}
\end{algorithm}

\begin{algorithm}                     
\begin{algorithmic}[1]
\algrestore{myalg}
\STATE{For each $j\in[K]$, solve the second-stage exploration LP:
\begin{align*}
\bx^{l,j} = \arg\max_{(x_1,\dots,x_K)} \ x_j &&  \\
\text{s.t.} \ \sum_{k\in[K]}U^{\mathsf{rew}}_{k}(T_{l-1})x_k&\ge \mathsf{J}_{l}^{\mathsf{PES}} &\\
\sum_{k\in[K]}L^{\mathsf{cost}}_{i,k}(T_{l-1}) x_k &\leq B_i & \forall i \in [d]\\
\sum_{k\in [K]}x_k &\leq T &  \\
x_k & \geq 0 & \forall\ k\in[K] 
\end{align*}
}
\STATE{For all $k\in[K]$, let $N_k^l=\frac{(t_{l}-t_{l-1})}{T}\sum_{j=1}^{K}\frac{1}{K}(\bx^{l,j})_k$. Let $\bm{z}_{T_{l-1}+1}=\bm{z}_{T_{l-1}}$. Starting from this action, choose each price vector $\bp_k$ for $\gamma N_k^l$ consecutive periods, $k\in[K]$ (we overlook the rounding issues here, which are easy to fix in regret analysis). Stop the algorithm once time horizon is met or one of the resources is exhausted.} 
\STATE{End epoch $l$.  Mark the last period in epoch $l$ as $T_l$.}
\ENDFOR
\STATE{For epoch $\nu(s,d)+1$ (the last epoch), let $\bar{q}_{j,k}(T_{\nu(s,d)})$ be the empirical mean demand of product $j$ sold at price vector $\bp_k$ up to period $T_{\nu(s,d)}$. Solve the following deterministic LP
\begin{align*}
\max_{(x_1,\dots,x_K)} \sum_{k\in[K]}\sum_{j\in[n]}p_{j,k} \ \bar{q}_{j,k}(T_{\nu(s,d)}) \ x_k &&& \\
\text{s.t.} \ \sum_{k\in[K]}\sum_{j\in[n]} a_{ij} \ \bar{q}_{j,k}(T_{\nu(s,d)}) \ x_k &\leq B_i && \forall\ i \in [d] \\
\sum_{k\in [K]}x_k &\leq T && \\
x_k & \geq 0 && \forall\ k\in[K],
\end{align*}
and find an optimal solution with the least number of non-zero variables, $\bx^*_{\tilde{\bq}}$. Let $N_k^{\nu(s,d)+1}=\frac{(T-t_{\nu(s,d)})}{T}{(\bx^*_{\tilde{\bq}})_k}$ for all $k\in[K]$. First let $\bm{z}_{T_{\nu(s,d)}+1}=\bm{z}_{T_{\nu(s,d)}}$. Start from this action, choose each price vector $\bp_k$ for $\gamma N_k^{\nu(s,d)+1}$ consecutive periods, $k\in[K]$ (we overlook the rounding issues here, which are easy to fix in regret analysis). Stop the algorithm once time horizon is met or one of the resources is exhausted. End epoch $\nu(s,d)+1$.}
\end{algorithmic}
\end{algorithm}

\subsubsection{Description of the algorithm.}

Our algorithm runs in an epoch schedule which improves the epoch schedule of the \textsf{SS-SE} algorithm (\citealt{simchi2023phase}). 
Specifically, our \mainalgo algorithm first computes an index $\nu(s,d)$ based on the switching budget $s$, the number of actions $K$, and the number of resource constraints $d$, then computes a series of fixed time points $\{t_l\}_{l=1}^{\nu(s,d)+1}$ according to formula (\ref{eq:points}) (see also \citealt{perchet2016batched,gao2019batched} for the use of such formulas in \mab), which provides guidance on how to divide the $T$ selling periods into $\nu(s,d)+1$ epochs.  
Compared with the \textsf{SS-SE} algorithm  which directly uses the pre-determined sequence $\{t_l\}_{l=1}^{\nu(s,d)+1}$ as its epoch schedule, our algorithm exhibits two notable differences in determining the epoch schedule.
First, the parameter $\nu(s,d)$ takes account of the number of resource constraints $d$. Second, our algorithm uses an \textit{adaptive} epoch schedule $\{T_l\}_{l=1}^{\nu(s,d)+1}$ rather than the pre-determined schedule $\{t_l\}_{l=1}^{\nu(s,d)+1}$ --- in particular, our algorithm decides the length of the next epoch only \textit{after} the current epoch ends, and the length would be determined by both $\{t_l\}_{l=1}^{\nu(s,d)+1}$ and the data collected so far.
Such an adaptive epoch schedule is crucial for our algorithm to achieve the desired theoretical guarantee\footnote{We provide a more detailed explanation for this point. Almost all existing \bnrm and \bwk literature assumes a ``null arm'' with zero reward and zero costs and allows the algorithms to repetitively switch to the null arm. While such an assumption is completely fine (and without loss of generality) in the classical \bnrm / \bwk setting, the behavior of repetitively switching to a null arm would cause a waste of switching budget in our limited-switch setting, and this would make the algorithm suboptimal. Consequently, our algorithm has to  plan for an ``early stopping'' of each epoch, which requires the epoch schedule to be adaptive.}.

During each epoch except for the last one, our algorithm strikes a balance between exploration and exploitation via a Two-Stage Linear Programming (\textsf{2SLP}) scheme. 
Specifically, our algorithm first builds high-probability upper and lower confidence bounds on the purchase probability of each price vector, based on the demand data collected so far.
Then, the algorithm solves a first-stage pessimistic LP, which is a ``pessimistic'' variant of the DLP studied in Section~\ref{sec:NRM}, with the reward of each action being as underestimated as possible and the consumption of each action being as overestimated as possible.
Intuitively, the optimal value of this pessimistic LP serves as a conservative estimate on how much accumulated revenue should be generated by a ``plausible enough'' policy. 
The algorithm then moves to the second stage, where it solves $K$ linear programs.
For any $j \in [K]$, the $j^{\text{th}}$ linear program considers how to execute each action, with an objective of exploring action $j$ as many periods as possible, subject to $d$ constraints on the inventory consumption, and an extra constraint on generating at least as much revenue as what the pessimistic LP suggests --- such a constraint ensures that the exploration of action $j$ cannot be too extensive to hurt the revenue.  
Different from the pessimism in the first stage, all the constraints in the second stage are specified in an ``optimistic'' manner, with the reward of each action being as overestimated as possible and the consumption of each action being as underestimated as possible. 
By doing so, the $j^{\text{th}}$ linear program implicitly encourages exploring action $j$ more (while still keeping its solution approximately plausible enough).  
Finally, our algorithm makes decisions by exploring each arm in a ``balanced'' fashion, i.e., it computes an average of the $K$ linear programming solutions obtained in the second stage, and then determines the total number of periods to execute each action in this epoch based on the average.

In the very last epoch, our algorithm implements Algorithm~\ref{alg:NRMUB} to conduct pure exploitation, by using the empirical distributions estimated from the data as the stochastic distributions of $\bQ$.
It is worth noting that, in the special case of $\nu(s,d)=1$, there are only two epochs, and  the \textsf{LS-2SLP} algorithm becomes essentially the same as the explore-then-exploit algorithm in \citet{besbes2012blind}, with some slight difference such as the epoch schedule's dependence on $K$. 
We will numerically compare these two algorithms in Section~\ref{sec:simulation}.

\subsubsection{Discussion of the \textsf{2SLP} scheme.} 
The \textsf{2SLP} scheme builds on the insights from the \textsf{Balanced Exploration} algorithm (\citealt{badanidiyuru2013bandits}), which extends the celebrated ``successive elimination'' idea by choosing over \textit{mixtures} of actions (rather than individual actions) and ensuring that the choice is ``balanced'' across actions. 
The \textsf{Balanced Exploration} algorithm is however computationally inefficient, as it conducts elimination over mixtures of actions in an explicit and exact manner (which requires one to solve infinitely many linear programs), and requires an ``approximate optimization over a (complicated) infinite-dimensional set'' step which the authors do not provide an implementation (see Remark 4.2 in \cite{badanidiyuru2018bandits}). 
The \textsf{2SLP} approach bypasses the computational difficulty inherent in \textsf{Balanced Exploration} by making the elimination procedure \textit{implicit} and \textit{relaxed}, and reducing the hard optimization tasks to simply solving $(K+1)$ linear programs. 
Such improvements are enabled by the careful design of the first-stage pessimistic LP and the second-stage exploration LP's.
Interestingly, the \textsf{2SLP} approach starts with pessimism in the first stage and turns to optimism in the second stage --- such a combination of pessimism and optimism seems novel and may be of independent interest. 
As a comparison, the UCB-type algorithm in \cite{agrawal2014bandits} is fully optimistic. We also note that a fully optimistic algorithm seems to face fundamental limitations in our problem; see the last part of Appendix \ref{sec:failure} for more discussions.


\subsubsection{Theoretical guarantees.}
The \mainalgo algorithm is a limited-switch algorithm. 
During each epoch except for the last one, the \textsf{LS-2SLP} policy chooses at most $K$ actions, thus making at most $K-1$ switches between them.
During the last epoch (when the algorithm does purely exploitation), since there are $d+1$ constraints in the deterministic LP, the optimal solution contains at most $d+1$ non-zero solutions.
Yet the last action executed during the second last epoch is not necessarily among the non-zero solutions, thus it requires at most $d+1$ switches.
There are $\nu(s,d) = \lfloor \frac{s-d-1}{K-1} \rfloor$ epochs before the last exploitation epoch, so there are at most $\nu(s,d) \cdot (K-1) + (d+1) \leq s$ switches, satisfying the definition of the $s$-switch learning policy.
We then establish the following upper bound on the regret incurred by the \textsf{LS-2SLP} policy.

\begin{theorem}
\label{thm:BlindNRMUB}
Let $\phi$ be the \mainalgo policy as suggested by Algorithm~\ref{alg:bsse}. Let $\underline{b}>0$ be any arbitrary constant. For any \textsf{BNRM} problem $\mathcal{P}=(T,\bm{B},K,d,n,P,A)$ with $T\ge1, d \ge 0, K>d+1$ and $\Bmin/T\ge\underline{b}$, $\phi$ is guaranteed to be a $s$-switch learning policy, and the regret incurred by $\phi$ satisfies
\begin{align*}
R^\phi_s(T)\le R^\phi(T)\le \left(\max\{c/\underline{b},c'\}\cdot \sqrt{n^3(d+1)\log[(d+1)KT]} K^{1-\frac{1}{2-2^{-\nu(s,d)}}}\log T\right)\cdot T^{\frac{1}{2-2^{-\nu(s,d)}}},
\end{align*}
where $\nu(s,d)=\left\lfloor\frac{s-d-1}{K-1}\right\rfloor$, and $c,c'>0$ are some absolute constants completely determined by $\amax,\pmax$.
\end{theorem}

It is worth noting that the above upper bound holds in a \textit{non-asymptotic} sense: it holds for all finite $T$ and $\bm{B}$, as long as $\Bmin/T$ is lower bounded by a positive constant $\underline{b}$.
The proof of Theorem~\ref{thm:BlindNRMUB} is deferred to Section~\ref{app:BlindNRMUB} in the appendix. 
Compared to prior analysis of uniform exploration-type strategies (e.g., \cite{simchi2023phase}) where key quantities such as $n_k(T_{l-1})$ (and thus the confidence bounds) admit simple analytical forms, our analysis of the \textsf{2SLP} addresses non-trivial technical challenges in dealing with non-analytical forms of $n_k(T_{l-1})$ and related quantities via LP relaxation analysis.



\subsection{Lower Bounds} 
\label{sec:BlindNRMLB}

In this section we prove a matching lower bound, which holds for both the $s$-switch regret and the overall regret.
This lower bound is based on the construction of a family of problem instances in the \lsbnrm setup, combined with non-trivial information-theoretic arguments.
Note that our lower bound is established for all $T,\bm{B},K,d,n,s$ (under certain weak conditions), which is substantially stronger than a single lower bound demonstrated for specific values of $T,\bm{B},K,d,n,s$.

\begin{theorem}
\label{thm:BlindNRMLB} 
Let $\underline{b}>0$ be an arbitrary constant. 
For any $T\ge 1, d\ge 0, K\ge 2(d+1)$, $n\ge d+1$ and $\bm{B}$ such that $B_i/T\in[\underline{b},1]$ for all $i\in[d]$, there exist $P,A$,  such that for the \bnrm problem $\mathcal{P}=(T,\bm{B},K,d,n,P,A)$, for any switching budget $s\ge0$ and any  $\phi\in\Phi[s]$, 
\begin{align*}
R^\phi(T)\ge R^\phi_s(T)\ge\left(\min\{c\underline{b},c'\}\cdot{(d+1)^{-3}K^{-\frac{3}{2}-\frac{1}{2-2^{-\nu(s,d)}}}}{(\log T)^{-\frac{5}{2}}}\right)\cdot T^{\frac{1}{2-2^{-\nu(s,d)}}},
\end{align*}
where $\nu(s,d)=\left\lfloor\frac{s-d-1}{K-1}\right\rfloor$, and $c,c'>0$ are some numerical constants that do not depend on any problem parameters.
\end{theorem}

The proof of Theorem~\ref{thm:BlindNRMLB} is highly non-trivial. We outline the proof idea below. The complete proof is deferred to Section~\ref{app:BlindNRMLB} in the appendix.

\textsc{Proof Idea.}
We construct a \bnrm problem $\mathcal{P}$ as follows. 
For any problem input $\bm{B} \in \bR^d$, define $b_1=B_1/T,\dots,b_d=B_d/T$. 
Let there be $n\ge d+1$ products. 
Let the $d\times n$ consumption matrix be
\begin{align*}
A= 2 \cdot
\bigl[
\bm{0}_{d\times1} \quad \text{diag}(b_1,\dots,b_d) \quad \bm{0}_{d\times (n-(d+1))}
\bigr],
\end{align*}
where $\text{diag}(b_1,\dots,b_d)$ stands for the $d\times d$ diagonal matrix whose diagonal entries are $b_1,\dots,b_d$. For any $j\in[n]$, $k\in[K]$, let the price be
\begin{align*}
p_{j,k}=\begin{cases}1,&\text{if } j=1,\\
0, & \text{otherwise.}
\end{cases}
\end{align*}
Based on the above \bnrm problem $\mathcal{P}$, we will construct different \bnrm instances by specifying different demand distributions $\bQ$.

We prove Theorem~\ref{thm:BlindNRMLB} even when we restrict $\bQ$ to Bernoulli demand distributions. 
Recall that we use $q_{j,k} = \bE[Q_{j,k}]$ to stand for the mean value of the distribution $Q_{j,k}$. 
When restricted to Bernoulli distributions, such a $q_{j,k}$ uniquely describes the distribution of $Q_{j,k}$. 
Thus every  $\bq \in[0,1]^{n\times K}$ uniquely determines a \bnrm instance  $\cI_{\bq}:=(T,\bm{B},K,d,n,P,A,s;\bq)$. 
Specifically, we parameterize $\bq$ by two vectors $\bmu_1\in[-\frac{1}{2},\frac{1}{2}]^{K}$ and $\bmu_2\in[-\frac{1}{2},\frac{1}{2}]^{K}$, such that for all $k\in[K]$ and $j\in[n]$,
\begin{align*}
q_{j,k;\bmu}=\begin{cases}\frac{1}{2}+\mu_{1,k},&\text{if }j=1,\\
\frac{1}{2}-\mu_{2,k},&\text{else if }j=(k-1)\%(d+1)+1,\\
\frac{1}{2}+\mu_{2,k},&\text{else if }j=k\%(d+1)+1,\\
\frac{1}{2}, &\text{else if }j\in[2,d+1],\\
0, &\text{else},
\end{cases}
\end{align*}
where $\%$ stands for the modulo operation (which returns the non-negative remainder of a division). 
In the lower bound proof, we will assign different values to $\bmu$ to construct different \bnrm instances. 
Below we will use $\cI_{\bmu}:=(T,\bm{B},K,d,n,P,A,s;\bmu)$ to stand for a \bnrm instance, which highlights the dependence on $\bmu$. 
Let $\mathsf{DLP}_{\bmu}$ denote the DLP as defined by \eqref{eqn:obj}, \eqref{eqn:constraint:inventory}, \eqref{eqn:constraint:time}, \eqref{eqn:constraint:NonNeg}, on instance $\cI_{\bmu}$:
\begin{align}
\mathsf{J}^{\mathsf{DLP}_{\bmu}} =\max_{\bx}  & ~\sum_{k\in[K]} \left(\frac{1}{2}+\mu_{1,k}\right)x_k  \notag \\
\text{s.t.} & ~b_i\left(\frac{1}{2}\sum_{k\in[K]} x_k + \sum_{k': k'\%(d+1)=i}\mu_{2,k'}x_{k'}- \sum_{k'': k''\%(d+1)=i+1} \mu_{2,k''}x_{k''}\right)\leq \frac{b_i T}{2}, ~~~~ \forall\ i \in [d], \notag \\
&~\sum_{k\in[K]}x_k\le T,\notag\\
&~x_k  \geq 0,~~~~ \forall k\in[K]. \notag
\end{align}
In Lemma \ref{lem:WorstCaseLP}, we conduct a thorough analysis of the above type of linear programs, and show that for a family of properly structured $\bmu$, $\mathsf{DLP}_{\bmu}$ is always non-degenerate --- this means that $\DSL$, the least number of non-zero components of any optimal solution to $\mathsf{DLP}_{\bmu}$, is equal to $d+1$. 
By Theorem~\ref{thm:NRMLB}, for any such \bnrm instance $\cI_{\bmu}$, even for a clairvoyant policy that knows $\bmu$ in advance, it needs to make at least $d$ switches to guarantee a sublinear regret (note that $\bmu$ is treated as a fixed quantity independent of $T$ in this statement); 
in Lemma \ref{lem:WorstCaseLP}, we strengthen the above statement and show that even when $\bmu$ is not fixed and can depend on $T$, any policy still needs to make at least $d$ switches to avoid a large regret. 
Moreover, since a learning policy $\phi$ does not know the value of $\bmu$ in advance, it has to make much more switches than the clairvoyant policy to learn $\bmu$. In the rest of the proof,  we take into account the effect of unknown demand distributions and show a lower bound  for both $R^\phi(T)$ and $R^\phi_s(T)$, under any switching budget $s$ --- the basic idea is that any limited-switch learning policy faces some fundamental difficulties in distinguishing similar but different $\bmu$, which necessarily leads to certain worst-case revenue loss, and we can measure it using certain linear programs.

Specifically, for any $s$-switch learning policy $\phi\in\Phi[s]$, we construct a class of \bnrm instances by adversarially choosing a class of $\bmu$, such that policy $\phi$ must incur an expected revenue loss of $\widetilde{\Omega}\left(T^{\frac{1}{2-2^{-\nu(s,d)}}}\right)$ under one of the constructed instances. A challenge here is that $\phi$ is an arbitrary and abstract $s$-switch policy --- we need  more information about $\phi$ to design $\bmu$. We address this challenge by developing an enhanced version of the  ``tracking the cover time'' argument. 
The ``tracking the cover time'' argument was originally introduced in \cite{simchi2019phase}, the conference version of \cite{simchi2023phase}, to derive regret lower bounds for \mab with limited switches. Their setting can be viewed as a special case of our problem when $d = 0$. Their approach involves tracking a sequence of ordered stopping times and constructing hard \mab instances based on algorithm-dependent realizations of these stopping times. 
Since we are proving a larger regret lower bound here (our lower bound  gradually increases as $d$ increases), we have to utilize the structure of resource constraints and incorporate their ``complexity'' into the construction of hard instances which lead to our lower bound. We thus develop novel techniques beyond \cite{simchi2019phase,simchi2023phase}, i.e., incorporating the complexity of resource constraints into the ``tracking the cover time'' argument, which requires us to strategically design $\bmu$ according to Lemma \ref{lem:WorstCaseLP} and utilize several LP benchmarks; see details in Section~\ref{app:BlindNRMLB} in the appendix.


\subsection{Implications}
\subsubsection*{$O(\log\log T)$ switches are sufficient for \bnrm and \bwk.} Plugging $s=(K-1)\log_2\log_2 T+d+1$ into Algorithm \ref{alg:bsse} (or Algorithm \ref{alg:bsse2}) and Theorem~\ref{thm:BlindNRMUB} (or Proposition~\ref{prop:BwKUB}), we obtain an algorithm that achieves the optimal $\widetilde{\Theta}(\sqrt{T})$ regret for the classical \bnrm problem, while using only $O(\log\log T)$ switching budget. 
Note that $\Omega(\log\log T)$ switching budget is necessary for obtaining the  $\widetilde{\Theta}(\sqrt{T})$ regret even in the simpler \mab setting, see \cite{simchi2023phase}. 
Our algorithm and result thus show that a total number of $\Theta(\log\log T)$ switching budget is necessary and sufficient to achieve the optimal $\widetilde{\Theta}(\sqrt{T})$ regret for the classical \bnrm problem. 
Compared with existing optimal algorithms that require $\Omega(T)$ switching cost in the worst case, our algorithm achieves a \emph{doubly exponential} improvement on the switching cost.

\subsubsection*{Regret impact of resource constraints.}

Combining Theorem~\ref{thm:BlindNRMUB} and Theorem~\ref{thm:BlindNRMLB}, we can see that for any switching budget $s$, the optimal  regret of the \lsbnrm problem is on the order of $\widetilde{\Theta}\left(T^{\frac{1}{2-2^{-\left\lfloor\frac{s-d-1}{K-1}\right\rfloor}}}\right)$.
This suggests that given a fixed switching budget $s$, an increase in the number of resources $d$ may result in an increase in the order of the optimal regret.
To the best our knowledge, this is the first result of such kind that explicitly characterizes how the dimension of the resource constraints makes a stochastic online learning problem ``harder''.
In other words, an increase in the number of resources increases the required number of switches to achieve a same regret bound.
Note that both the classical \mab problem and the \bnrm problem studied in the literature essentially exhibit the same optimal regret rate on the order of $\widetilde{\Theta}(\sqrt{T})$, where the regret rate is not affected by the dimension of the resource constraints.
Our results indicate a separation of the optimal regret rates associated with the dimension of the resource constraints, due to the existence of a switching constraint.

\subsubsection*{Managerial implications.}
There are two managerial implications.
First, the length of each epoch is increasing, i.e., more of the price changes occur early in the selling season. 
This implication suggests managers should be more cautious in the earlier phase of the selling season.
Informally, this is because in the earlier epochs, we do not want to incur huge regret by sticking to each price vector for too long.
Later on for each epoch, we have relatively betteintur understanding of the underlying demand distributions, 
so we can are more confident in choosing each price vector
for a longer duration of time.
Second, our algorithm crucially relies on constructing upper and lower confidence bounds for both revenue and cost of each price vector.
We then use such confidence bounds to solve some linear programs, which would suggest which prices to use.
Practitioners should keep in mind that for an inventory-constrained problem, it is not necessarily the ratio between revenue and cost, but both revenue and cost that matter.

\section{Simulation Study} 
\label{sec:simulation}

We consider the blind network revenue management problem and follow the same setup as in \citet{besbes2012blind, ferreira2018online}.\footnote{We do not impose a switching constraint here. We note that when there is a switching constraint, many benchmark algorithms would incur $\Omega(T)$ regret, and our algorithms would naturally be the best choice.}
Suppose there are 2 products and 3 resources.
Product 1 consumes 1 unit of resource 1, 3 units of resource 2, and no unit of resource 3; Product 2 consumes 1 unit of resource 1, 1 unit of resource 2, and 5 units of resource 3.
So the first column of the demand consumption matrix is $(1,3,0)^\top$, and the second column is $(1,1,5)^\top$.
The set of feasible prices is given by $(p_1, p_2) \in \{(1, 1.5),(1, 2),(2, 3),(4, 4),(4, 6.5)\}$.
We vary the initial inventory to be one of the two following, $(B_1, B_2, B_3) \in \left\{(0.3, 0.5, 0.7) \cdot T, (1.5, 1.2, 3) \cdot T \right\}$, which we refer to as small inventory scenario and large inventory scenario, respectively.
For each inventory scenario, we vary the time horizon between $\{1000, 2000, ..., 10000\}$.
Finally, we consider three different demand models: a linear model, an exponential model and a logit model.

\begin{enumerate}
\item Linear model: $q_1 = \max\{0,0.8 - 0.15 p_1\}$, $q_2 = \max\{0,0.9 - 0.3 p_2\}$.
\item Exponential model: $q_1 = 0.5 \exp(-0.5 p_1)$, $q_2 = 0.9 \exp(- p_2)$.
\item Logit model: $q_1 = \frac{\exp(-p_1)}{1+\exp(-p_1)+\exp(-p_2)}$, $q_2 = \frac{\exp(-p_2)}{1+\exp(-p_1)+\exp(-p_2)}$
\end{enumerate}

\begin{figure}[!htb]
\footnotesize
\begin{subfigure}{.47\textwidth}
\centering
\includegraphics[width=\textwidth]{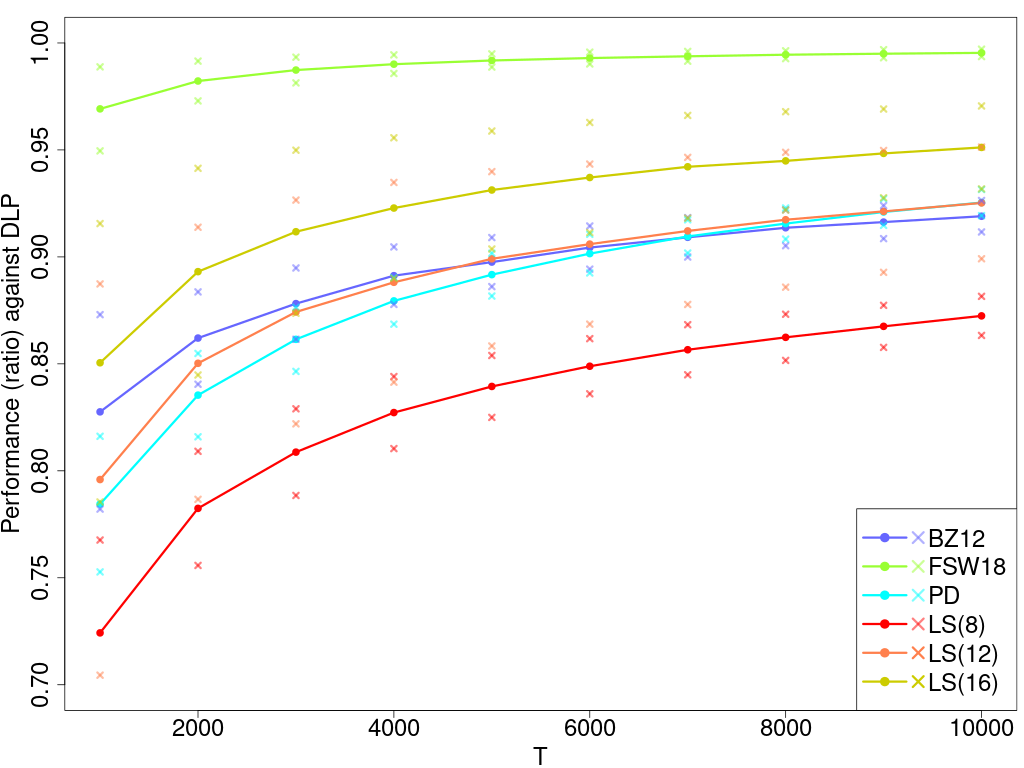}
\caption{\footnotesize Linear demand, small inventory.}
\end{subfigure}\hfill
\begin{subfigure}{.47\textwidth}
\centering
\includegraphics[width=\textwidth]{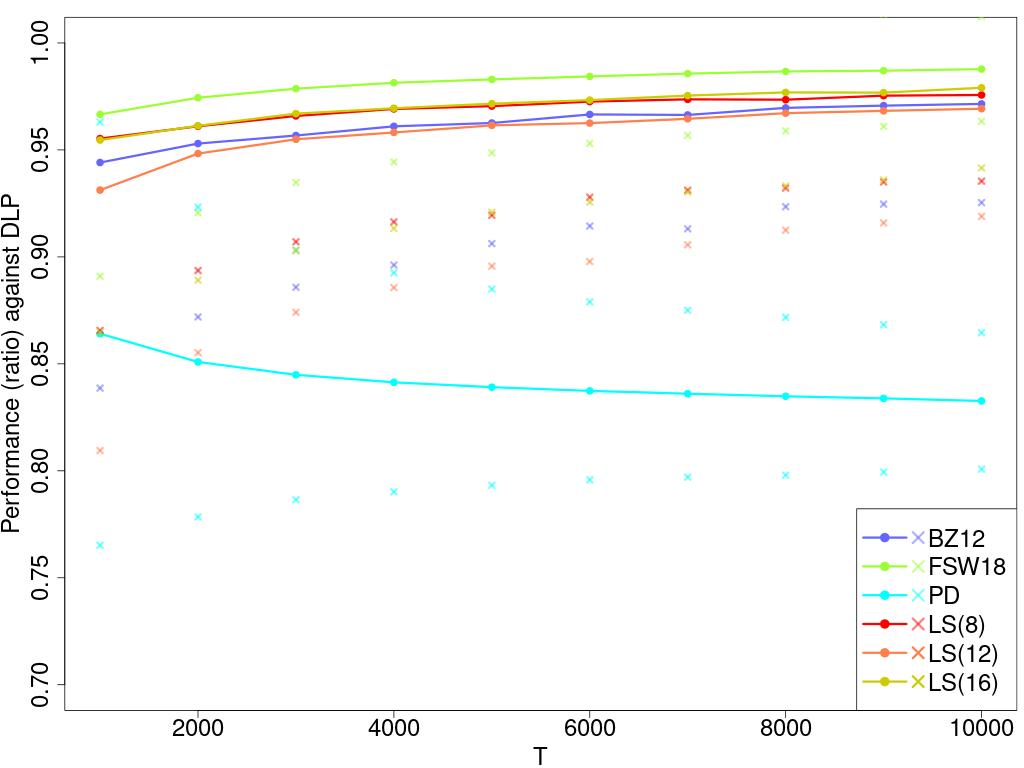}
\caption{\footnotesize Linear demand, large inventory.}
\end{subfigure}

\begin{subfigure}{.47\textwidth}
\centering
\includegraphics[width=\textwidth]{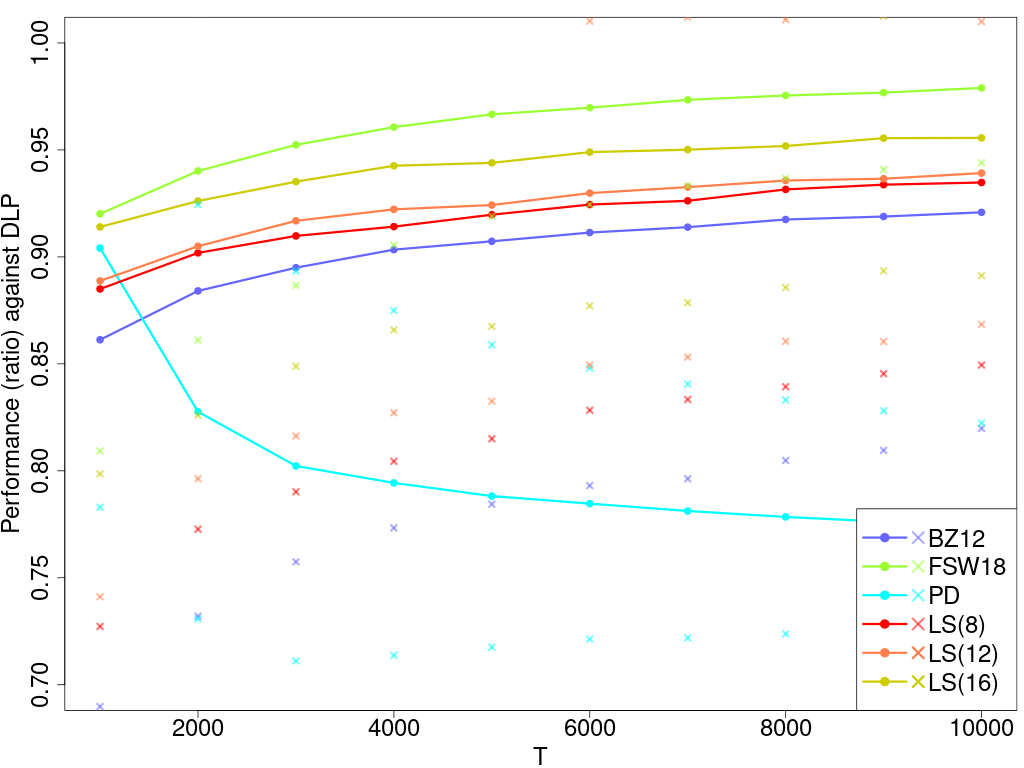}
\caption{\footnotesize Exponential demand, small inventory.}
\end{subfigure}\hfill
\begin{subfigure}{.47\textwidth}
\centering
\includegraphics[width=\textwidth]{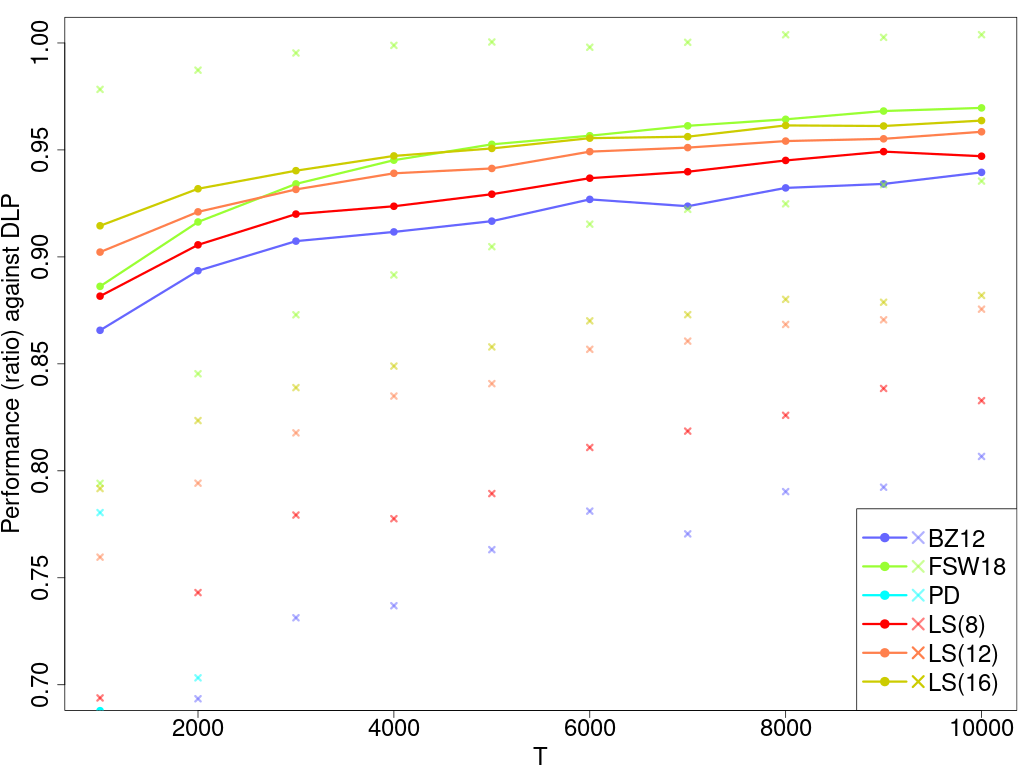}
\caption{\footnotesize Exponential demand, large inventory.}
\end{subfigure}

\begin{subfigure}{.47\textwidth}
\centering
\includegraphics[width=\textwidth]{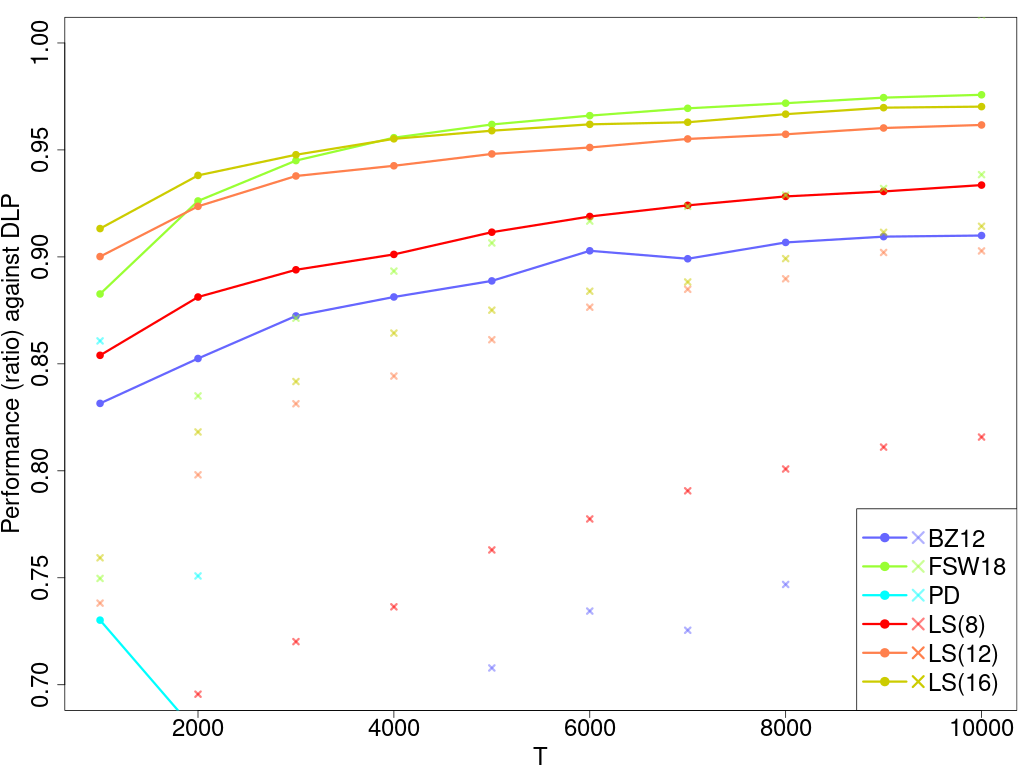}
\caption{\footnotesize Logit demand, small inventory.}
\end{subfigure}\hfill
\begin{subfigure}{.47\textwidth}
\centering
\includegraphics[width=\textwidth]{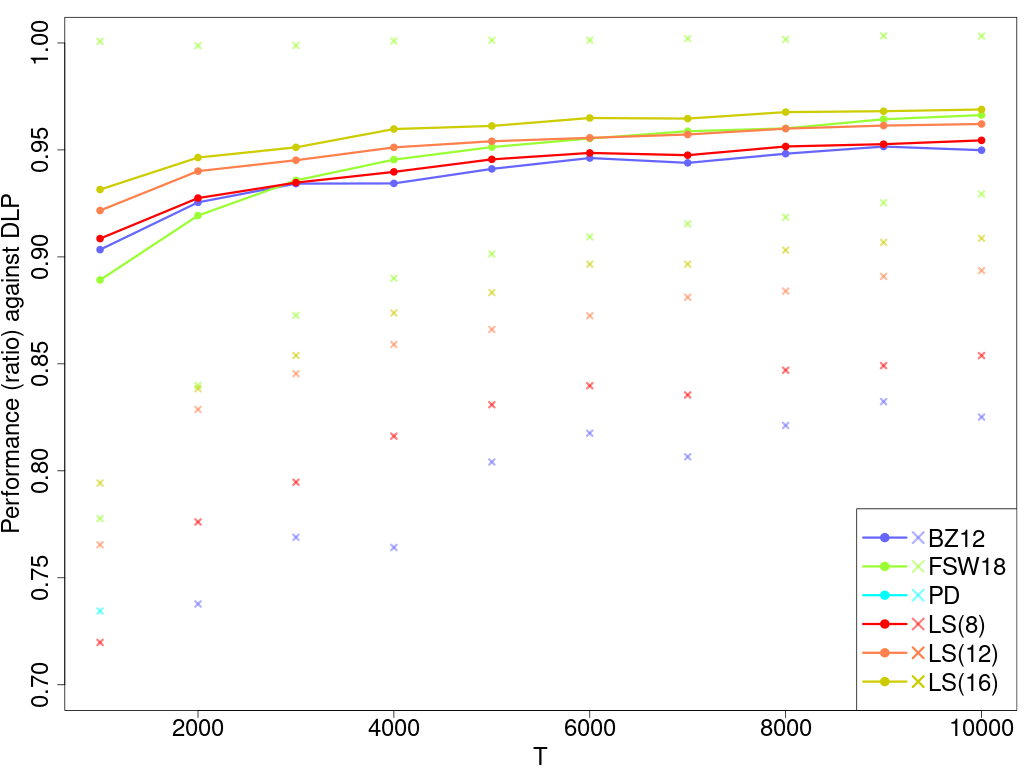}
\caption{\footnotesize Logit demand, large inventory.}
\end{subfigure}
\caption{Numerical results under three different models and two inventory levels when there are $K=5$ price vectors.}
\label{fig:computational}
{\scriptsize {\it Note:} The performance of algorithms are normalized relative to the DLP upper bound, thus between 0 and 1. The cross marks represent the $95\%$ confidence intervals ($\pm 1.96$ standard error).}
\end{figure}

We consider three different classes of algorithms, and evaluate their performance in Figure~\ref{fig:computational}.
In each sub-figure, we normalize all the earned revenue by the corresponding DLP upper bound, so that the ratio is always upper bounded by 1.

The first algorithm we consider is proposed in \citet{besbes2012blind}, which we refer to as \textbf{BZ12}.
The authors split the time horizon into two phases and conduct exploration in the first phase and exploitation in the second phase.
To make the comparison fair, we conduct the ``limited switches'' procedure as we described in Algorithm~\ref{alg:NRMUB}, which is to continue using the last price vector used in the ``exploration'' phase as the first price vector used in the ``exploitation'' phase.
The blue solid line is \textbf{BZ12} without updating inventory.

The second algorithm we consider is proposed in \citet{ferreira2018online}, which we refer to as \textbf{FSW18}.
The authors propose a Thompson Sampling algorithm that involves solving a linear program in each time period.
We choose the uniform prior and use the Beta distribution for the underlying purchase probability.
There is no guarantee of how many times that the policy switches between price vectors.
The green solid line is \textbf{FSW18} without updating inventory.

The third algorithm we consider is proposed in \citet{badanidiyuru2013bandits}.
The authors propose a general primal-dual algorithm that involves updating the dual variables in an exponential weighting fashion in each time period.
There is no guarantee of how many times that the policy switches between price vectors.
Moreover, when there exists a price vector such that the purchase probabilities are all zero or near-zero for all products, the numerical performance of this algorithm is unstable, and often quite poor\footnote{This is because when the prices are so high, the ``Lower Confidence Bound'' estimate of the cost is very close to zero, thus resulting the bang-per-buck ratio to be infinity. So in the simulation, the primal-dual algorithm spends a significant amount of periods in choosing the high-price vectors. If such high-price vectors turn out to be efficient, then the algorithm performs well (linear demand cases as in Figure~\ref{fig:computational}); if not, the algorithm performs poorly (exponential and logit demand cases as in Figure~\ref{fig:computational}). This suggests that the general \bwk algorithms need to cater to the \bnrm structure in order to achieve better performance.}.
The solid cyan line is the primal-dual algorithm which we refer to as \textbf{PD}.
Since the performance ratio of \textbf{PD} is below $70\%$, the solid cyan line is not shown in the exponential and logit demand cases in Figure~\ref{fig:computational}.

Finally we run our \mainalgo algorithm denoted as \textbf{LS(s)}, parameterized by $s$, under different levels of switching budgets. In our experiments we simply set the parameter $\gamma=1$. 
The red, orange, and yellow solid lines are under switching budgets of $s=8, 12, 16$, respectively, without updating inventory. 
It is worth noting that an execution of our algorithm does not necessarily use all the switching budget; see Table~\ref{tbl:NSwitchesOneExample}.
We guarantee that the total number of switches does not exceed such a switching budget.
It is not surprising that when we increase the switching budget $s$, the performance of our proposed algorithm improves.

We also calculated and compared all the algorithms with their second version to update inventory\footnote{The inventory-updating version of the \textsf{LS-2SLP} algorithm is presented in Algorithm \ref{alg:update} of Appendix \ref{sec:AdditionalAlgos}.} (except the primal-dual algorithm \textbf{PD}, which does not have an updating inventory version).
For most cases, the inventory updating version of all the algorithms are slightly better than the no inventory updating version.
For the sake of brevity, we present here only the no inventory updating version, and defer presenting the inventory updating version in Section~\ref{sec:AdditionalSimulations} in the Appendix.

As we see from Figure~\ref{fig:computational}, in all the scenarios, the performance of learning algorithms all increase as the length of horizon increases.
In all cases, our proposed \mainalgo policy has better performance as the switching budget increases.
For most cases, our proposed policy outperforms \textbf{BZ12}.
{ Even when our policy has a switching budget of $s=8$, in which case our policy degenerates into a similar policy as \textbf{BZ12} with only one exploration phase and one exploitation phase and having the same theoretical regret rate, our policy has a better performance than \textbf{BZ12}.
This is because our algorithms take into account the impact of $K$, so that the length of the exploration phase suggested by our \textbf{LS(8)} algorithm is different from that suggested by \textbf{BZ12}.}

The Thompson Sampling algorithms as proposed in \textbf{FSW18} often have the best performance in all six scenarios.
But when our policy is endowed with more switching budgets, the performance of our policy is comparable to, and sometimes even better than, the Thompson Sampling algorithm.

Next, we compare the total number of switches made.
We report in Table~\ref{tbl:NSwitchesOneExample} in the Appendix the number of total switches made, averaged across all simulations.
In all six scenarios, the Thompson Sampling algorithms as proposed in \textbf{FSW18} make the most number of switches.
The algorithms as proposed in \textbf{BZ12} always make $4 \sim 5$ switches, because they only has one phase of exploration --- and then directly go to exploitation phase.
Our algorithms, when endowed with a switching budget of $8$, have one exploration phase as well, thus also making $4 \sim 5$ switches.
When the switching budget is larger, our algorithms take such advantages by making more switches.
It is worth noting that in all six scenarios, our algorithms do not always necessarily use all the switching budget.

To conclude this section, we conduct a new simulation study on a slightly different setup with a larger set of feasible prices.
The set of feasible prices is given by $(p_1,p_2) \in \{(0.5,0.5)$, $(0.5,0.8)$, $(0.5,1)$, $(0.5,1.5)$, $(0.8,0.8)$, $(0.8,1)$, $(0.8,1.5)$, $(1,1.5)$, $(1,2)$, $(2,3)$, $(2,4)$, $(4,4)$, $(4,6.5)$, $(4,8), (5,8)\}$.
We again vary the initial inventory to be either the small inventory scenario or the large inventory scenario.
And we consider three different demand models: a linear model, an exponential model and a logit model.

\begin{figure}[!htb]
\footnotesize 
\begin{subfigure}{.47\textwidth}
\centering
\includegraphics[width=\textwidth]{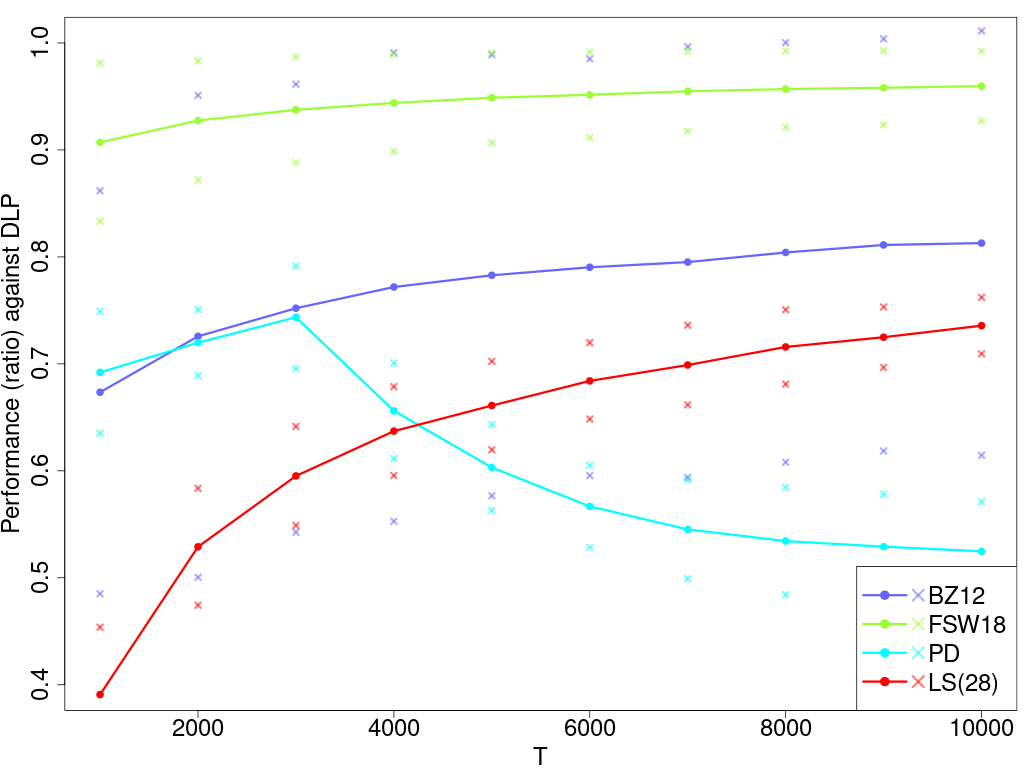}
\caption{\footnotesize Linear demand, small inventory.}
\end{subfigure}\hfill
\begin{subfigure}{.47\textwidth}
\centering
\includegraphics[width=\textwidth]{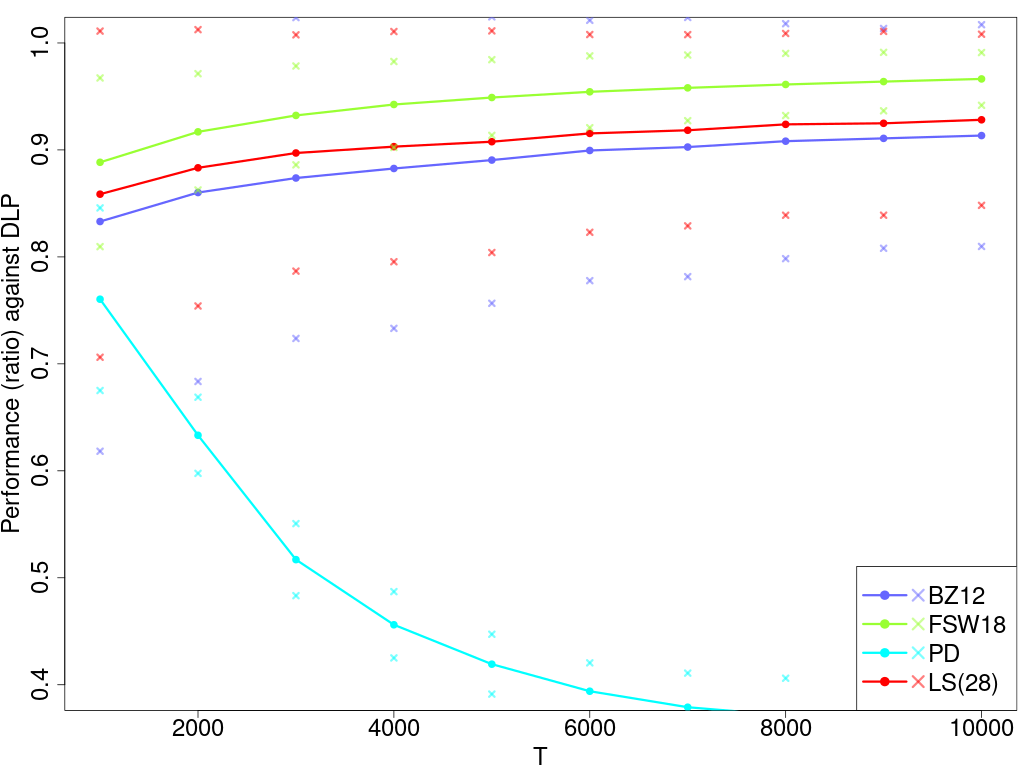}
\caption{\footnotesize Linear demand, large inventory.}
\end{subfigure}

\begin{subfigure}{.47\textwidth}
\centering
\includegraphics[width=\textwidth]{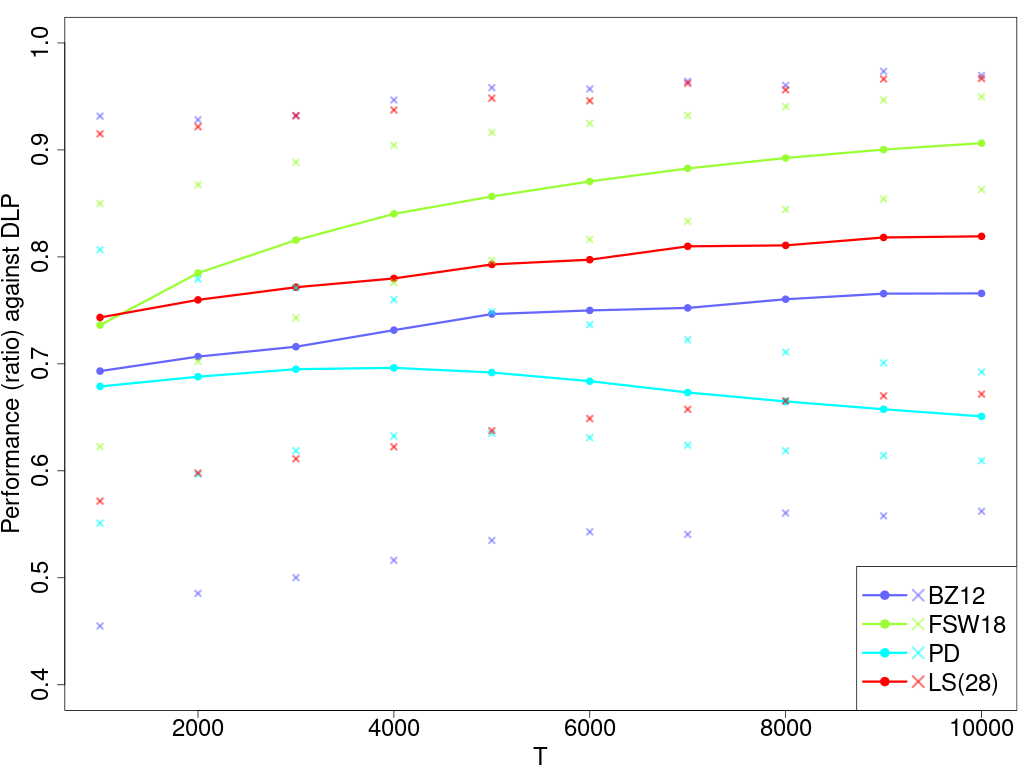}
\caption{\footnotesize Exponential demand, small inventory.}
\end{subfigure}\hfill
\begin{subfigure}{.47\textwidth}
\centering
\includegraphics[width=\textwidth]{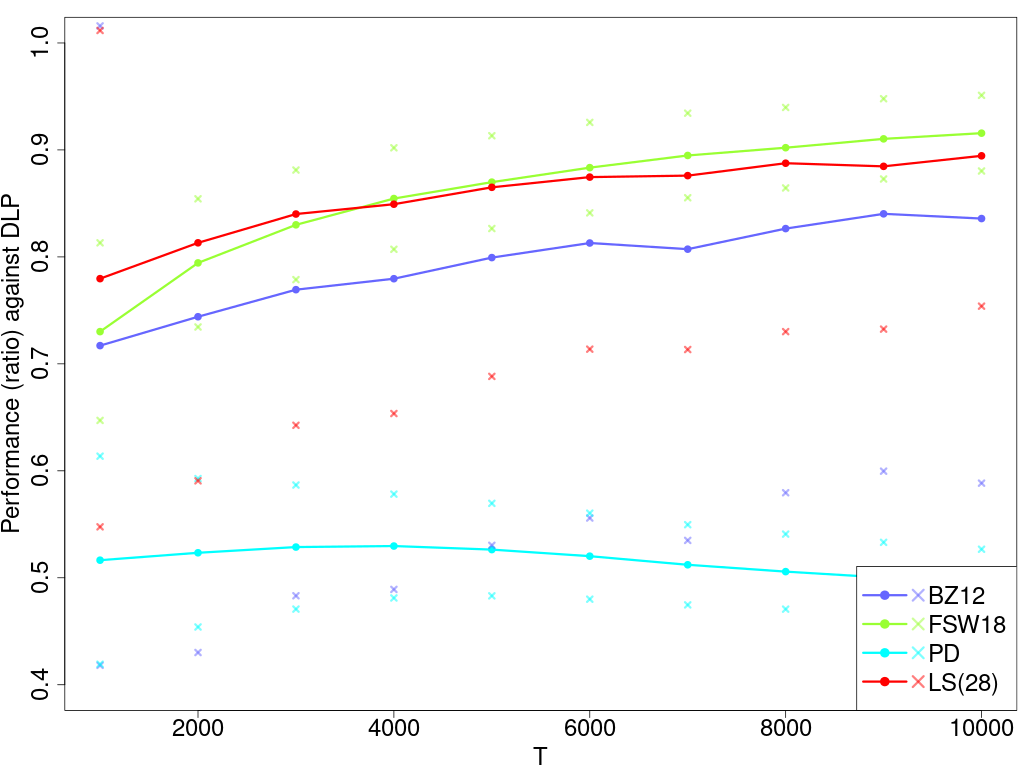}
\caption{\footnotesize Exponential demand, large inventory.}
\end{subfigure}

\begin{subfigure}{.47\textwidth}
\centering
\includegraphics[width=\textwidth]{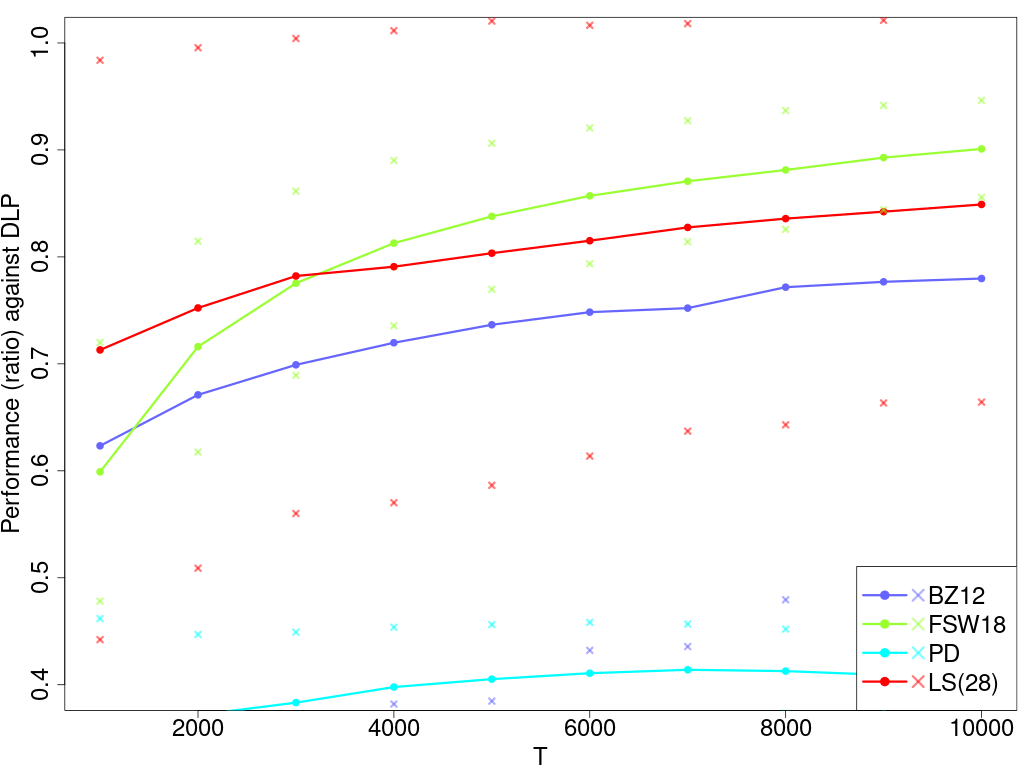}
\caption{\footnotesize Logit demand, small inventory.}
\end{subfigure}\hfill
\begin{subfigure}{.47\textwidth}
\centering
\includegraphics[width=\textwidth]{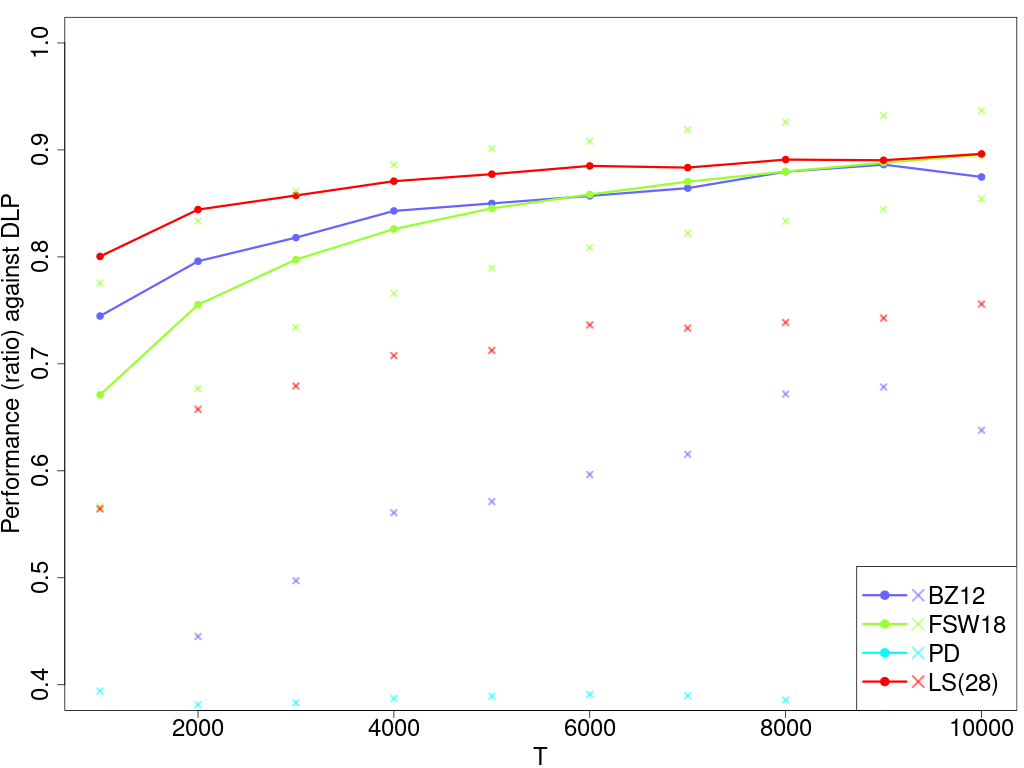}
\caption{\footnotesize Logit demand, large inventory.}
\end{subfigure}
\caption{Numerical results under three different models and two inventory levels when there are $K=15$ price vectors.}
\label{fig:computational:K=15}
{\footnotesize {\it Note:} The performance of algorithms are normalized relative to the DLP upper bound, thus between 0 and 1. The cross marks represent the $95\%$ confidence intervals ($\pm 1.96$ standard error).}
\end{figure}

The simulation results are shown in Figure~\ref{fig:computational:K=15}.
Figure~\ref{fig:computational:K=15} shows our \mainalgo algorithm parameterized by $s=28$.
The value of this parameter is selected to be larger because the number of discrete price vectors $K=15$ is much larger than the number of price vectors in the first simulation setup.
Figure~\ref{fig:computational:K=15} also shows the other algorithms proposed in the literature.
To compare the performances of the algorithms, our proposed \mainalgo policy outperforms \textbf{BZ12} in most cases, especially when the inventory is large.
Our proposed \mainalgo policy is also comparable to \textbf{FSW18} in a few cases.
Only in the linear demand with small inventory our proposed \mainalgo policy does not perform well compared to the benchmark algorithms.

The simulation results suggest that, our proposal of \textsf{2SLP} is effective when the number of price vectors is large.
Yet practitioners need to allocate a large enough switching budget when the number of price vectors is large (which is necessary, as illustrated by our Theorem \ref{thm:BlindNRMLB}).
When the switching budget is of a primary concern to practitioners, we need to leverage some structural relationship between price and demand.
We refer to Section~\ref{sec:LinearDemands} for a different algorithm that is specialized for the linear demand setup.

\section{Discussions on the Continuous Price Setup Under Linear Demand}
\label{sec:LinearDemands}

In the previous sections, we have studied the \lsbnrm setup when the decision maker chooses from a discrete, finite set of $K$ candidate price vectors, and have analyzed the performance of \mainalgo in the discrete price setup.
In this section, we extend our model to the continuous price setup under linear demand.
This setup has been studied in many papers, e.g., \citet{besbes2012blind, keskin2014dynamic, ferreira2018online}.

In each period $t$, a decision maker can post prices for the $n$ products by selecting a price vector from a continuous range $P_c:=[\pmin,\pmax]^n$, where $\pmax>\pmin\ge0$ are constants.
Given price vector $\bm{z}_t$, the realized demands for the $n$ products in period $t$ are represented by a $n$-dimensional vector $\bQ_t(\bm{z}_t)$, which are generated from $\bQ_{t}(\bm{z}_t) = \bm{\alpha} + \beta \bm{z}_t + \bm{\epsilon}_t$; here $\bm{\alpha}$ is a $n$-dimensional vector, $\beta$ is a $n \times n$ matrix, and $\bm{\epsilon}_t$ is a $n$-dimensional vector.
We assume that the coefficients $(\bm{\alpha}, \beta) \in \Theta \subseteq [-C_\theta,C_\theta]^{n \times (n+1)}$ are unknown but fixed quantities over a bounded region $\Theta$, where $C_\theta>0$ is a constant.
We assume that there exists a (possibly unknown) price vector $\bar{\bm{p}} \in P_c$ such that the average demand consumption strictly respects the resource constraints, that is, there exists $\bar{\bm{p}} \in P_c$ and a positive constant $C_\delta > 0$ such that $A(\bm{\alpha} + \beta \bar{\bm{p}}) \leq \bm{B}/T - C_\delta \bm{1}_n$, where $\bm{1}_n$ stands for a length-$n$ vector with all elements equal to one. 
We also assume that all components of $\bm{\epsilon}_t$ are identically and independently distributed (i.i.d.) standard sub-Gaussian random variables, for all $ t\in[T]$. 

In the known distribution case, there is one notable distinction between the discrete price setup and the continuous price setup.
In the discrete price setup, as suggested by Theorem~\ref{thm:NRMLB}, it typically requires some non-negligible number of price vectors (up to $d$ the number of resources) for a policy to be asymptotically optimal.
Yet in the continuous price setup, if the demand function satisfies the regularity conditions, using one single price suffices to be asymptotically optimal; see \citet{gallego1994optimal, jasin2014reoptimization, wang2014close}.

In the unknown distribution case, we design a new algorithm \textsf{LS-ETE}.
The algorithm can be seen as a variant of the explore-then-exploit algorithm of \cite{besbes2012blind}; compared to the algorithm of \cite{besbes2012blind} which only achieves $\widetilde{O}(T^{(n+2)/(n+3)})$ regret, \textsf{LS-ETE} is able to achieve much better performance, since it utilizes the linear structure of the demand model to conduct estimation and optimization; see Algorithm \ref{alg:lsete} for the details. 
The regret guarantee of the algorithm is provided in Theorem~\ref{thm:linearBNRM} below; see Appendix~\ref{app:linearBNRM} for the proof.




\begin{algorithm}[htbp]
\caption{Limited-Switch Learning via Explore-Then-Exploit (\textsf{LS-ETE})}
\label{alg:lsete}
\leftline{{\bf Input:} Problem parameters $(T,\bm{B},d,n,P_C,A)$; switching budget $s$.} 
\leftline{{\bf Initialization:} Define $ t_1=\lfloor{T^{2/3}}\rfloor$ and $C_{\text{lin}}=15\amax(\pmax+1)\max\{C_\theta,1\}/\sqrt{\min\{(\pmax-\pmin)^2,1\}}$.}
\leftline{Set $\gamma=\max\big\{1- C_{\text{lin}}n^{3}\sqrt{\log[n(d+1)T]}{{T}^{2/3}}/{\Bmin},0\big\}$.} 
\leftline{{\bf Notation:} Let $T_1$ denote the ending period of epoch $1$ (to be determined by the algorithm).}
Let $\bm{z}_t\in P_c$ denote the algorithm's selected price vector at period $t$. Let $\bm{e}_j\in\{0,1\}^n$ be the unit vector whose $j$-th component is 1 and all other components are 0. Let $\bm{1}=(1,\dots,1)\in\{0,1\}^n$.\\
{{\bf Policy:}}
\begin{algorithmic}[1]
\STATE{\textbf{Epoch 1}: Let $\bm{z}_{1}=\pmin\bm{1}$. Choose this price vector for $\gamma t_1/(n+1)$ consecutive periods (we overlook the rounding issues here, which are easy to fix in regret analysis).
For $j=1,\dots,n$, choose each price vector
\[\pmin\bm{1}+(\pmax-\pmin)\bm{e}_j\] for $\gamma t_1/(n+1)$ consecutive periods. Stop the algorithm once the time horizon is met, or one of the resources is exhausted; otherwise, end epoch 1 with $T_1=\gamma t_1$.} 
\STATE{\textbf{Epoch 2}: Obtain the least squares estimates of $(\bm{\alpha},{\beta})$ by computing
\begin{align*}
(\hat{\bm{\alpha}},\hat{\beta}):=\arg\min_{(\bm{\alpha},\beta)\in \Theta}\sum_{t=1}^{T_1}\|\bm{Q}_t(\bm{z}_t)-\bm{\alpha}-\beta \bm{z}_t\|_2^2.
\end{align*}
Solve the following deterministic LP \vspace{-5mm}
\begin{align*}
\max_{\bp} \ \bp^\top(\hat{\bm{\alpha}}+\hat{\beta}\bp) &\\
\text{s.t.} \ A(\hat{\bm{\alpha}}+\hat{\beta}\bp) &\leq \bm{B}/T \\
\bp & \in P_c,
\end{align*}
and find an optimal solution $\bp^*_{\hat{\bm{\alpha}},\hat{\beta}}$. Let $\bm{z}_{T_{1}+1}=\bp^*_{\hat{\bm{\alpha}},\hat{\beta}}$. Keep choosing this price vector at every period. Stop the algorithm once the time horizon is met, or one of the resources is exhausted. End epoch 2.}
\end{algorithmic}
\end{algorithm}

\begin{theorem}
\label{thm:linearBNRM}
Let $\phi$ be the \textsf{LS-ETE} policy as suggested by Algorithm \ref{alg:lsete}. Let $\underline{b}>0$ be an arbitrary constant. For any \textsf{BNRM} problem under linear demand, $\mathcal{P}=(T,\bm{B},d,n,P_c,A)$ with $T\ge1, d \ge 0, n\ge1$ and $\Bmin/T\ge\underline{b}$, $\phi$ is guaranteed to be a $s$-switch learning policy, and the regret incurred by $\phi$ satisfies
\begin{align*}
R^\phi_s(T)\le R^\phi(T)\le 
\begin{cases}
{\max\{c/\underline{b},c'\}n^5\sqrt{\log[n(d+1)T]}}\cdot T^{2/3}, & \text{if }s\ge n+1,\\
c n^2\cdot T, & \text{if }s< n+1.
\end{cases}
\end{align*}
where $c,c'$ are some absolute constants completely determined by $\pmin,\pmax,\amax,C_\theta,C_\delta$.
\end{theorem}

Comparing the regret guarantee in the continuous price setup (Theorem \ref{thm:linearBNRM}) with the regret guarantee in the discrete price setup (see Section \ref{sec:BlindNRM}), we make two observations.

First, in the discrete price setup, we need at least $K+d$ number of price changes to achieve $\widetilde{O}(T^{2/3})$ or any sublinear regret (here $d$ plays an important role); see Figure~\ref{fig:BNRMRegret}. 
In the continuous price setup, we can use $n+1$ number of price changes to achieve $\widetilde{O}(T^{2/3})$ regret (here $d$ does not play an important role). 
This comparison implies that, when the number of price vectors $K$ is large, or when the number of resources $d$ is large, it is more beneficial to model the problem as a continuous price problem and leverage the structural relationship to facilitate learning.
On the other hand, when the number of products $n$ is large, learning in the continuous price setup is more challenging, and a discrete price model is more suitable for meaningful results.

Second, an advantage of \mainalgo in the discrete price setup is that, as $s$ increases, regret becomes smaller than $\widetilde{O}(T^{2/3})$. 
In particular, the regret becomes $\widetilde{O}(T^{1/2})$ when $s=\Omega(\log\log T)$. Can similar improvements of regret be obtained in the continuous price setup? The answer is yes. In fact, we find that a straightforward generalization of \mainalgo to the linear demand setting (where the upper and lower confidence bounds of revenue/cost associated with each price are computed by doing maximization and minimization over the confidence ellipsoid of underlying parameters) can achieve $\widetilde{O}(T^{1/2})$ regret in the continuous price setup, using $s=O(\log\log T)$ switches. 
Unfortunately, the generalized \mainalgo approach becomes computationally \emph{inefficient} in the continuous price setup, as calculating the upper and lower confidence bounds for every price requires enumerations over $P_c$. 
This is why we choose to present the simple and more practical \lsete algorithm in the continuous price setup. 
Designing a limited-switch, statistically optimal, and computationally efficient algorithm for the continuous price setup remains an open question.

The above two observations highlight the theoretical differences and practical trade-offs between solving the \lsbnrm problem under discrete and continuous price setups.

\section{Concluding Remarks}

\label{sec:Conclusions}

In this paper, we have studied the blind network revenue management problem under the constraint that the decision maker cannot change the price vector more than a limited number of times.
Our results suggest that the best-achievable regret, as a function of the switching budget, can be characterized by a resource-dependent piecewise-constant function of the switching budget.
We recommend using our algorithms when decision makers simultaneously face uncertainty, resource constraints, and switching constraints.

We conclude this paper by acknowledging two limitations which could lead to future research questions.
First, if we compare Theorem~\ref{thm:BlindNRMUB} and Theorem~\ref{thm:BlindNRMLB}, there is a gap in the dependence on $K$ the number of feasible price vectors.
This could be an issue in situations where the number of feasible price vectors $K$ is large in some practical situations.
It is an interesting future research question to close this gap, as the proof techniques from \citet{simchi2023phase} does not directly close the gap when there are resource constraints.
Simulation results suggest that, our proposal of \textsf{2SLP} is effective when the number of price vectors is large.
Yet practitioners need to allocate a large enough switching budget.
When the switching budget is of a primary concern to practitioners, we propose to model the prices as coming from a continuous range, and use a different algorithm that is specialized for the linear demand model. We leave designing optimal and efficient limited-switch algorithms for more general demand models to future work.

Second, we note that in Theorem~\ref{thm:BlindNRMLB} the choice of the parameters are not fully general.
We allow for general choices of $\bm{B}$ the initial inventory and $n$ the number of products, but we cannot allow for general choices of $P$ the price vectors and $A$ the consumption structure.
It is an interesting future research question to discuss how general choices of $P$ and $A$ would impact the hardness of the \bnrm problem.

\bibliographystyle{informs2014} 
\bibliography{bibliography} 

\clearpage

\ECSwitch

\ECHead{Online Appendix}

\section{Summary of Mathematical Notation}
\label{sec:MathNotation}
Let $\bN$, $\bR$ and $\bR_+$ be the set of positive integers, real numbers and non-negative real numbers, respectively.
For any $N \in \bN$, define $[N] = \{1,2,...,N\}$.
We use bold font letters for vectors, where we do not explicitly indicate how large the dimension is.
For any vector $\bm{x} \in \bR^K$, let $\left\| \bm{x} \right\|_0 = \sum_{k \in [K]} \bI\{x_k \ne 0\}$ be the $L_0$ norm of $\bm{x}$, i.e. the number of non-zero elements in vector $\bm{x}$.
For any vector $\bm{x} \in \bR^K$ and any $k \in [K]$, let $(\bm{x})_k$ be the $k^{\text{th}}$ element of vector $\bm{x}$.
For any positive real number $x \in \bR_+$, let $\lfloor x \rfloor$ be the largest integer that is smaller or equal to $x$; for any non-positive real number $x \in \bR \setminus \bR_+$, let $\lfloor x \rfloor = 0$. 
For any set $X$, let $\Delta(X)$ be the set of all probability distributions over $X$. We  use big $O,\Omega,\Theta$ notation to hide constant factors, and use $\widetilde{O},\widetilde{\Omega},\widetilde{\Theta}$ notation to hide both constant and logarithmic factors.

\section{Additional Discussions}

\subsection{Additional Discussions of the Network Revenue Management Problem}

In this Section, we further discuss the network revenue management (\nrm) problem.

For the \nrm problem, it is well known that the Deterministic Linear Programs (DLP's) provide upper bounds on the revenue generated from any admissible policy, in both the distributionally known and distributionally unknown cases. 
We define the DLP's in Section~\ref{sec:NRMOverview}. 
In Section~\ref{sec:BlindNRM} when we discussed the distributionally unknown case, such DLP's serve as a useful bridge to analyze the algorithm's performance and to establish fundamental limits for \lsbnrm.

\label{sec:additional:NRM}
\begin{figure}[!tb]
\centering
\includegraphics[width=0.75\textwidth,trim={4 15.5 12 56},clip]{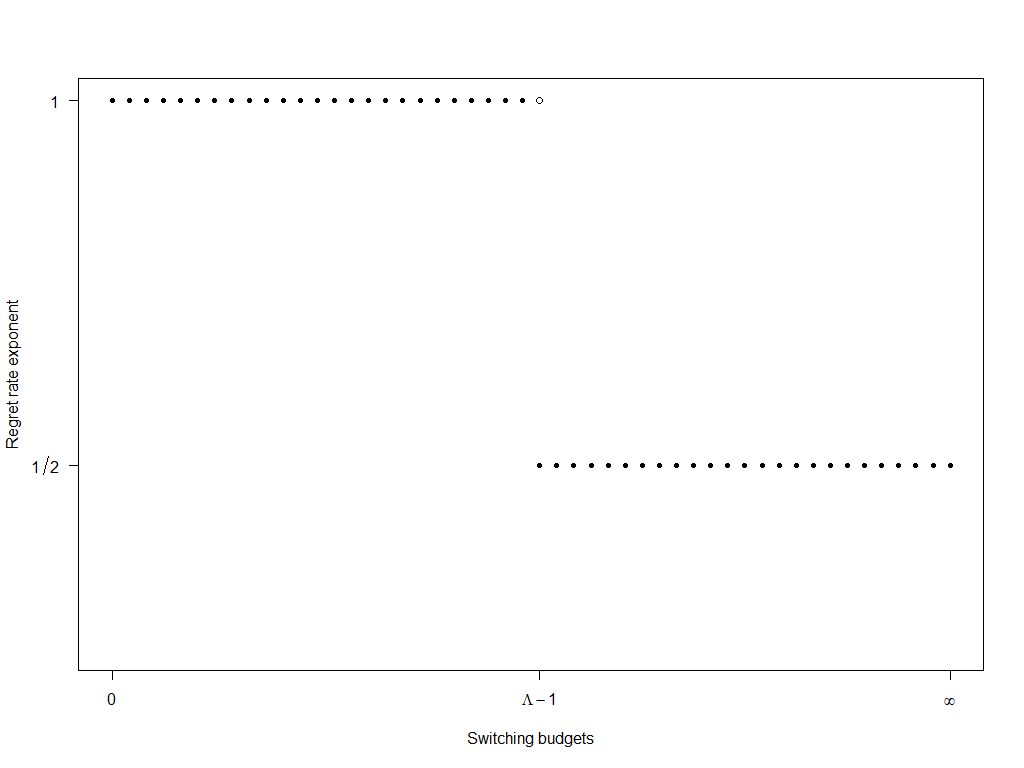}
\caption{The intrinsic gap on the optimal regret $\Rev(\pi^*[\infty])-\Rev(\pi^*[s])$ in the distributionally-known setup} 
\label{fig:NRMRegret}
{\footnotesize {\it Note:} While the $\Theta(T)$ bound is tight for all $s<\DSL-1$, the $\widetilde{O}(\sqrt{T})$ bound shown for $s\ge\DSL-1$ is not necessarily tight for all $s\ge\DSL-1$; characterizing the exact rate of $\Rev(\pi^*[\infty])-\Rev(\pi^*[s])$ for every $s\ge\DSL-1$ is an interesting future direction. There is a deep line of literature that improves the $O(\sqrt{T})$ result obtained from standard techniques, such as \citet{arlotto2020logarithmic, bumpensanti2020re, jasin2012re, jasin2014reoptimization, wang2022constant}. Yet to the best of our knowledge, none imposes a limited switches constraint.}
\end{figure}

We present lower and upper bounds on the regret for the distributionally known case in Sections~\ref{sec:NRMLB} and~\ref{sec:NRMUB}, respectively. 
Combining both results, we show an intrinsic gap between linear regret and sublinear regret, depending on the switching budget. 
See Figure~\ref{fig:NRMRegret}. 
When the DLP is non-degenerate, the number of resource constraints $d$ is the minimum required number of switches to achieve a sublinear regret; when the switching budget is strictly below $d$, a linear regret is inevitable. 
This explains why $d$ appears in the regret rate in the \lsbnrm problem, as it appears in the last epoch during the pure exploitation phase.

For the upper bound result introduced in Section~\ref{sec:NRMUB}, it is worth noting the following subtle difference, that the static control policy \citep{gallego1997multiproduct, cooper2002asymptotic, maglaras2006dynamic, liu2008choice} achieves a similar $O(\sqrt{T})$ regret in a similar setup, when the selling horizon stops immediately if the total cumulative demand of any resource exceeds its initial inventory.
In our setup the stopping criterion is different\footnote{Following each trajectory of randomness, the ungenerous stopping criterion stops earlier than the generous criterion, hence the regret is larger.
As a result, our upper bound under the ungenerous stopping criterion is not a direct implication of \citet{gallego1997multiproduct}.}, which requires slightly different techniques.
The static control policy was also used in \citet{besbes2012blind} to prove a similar result in the \nrm setup under the same stopping criterion.
However, their proof technique critically requires the maximum price to be bounded, which does not generalize to the \stp setup.
Instead, we followed the ideas from \citet{hajiaghayi2007automated, ma2021dynamic, balseiro2019dynamic} to prove a slightly different result.

Sections~\ref{sec:NRMLB} and~\ref{sec:NRMUB} provide techniques that appear in the analysis of the \lsbnrm problem. 
In particular, algorithmically, the use of a discounting factor $\gamma$ and the action allocation rule in the last epoch of Algorithm~\ref{alg:bsse} in the distrbutionally unknown case borrow ideas from Algorithm~\ref{alg:NRMUB} in the distributionally known case. 
And to prove the lower bound, the construction of the hard instances in Theorem~\ref{thm:BlindNRMLB} in the distrbutionally unknown case builds on the instance-dependent lower bound proved in Theorem~\ref{thm:NRMLB} in the distrbutionally known case.

\subsection{Regret Equivalence between Limited Switches and Limited Adaptivity for the Blind Network Revenue Management Problem}
\label{sec:equi} 

We establish regret equivalence results between limited switching budgets and limited adaptivity.
Limited switching budgets is the hard constraint that one cannot change actions more than a fixed number of times.
Limited adaptivity, as originally introduced in \cite{dean2005adaptivity} for stochastic packing and stochastic knapsack problems and in \cite{perchet2016batched} for stochastic bandit problems, is the hard constraint that one cannot collect feedbacks and adapt to the feedbacks more than a fixed number of times.
As a concrete example of limited-adapativity problems, the $M$-batched multi-armed bandit problem is a variant of the classical \mab problem where the decision maker must split her learning process into $M$ batches and is only able to observe data  from a given batch after the entire batch is completed, see \cite{perchet2016batched,gao2019batched}. 

An algorithm with a limited switching budget can keep track of the demand feedbacks in each period, yet constrained on the changes in actions; while an algorithm with limited adaptivity can prescribe a trajectory of different prices with unlimited changes, yet without knowing the status of the system.
Our results imply that when one ability (switching budget or adaptivity) is constrained, we do not need the other ability more than necessary.

Our limited switches policy as described in Algorithm~\ref{alg:bsse} is not only a limited-switch policy, but also a limited-adaptivity policy --- throughout the entire $T$ periods, it only queries the collected data for at most $\nu(s,d)$ times.
Indeed, the policy runs in at most $\nu(s,d)+1$ epochs, and within each epoch the policy's actions can be pre-determined at the beginning of the epoch.
Our policy can thus be treated as an $M$-batched policy, providing  $\widetilde{O}(T^{\frac{1}{2-2^{1-M}}})$ regret for the $M$-batched \bnrm{} / \bwk problem, which generalizes the previous upper bounds and matches the previous lower bounds for the batched bandit problem without resource constraints (\citealt{perchet2016batched,gao2019batched}). We thus characterize the  $\widetilde{\Theta}(T^{\frac{1}{2-2^{1-M}}})$  optimal regret for the $M$-batched \bnrm problem. We make two interesting observations: (1) in contrast to the optimal regret rate for the limited-switch problem, the optimal regret rate for the limited-adaptivity (batched) \bnrm problem does not change with respect to the number of resource constraints; (2) there exists a clean formula $M=\nu(s,d)+1$ directly linking the optimal regret for the limited-switch and limited-adaptivity  \bnrm problems. 
We state the following results:

\begin{statement}
For any \lsbnrm problem with switching budget $s$ and number of resources $d$,  there exists a number $M=\nu(s,d)+1$, such that the 
corresponding  $M$-batched \bnrm problem exhibits the same optimal regret rate (in terms of $T$).
\end{statement}

Note that \cite{simchi2023phase} established a similar ``regret equivalence'' result between limited switches and limited adaptivity for the classical \mab problem without resource constraints.
Our regret equivalence result is a generalization of their result to the resource-constrained problems.

\subsection{Failures of Some Existing Learning Strategies}\label{sec:failure}

\subsubsection*{Failure of Eliminating Individual Actions.}
Let $d=2,n=2$, $A=I_{2\times2}$, $\bm{B}=(T/4,T/4)^\top$. All demand distributions are Bernoulli distributions. Let $K=3$ and $(\bp_1=(1,1)^{\top},\bq_{:,1}=(\frac{1}{4},\frac{1}{4}+\Delta)^\top)$, $(\bp_2=(1,1.1)^{\top},\bq_{:,2}=(\frac{1}{2},0)^\top), (\bp_3=(1.1,1)^{\top},\bq_{:,3}=(0,\frac{1}{2}+\Delta')^\top)$ where $\Delta,\Delta'$ are small perturbations close to 0. If an algorithm tries to eliminate actions based on their ``estimated individual performance when played alone over $T$ rounds'', then both action 2 and action 3 will be quickly eliminated, as their individual performance is much worse than action 1 (due to the early exhaustion of resource 1 and resource 2, respectively). However, when $\Delta<0$ and $\Delta'=0$, playing each of action 2 and action 3 for approximately $T/2$ rounds would be an approximately optimal strategy, yielding $\Omega(T|\Delta|)$ more total expected revenue than playing action 1 alone. This indicates that any algorithm that eliminates action 2 or action 3 too early would fail to achieve the optimal regret rate.

\subsubsection*{Failure of Uniform Exploration.}Let $d=1,n=2,A=[1,0],\bm{B}=T/11$. All demand distributions are Bernoulli distributions.  Let $K=2$ and $(\bp_1=(1,0)^{\top},\bq_{:,1}=(\frac{1}{10},\frac{1}{10})^\top)$, $(\bp_2=(2,1)^{\top},\bq_{:,2}=(0,\frac{1}{20})^\top)$. On this instance, one can show that any effective learning algorithm must play action 1 for approximately $\frac{10}{11}T$ periods and play action 2 for approximately  $\frac{1}{10}T$ periods, over $T$ periods. Consider a \cite{simchi2023phase}-type algorithm which uses the same epoch schedule as our Algorithm \ref{alg:bsse} but---instead of using the \textsf{2SLP} approach to explore the two actions---it simply explores each action for an equal number of periods in epochs 1 to $\nu(s,d)$; at the last epoch $\nu(s,d)+1$, the algorithm behaves the same as our Algorithm \ref{alg:bsse} and plays the two actions with
 the right proportion (determined by the DLP). Consider the case where $\nu(s,d)\ge\log_2\log_2(T/K)$. By  \eqref{eq:large-r}, $t_{\nu(s,d)}\ge T/4$. This means that before the last epoch\footnote{Here we assume no early termination before the last epoch; more formally, one can show that early termination happens with low probability.}, action 2 has already been played for at least $(T/4)/2=T/8$ periods, which significantly exceeds the right proportion of $T/11$ periods over $T$ periods. As a result, the uniform exploration algorithm cannot be effective (in fact, one can show that it incurs linear regret).

\subsubsection*{Failure of Epoch-Based Optimism.} After looking at Algorithm \ref{alg:bsse}, one might wonder whether it would be possible to use a single-stage optimistic LP in the spirit of \cite{agrawal2014bandits} (which solves an “optimistic” variant of the DLP studied in Section \ref{sec:NRM}, with the reward of each action being as overestimated
as possible and the consumption of each action being as underestimated as possible) to replace our two-stage LPs to conduct exploration in epochs 2 to $\nu(s,d)$, while keeping the epoch schedule, the first epoch, and the last epoch of the algorithm unchanged. We call such an algorithm the ``Epoch-Based Optimism'' algorithm. At a first glance, this algorithm has the advantage of playing only $(d+1)$ actions rather than $K$ actions in each epoch 2 to $\nu(s,d)$, thus could significantly save the switching cost. Unfortunately, this algorithm does not work, as optimism requires the epoch schedule to be significantly more frequent than ours. In fact, \emph{all} existing optimism algorithms in bandit literature requires at least $\widetilde{\Omega}(\log T)$ epochs to achieve non-trivial guarantees, while our algorithm can achieve non-trivial (and essentially optimal under the switching or adaptivity constraint) guarantees for any finite or $O(\log\log T)$ epochs. In other words, while optimism could save switches within each epoch, it necessitates the need to have exponentially more epochs, which eventually makes the total number of switches much larger.

In what follows, we provide a simple example where the Epoch-Based Optimism algorithm fails with our epoch schedule, even when $d=0$ (i.e., there is no resource constraint). Let $K=10$ and $\nu(s,d)+1=3$. $d=0$ implies that there is a single optimal action. The Epoch-Based Optimism algorithm will choose $10$ actions uniformly in epoch 1, choose $d+1=1$ action in epoch 2, and choose $d+1=1$ action in epoch 3. This means there are at least 8 actions that are not explored at all in epoch 2 and epoch 3. Unfortunately, the length of epoch 1 ($\widetilde{\Theta}(K^{3/7}T^{4/7})$) is not long enough to ensure that the optimal action can be chosen in epoch 2 or 3, as $O(T^{-2/7})$ perturbations to coordinates of $\bq$ are statistically indistinguishable based on the observations from epoch 1. One can show that $\widetilde{\Omega}(T^{5/7})$ regret is unavoidable for the Epoch-Based Optimism algorithn, which is worse than Algorithm \ref{alg:bsse}'s guarantee of $\widetilde{O}(T^{4/7})$. This example illustrates that fact that optimism is only effective when the epoch schedule becomes frequent enough.

\section{Problem Formulation of the Bandits with Knapsacks Problem}
\label{sec:BwK}

In this section, we review the bandits with knapsacks (\bwk) problem and introduce its variant, the bandits with knapsacks under limited switches (\lsbwk).
We explain the relations between \bnrm vs. \bwk, and the relations between \lsbnrm vs. \lsbwk in Section~\ref{sec:Relations}.

The \bwk problem is a general learning framework introduced in \cite{badanidiyuru2013bandits} and has been later on 
extensively studied in the machine learning literature. It generalizes the classical \mab problem and includes the \bnrm problem as a special case.
See \cite{badanidiyuru2013bandits} for a survey.

\subsubsection*{\bwk Setup.}
Similar to the \bnrm problem, let there be a discrete, finite time horizon with $T$ periods.
Time starts from period $1$ and ends in period $T$.
Unlike \bnrm, there is no ``product'' nor ``consumption matrix'' in \bwk.
Let there be $d$ different resources, each endowed with finite initial capacity $B_i$, $\forall i \in [d]$.

Recall that, in the \bnrm setup and in each period $t$, the decision maker pulls one arm from a finite set of $K$ distinct price vectors, which we have denoted using $\bm{z}_t \in \{\bp_1, ..., \bp_K\}$.
In the \bwk setup and in each period $t$, the decision maker pulls one arm from a finite set of $K$ distinct arms, which, with a little reload of notation, we denote using $z_t \in [K]$.
Each time an arm $k\in[K]$ is pulled, a random reward $R_{k} \in [0,\rmax]$ is received at a random cost $C_{i,k} \in [0,\cmax]$ of each resource $i$, which we denote using the cost vector $\bm{C}_{k} \in [0,\cmax]^d$. 
The distributions of both the random reward and the random cost vector are fixed but unknown to the decision maker, and have to be sequentially learned over time. Denote $r_k := \bE[R_k]$, and $c_{i,k} := \bE[C_{i,k}]$. 
The decision maker stops at the earlier time when one or more resource constraint is violated, or when the time horizon ends.
We use $\mathcal{I} = (T,\bm{B},K,d;\bm{R},\bm{C})$ to stand for one instance of the problem.

\subsubsection*{Regime for Regret Analysis.}

Similar to the \bnrm problem, we derive non-asymptotic bounds on the regret of policies in terms of the number of time periods $T$. 
For all of our results, we adopt the following regret analysis regime: there exists an arbitrary constant $\underline{b}$, such that $\Bmin \geq \underline{b} T$.
Such a regime is similar to the standard regime for regret analysis in the \bwk literature; see, e.g., \citet{badanidiyuru2013bandits,agrawal2014bandits}.

Following the literature, we assume $\cmax,\rmax$ are all constants that do not depend on $T$ or $\bm{B}$. 
The other parameters $K$ and $d$ do not depend on $T$ or $\bm{B}$, either.
Yet we write out our regret bounds' exact dependence on $K$ and $d$ in our main Theorems and all the proofs. 
Obtaining regret upper and lower bounds that are tight on the orders of $K$ and $d$ is an interesting future direction.

\subsubsection*{New Constraint to \bwk.}
We model the business constraint of limited changes of arms as a hard constraint, and define the \lsbwk problems as the \bwk problems with an extra limited switches constraint.
Specifically, on top of the initial resource capacities, the decision maker is initially endowed with a fixed number of switching budget $s$, to change the arm from one to another.
When two consecutive arm pulls are different, i.e., $z_t \ne z_{t+1}$, one unit of switching budget is consumed.
When there is no switching budget remaining, the decision maker cannot change the arm anymore, and has to keep pulling the last arm pulled.

There are other ways to model the business constraint of limited switches, but all are beyond the scope of this paper; see Section~\ref{sec:SwitchingBudget} for more discussions.
We can view the \bwk problems as the \lsbwk problems under an infinite switching budget.
Since a limited switching budget restricts the family of feasible policies, any feasible policy for the \lsbwk problem is also a feasible policy for the \bwk problem.

\subsection{Comparison of The Models}
\label{sec:Relations}

The \bnrm problem and the \bwk problem are closely related.
The distinct price vectors in the \bnrm problem corresponds to the distinct arms in the \bwk problem.
The revenue of one price vector, $\sum_{j=1}^n Q_{j,k} p_{j,k}$, corresponds to the reward of any arm, $R_k$.
And for each resource $i \in [d]$, the consumption of one price vector, $\sum_{j=1}^n Q_{j,k} a_{ij}$, corresponds to the cost of any arm, $C_{i,k}$. 
The \bwk problem is more general than the \bnrm problem, in the sense that for any fixed arm $k\in[K]$, the reward $R_k$ and any cost $C_{i,k}$ can have an arbitrary and unknown relationship.
But in the \bnrm problem, the revenue $\sum_{j=1}^n Q_{j,k} p_{j,k}$ and any consumption $\sum_{j=1}^n Q_{j,k} a_{ij}$ are correlated through the random demand $Q_{j,k}$ (intuitively, the revenue earned is proportional to the consumption of resources). The \bnrm problem can thus be understood as a special case of the \bwk problem with a specific reward-cost structure.

Due to this reason, under the distributionally know case, all the results presented in Section~\ref{sec:NRM} for the \nrm problem either directly apply, or easily generalize to the \stp problem, which is the distributionally known version of the \bwk problem that we discuss in Section~\ref{sec:stp} in the Appendix.

Similarly, the \lsbwk problem is also more general than the \lsbnrm problem.
Consequently, establishing an upper bound on regret for \lsbwk is more challenging than establishing an upper bound on regret for \lsbnrm; 
meanwhile, establishing a lower bound on regret for \lsbnrm is more challenging than establishing a lower bound on regret for \lsbwk, as one has to construct hard problem instances without breaking the specific structure of \lsbnrm (this can be highly non-trivial, as illustrated in Section \ref{sec:lower-tech}). In this paper, since we aim to derive results that are strong in terms of both upper and lower bounds, we deal with the two models rather than only one of them.

\subsection{Related Modeling Components}

We survey other related modeling components that have appeared in the literature, including the stopping criterion, performance metric, and the modeling of limited switches.

\subsubsection{Stopping Criterion.} \label{sec:StoppingCriterion}
At each point in time, as long as the remaining inventory for any resource is zero, the selling horizon stops.
This stopping criterion is standard in the \bnrm and \bwk literature, see \citet{besbes2012blind,badanidiyuru2013bandits}.
We refer to this stopping criterion as the ``ungenerous'' stopping criterion.

There is a second stopping criterion that is common in the revenue management literature when the stochastic distribution is known.
This setup assumes time horizon never stops.
Even if some resources are stocked-out, the decision maker continues to generate revenue from products that does not use the stocked-out resources.
Either the admissible policy eliminates the possibility to sell stocked-out resources \citet{gallego1997multiproduct, rusmevichientong2020dynamic}, or the realized demand in each period is simply the minimum between the remaining inventory and the generated demand \citet{ma2021dynamic}.
We refer to this stopping criterion as the ``generous'' stopping criterion.

Following each trajectory of randomness, the ungenerous stopping criterion stops earlier than the generous stopping criterion, hence the regret is larger.

\subsubsection{Performance Metric.} \label{sec:PerformanceMetric}
We adopt the regret as our performance metric.
Regret is defined as the worst-case expected revenue loss over all possible demand distributions $\bQ$ (or over all possible $(\bm{R},\bC)$ in the \bwk setup). 
We study two regret notions in this setup: the first one, referred to as the \textit{$s$-switch regret}, is the (worst-case) gap of expected revenue between our proposed policy (with a finite switching budget) and the optimal clairvoyant policy with the same switching budget; the second one, referred to as the \textit{overall regret}, is the (worst-case) gap of expected revenue between our proposed policy (with a finite switching budget) and the optimal clairvoyant policy with an infinite switching budget. 
Here, a clairvoyant policy is endowed with perfect knowledge of the distributions, but not the exact realizations. 
While the overall regret is clearly larger than the $s$-switch regret, interestingly, we show that they are in the same order.

There are other performance metrics that are common in the revenue management literature when the stochastic distribution is known, see, e.g., \citet{adelman2007dynamic, jasin2014reoptimization, chen2018beating, vera2019bayesian}.
But all are beyond the scope of this paper.

\subsubsection{Modeling Limited Switches.} \label{sec:SwitchingBudget}
We model the switching budget as a hard constraint that cannot be violated, which is common in the literature.
\cite{cheung2017dynamic} consider a dynamic pricing model where the demand function is unknown but belongs to a known finite set, and a pricing policy makes limited number of price changes.
\cite{chen2019parametric} studies a multi-period stochastic joint inventory replenishment and pricing problem with unknown demand and limited price changes.
\cite{simchi2023phase} consider the stochastic \mab problem with a general switching constraint. \cite{chen2020data} consider the dynamic pricing and inventory
control problem in the face of censored demand.
All the above papers adopt the same modeling approach, yet none of the above papers considers the existence of non-replenishable resource constraints.

There is one other prevalent way of modeling, which models the business constraint as incurring switching costs; see \citet{agrawal1988asymptotically, agrawal1990multi, cesa2013online}, and \citet{jun2004survey} for a survey.
Most papers from this stream of research penalize switching costs into the objective function of the reward calculation.
That is, the objective is to minimize a convex combination of the regret incurred, plus the switching costs. Since we treat the switching budget as a hard constraint that can never be violated, we face a more challenging learning task.

\section{The Stochastic Packing Problem}
\label{sec:stp}

In this section, we introduce and discuss the stochastic packing (\stp) problem.
The \stp problem is a generalization of the \nrm problem, and the distributionally known version of the \bwk problem.
Following the notations that were introduced in Section~\ref{sec:BwK}, we introduce the following definitions.

For any problem instance $\mathcal{I} = (T,\bm{B},K,d;\bm{R},\bm{C})$, we adopt the general notation $\pi: \bR^d \times [s] \times [T] \to \Delta([K])$ to denote any policy with the full information about stochastic distributions, which suggests a (possibly randomized) price vector to use given the remaining inventory, remaining switching budget, and the remaining periods.
For any $s \in \bN$, let $\Pi[s]$ be the set of policies that changes prices for no more than $s$ times on this problem instance $\cI$.
For any $s,s' \in \bN$ such that $s \leq s'$, we know that $\Pi[s] \subseteq \Pi[s']$.
Let $\Pi[\infty]$ be the set of policies with an infinite switching budget (which is the set of all possible policies).
Let $\Rev(\pi)$ be the expected revenue that policy $\pi$ generates on this problem instance $\cI$.
Let $\pi^*[s]\in\argmax_{\pi\in\Pi[s]}\Rev(\pi)$ be one of the optimal dynamic policies with switching budget $s$, and $\pi^*[\infty]$ be one of the optimal dynamic policies with an infinite switching budget (i.e., without a switching constraint).

Next we introduce the deterministic linear programs (DLP) for the stochastic packing problem.
For any $\mathcal{I} = (T,\bm{B},K,d;\bm{R},\bm{C})$, the literature have studied the following DLP, which we refer to as DLP-G.
\begin{align}
\DLPG = \max_{(x_1,\dots,x_K)} \sum_{k\in[K]} \ r_{k} \ \ x_k &&& \label{eqn:obj-G} \\
\text{s.t.} \ \sum_{k\in[K]} \ c_{i,k} x_k &\leq B_i && \forall\ i \in [d] \label{eqn:constraint:inventory-G} \\
\sum_{k\in [K]}x_k &\leq T && \label{eqn:constraint:time-G} \\
x_k & \geq 0 && \forall\ k\in[K] \label{eqn:constraint:NonNeg-G}
\end{align}
Since such a linear program is a packing LP, this generalization of the \nrm problem is also referred to as the stochastic packing (\stp) problem.

Recall that for the \bnrm problem, the set of optimal solutions to the DLP is denoted as $X^* = \arg\max_{\bm{x} \in \bR^K} \{ \eqref{eqn:obj} \left| \eqref{eqn:constraint:inventory}, \eqref{eqn:constraint:time}, \eqref{eqn:constraint:NonNeg} \text{ are satisfied} \right.\}$.
With a little abuse of notation, let the set of optimal solutions to the DLP-G be the same notation $X^* = \arg\max_{\bm{x} \in \bR^K} \{ \eqref{eqn:obj-G} \left| \eqref{eqn:constraint:inventory-G}, \eqref{eqn:constraint:time-G}, \eqref{eqn:constraint:NonNeg-G} \text{ are satisfied} \right.\}$.
The distinction between DLP and DLP-G should be clear from the context.

Similar to the \bnrm setup, let $\DSL = \min \{ \left\| \bx \right\|_0 \left| \bx \in X^* \right.\}$ be the least number of non-zero variables of any optimal solution.
Let $\cDSL = \arg\min \{ \left\| \bx \right\|_0 \left| \bx \in X^* \right.\}$ be the set of such solutions.
For any $\bx^* \in \cDSL$, let $\cZ(\bx^*) = \{ k \in [K] \left| \bx^* \ne 0 \right.\} \subseteq [K]$ be the subset of dimensions that are non-zero in $\bx^*$.
Note that $\DSL$ is an instance-dependent quantity such that $\DSL \leq d+1$, where $d+1$ is the number of all constraints (resource constraints and time constraint) in the linear program.
When DLP-G is non-degenerate, then equality holds and $\DSL = d+1$.

For the \stp problem, we are able to show the same upper and lower bounds on the regret as in Section~\ref{sec:NRM} for the \nrm problem.
In particular, for the lower bound, since the \nrm problem is a special case of the \stp problem, the lower bound construction in Theorem~\ref{thm:NRMLB} directly applies to the \stp problem setup.
So the lower bound result holds.

\begin{proposition}
\label{coro:stpLB}
Let $\bm{b}=\bm{B}/T$ (i.e., $b_1= B_1/T ,\dots,b_d= B_d/T$) be any arbitrary vector of constants.
For any distributions $\bm{R},\bm{C}$ and any \stp instance with $\mathcal{I} = (T,\bm{B},K,d;\bm{R}, \bm{C})$ with $\bm{B} = \bm{b} T$, there is an associated $\DSL$ number (defined in Section \ref{sec:stp} above), such that any policy $\pi \in \Pi[\DSL-2]$ earns an expected revenue:
\begin{align*}
\Rev(\pi) \leq \DLPG - c\cdot T,
\end{align*}
where $c>0$ is some distribution-dependent constant that possibly depends on $\bm{b}$ but does not depend on $T$.
\end{proposition}

As a direct implication of Proposition~\ref{coro:stpLB}, we combine the inequality in Proposition~\ref{coro:stpLB} with the known fact that $\Rev(\pi^*[\infty]) \ge \DLPG - O(\sqrt{T})$ and have
\begin{align*}
\Rev(\pi) \leq \Rev(\pi^*[\infty]) - \Omega(T).
\end{align*}
That is, the regret scales linearly with $(T,\bm{B})$ when other parameters are fixed.
Note that, only Proposition~\ref{coro:stpLB} requires this linear scaling regime; all the other theorems in this paper are described under a more general asymptotic regime (see discussions in Section~\ref{sec:Considered}).

For the upper bound, we can easily extend the results in Theorem~\ref{thm:NRMUB} to the more general \stp problem setup.
We first describe the Tweaked LP Policy as in Algorithm~\ref{alg:stpUB}.


\begin{algorithm}[!tb]
\caption{Tweaked LP Policy for the \stp Problem.}
\label{alg:stpUB}
\leftline{{\bf Input:} $\mathcal{I} = (T,\bm{B},K,d;\bm{R},\bm{C})$}
\leftline{{\bf Policy:}}
\begin{algorithmic}[1]
\STATE{Define $\gamma = \max\big\{1 - \frac{2 \cmax}{\Bmin} \sqrt{T \log{T}},0\big\}$.}
\STATE{Solve the DLP-G as defined by \eqref{eqn:obj-G}, \eqref{eqn:constraint:inventory-G}, \eqref{eqn:constraint:time-G}, and \eqref{eqn:constraint:NonNeg-G}. Find an optimal solution with the least number of non-zero variables, $\bx^* \in \cDSL$.}
\STATE{Arbitrarily choose any permutation $\sigma: [\DSL] \to \cZ(\bx^*)$ from all $(\DSL)!$ possibilities.}
\STATE{Execute: In the \stp setup, pull arm $\sigma(1)$ for the first $\gamma \cdot x^*_{\sigma(1)}$ periods, then $\sigma(2)$ for the next $\gamma \cdot x^*_{\sigma(2)}$ periods, ..., and finally $\sigma(\DSL)$ for the last $T - \gamma \cdot \sum_{l=1}^{\DSL-1}x^*_{\sigma(l)}$ periods.}
\end{algorithmic}
\end{algorithm}

\begin{proposition}
\label{prop:stpUB}
Let $\underline{b}>0$ be an arbitrary constant.
For any \stp instance $\mathcal{I} = (T,\bm{B},K,d;\bm{R},\bm{C})$ with $T\ge1, d \ge 0, K>d+1$ and $\Bmin/T\ge\underline{b}$, any policy $\pi$ as defined in Algorithm~\ref{alg:stpUB} satisfies $\pi\in\Pi[\DSL-1]$ and earns an expected reward:
\begin{align*}
\Rev(\pi) & \geq \DLP-\max\{c/\underline{b}, c'd\} \sqrt{T\log T} \\
& \geq \Rev(\pi^*[\infty])-\max\{c/\underline{b}, c'd\} \sqrt{T\log T}
\end{align*}
where $c,c'>0$ are some absolute constants completely determined by $\rmax,\cmax$.
\end{proposition}

The above upper bound holds for all instances. 
We outline two key steps here and defer the details of our proof to Appendix~\ref{app:NRMUB}.
In the first step, we show that with high probability, the selling horizon never stops earlier than the last period $T$.
Second, conditioning on this high probability event, the expected revenue is at least $\gamma$ fraction of the LP objective.
Combining the two steps together we know that the total loss in rewards is upper bounded by $(1 - \gamma)\DLPG$.

\section{Bandits with Knapsacks under Limited Switches} 
\label{sec:BwKLS}

In this section, we study the \lsbwk problem, introduce an efficient algorithm, and provide matching upper and lower bounds of the optimal regret.
Since we have introduced the definitions and results for the \lsbnrm problem in Section~\ref{sec:BlindNRM}, we mainly introduce the \lsbwk model counterparts only when the notation is different.

\subsubsection*{Learning Policies and Clairvoyant Policies.}
Recall that in Section~\ref{sec:BlindNRM}, we distinguish between a \bnrm \textit{instance} $\mathcal{I}=(T,\bm{B}, K,d,n,P,A;\bQ)$ and a \bnrm \textit{problem} $\mathcal{P}=(T,\bm{B}, K,d,n,P,A)$ based on whether the underlying demand distributions $\bQ$ are specified or not. 
In this section, we distinguish between a \bwk \textit{instance} $\mathcal{I}=(T,\bm{B}, K,d;\bm{R},\bm{C})$ and a \bwk \textit{problem} $\mathcal{P}=(T,\bm{B}, K,d)$ based on whether the underlying demand distributions $\bm{R}, \bm{C}$ are specified or not. 
Consider a \bwk problem $\mathcal{P}=(T,\bm{B},K,d)$.
Let $\phi$ denote {any} non-anticipating learning policy; specifically, $\phi$ consists of a sequence of (possibly randomized) decision rules $(\phi^t)_{t\in[T]}$, where each $\phi^t$ establishes a probability kernel acting from the space of historical actions and observations in periods $1,\dots,t-1$ to the space of actions at period $t$. 
For any $s \in \bN$, let $\Phi[s]$ be the set of learning policies that change price vectors for no more than $s$ times almost surely under all possible demand distributions $\bm{R},\bm{C}$.
For any $s,s' \in \bN$ such that $s \leq s'$, $\Phi[s] \subseteq \Phi[s']$.
Let $\Phi[\infty]$ be the set of all admissible learning policies. 
Let $\Rev_{\bm{R},\bm{C}}(\phi)$ be the expected reward that a learning policy $\phi$ generates under demand distributions $\bm{R},\bm{C}$.

As we have defined in Section~\ref{sec:Considered}, $\pi$ refers to a \textit{clairvoyant} policy with full distributional information about the true distributions $\bm{R},\bm{C}$.
For any $s \in \bN$, let ${\Pi_{\bm{R},\bm{C}}}[s]$ be the set of clairvoyant policies that change price vectors for no more than $s$ times under the true distributions $\bm{R},\bm{C}$. 
For any $s,s' \in \bN$ such that $s \leq s'$, $\Pi_{\bm{R},\bm{C}}[s] \subseteq \Pi_{\bm{R},\bm{C}}[s']$.
Let $\Pi_{\bm{R},\bm{C}}[\infty]$ be the set of all admissible clairvoyant policies.
Let $\Rev_{\bm{R},\bm{C}}(\pi)$ be the expected revenue that a clairvoyant policy $\pi\in\Pi_{\bm{R},\bm{C}}$ generates under distributions $\bm{R},\bm{C}$.
Let $\pi_{\bm{R},\bm{C}}^*[s] \in \arg\sup_{\pi\in\Pi_{\bm{R},\bm{C}}[s]}\Rev(\pi)$ be one optimal clairvoyant policy with switching budget $s$, and $\pi_{\bm{R},\bm{C}}^*$ be one of the optimal dynamic policies with an infinite switching budget (i.e., without a switching constraint).

\subsubsection*{Performance Metrics.}
The performance of an $s$-switch learning policy $\phi\in\Phi[s]$ is measured against the performance of the optimal $s$-switch clairvoyant policy $\pi_{\bm{R},\bm{C}}^*[s]$.
Specifically, for any \bnrm problem $\mathcal{P}$ and switching budget $s$, we define the \textit{$s$-switch regret} of a learning policy $\phi\in\Phi[s]$ as the worst-case difference between the expected revenue of the optimal $s$-switch clairvoyant policy  $\pi_{\bm{R},\bm{C}}^*[s]$ and the expected revenue of policy $\phi$: 
\begin{align*}
R_s^\phi(T)& =\sup_{\bm{R},\bm{C}}\left\{\Rev_{\bm{R},\bm{C}}(\pi_{\bm{R},\bm{C}}^*[s])-\Rev_{\bm{R},\bm{C}}(\phi)\right\}.
\end{align*}
We also measure the performance of policy $\phi$ against the performance of the optimal unlimited-switch clairvoyant policy $\pi_{\bm{R},\bm{C}}^*$. 
Specifically, we define the \textit{overall regret} of a learning policy $\phi\in\Phi[s]$ as the worst-case difference between the expected revenue of the optimal unlimited-switch clairvoyant policy $\pi_{\bm{R},\bm{C}}^*$ and the expected revenue of the policy $\phi$:
\begin{align*}
{R}^\phi(T) & =\sup_{\bm{R},\bm{C}}\left\{\Rev_{\bm{R},\bm{C}}(\pi_{\bm{R},\bm{C}}^*)-\Rev_{\bm{R},\bm{C}}(\phi)\right\}.
\end{align*}
Intuitively, the $s$-switch regret $R_s^\phi(T)$ measures the ``informational revenue loss'' due to not knowing the demand distributions, while the overall regret $R^\phi(T)$ measures the ``overall revenue loss'' due to not knowing the demand distributions and not being able to switch freely. Clearly, the overall regret $R^\phi(T)$ is always larger than the $s$-switch regret $R_s^\phi(T)$.
Interestingly, as we will show later, for all $s$, ${R}^\phi(T)$ and ${R}_s^\phi(T)$ are always in the same order in terms of the dependence on $T$.

\begin{algorithm}[htbp]
\caption{ Limited-Switch Learning via Two-Stage Linear Programming (\textsf{LS-2SLP}) for \lsbwk}
\label{alg:bsse2}
\leftline{{\bf Input:} Problem parameters $(T,\bm{B},K,d)$; switching budget $s$.} 
\leftline{{\bf Initialization:} Calculate $\nu(s,d)=\left\lfloor\frac{s-d-1}{K-1}\right\rfloor$. Define $t_0=0$ and}
\[
t_l=\left\lfloor K^{1-\frac{2-2^{-(l-1)}}{2-2^{-\nu(s,d)}}}T^{\frac{2-2^{-(l-1)}}{2-2^{-\nu(s,d)}}}\right\rfloor,~~\forall l=1,\dots,\nu(s,d)+1.
\]
{Set $\gamma=\max\bigg\{1- 17\frac{\cmax\sqrt{(d+1)\log[(d+1)KT]}\log T}{\Bmin}{t_1},0\bigg\}$.} \\
\leftline{{\bf Notation:} {Let $T_l$ denote the ending period of epoch $l$ (which will be determined by the algorithm).}}
{Let $z_t$ denote the algorithm's action at period $t$. Let $z_0 \in [K]$ be a random action.}\\
{{\bf Policy:}}
\begin{algorithmic}[1]
\FOR{epoch $l=1,\dots,\nu(s,d)$}
\IF{$l=1$}
\STATE{Set $T_0=L^{\mathsf{rew}}_k(0)=L^{\mathsf{cost}}_{i,k}(0)=0$ and $U^{\mathsf{rew}}_k(0)=U^{\mathsf{cost}}_{i,k}(0)=\infty$ for all $i\in[d],k\in[K]$.}
\ELSE
\STATE{Let $n_{k}(T_{l-1})$ be the total number of periods that action $k$ is chosen, up to period $T_{l-1}$. Calculate $\bar{c}_{i,k}(T_{l-1})$ to be the empirical average consumption of resource $i$ by selecting arm $k$, up to period $T_{l-1}$; Calculate $\bar{r}_{k}(T_{l-1})$ to be the empirical average reward by selecting arm $k$, up to period $T_{l-1}$. Calculate confidence radius $\radius_k(T_{l-1})=\sqrt{\frac{\log\left[(d+1)KT\right]}{n_k(T_{l-1})}}$ and 
\[
\begin{cases}U^{\mathsf{rew}}_{k}(T_{l-1})=\min\left\{\bar{r}_{j,k}(T_{l-1})+ \rmax \radius_k(T_{l-1}),U^{\mathsf{rew}}_k(T_{l-2})\right\}
,\\L^{\mathsf{rew}}_{k}(T_{l-1})=\max\left\{\bar{r}_{j,k}(T_{l-1})- \rmax \radius_k(T_{l-1}),L_k^{\mathsf{rew}}(T_{l-2})\right\},\end{cases}~~\forall k\in[K],
\]
\[
\begin{cases}U^{\mathsf{cost}}_{i,k}(T_{l-1})=\min\left\{\bar{c}_{i,k}{(T_{l-1})}+ \cmax \radius_k(T_{l-1}),U^{\mathsf{cost}}_{i,k}(T_{l-2})\right\}
,\\L^{\mathsf{cost}}_{i,k}(T_{l-1})=\max\left\{\bar{c}_{i,k}{(T_{l-1})}- \cmax \radius_k(T_{l-1}),L_{i,k}^{\mathsf{cost}}(T_{l-2})\right\},\end{cases}~~\forall i\in[d],\forall k\in[K].
\]}
\ENDIF
\STATE{Solve the first-stage pessimistic LP:
\begin{align*}
\mathsf{J}^{\mathsf{PES}}_{l} = \max_{(x_1,\dots,x_K)} \sum_{k\in[K]}L^{\mathsf{rew}}_k(T_{l-1})  x_k &&  \\
\text{s.t.} \ \sum_{k\in[K]}U_{i,k}^{\mathsf{cost}}(T_{l-1})  x_k &\leq B_i & \forall i \in [d] \\
\sum_{k\in [K]}x_k &\leq T &  \\
x_k & \geq 0 & \forall k\in[K] 
\end{align*}}
\algstore{myalg}
\end{algorithmic}
\end{algorithm}

\begin{algorithm}                     
\begin{algorithmic}[1]
\algrestore{myalg}
\STATE{For each $j\in[K]$, solve the second-stage exploration LP:
\begin{align*}
\bx^{l,j} = \arg\max_{(x_1,\dots,x_K)} \ x_j &&  \\
\text{s.t.} \ \sum_{k\in[K]}U^{\mathsf{rew}}_{k}(T_{l-1})x_k&\ge \mathsf{J}_{l}^{\mathsf{PES}} &\\
\sum_{k\in[K]}L^{\mathsf{cost}}_{i,k}(T_{l-1}) x_k &\leq B_i & \forall i \in [d]\\
\sum_{k\in [K]}x_k &\leq T &  \\
x_k & \geq 0 & \forall\ k\in[K] 
\end{align*}
}
\STATE{For all $k\in[K]$, let $N_k^l=\frac{(t_{l}-t_{l-1})}{T}\sum_{j=1}^{K}\frac{1}{K}(\bx^{l,j})_k$. Let $z_{T_{l-1}+1}=z_{T_{l-1}}$. Starting from this arm, Select each arm $k$ for $\gamma N_k^l$ consecutive periods, $k\in[K]$ (we overlook the rounding issues here, which are easy to fix in regret analysis). Stop the algorithm once time horizon is met or one of the resources is exhausted.} 
\STATE{End epoch $l$.  Mark the last period in epoch $l$ as $T_l$.}
\ENDFOR
\STATE{For epoch $\nu(s,d)+1$ (the last epoch), 
calculate $\bar{c}_{i,k}(T_{\nu(s,d)})$ to be the empirical average consumption of resource $i$ by selecting arm $k$, up to period $T_{\nu(s,d)}$; Calculate $\bar{r}_{k}(T_{\nu(s,d)})$ to be the empirical average reward by selecting arm $k$, up to period $T_{\nu(s,d)}$. Solve the following deterministic LP
\begin{align*}
\max_{(x_1,\dots,x_K)} \sum_{k\in[K]} \bar{r}_{j,k}(T_{\nu(s,d)}) \ x_k &&& \\
\text{s.t.} \ \sum_{k\in[K]} \bar{r}_{j,k}(T_{\nu(s,d)}) \ x_k &\leq B_i && \forall\ i \in [d] \\
\sum_{k\in [K]}x_k &\leq T && \\
x_k & \geq 0 && \forall\ k\in[K],
\end{align*}
and find an optimal solution with the least number of non-zero variables, $\bx^*_{\tilde{\bq}}$. Let $N_k^{\nu(s,d)+1}=\frac{(T-t_{\nu(s,d)})}{T}{(\bx^*_{\tilde{\bq}})_k}$ for all $k\in[K]$. First let $z_{T_{\nu(s,d)}+1}=z_{T_{\nu(s,d)}}$. Start from this arm, choose each arm $k$ for $\gamma N_k^{\nu(s,d)+1}$ consecutive periods, $k\in[K]$ (we overlook the rounding issues here, which are easy to fix in regret analysis). Stop the algorithm once time horizon is met or one of the resources is exhausted. End epoch $\nu(s,d)+1$.}
\end{algorithmic}
\end{algorithm}

\subsection{Upper Bounds}

In this section, we describe the Limited-Switch Learning via Two-Stage Linear Programming (\textsf{LS-2SLP}) algorithm as Algorithm~\ref{alg:bsse2} in the \lsbwk setup.
Note that, Algorithm~\ref{alg:bsse} should not be directly viewed as a special case of Algorithm~\ref{alg:bsse2}, as it utilizes \bnrm's feedback structure and can have much better empirical performance in the \lsbnrm setup.




We analyze the performance of Algorithm~\ref{alg:bsse2} as follows.

\begin{proposition}
\label{prop:BwKUB}
Let $\phi$ be the \mainalgo policy as suggested by Algorithm~\ref{alg:bsse2}.
Let $\underline{b}>0$ be an arbitrary constant. For any \textsf{BwK} problem $\mathcal{P}=(T,\bm{B},K,d)$ with $T\ge1, d \ge 0, K>d+1$ and $\Bmin/T\ge\underline{b}$, $\phi$ is guaranteed to be a $s$-switch learning policy, and the regret incurred by $\phi$ satisfies
\begin{align*}
R^\phi_s(T)\le R^\phi(T)\le \left(\max\{c/\underline{b},c'\}\cdot \sqrt{(d+1)\log[(d+1)KT]} K^{1-\frac{1}{2-2^{-\nu(s,d)}}}\log T\right)\cdot T^{\frac{1}{2-2^{-\nu(s,d)}}},
\end{align*}
where $\nu(s,d)=\left\lfloor\frac{s-d-1}{K-1}\right\rfloor$, and $c,c'>0$ are some absolute constants completely determined by $\rmax,\cmax$.
\end{proposition}

It is worth noting that the above upper bound holds in a \textit{non-asymptotic} sense: it holds for all finite $T$ and $\bm{B}$, as long as $\Bmin/T$ is lower bounded by a positive constant $\underline{b}$.

\subsection{Lower Bounds}
As we have discussed in Section~\ref{sec:Relations}, the lower bound construction in Theorem~\ref{thm:BlindNRMLB} for the \lsbnrm problem directly applies to the \lsbwk problem.
So the lower bound result holds.

\begin{corollary}
\label{coro:BwKLB}
Let $\underline{b}>0$ be an arbitrary constant. 
For any $T\ge 1, d\ge 0, K\ge 2(d+1)$ and $\bm{B}$ such that $B_i/T\in[\underline{b},1]$ for all $i\in[d]$,  for the \bwk problem $\mathcal{P}=(T,\bm{B},K,d)$, for any switching budget $s\ge0$ and any  $\phi\in\Phi[s]$, 
\begin{align*}
R^\phi(T)\ge R^\phi_s(T)\ge\left(\min\{c\underline{b},c'\}\cdot{(d+1)^{-3}K^{-\frac{3}{2}-\frac{1}{2-2^{-\nu(s,d)}}}}{(\log T)^{-\frac{5}{2}}}\right)\cdot T^{\frac{1}{2-2^{-\nu(s,d)}}},
\end{align*}
where $\nu(s,d)=\left\lfloor\frac{s-d-1}{K-1}\right\rfloor$, and $c,c'>0$ are some numerical constants that do not depend on any problem parameters.
\end{corollary}

\subsection{Regret Equivalence between Limited Switches and Limited Adaptivity.}
Similar to the discussions in Section~\ref{sec:equi}, we establish the regret equivalence between limited switches and limited adaptivity for the \bwk problem.

\begin{statement}
For any \lsbwk problem with switching budget $s$ and number of resources $d$,  there exists a number $M=\nu(s,d)+1$, such that the corresponding $M$-batched \bwk problem exhibits the same optimal regret rate (in terms of $T$).
\end{statement}

\section{Proof of Theorem~\ref{thm:NRMLB} and Proposition~\ref{coro:stpLB}.}\label{app:NRMLB}

\proof{Proof of Theorem~\ref{thm:NRMLB}.}
For any problem instance $\mathcal{I} = (T,\bm{B},K,d,n,P,A;\bm{Q})$.
Any policy $\pi \in \Pi[\DSL-2]$ only selects no more than $\DSL-1$ many price vectors.
For any $k \in [K]$, 
let $\tau_{k}$ be the total number of periods that price $\bp_{k}$ is offered during the selling horizon, under policy $\pi$.
Notice that $\tau_{k}$ is  a random variable, i.e., it is determined by the random trajectory of demand realization and action selection. 

Now denote $Y_{j,k}$ as the random amount of product $j$ sold, during the $\tau_k$ periods that price vector $k$ is offered.
Here $\tau_{k}$ is a random amount, so we cannot directly use Hoeffding inequality to connect $Y_{j,k}$ with $\tau_{k} q_{j, k}$.
But we can adapt the clean event analysis trick from Chapter 1.3 of \cite{slivkins2019introduction}.
Suppose there was a tape of length $T$ for each product $j\in[n]$ and each price vector $k\in[K]$, with each cell independently sampled from the distribution of $Q_{j,k}$.
This tape serves as a coupling of the random demand: in each period $t$ if price vector $k$ is offered, we simply generate a demand of each product $j\in[n]$ from the $t^{\text{th}}$ cell of the tape associated with product $j$ and price vector $k$. Let $Y_{j,k}(t)$ denote the random amount of product $j$ sold, during the first $t$ periods that the price vector $k$ is offered. 
Now we can use Hoeffding inequality on each reward tape:
$$\forall k, \forall j, \forall t, \Pr\left( \left| Y_{j,k}(t) - t q_{j,k} \right| \leq \sqrt{3 t \log{T}} \right) \geq 1 - \frac{2}{T^6}.$$
Denote the following ``clean event'' $E$:
\begin{align*}
\forall k, \forall j, \forall t, \left| Y_{j,k}(t) - t q_{j,k} \right| \leq \sqrt{3 t \log{T}}.
\end{align*}
Using a union bound we have:
$$\Pr\left( \forall k, \forall j, \forall t, \left| Y_{j,k}(t) - t q_{j,k} \right| \leq \sqrt{3 t \log{T}} \right) \geq 1 - \frac{2}{T^3}$$
because $K,n$ are both less than $T$, and each arm cannot be pulled longer than $T$ periods.
The happening of such event implies that
\[
\forall j, \forall k, \left| Y_{j,k} - \tau_k q_{j,k} \right| \leq \sqrt{3 \tau_k \log{T}},
\]
i.e., 
the realized demands are close to the expected demands, suggesting that we can use LP to approximately bound the revenue generated by any policy $\pi \in \Pi[\DSL-2]$.

Specifically, we make the following arguments. 
On one hand, if we focus on the usage of any price vector indexed by $k \in [K]$, the total revenue is $\sum_{j \in [n]} Y_{j,k} p_{j,k}$. Thus, conditional on $E$, the total  revenue generated by policy $\pi$ during the entire horizon can be upper bounded by
\begin{align*}
\sum_{k\in [K]} \sum_{j \in [n]} Y_{j,k} p_{j,k} &\leq \sum_{k \in [K]} \sum_{j \in [n]} (q_{j,k} \tau_{k} + \sqrt{3T\log{T}}) p_{j,k}\\& \leq ( \sum_{k \in [K]} \sum_{j \in [n]} q_{j,k} \tau_{k} p_{j,k} ) + nd \pmax \sqrt{3T\log{T}},
\end{align*}
where the last inequality follows from $\pi\in\Pi[\DSL-2]$ and $\DSL \leq d+1$.

On the other hand, the consumption of resource $i$ must not violate the resource constraints.
$$\sum_{k \in [K]} \sum_{j \in [n]} Y_{j,k} a_{ij} \leq B_i.$$
Lower bounding $Y_{j,k}$ by $q_{j,k} \tau_{k} - \sqrt{3T\log{T}}$ we have
$$\sum_{k \in [K]} \sum_{j \in [n]}  a_{ij}q_{j,k}\tau_k \leq B_i + \sum_{k \in [K]} \sum_{j \in [n]} \sqrt{3T\log{T}} \leq B_i + nd \sqrt{3T\log{T}}.$$
So conditional on $E$, any policy $\pi\in\Pi[\DSL-2]$ always satisfies the following constraints:
\begin{align*} 
\sum_{k\in[K]}\sum_{j\in[n]} a_{ij} \ q_{j,k} \ \tau_k &\leq B_i + n d \sqrt{3T\log{T}} && \forall\ i \in [d] \\
\sum_{k\in [K]}\tau_k &\leq T && \\
\tau_k & \geq 0 && \forall\ k\in[K] \\
\sum_{k=1}^K \bI\{\tau_k > 0\} & \leq \DSL - 1, &&
\end{align*}
with its total revenue upper bounded by 
\begin{align*}
\sum_{k \in [K]} \sum_{j \in [n]} p_{j,k}q_{j,k} \tau_{k}   + nd \pmax \sqrt{3T\log{T}}.
\end{align*}


Recall that the optimal solution to the DLP uses $\DSL$ price vectors.
We have used $\mathcal{Z}(\bm{x}^*)$ to denote the set of price indices that are non-zero in the optimal solution to the DLP.
Note that, any policy $\pi \in \Pi[\DSL-2]$ must select no more than $\DSL-1$ price vectors.
So no matter which $\bm{x}^* \in \mathcal{X}$ one picks, there must exist an $l \in \mathcal{Z}(\bm{x}^*)$ such that under $\pi$, $\tau_l = 0$.
For any $l \in \mathcal{Z}(\bm{x}^*)$, define a family of mixed integer linear programs parameterized by $l$,
\begin{align*}
\bm{\mathsf{(DLP_l)}} \quad \DLP_{l} = \max_{(x_k)_{k \in [K]}} \sum_{k\in[K]}\sum_{j\in[n]}p_{j,k} \ q_{j,k} \ x_k &&& \\
\text{s.t.} \ \sum_{k\in[K]}\sum_{j\in[n]} a_{ij} \ q_{j,k} \ x_k &\leq B_i && \forall\ i \in [d] \\
\sum_{k\in [K]}x_k &\leq T && \\
x_l & = 0 && \\
x_k & \geq 0 && \forall\ k\in[K] \\
\sum_{k\in[K]} \bI\{x_k > 0\} & \leq \DSL - 1, &&
\end{align*}
such that this family of linear programs use no more than $(\DSL - 1)$ non-zero variables.
Now construct the following LP's, which we denote as ``perturbed LP's'':
\begin{align*}
\bm{\mathsf{(Perturbed \ DLP_l)}} \quad \mathsf{J}^\mathsf{Perturbed}_{l} = \max_{(x_k)_{k \in [K]}} \sum_{k\in[K]}\sum_{j\in[n]}p_{j,k} \ q_{j,k} \ x_k &&& \\
\text{s.t.} \ \sum_{k\in[K]}\sum_{j\in[n]} a_{ij} \ q_{j,k} \ x_k &\leq B_i + nd \sqrt{3T\log{T}} && \forall\ i \in [d] \\
\sum_{k\in [K]}x_k &\leq T && \\
x_l & = 0 && \\
x_k & \geq 0 && \forall\ k\in[K] \\
\sum_{k\in[K]} \bI\{x_k > 0\} & \leq \DSL - 1. &&
\end{align*}
Since from each solution $\bm{x}^*$ of the $\mathsf{Perturbed \ DLP_l}$, we can find a corresponding discounted solution $\bm{x}^* / (1+(nd\sqrt{3T\log{T}}) / \Bmin)$ that is feasible to the $\mathsf{DLP_l}$.
This suggests that $\mathsf{J}^\mathsf{Perturbed}_l \leq \DLP_l \cdot (1+(nd\sqrt{3T\log{T}}) / \Bmin)$, because $\mathsf{DLP_l}$ is a maximization problem.

Next we define an instance-dependent gap between the maximum objective value of $\mathsf{DLP_l}$, and the objective value of DLP.
Let $\Delta = (\DLP - \max_{l \in \mathcal{Z}(\bm{x}^*)} \DLP_l) / \DLP$ be such an instance-dependent gap normalized by $\DLP$. Importantly, while $\DLP$ scales linearly with $T$ and $\bm{B}$, $\Delta$ remains fixed as $T$ and $\bm{B}$ grow.





Putting everything together, we obtain the following result: conditional on event $E$ that happens with probability at least $1 - \frac{2}{T^3}$, for any policy $\pi\in\Pi[\DSL-2]$ and any possible realization  of $(\tau_1,\dots,\tau_K)$,  the total revenue collected during the selling horizon is  upper bounded by
\begin{align*}
 \max_{l \in \mathcal{Z}(\bm{x}^*)} \mathsf{J}^\mathsf{Perturbed}_l + nd \sqrt{3T\log{T}} \pmax 
& \leq \max_{l \in \mathcal{Z}(\bm{x}^*)} \DLP_l \cdot (1+\frac{nd\sqrt{3T\log{T}}}{\Bmin}) + nd \sqrt{3T\log{T}} \pmax \\
& \leq (\DLP - \Delta \DLP) \cdot (1+\frac{nd\sqrt{3T\log{T}}}{\Bmin}) + nd \sqrt{3T\log{T}} \pmax,
\end{align*}
which suggests that 
\[\Rev(\pi) \leq (\DLP - \Delta \DLP) \cdot (1+\frac{nd\sqrt{3T\log{T}}}{\Bmin}) + nd \sqrt{3T\log{T}} \pmax +1=\DLP - \Omega(T).\]
\Halmos
\endproof

\proof{Proof of Proposition~\ref{coro:stpLB}.}
For any problem instance $\mathcal{I} = (T,\bm{B},K,d;\bm{C}, \bm{R})$, we consider an arbitrary policy $\pi \in \Pi[\DSL-2]$ that only selects no more than $\DSL-1$ many arms.
For any $k \in [K]$, 
let $\tau_{k}$ be the total number of periods that action ${k}$ is offered during the selling horizon, under policy $\pi$.
Notice that $\tau_{k}$ is  a random variable, i.e., it is determined by the random trajectory of reward and cost realization and action selection. 

Now denote $C^s_{i,k}$ as the random amount of resource $i$ consumed, during the $\tau_k$ periods that arm $k$ is pulled;
denote $R^s_{k}$ as the random amount of rewards generated, during the $\tau_k$ periods that arm $k$ is pulled.
Here $\tau_{k}$ is a random amount, so we cannot directly use Hoeffding inequality.
But again we can use the ``reward tape'' trick demonstrated in the previous proof under the \nrm setup.  Let $C_{i,k}^s(t)$ denote the random amount of resource $i$ consumed, during the first $t$ periods that the arm $k$ is pulled; let $R^s_{k}{(t)}$ denote the random amount of rewards generated, during the first $t$ periods that arm $k$ is pulled.
Now we can use Hoeffding inequality on each reward tape: 
\begin{align*}
\forall k, \forall i, \forall t, & \Pr\left( \left| C^s_{i,k}(t) - t c_{i,k} \right| \leq \cmax \sqrt{3 t \log{T}} \right) \geq 1 - \frac{2}{T^6}; \\
\forall k, \forall t, & \Pr\left( \left| R^s_{k}(t) - t r_{k} \right| \leq \rmax \sqrt{3 t \log{T}} \right) \geq 1 - \frac{2}{T^6}.
\end{align*}
Denote the following event $E$:
\begin{align*}
\forall k, \forall i, \forall t, & \left| C^s_{i,k}(t) - t c_{i,k} \right| \leq \cmax \sqrt{3 t \log{T}}; \\
\forall k, \forall t, & \left| R^s_{k}(t) - t r_{k} \right| \leq \rmax \sqrt{3 t \log{T}}. 
\end{align*}
Using a union bound we have:
$$\Pr\left( E \right) \geq 1 - \frac{4}{T^3}$$
because $K,d$ are both less than $T$, and each arm cannot be pulled longer than $T$ periods.
The happening of such event implies that 
\begin{align*}
\forall k, \forall i,  & \left| C^s_{i,k} - \tau_k c_{i,k} \right| \leq \cmax \sqrt{3 \tau_k \log{T}}, \\
\forall k,& \left| R^s_{k} - \tau_k r_{k} \right| \leq \rmax \sqrt{3 \tau_k \log{T}},
\end{align*}
i.e., 
the realized rewards and costs are close to the expected values, suggesting that we can use LP to approximately bound the total reward collected by any policy $\pi \in \Pi[\DSL-2]$.

Specifically, we make the following arguments. 
On one hand, conditional on $E$, for any realization  of $(\tau_1,\dots,\tau_K)$, the total reward collected by policy $\pi$ during the entire horizon can be upper bounded by
\begin{align*}
\sum_{k \in [K]} R^s_{k} & \leq \sum_{k \in [K]} ( r_{k} \tau_{k} + \rmax \sqrt{3T\log{T}} ) \\
& \leq ( \sum_{k \in [K]} r_{k} \tau_{k} ) + d \rmax \sqrt{3T\log{T}},
\end{align*}
where the last inequality follows from $\pi\in\Pi[\DSL-2]$ and $\DSL \leq d+1$.

On the other hand, the consumption of each resource $i$ must not violate the resource constraints.
$$\sum_{k \in [K]} C^s_{i,k} \leq B_i.$$
Lower bounding $C^s_{i,k}$ by $( \tau_{k} c_{i, k} - \cmax \sqrt{3T\log{T}} )$ we have
$$\sum_{k \in [K]} \tau_{k} c_{i,k} \leq B_i + \sum_{k \in [K]} \cmax \sqrt{3T\log{T}} \leq B_i + d \cmax \sqrt{3T\log{T}}.$$
So conditional on $E$, any policy $\pi\in\Pi[\DSL-2]$ always satisfies the following constraints:
\begin{align*} 
\sum_{k\in[K]}r_{i,k} \ \tau_k &\leq B_i + d\cmax \sqrt{3T\log{T}} && \forall\ i \in [d] \\
\sum_{k\in [K]}\tau_k &\leq T && \\
\tau_k & \geq 0 && \forall\ k\in[K] \\
\sum_{k=1}^K \bI\{\tau_k > 0\} & \leq \DSL - 1, &&
\end{align*}
with its total collected reward upper bounded by 
\begin{align*}
\sum_{k \in [K]} r_{i,k} \tau_{k}   + d\rmax \sqrt{3T\log{T}}.
\end{align*}

Recall that the optimal solution to the DLP-G uses $\DSL$ many prices.
We have used $\mathcal{Z}(\bm{x}^*)$ to denote the set of price indices that are non-zero in the optimal solution to the DLP-G.
Note that, any policy $\pi \in \Pi[\DSL-2]$ must select no more than $\DSL-1$ price vectors.
So no matter which $\bm{x}^*$ one selects, there must exist an $l \in \mathcal{Z}(\bm{x}^*)$ such that under $\pi$, $\tau_l = 0$.
For any $l \in \mathcal{Z}(\bm{x}^*)$, define a family of mixed integer linear programs parameterized by $l$,
\begin{align*}
\bm{\mathsf{(DLP_l-G)}} \quad \DLPG_{l} = \max_{(x_k)_{k \in [K]}} \sum_{k\in[K]} r_k \ x_k &&& \\
\text{s.t.} \ \sum_{k\in[K]} c_{i,k} \ x_k &\leq B_i && \forall\ i \in [d] \\
\sum_{k\in [K]}x_k &\leq T && \\
x_l & = 0 && \\
x_k & \geq 0 && \forall\ k\in[K] \\
\sum_{k\in[K]} \bI\{x_k > 0\} & \leq \DSL - 1, &&
\end{align*}
such that this family of linear programs use no more than $(\DSL - 1)$ non-zero variables.
Now construct the following LP's, which we denote as ``perturbed LP's'':
\begin{align*}
\bm{\mathsf{(Perturbed \ DLP_l-G)}} \quad \mathsf{J}^\mathsf{Perturbed-G}_l = \max_{(x_k)_{k \in [K]}} \sum_{k\in[K]} r_{k} \ x_k &&& \\
\text{s.t.} \ \sum_{k\in[K]} c_{i,k} \ x_k &\leq B_i + d \cmax \sqrt{3T\log{T}} && \forall\ i \in [d] \\
\sum_{k\in [K]}x_k &\leq T && \\
x_l & = 0 && \\
x_k & \geq 0 && \forall\ k\in[K] \\
\sum_{k\in[K]} \bI\{x_k > 0\} & \leq \DSL - 1. &&
\end{align*}
Since from each solution $\bm{x}^*$ of the $\mathsf{Perturbed \ DLP_l-G}$, we can find a corresponding discounted solution $\bm{x}^* / (1+\frac{d \cmax \sqrt{3T\log{T}}}{\Bmin})$ that is feasible to the $\mathsf{DLP_l-G}$.
This suggests that $\mathsf{J}^\mathsf{Perturbed}_l \leq \DLPG_l \cdot (1+\frac{d \cmax \sqrt{3T\log{T}}}{\Bmin})$, because DLP-G is a maximization problem.

Next we define an instance-dependent gap between the maximum objective value of $\mathsf{DLP_l}$, and the objective value of DLP-G.
Let $\Delta = (\DLPG - \max_{l \in \mathcal{Z}(\bm{x}^*)} \DLPG_l) / \DLPG$ be such an instance-dependent gap normalized by $\DLPG$. 
Importantly, while $\DLP$ scales linearly with $T$ and $\bm{B}$, $\Delta$ remains fixed as $T$ and $\bm{B}$ grow.

Putting everything together, we obtain the following result: conditional on event $E$ that happens with probability at least $1 - \frac{4}{T^3}$, for any policy $\pi\in\Pi[\DSL-2]$ and any possible realization  of $(\tau_1,\dots,\tau_K)$,  the total collected reward is  upper bounded by
\begin{align*}
 \max_{l \in \mathcal{Z}(\bm{x}^*)} \mathsf{J}^\mathsf{Perturbed-G}_l + d \rmax \sqrt{3T\log{T}} 
& \leq \max_{l \in \mathcal{Z}(\bm{x}^*)} \DLPG_l \cdot (1+\frac{d \cmax \sqrt{3T\log{T}}}{\Bmin}) + d \rmax \sqrt{3T\log{T}} \\
& \leq (\DLPG - \Delta \DLPG) \cdot (1+\frac{d \cmax \sqrt{3T\log{T}}}{\Bmin}) + d \rmax \sqrt{3T\log{T}},
\end{align*}
which suggests that 
\[\Rev(\pi) \leq (\DLPG - \Delta \DLPG) \cdot (1+\frac{d \cmax \sqrt{3T\log{T}}}{\Bmin}) + d \rmax \sqrt{3T\log{T}} +1=\DLP - \Omega(T).\] \Halmos
\endproof

\section{Proof of Theorem~\ref{thm:NRMUB} and Proposition~\ref{prop:stpUB}}
\label{app:NRMUB}

In this section, we prove Theorem~\ref{thm:NRMUB} under the \nrm setup and Proposition~\ref{prop:stpUB} under the \stp setup.
For better exposition, we prove them two times under the two setups.
{
In the proof of Theorem~\ref{thm:NRMUB}, we only consider the case when 
\begin{align*}
\frac{T}{\log{T}} > \frac{4 \amax^2 n}{\underline{b}^2},
\end{align*}
which implies $\gamma > 0$. 
Otherwise, if 
\begin{align*}
\frac{T}{\log{T}} \leq \frac{4 \amax^2 n}{\underline{b}^2},
\end{align*}
the proof of Theorem~\ref{thm:NRMUB} becomes straightforward: $T$ being sufficiently small ensures that the regret upper bound in Theorem~\ref{thm:NRMUB} exceeds a constant multiple of $nT$, thus holding trivially.
In the proof of Proposition~\ref{prop:stpUB}, we only consider the case when
\begin{align*}
\frac{T}{\log{T}} > \frac{4 \cmax^2}{\underline{b}^2},
\end{align*}
which implies $\gamma > 0$. 
Otherwise, if
\begin{align*}
\frac{T}{\log{T}} \leq \frac{4 \cmax^2}{\underline{b}^2},
\end{align*}
the proof of Proposition~\ref{prop:stpUB} becomes straightforward: $T$ being sufficiently small ensures that the regret upper bound in Proposition~\ref{prop:stpUB} exceeds a constant multiple of $T$, thus holding trivially.
}

\proof{Proof of Theorem~\ref{thm:NRMUB}.}
Let $\pi$ be any policy suggested in Algorithm~\ref{alg:NRMUB}.
Let $\bx^*$ be the associated optimal solution.
We prove Theorem~\ref{thm:NRMUB} by comparing the expected revenue earned by Algorithm~\ref{alg:NRMUB} against a virtual policy $\pi^\mathsf{v}$.
This virtual policy $\pi^\mathsf{v}$ mimics Algorithm~\ref{alg:NRMUB} in steps 1--3.
But in step 4, it sets the price vector to be $\bp_{\sigma(1)}$ for the first $\gamma \cdot x^*_{\sigma(1)}$ periods, then $\bp_{\sigma(2)}$ for the next $\gamma \cdot x^*_{\sigma(2)}$ periods, ..., $\bp_{\sigma(\DSL)}$ for the next $\gamma \cdot x^*_{\sigma(\DSL)}$ periods, and finally $\bp_{\infty}$ for the last $(T - \gamma \cdot \sum_{l=1}^{\DSL} x^*_{\sigma(l)})$ periods.
Here $\bp_{\infty}$ serves as a shut-off price, under which $Q_j(\bp_{\infty}) = 0, \forall j \in [n]$.

Policy $\pi^{\mathsf{v}}$ is virtual because it requires a shut-off price $\bp_{\infty}$ that may or may not be available.
Moreover, it requires $\DSL$ many price changes, which is more than $\DSL-1$ many changes as suggested in Algorithm~\ref{alg:NRMUB}.

Policy $\pi^{\mathsf{v}}$ serves to bridge our analysis.
It breaks our Theorem~\ref{thm:NRMUB} into two inequalities that we will prove separately.
\begin{align}\label{eq:known-ub}
\Rev(\pi) \geq \Rev(\pi^{\mathsf{v}}) \geq \left(1 - 2 \amax \sqrt{\frac{n T \log{T}}{\Bmin^2}} - \frac{d}{T^2} \right) \cdot \DLP
\end{align}

For any policy $\pi$ as defined in Algorithm~\ref{alg:NRMUB} and its associated virtual policy $\pi^{\mathsf{v}}$, they both solve the same DLP and have the same optimal solution.
To prove the first inequality, note that both $\pi$ and $\pi^{\mathsf{v}}$ commit to the same prices in the first $\tilde{T} := \gamma \cdot \sum_{l=1}^{\DSL} x^*_{\sigma(l)}$ time periods, and earns the same revenue following each trajectory of random demand.
At the end of period $\tilde{T}$, policy $\pi$ still commits to $\bp_{\sigma(\DSL)}$, while policy $\pi^{\mathsf{v}}$ makes one change and sets $\bp_{\infty}$.
At the end of period $\tilde{T}$, if the selling horizon has ended due to inventory stock-outs, then either policy earns zero revenue, so $\pi$ and $\pi^{\mathsf{v}}$ makes no difference.
If the selling horizon has not ended and there is remaining inventory for any resource, then policy $\pi$ earns non-negative revenue, while $\pi^{\mathsf{v}}$ earns zero by setup a shut-off price.
Following each trajectory of random demand, policy $\pi$ earns more revenue than $\pi^{\mathsf{v}}$.
As a result, $\Rev(\pi)\geq\Rev(\pi^{\mathsf{v}})$.

To prove the second inequality, we introduce the following notation.
Let $\bI_{t,k}, \forall k \in [K], t \in [T]$ be an indicator of whether or not policy $\pi^{\mathsf{v}}$ offers price $\bp_k$ in period $t$.
\begin{align*}
\bI_{t,k} = \left\{
\begin{aligned}
& 1, && \text{if } k=\sigma(1), t \leq \gamma x^*_{\sigma(1)}; \\
& 1, && \text{if } \exists 1< l_0 \leq \DSL, s.t. \sigma(l_0)=k, \gamma \sum_{l=1}^{l_0-1} x^*_{\sigma(l)} < t \leq \gamma \sum_{l=1}^{l_0} x^*_{\sigma(l)}; \\
& 0, && \text{otherwise}
\end{aligned}
\right.
\end{align*}
$\bI_{t,k}$ is deterministic once policy $\pi^{\mathsf{v}}$ is determined.

Under policy $\pi^{\mathsf{v}}$, following each trajectory of random demand, we define the length of the effective selling horizon $\tau$ as a function of a stopping time $t_0$:
\begin{align*}
\tau = \tilde{T} \wedge \min\left\{ t_0 - 1\left| \exists i, s.t. \ \sum_{t=1}^{t_0} \sum_{k \in [K]} \bI_{t,k} \sum_{j \in [n]} Q_{j,k} \cdot a_{ij} > B_i \right. \right\}
\end{align*}
The effective selling horizon is the minimum between (i) last period before the cumulative demand of any resource exceeds its initial inventory, and (ii) the last period before policy $\pi^{\mathsf{v}}$ switches to the shut-off price.

Let $D_{t,i}$ be the remaining inventory of resource $i$ at the end of period $t$.
Under this notation, $D_{0,i} = B_i$.
Note that $D_{t,i}$ are random variables, and during the effective selling horizon, inventory updates in the following fashion
\begin{align}
\label{eqn:InventoryUpdate}
\forall t \in [\tau], D_{t,i} = D_{t-1,i} - \sum_{k \in [K]} \bI_{t,k} \sum_{j\in[n]} Q_{j,k} \cdot a_{ij} \geq 0
\end{align}

Now we calculate the expected revenue.
\begin{align}
\Rev(\pi^{\mathsf{v}}) & \geq \bE_{Q_{j,k}} \left[ \sum_{t=1}^{\tilde{T}} \bI_{\{\forall i, D_{t-1, i} \geq n \amax\}} \sum_{k\in[K]} \bI_{t,k} \sum_{j\in[n]} p_{j,k} Q_{j,k} \right] \nonumber \\
& = \sum_{t=1}^{\tilde{T}} \Pr(\forall i, D_{t-1, i} \geq n \amax) \sum_{k\in[K]} \bI_{t,k} \sum_{j\in[n]} p_{j,k} q_{j,k} \nonumber \\
& \geq \Pr(\forall i, D_{\tilde{T}, i} \geq n \amax) \sum_{t=1}^{\tilde{T}} \sum_{k\in[K]} \bI_{t,k} \sum_{j\in[n]} p_{j,k} q_{j,k} \label{eqn:RevCrudeLB}
\end{align}
We explain the inequalities.
The first inequality is because we only focus on the revenue earned if event $\{ \forall i, D_{t-1, i} \geq n \amax \}$ happens, while ignoring the revenue earned if event $\{ \forall i, D_{t-1, i} \geq n \amax \}$ does not happen; and when event $\{ \forall i, D_{t-1, i} \geq n \amax \}$ does happen, the maximum amount of any resource $i$ demanded in one single period cannot exceed $n \amax$.
The first equality is expanding the expectations, where we use the fact that $\bI_{\{\forall i, D_{t-1, i} \geq n \amax\}}$ and $\sum_{k\in[K]} \bI_{t,k} \sum_{j\in[n]} p_{j,k} Q_{j,k}$ are independent, because the indicator is a random event happening up to period $t-1$, while the summation term is a random amount happening in period $t$.
The third inequality is due to \eqref{eqn:InventoryUpdate}, $D_{t,i}$ is decreasing in $t$, so $D_{t-1,i} \geq D_{\tilde{T},i}$.

In this block of inequalities \eqref{eqn:RevCrudeLB}, the summation term $$\sum_{t=1}^{\tilde{T}} \sum_{k\in[K]} \bI_{t,k} \sum_{j\in[n]} p_{j,k} q_{j,k} = \sum_{l=1}^{\DSL} \sum_{k \in \{\sigma(l)=k\}} \gamma x^*_{k} \sum_{j\in[n]} p_{j,k} q_{j,k} = \gamma \DLP,$$ where the indicators $\bI_{t,k}$ locate which $k$ counts into this summation. 

The next thing we do is to lower bound the probability term in \eqref{eqn:RevCrudeLB}. 
Note that, for any $i\in[d]$,
\begin{align*}
\bE \left[ \sum_{t=1}^{\tilde{T}} \sum_{k \in [K]} \bI_{t,k} \sum_{j\in[n]} Q_{j,k} \cdot a_{ij} \right] & = \sum_{t=1}^{\tilde{T}} \sum_{k \in [K]} \bI_{t,k} \sum_{j\in[n]} q_{j,k} \cdot a_{ij} \\
& = \sum_{l=1}^{\DSL} \sum_{k \in \{\sigma(l)=k\}} \gamma x^*_{k} \sum_{j\in[n]} q_{j,k} a_{ij} \\
& \leq \gamma B_i \\
& < B_i - n \amax,
\end{align*}
where the last (strict) inequality is because we plug in $\gamma = 1 - 2 \amax \sqrt{\frac{n T \log{T}}{\Bmin^2}}$.

This above inequality suggests that for any $i \in[d]$, the expected cumulative demand generated up till period $\tilde{T}$ is strictly less than $B_i - n \amax$.
So we can use concentration inequalities.
\begin{align}
\Pr(\forall i, D_{\tilde{T}, i} \geq n \amax) & = 1 - \Pr\left( \exists i, s.t. \sum_{t=1}^{\tilde{T}} \sum_{k \in [K]} \bI_{t,k} \sum_{j\in[n]} Q_{j,k} \cdot a_{ij} \geq B_i - n \amax\right) \nonumber \\
& \geq 1 - \sum_{i \in [d]} \Pr\left( \sum_{t=1}^{\tilde{T}} \sum_{k \in [K]} \bI_{t,k} \sum_{j\in[n]} Q_{j,k} \cdot a_{ij} - \gamma B_i \geq (1-\gamma) B_i - n \amax\right) \nonumber \\
& \geq 1 - \sum_{i \in [d]} \exp{\left(-\frac{2((1-\gamma) B_i - n \amax)^2}{\amax^2 nT}\right)} \nonumber \\
& \geq 1 - d \exp{\left(-\frac{2((1-\gamma) \Bmin - n \amax)^2}{\amax^2 nT}\right)} \nonumber \\
& \geq 1 - \frac{d}{T^2} \label{eqn:InventoryConcentration}
\end{align}
where the first inequality is due to union bound;
the second inequality is due to Hoeffding inequality, $Q_{j,k} a_{ij}$ is bounded by $\amax$, and there are no more than $n \cdot T$ such terms;
the third inequality is because we lower bound each $B_i$ by $\Bmin$;
the last inequality is when we plug in $1 - \gamma = 2 \amax \sqrt{\frac{n T \log{T}}{\Bmin^2}}$, and we know that $T>n$.

Putting \eqref{eqn:InventoryConcentration} into \eqref{eqn:RevCrudeLB}, and using the fact that $\DLP\le n\pmax T$, we have
\begin{align*}
\Rev(\pi^{\mathsf{v}}) & \geq \DLP - \left(\frac{2 \amax \pmax}{\underline{b}} \sqrt{n^3} \sqrt{T\log T} + \pmax d n \frac{1}{T}\right) \\
& \geq \Rev(\pi^*[\infty])- \max\left\{\frac{3 \amax \pmax}{\underline{b}}, 3\pmax d \right\} \sqrt{n^3} \sqrt{T\log T}
\end{align*}
which finishes the proof.
\Halmos
\endproof

\proof{Proof of Proposition~\ref{prop:stpUB}.}
Let $\pi$ be any policy suggested in Algorithm~\ref{alg:stpUB} under \stp Setup.
Let $\bx^*$ be the associated optimal solution.
We prove Proposition~\ref{prop:stpUB} by comparing the expected revenue earned by Algorithm~\ref{alg:stpUB} against a virtual policy $\pi^\mathsf{v}$.
This virtual policy $\pi^\mathsf{v}$ mimics Algorithm~\ref{alg:stpUB} in steps 1--3.
But in step 4, it pulls arm $\sigma(1)$ for the first $\gamma \cdot x^*_{\sigma(1)}$ periods, then arm $\sigma(2)$ for the next $\gamma \cdot x^*_{\sigma(2)}$ periods, ..., $\sigma(\DSL)$ for the next $\gamma \cdot x^*_{\sigma(\DSL)}$ periods, and finally halts for the last $(T - \gamma \cdot \sum_{l=1}^{\DSL} x^*_{\sigma(l)})$ periods pulling no arms.
Such a halting notion was introduced in \citet{badanidiyuru2013bandits}.

Policy $\pi^{\mathsf{v}}$ serves to bridge our analysis.
It breaks our Proposition~\ref{prop:stpUB} into two inequalities that we will prove separately.
\begin{align}\label{eq:known-ubg}
\Rev(\pi) \geq \Rev(\pi^{\mathsf{v}}) \geq \left(1 - \frac{2 \cmax}{\Bmin} \sqrt{T \log{T}} - \frac{d}{T^2} \right) \cdot \DLPG
\end{align}

For any policy $\pi$ as defined in Algorithm~\ref{alg:stpUB} and its associated virtual policy $\pi^{\mathsf{v}}$, they both solve the same DLP-G and have the same optimal solution.
To prove the first inequality, note that both $\pi$ and $\pi^{\mathsf{v}}$ pull the same arm in the first $\tilde{T} := \gamma \cdot \sum_{l=1}^{\DSL} x^*_{\sigma(l)}$ time periods, and earns the same revenue following each trajectory of random demand.
At the end of period $\tilde{T}$, policy $\pi$ still pulls arm $\sigma(\DSL)$, while policy $\pi^{\mathsf{v}}$ halts.
At the end of period $\tilde{T}$, if the selling horizon has ended due to inventory stock-outs, then both policies earn zero reward, so $\pi$ and $\pi^{\mathsf{v}}$ make no difference.
If the selling horizon has not ended and there is remaining inventory for some resource, then policy $\pi$ earns non-negative reward, while $\pi^{\mathsf{v}}$ halts and earns zero.
Following each trajectory of random demand, policy $\pi$ earns more revenue than $\pi^{\mathsf{v}}$.
As a result, $\Rev(\pi)\geq\Rev(\pi^{\mathsf{v}})$.

To prove the second inequality, we introduce the following notation.
Let $\bI_{t,k}, \forall k \in [K], t \in [T]$ be an indicator of whether or not policy $\pi^{\mathsf{v}}$ pulls arm $k$ in period $t$.
\begin{align*}
\bI_{t,k} = \left\{
\begin{aligned}
& 1, && \text{if } k=\sigma(1), t \leq \gamma x^*_{\sigma(1)}; \\
& 1, && \text{if } \exists 1< l_0 \leq \DSL, s.t. \sigma(l_0)=k, \gamma \sum_{l=1}^{l_0-1} x^*_{\sigma(l)} < t \leq \gamma \sum_{l=1}^{l_0} x^*_{\sigma(l)}; \\
& 0, && \text{otherwise}
\end{aligned}
\right.
\end{align*}
$\bI_{t,k}$ is deterministic once policy $\pi^{\mathsf{v}}$ is determined.

Under policy $\pi^{\mathsf{v}}$, following each trajectory of random demand, we define the length of the effective selling horizon $\tau$ as a function of a stopping time $t_0$:
\begin{align*}
\tau = \tilde{T} \wedge \min\left\{ t_0 - 1\left| \exists i, s.t. \ \sum_{t=1}^{t_0} \sum_{k \in [K]} \bI_{t,k} C_{i,k} > B_i \right. \right\}
\end{align*}
The effective selling horizon is the minimum between (i) last period before the cumulative demand of any resource exceeds its initial inventory, and (ii) the last period before policy $\pi^{\mathsf{v}}$ halts.

Let $D_{t,i}$ be the remaining inventory of resource $i$ at the end of period $t$.
Under this notation, $D_{0,i} = B_i$.
Note that $D_{t,i}$ are random variables, and during the effective selling horizon, inventory updates in the following fashion
\begin{align}
\label{eqn:InventoryUpdate-G}
\forall t \in [\tau], D_{t,i} = D_{t-1,i} - \sum_{k \in [K]} \bI_{t,k} C_{i,k} \geq 0
\end{align}

Now we calculate the expected revenue.
\begin{align}
\Rev(\pi^{\mathsf{v}}) & \geq \bE_{Q_{j,k}} \left[ \sum_{t=1}^{\tilde{T}} \bI_{\{\forall i, D_{t-1, i} \geq \cmax\}} \sum_{k\in[K]} \bI_{t,k} R_k \right] \nonumber \\
& = \sum_{t=1}^{\tilde{T}} \Pr(\forall i, D_{t-1, i} \geq \cmax) \sum_{k\in[K]} \bI_{t,k} R_k \nonumber \\
& \geq \Pr(\forall i, D_{\tilde{T}, i} \geq \cmax) \sum_{t=1}^{\tilde{T}} \sum_{k\in[K]} \bI_{t,k} R_k \label{eqn:RevCrudeLB-G}
\end{align}
We explain the inequalities.
The first inequality is because we only focus on the revenue earned if event $\{ \forall i, D_{t-1, i} \geq \cmax \}$ happens, while ignoring the revenue earned if event $\{ \forall i, D_{t-1, i} \geq \cmax \}$ does not happen; and when event $\{ \forall i, D_{t-1, i} \geq \cmax \}$ does happen, the maximum amount of any resource $i$ demanded in one single period cannot exceed $\cmax$.
The first equality is expanding the expectations, where we use the fact that $\bI_{\{\forall i, D_{t-1, i} \geq \cmax\}}$ and $\sum_{k\in[K]} \bI_{t,k} R_k$ are independent, because the indicator is a random event happening up to period $t-1$, while the summation term is a random amount happening in period $t$.
The third inequality is due to \eqref{eqn:InventoryUpdate-G}, $D_{t,i}$ is decreasing in $t$, so $D_{t-1,i} \geq D_{\tilde{T},i}$.

In this block of inequalities \eqref{eqn:RevCrudeLB-G}, the summation term
\begin{align*}
\sum_{t=1}^{\tilde{T}} \sum_{k\in[K]} \bI_{t,k} R_k = \sum_{l=1}^{\DSL} \sum_{k \in \{\sigma(l)=k\}} \gamma x^*_{k} R_k = \gamma \DLPG,
\end{align*}
since the indicators $\bI_{t,k}$ locate which $k$ counts into this summation. So the next thing we do is to lower bound the probability term, $\Pr(\forall i, D_{\tilde{T}, i} \geq \cmax)$.

Note that for any $i\in[d]$,
\begin{align*}
\bE \left[ \sum_{t=1}^{\tilde{T}} \sum_{k \in [K]} \bI_{t,k} C_{i,k} \right] & = \sum_{t=1}^{\tilde{T}} \sum_{k \in [K]} \bI_{t,k} c_{i,k} \\
& = \sum_{l=1}^{\DSL} \sum_{k \in \{\sigma(l)=k\}} \gamma x^*_{k} c_{i,k} \\
& \leq \gamma B_i \\
& < B_i - \cmax
\end{align*}
where the last inequality is because we plug in $\gamma = 1 - \frac{2 \cmax}{\Bmin} \sqrt{T \log{T}}$, and that $2\sqrt{T \log{T}} > 1$.

This above inequality suggests that for any $i \in[d]$, the expected cumulative demand generated up till period $\tilde{T}$ is strictly less than $B_i - \cmax$.
So we can use concentration inequalities.
\begin{align}
\Pr(\forall i, D_{\tilde{T}, i} \geq \cmax) & = 1 - \Pr\left( \exists i, s.t. \sum_{t=1}^{\tilde{T}} \sum_{k \in [K]} \bI_{t,k} C_{i,k} \geq B_i - \cmax\right) \nonumber \\
& \geq 1 - \sum_{i \in [d]} \Pr\left( \sum_{t=1}^{\tilde{T}} \sum_{k \in [K]} \bI_{t,k} C_{i,k} - \gamma B_i \geq (1-\gamma) B_i - \cmax\right) \nonumber \\
& \geq 1 - \sum_{i \in [d]} \exp{\left(-\frac{2((1-\gamma) B_i - \cmax)^2}{\cmax^2 nT}\right)} \nonumber \\
& \geq 1 - d \exp{\left(-\frac{2((1-\gamma) \Bmin - \cmax)^2}{\cmax^2 nT}\right)} \nonumber \\
& \geq 1 - \frac{d}{T^2} \label{eqn:InventoryConcentration-G}
\end{align}
where the first inequality is due to union bound;
the second inequality is due to Hoeffding inequality, $C_{i,k}$ is bounded by $\cmax$;
the third inequality is because we lower bound each $B_i$ by $\Bmin$;
the last inequality is when we plug in $1 - \gamma = \frac{2 \cmax}{\Bmin} \sqrt{T \log{T}}$, and that $\sqrt{T\log{T}}>1$.

Putting \eqref{eqn:InventoryConcentration-G} into \eqref{eqn:RevCrudeLB-G} , and using the fact that $\DLPG\le\rmax T$, we have
\begin{align*}
\Rev(\pi^{\mathsf{v}}) & \geq \DLPG - \left(\frac{2 \cmax \rmax}{\underline{b}} \sqrt{T\log T} + \rmax d \frac{1}{T}\right) \\
& \geq \Rev(\pi^*[\infty])- \max\left\{\frac{3 \cmax \rmax}{\underline{b}}, 3\rmax d \right\} \sqrt{T\log T}
\end{align*}
which finishes the proof.
\Halmos
\endproof

\section{Proof of Theorem~\ref{thm:BlindNRMUB} and Proposition~\ref{prop:BwKUB}}
\label{app:BlindNRMUB}

\subsection{Preliminary Lemma}
\begin{lemma}\label{lem:points}Let ${(t_l)}_{l\in[\nu(s,d)+1]}$ be defined by \eqref{eq:points}. Then for any $l\in[\nu(s,d)+1]$, we have
\[\sum_{r\in[l]}\frac{t_r-t_{r-1}}{\sqrt{t_{r-1}}}\le \frac{8\log T}{\sqrt{K}}t_1.\]
\end{lemma}
\proof{Proof of Lemma \ref{lem:points}.}
For any $r\in[l]$, we have
\begin{align}\label{eq:small-r}
    \frac{t_r-t_{r-1}}{\sqrt{t_{r-1}}}&\le \frac{t_r}{\sqrt{t_{r-1}}}\le \frac{K(T/K)^{\frac{2-2^{-(r-1)}}{2-2^{-\nu(s,d)}}}}{\sqrt{K(T/K)^{\frac{2-2^{-(r-2)}}{2-2^{-\nu(s,d)}}}/2}}\le\sqrt{2K}(T/K)^{\frac{1}{2-2^{-\nu(s,d)}}}\le2\sqrt{\frac{2}{K}}t_1.
\end{align}
For any $r\in[l]$ and $r\ge\log_2\log_2(T/K)+1$, we have
\begin{align}\label{eq:large-r}
    t_{r-1}\ge\frac{1}{2}K(T/K)^{\frac{2-2^{-(r-2)}}{2-2^{-\nu(s,d)}}}\ge\frac{1}{2}K(T/K)^{1-2^{-(r-1)}}\ge\frac{1}{2}K\frac{(T/K)}{2}\ge{T/4}.
\end{align}
Moreover, we have
\begin{align}\label{eq:t1}
    t_1\ge\frac{1}{2}K(T/K)^{\frac{1}{2-2^{-\nu(s,d)}}}\ge\frac{1}{2}\sqrt{KT}.
\end{align}

Using the above three inequalities, we have
    \begin{align*}
    &\sum_{r\in[l]}\frac{t_r-t_{r-1}}{\sqrt{t_{r-1}}}\\
    &=\sum_{r\in[l]}\bI(r< \log_2\log_2(T/K)+1)\frac{t_r-t_{r-1}}{\sqrt{t_{r-1}}}+\sum_{r\ge [l]}\bI(r\ge \log_2\log_2(T/K)+1)\frac{t_r-t_{r-1}}{\sqrt{t_{r-1}}}\\
    &\le(\log_2\log_2(T/K)+1)2\sqrt{\frac{2}{K}}t_1+\sum_{r\ge [l]}\bI(r\ge \log_2\log_2(T/K)+1)\frac{t_r-t_{r-1}}{\sqrt{t_{r-1}}}\\
    &\le (\log_2\log_2(T/K)+1)2\sqrt{\frac{2}{K}}t_1+\sum_{r\in[l]}\frac{t_r-t_{r-1}}{\sqrt{T/4}}\\
    &\le (\log_2\log_2(T/K)+1)2\sqrt{\frac{2}{K}}t_1+2\sqrt{T}\\
    &\le (\log_2\log_2(T/K)+1)2\sqrt{\frac{2}{K}}t_1+\frac{4}{\sqrt{K}}t_1\\
    &\le\frac{8\log T}{\sqrt{K}}t_1,
\end{align*}
where the first inequality follows from \eqref{eq:small-r}, the second inequality follows from \eqref{eq:large-r}, and the second-to-last inequality follows from \eqref{eq:t1}.
\Halmos
\endproof

\subsection{Proof of Theorem~\ref{thm:BlindNRMUB} and Proposition~\ref{prop:BwKUB}}

In this section we prove Theorem~\ref{thm:BlindNRMUB} under the \bnrm setup and Proposition~\ref{prop:BwKUB} under the \bwk setup.
For better exposition we prove them two times under the two setups.
{ 
In the proof of Theorem~\ref{thm:BlindNRMUB}, we only consider the case when 
\begin{align*}
T > \frac{17 \amax \sqrt{n(d+1)\log[(d+1)KT]} K^{1-\frac{1}{2-2^{-\nu(s,d)}}} T^{\frac{1}{2-2^{-\nu(s,d)}}} \log{T}}{\underline{b}},
\end{align*}
which implies $\gamma > 0$. 
Otherwise, if 
\begin{align*}
T \leq \frac{17 \amax \sqrt{n(d+1)\log[(d+1)KT]} K^{1-\frac{1}{2-2^{-\nu(s,d)}}} T^{\frac{1}{2-2^{-\nu(s,d)}}} \log{T}}{\underline{b}},
\end{align*}
the proof of Theorem~\ref{thm:BlindNRMUB} becomes straightforward: $T$ being sufficiently small ensures that the regret upper bound in Theorem~\ref{thm:BlindNRMUB} exceeds a constant multiple of $nT$, thus holding trivially.
In the proof of Proposition~\ref{prop:BwKUB}, we only consider the case when
\begin{align*}
T > \frac{17 \cmax \sqrt{(d+1)\log{[(d+1)KT]}} K^{1-\frac{1}{2-2^{-\nu(s,d)}}} T^{\frac{1}{2-2^{-\nu(s,d)}}} \log{T}}{\underline{b}},
\end{align*}
which implies $\gamma > 0$. 
Otherwise, if
\begin{align*}
T \leq \frac{17 \cmax \sqrt{(d+1)\log{[(d+1)KT]}} K^{1-\frac{1}{2-2^{-\nu(s,d)}}} T^{\frac{1}{2-2^{-\nu(s,d)}}} \log{T}}{\underline{b}},
\end{align*}
the proof of Proposition~\ref{prop:BwKUB} becomes straightforward: $T$ being sufficiently small ensures that the regret upper bound in Proposition~\ref{prop:BwKUB} exceeds a constant multiple of $T$, thus holding trivially.
}

\proof{Proof of Theorem~\ref{thm:BlindNRMUB}.}
Consider the DLP defined by \eqref{eqn:obj}, \eqref{eqn:constraint:inventory}, \eqref{eqn:constraint:time}, and \eqref{eqn:constraint:NonNeg}. 
Let $\mathsf{DLP}_{\bQ}$ denote the DLP with the underlying distributions being $\bQ$. 
Let $\DLP_{\bQ}$ denote the optimal objective value of $\mathsf{DLP}_{\bQ}$.

Let $\tilde{l}$ be the last epoch in the execution of policy $\phi$, and let $\tau$ be the last period before the policy $\phi$ stops. We know that $\tau+1$ is a stopping time and we have $T_{\tilde{l}-1}< \tau\le t_{\tilde{l}}\le T$. Since $\phi$ makes at most $(K-1)(\tilde{l}-1)$ switches before $T_{\tilde{l}-1}$ and makes at most $d+1$ switches after $T_{\tilde{l}-1}$, its total number of switches is always upper bounded by $(K-1)\nu(s,d)+(d+1)\le s$.

We use a coupling argument for the regret analysis. Consider a virtual policy $\phi^{\mathsf{v}}$ that runs under exactly the same demand realization process and acts exactly the same as $\phi$ until period $\tau$, but keeps running until the end of epoch $\nu(s,d)+1$ regardless of the resource constraints. 
Without conflicts to the previously defined notation in Algorithm \ref{alg:bsse}, for each sample path of the action and demand realization process under the execution of policy $\phi^{\mathsf{v}}$, let $T_l$ denote the last period of epoch $l$ ($l\in[\nu(s,d)+1]$), let $n_{k}(t)$ be the total number of periods that price vector $\bp_k$ is chosen during period 1 to $t$ ($k\in[K], t\in[T_{\nu(s,d)+1}]$), and let $\bar{q}_{j,k}(t)$ be the average realized demand of product $j$ sold at price vector $\bp_k$ during period 1 to $t$, on this sample path ($j\in[n], k\in[K], t\in[T_{\nu(s,d)+1}]$). 
For all $t\in[T_{\nu(s,d)+1}]$, define the confidence radius as
\[
\radius_k(t)=\sqrt{\frac{\log [(d+1)KT]}{n_k(t)}},~~\forall k\in[K].
\]
For notational simplicity, define
\[
\begin{cases}U^{\mathsf{rew}}_{k}(T_{l})=\min\left\{\sum_{j\in[n]}p_{j,k}\bar{q}_{j,k}(T_{l})+||\bp_k||_2\radius_k(T_{l}),U^{\mathsf{rew}}_k(T_{l-1})\right\}
,\\L^{\mathsf{rew}}_{k}(T_{l})=\max\left\{\sum_{j\in[n]}p_{j,k}\bar{q}_{j,k}(T_{l})-||\bp_k||_2\radius_k(T_{l}),L_k^{\mathsf{rew}}(T_{l-1})\right\},\end{cases}~~\forall k\in[K],
\]
\[
\begin{cases}U^{\mathsf{cost}}_{i,k}(T_{l})=\min\left\{\sum_{j\in[n]}a_{ij}\bar{q}_{j,k}{(T_{l})}+||A_i||_2\radius_k(T_{l}),U^{\mathsf{cost}}_{i,k}(T_{l-1})\right\}
,\\L^{\mathsf{cost}}_{i,k}(T_{l})=\max\left\{\sum_{j\in[n]}a_{ij}\bar{q}_{j,k}{(T_{l})}-||A_i||_2\radius_k(T_{l}),L_{i,k}^{\mathsf{cost}}(T_{l-1})\right\},\end{cases}~~\forall i\in[d],\forall k\in[K].
\]
for $l=\nu(s,d)$.

Define the \textit{clean event} $\mathcal{E}$ as
$$\left\{\forall i\in[d], k\in[K],  t\in[T_{\nu(s,d)+1}],~\begin{cases}\left|\sum_{j\in[n]}p_{j,k}\bar{q}_{j,k}(t)-\sum_{j\in[n]}p_{j,k}q_{j,k}\right|\le||\bp_k||_2\radius_k(t),\\\left|\sum_{j\in[n]}a_{ij}\bar{q}_{j,k}(t)-\sum_{j\in[n]}a_{ij}q_{j,k}\right|\le ||A_i||_2\radius_k(t).
\end{cases} 
 \right\}.$$
By the Hoeffding's inequality for general bounded random variables (see Theorem A.1 in \cite{slivkins2019introduction}) and a standard union bound argument (see Chapter 1.3.1 in \cite{slivkins2019introduction}), we have $\text{Pr}(\mathcal{E})\ge 1-\frac{2}{(d+1)KT}$ under distributions $\bQ$ and policy $\phi^{\mathsf{v}}$. Since the clean event happens with very high probability, we can just focus on a \textit{clean execution} of policy $\phi^{\mathsf{v}}$: an execution in which the clean event holds. 
Conditional on the clean event,  it holds that $\forall i\in[d], k\in[K],  l\in[\nu(s,d)]$,
\begin{align}\label{eqn:BlindNRM-UB-0}
\begin{cases}
    U_{k}^{\mathsf{rew}}(T_l)-2||\bp_k||_2\radius_k(t)\le L^{\mathsf{rew}}_{k}(T_l)\le\sum_{j\in[n]}p_{j,k}q_{j,k}\le U_{k}^{\mathsf{rew}}(T_l)\le L^{\mathsf{rew}}_{k}(T_l)+2||\bp_k||_2\radius_k(t),\\
    U_{i,k}^{\mathsf{cost}}(T_l)-2||A_i||_2\radius_k(t)\le L^{\mathsf{cost}}_{i,k}(T_l)\le\sum_{j\in[n]}a_{ij}q_{j,k}\le U_{i,k}^{\mathsf{cost}}(T_l)\le L^{\mathsf{cost}}_{i,k}(T_l)+2||A_i||_2\radius_k(t).
\end{cases}
\end{align}
In the rest of the proof, we always assume that $\mathcal{E}$ holds.

For all $i\in[d]$ and  $l\in[\nu(s,d)+1]$, we have
\begin{align}\label{eqn:BlindNRM-UB-1}
&\sum_{k\in[K]}\sum_{j\in[n]} a_{ij} \bar{q}_{j,k}(T_{l})n_{k}(T_{l})\notag\\
\le &\sum_{k\in[K]}\sum_{j\in[n]}a_{ij}q_{j,k}n_k(T_{l})+\sum_{k\in[K]}||A_{i}||_2\radius_k(T_l)n_k(T_{l})\notag\\
=&\sum_{k\in[K]}\sum_{j\in[n]}a_{ij}q_{j,k}n_k(T_{l})+||A_{i}||_2\sqrt{\log[(d+1)KT]}\sum_{k\in[K]}\sqrt{n_k(T_{l})}\notag\\
\le&\sum_{k\in[K]}\sum_{j\in[n]}a_{ij}q_{j,k}n_k(T_{l})+||A_{i}||_2\sqrt{\log[(d+1)KT]}\sqrt{K{T_{l}}},
\end{align}
where the last inequality is by Jensen's inequality. Similarly, for all $i\in[d]$ and  $l\in[\nu(s,d)+1]$, we also have
\begin{align*}
    \sum_{k\in[K]}\sum_{j\in[n]} a_{ij} \bar{q}_{j,k}(T_{l})n_{k}(T_{l})\ge\sum_{k\in[K]}\sum_{j\in[n]}a_{ij}q_{j,k}n_k(T_{l})-||A_{i}||_2\sqrt{\log[(d+1)KT]}\sqrt{K{T_{l}}}.
\end{align*}

For notational convenience, define  $\bx^{\nu(s,d)+1,i}=\bx^*_{\tilde{\bQ}}$ for all $i\in[K]$. Then we have $n_{k}(T_{l})=\sum_{r=1}^{l} \gamma N_k^r=\gamma\sum_{r=1}^l\left(\frac{t_{r}-t_{r-1}}{T}\sum_{o=1}^{K}\frac{(\bx^{r,o})_k}{K}\right)$ for all $l\in[\nu(s,d)+1]$. Then for all $i\in[d]$ and  $l\in[\nu(s,d)+1]$, we have
\begin{align}\label{eqn:BlindNRM-UB-2}
&\sum_{k\in[K]}\sum_{j\in[n]}a_{ij}q_{j,k}n_{k}(T_l)\notag\\
=&\gamma\sum_{r\in[l]}\frac{t_r-t_{r-1}}{T}\sum_{k\in[K]}\sum_{j\in[n]}a_{ij}q_{j,k}\sum_{o\in[K]}\frac{(\bx^{r,o})_k}{K}\notag\\
\le&\gamma\sum_{r\in[l]}\frac{t_r-t_{r-1}}{T}\sum_{k\in[K]}\left(L_{i,k}^{\mathsf{cost}}(T_{l-1})+2||A_i||_2\radius_k(T_{l-1})\right)\sum_{o\in[K]}\frac{(\bx^{r,o})_k}{K}\notag\\
=&\gamma\sum_{r\in[l]}\frac{t_r-t_{r-1}}{T}\sum_{o\in[K]}\frac{\sum_{k\in[K]}L_{i,k}^{\mathsf{cost}}(T_{l-1})(\bx^{r,o})_k}{K}+2\gamma||A_i||_2\sum_{r\in[l]}\frac{t_r-t_{r-1}}{T}\sum_{k\in[K]}\radius_k(T_{r-1})\sum_{o\in[K]}\frac{(\bx^{r,o})_k}{K}\notag\\
\le&\gamma\sum_{r\in[l]}\frac{t_r-t_{r-1}}{T}\sum_{o\in[K]}\frac{B_i}{K}+2\gamma||A_i||_2\sum_{r\in[l]}\frac{t_r-t_{r-1}}{T}\sum_{k\in[K]}\radius_k(T_{r-1})\sum_{o\in[K]}\frac{(\bx^{r,o})_k}{K}\notag\\
=&\gamma \frac{t_l}{T}B_i+2\gamma||A_i||_2\sum_{r\in[l]}\frac{t_r-t_{r-1}}{T}\sum_{k\in[K]}\radius_k(T_{r-1})\sum_{o\in[K]}\frac{(\bx^{r,o})_k}{K},
\end{align}
where the first inequality is by (\ref{eqn:BlindNRM-UB-0}), and the second inequality is by the  feasibility of $\bx^{r,o}$ to the second-stage exploration LP of epoch $r$. We now bound $\sum_{k\in[K]}\radius_k(T_{r-1})(\bx^{r,o})_k$. Since $[L^{\mathsf{rew}}_{k}(T_{r-1}),U_{k}^{\mathsf{rew}}(T_{r-1})]\subset [L^{\mathsf{rew}}_{k}(T_{r-2}),U_{k}^{\mathsf{rew}}(T_{r-2})]\subset\cdots\subset [L^{\mathsf{rew}}_{k}(T_{0}),U_{k}^{\mathsf{rew}}(0)]$ and $[L^{\mathsf{cost}}_{i,k}(T_{r-1}),U_{i,k}^{\mathsf{cost}}(T_{r-1})]\subset [L^{\mathsf{cost}}_{i,k}(T_{r-2}),U_{i,k}^{\mathsf{cost}}(T_{r-2})]\subset\cdots\subset [L^{\mathsf{cost}}_{i,k}(T_{0}),U_{i,k}^{\mathsf{cost}}(0)]$, for the first-stage pessimistic LP in Algorithm \ref{alg:bsse}, $\mathsf{J}_l^{\mathsf{PES}}$ is non-decreasing as $l$ grows, which in turns makes the constraints in the second-stage exploration LPs more stringent as $l$ grows, leading to non-increasing $\bx^{l,j}$ as $l$ grows $(\forall j\in[K])$. This ensures that $(\bx^{i,k})_k\ge(\bx^{r,k})_k\ge(\bx^{r,o})_k$ for all $i< r,o\in[K],k\in[K]$. Hence
\[
n_k(T_{r-1})=\sum_{i=1}^{r-1}\gamma N_k^i\ge\gamma\sum_{i=1}^{r-1}\frac{t_i-t_{i-1}}{T}\frac{(\bx^{i,k})_k}{K}
\ge\gamma\sum_{i=1}^{r-1}\frac{t_i-t_{i-1}}{T}\frac{(\bx^{r,o})_k}{K}=\gamma\frac{t_{r-1}}{T}\frac{(\bx^{r,o})_k}{K}.
\]
Thus
\begin{align}\label{eqn:BlindNRM-UB-3}
&\sum_{k\in[K]}\radius_k(T_{r-1})\sum_{o\in[K]}\frac{(\bx^{r,o})_k}{K}\notag\\
=&\sum_{o\in[K]}\sum_{k\in[K]}\sqrt{\frac{\log [(d+1)KT]}{n_k(T_{r-1})}}\frac{(\bx^{r,o})_k}{K}\notag\\
\le&\sum_{o\in[K]}\sqrt{\frac{T\log [(d+1)KT]}{\gamma K  t_{r-1}}}\sum_{k\in[K]}\sqrt{{(\bx^{r,o})_k}}\notag\\
\le&\sum_{o\in[K]}\sqrt{\frac{T\log [(d+1)KT]}{\gamma Kt_{r-1}}}\sqrt{(d+1)T}\notag\\
=& T\sqrt{\frac{(d+1)K\log [(d+1)KT]}{\gamma t_{r-1}}},
\end{align}
where the last inequality is by the fact that at most $(d+1)$ components of $\bx^{r,o}$ are non-zero and by Jensen's inequality.
Combining (\ref{eqn:BlindNRM-UB-2}) and (\ref{eqn:BlindNRM-UB-3}), we have
\begin{align}\label{eqn:BlindNRM-UB-4}
&\sum_{k\in[K]}\sum_{j\in[n]}a_{ij}q_{j,k}n_{k}(T_l)\notag\\\le& \gamma\frac{t_l}{T}B_i+2||A_i||_2\sqrt{\gamma(d+1)K\log[(d+1)KT]}\sum_{r\in[l]}\frac{t_r-t_{r-1}}{\sqrt{t_{r-1}}}\notag\\
\le& \gamma\frac{t_l}{T}B_i+\left(16||A_i||_2 \sqrt{(d+1)\log [(d+1)KT]}\log T\right)t_1\notag\\\le& \gamma\frac{t_l}{T}B_i+\left(16\amax \sqrt{n(d+1)\log [(d+1)KT]}\log T\right)t_1,
\end{align}
where the second inequality follows from Lemma \ref{lem:points}.
Combining (\ref{eqn:BlindNRM-UB-1}) and (\ref{eqn:BlindNRM-UB-4}), we know that for all $i\in[d]$ and $l=\nu(s,d)+1$,
\[
\sum_{k\in[K]}\sum_{j\in[n]} a_{ij} \bar{q}_{j,k}(T_{\nu(s,d)+1})n_{k}(T_{\nu(s,d)+1})\le  \gamma\frac{B_i}{T}t_{\nu(s,d)+1}+\left(17\amax \sqrt{n(d+1)\log [(d+1)KT]}\log T\right)t_1\le B_i.
\]

The above inequality indicates that conditional on the clean event, policy $\phi^{\mathsf{v}}$ will not violate any resource constraint. By the coupling relationship between $\phi$ and $\phi^\mathsf{v}$, we know that $\tau=T_{\nu(s,d)+1}$ conditional on the clean execution of policy $\phi^{\mathsf{v}}$. Thus conditional on the clean event, the total revenue collected by policy $\phi$ is
\begin{align}\label{eqn:BlindNRM-UB-5}
    &\sum_{k\in[K]}\sum_{j\in[n]}p_{j,k}\bar{q}_{j,k}(T_{\nu(s,d)+1})n_k(T_{\nu(s,d)+1})\notag\\
    \ge&\sum_{k\in[K]}\sum_{j\in[n]}p_{j,k}{q}_{j,k}n_k(T_{\nu(s,d)+1})-\sum_{k\in[K]}||\bp_k||_2\radius_k(T_{\nu(s,d)+1})n_k(T_{\nu(s,d)+1})\notag\\
    \ge&\sum_{k\in[K]}\sum_{j\in[n]}p_{j,k}{q}_{j,k}n_k(T_{\nu(s,d)+1})-\pmax\sqrt{n\log[(d+1)KT]}\sqrt{KT_{\nu(s,d)+1}}.
\end{align}
We now bound $\sum_{k\in[K]}\sum_{j\in[n]}p_{j,k}{q}_{j,k}n_k(T_{\nu(s,d)+1})$ conditional on the clean event. We have
\begin{align}\label{eqn:BlindNRM-UB-6}
&\sum_{k\in[K]}\sum_{j\in[n]}p_{j,k}{q}_{j,k}n_k(T_{\nu(s,d)+1})\notag\\
=&\gamma\sum_{r\in[\nu(s,d)+1]}\frac{t_r-t_{r-1}}{T}\sum_{k\in[K]}\sum_{j\in[n]}p_{j,k}q_{j,k}\sum_{o\in[K]}\frac{(\bx^{r,o})_k}{K}\notag\\
\ge&\gamma\sum_{r\in[\nu(s,d)+1]}\frac{t_r-t_{r-1}}{T}\sum_{k\in[K]}\left(U_{k}^{\mathsf{rew}}(T_{r-1})-2||\bp_k||_2\radius_k(T_{r-1})\right)\sum_{o\in[K]}\frac{(\bx^{r,o})_k}{K}\notag\\
\geq&\gamma\sum_{r\in[\nu(s,d)+1]}\frac{t_r-t_{r-1}}{T}\sum_{o\in[K]}\frac{\sum_{k\in[K]}U_{k}^{\mathsf{rew}}(T_{r-1})(\bx^{r,o})_k}{K} \notag \\
& \qquad \qquad -2\gamma\pmax\sqrt{n}\sum_{r\in[\nu(s,d)+1]}\frac{t_r-t_{r-1}}{T}\sum_{k\in[K]}\radius_k(T_{r-1})\sum_{o\in[K]}\frac{(\bx^{r,o})_k}{K}\notag\\
\ge&\gamma\sum_{r\in[\nu(s,d)+1]}\frac{t_r-t_{r-1}}{T}\sum_{o\in[K]}\frac{\mathsf{J}^{\mathsf{PES}}_r}{K}-2\gamma\pmax\sqrt{n}\sum_{r\in[\nu(s,d)+1]}\frac{t_r-t_{r-1}}{T}\sum_{k\in[K]}\radius_k(T_{r-1})\sum_{o\in[K]}\frac{(\bx^{r,o})_k}{K}\notag\\
\ge&\gamma \DLP_{\bQ}-\gamma\sum_{r\in[\nu(s,d)+1]}\frac{t_r-t_{r-1}}{T}\left(\DLP_{\bQ}-\mathsf{J}^{\mathsf{PES}}_r\right)-\left(16\pmax \sqrt{n(d+1)\log [(d+1)KT]}\log T\right)t_1,
\end{align}
where the last inequality follows from (\ref{eqn:BlindNRM-UB-3}) and Lemma \ref{lem:points}, which imply
\begin{align*}
&2\gamma\pmax\sqrt{n}\sum_{r\in[l]}\frac{t_r-t_{r-1}}{T}\sum_{k\in[K]}\radius_k(T_{r-1})\sum_{o\in[K]}\frac{(\bx^{r,o})_k}{K}\\\le&2\pmax\sqrt{n}\sqrt{\gamma(d+1)K\log [(d+1)KT]}\sum_{r\in[l]}\frac{t_r-t_{r-1}}{ \sqrt{t_{r-1}}}\\
\le&\left(16\pmax\sqrt{n}\sqrt{\gamma(d+1)\log[(d+1)KT]}\log T\right)t_1.
\end{align*}
Let $\bx^*_{\bQ}$ denote an optimal solution to $\mathsf{DLP}_{\bQ}$, then \[\min_{i\in[d]}\frac{B_i}{\sum_{k\in[K]}U_{i,k}^{\mathsf{cost}}(T_{r-1})(\bx^*_{\bQ})_k}\cdot\bx^*_{\bQ}\] is a feasible solution to the first-stage pessimistic LP at epoch $r\in[\nu(s,d)+1]$. Thus
\begin{align}\label{eq:LP-bound}
\mathsf{J}^{\mathsf{PES}}_r\ge&\left(\min_{i\in[d]}\frac{B_i}{\sum_{k\in[K]}U_{i,k}^{\mathsf{cost}}(T_{r-1})(\bx^*_{\bQ})_k}\right)\sum_{k\in[K]}L^{\mathsf{rew}}_k(T_{l-1})  (\bx^*_{\bQ})_k\notag\\
\ge&\left(\min_{i\in[d]}\frac{B_i}{\sum_{k\in[K]}U_{i,k}^{\mathsf{cost}}(T_{r-1})(\bx^*_{\bQ})_k}\right)\sum_{k\in[K]}\left(\sum_{j\in[n]}p_{j,k}q_{j,k}-||\bp_k||_2\radius_k(T_{l-1})\right)  (\bx^*_{\bQ})_k\notag\\
\ge&\left(\min_{i\in[d]}\frac{B_i}{\sum_{k\in[K]}U_{i,k}^{\mathsf{cost}}(T_{r-1})(\bx^*_{\bQ})_k}\right)\DLP_{\bQ}-\pmax\sqrt{n}\sum_{k\in[K]}\radius_k(T_{r-1})(\bx^*_{\bQ})_k.
\end{align}
For all $r\in[\nu(s,d)+1]$, $i\in[d]$,
\begin{align}\label{eq:LP-bound-sp1}
\sum_{k\in[K]}U_{i,k}^{\mathsf{cost}}(T_{r-1})(\bx^*_{\bQ})_k\le&\sum_{k\in[K]}\left(\sum_{j\in[n]}a_{ij}q_{j,k}+2||A_i||_2\radius_k(T_{r-1})\right)(\bx^*_{\bQ})_k\notag\\
\le& B_i+2||A_i||_2\sum_{k\in[K]}\radius_k(T_{r-1})(\bx^*_{\bQ})_k
\end{align}
By (\ref{eqn:BlindNRM-UB-0}), we know that $\forall i\in[r-1]$, $\DLP_{\bQ}\ge\mathsf{J}^{\mathsf{PES}}_i$ and $\bx^*_{\bQ}$ is feasible to the second-stage exploration LP of epoch $i$, thus $(\bx^{i,k})_k\ge(\bx^*_{\bQ})_k$ for all $k\in[K]$.
Therefore, we have
\[
n_k(T_{r-1})=\sum_{i=1}^{r-1}\gamma N_k^i\ge\gamma\sum_{i=1}^{r-1}\frac{t_i-t_{i-1}}{T}\frac{(\bx^{i,k})_k}{K}
\ge\gamma\sum_{i=1}^{r-1}\frac{t_i-t_{i-1}}{T}\frac{(\bx^*_{\bQ})_k}{K}=\gamma\frac{t_{r-1}}{T}\frac{(\bx^{*}_{\bQ})_k}{K},
\]
Thus
\begin{align}\label{eq:LP-bound-sp2}
&\sum_{k\in[K]}\radius_k(T_{r-1}){(\bx^{*}_{\bQ})_k}\notag\\
=&\sum_{k\in[K]}\sqrt{\frac{\log [(d+1)KT]}{n_k(T_{r-1})}}{(\bx^{*}_{\bQ})_k}\notag\\
\le&\sqrt{\frac{KT\log [(d+1)KT]}{\gamma  t_{r-1}}}\sum_{k\in[K]}\sqrt{{{(\bx^{*}_{\bQ})_k}}}\notag\\
\le&\sqrt{\frac{KT\log [(d+1)KT]}{\gamma t_{r-1}}}\sqrt{(d+1)T}\notag\\
=& T\sqrt{\frac{(d+1)K\log [(d+1)KT]}{\gamma t_{r-1}}},
\end{align}
{where the last inequality follows from the fact that there are at most $(d+1)$ non-zero components of $\bx^*_{\bQ}$ and Jensen's inequality.}
Combining (\ref{eq:LP-bound}), (\ref{eq:LP-bound-sp1}) and (\ref{eq:LP-bound-sp2}), we have
\begin{align}\label{eq:LP-bound-f}
&\DLP_{\bQ}-\mathsf{J}_r^{\mathsf{PES}}\notag\\\le&
\frac{2\amax\sqrt{n(d+1)K\log[(d+1)KT]}\frac{T}{\sqrt{\gamma t_{r-1}}}}{\Bmin}\DLP_{\bQ}+\pmax\sqrt{n(d+1)K\log[(d+1)KT]}\frac{T}{\sqrt{\gamma t_{r-1}}}\notag\\
\le&\sqrt{n(d+1)K\log[(d+1)KT]}\left(2\amax\frac{\DLP_{\bQ}}{\Bmin}+\pmax\right)\frac{T}{\sqrt{\gamma t_{r-1}}}
\end{align}
for all $r\in[\nu(s,d)+1]$. 
Combining (\ref{eq:LP-bound-f}) and (\ref{eqn:BlindNRM-UB-6}), we have
\begin{align}\label{eqn:BlindNRM-UB-7}
&\sum_{k\in[K]}\sum_{j\in[n]}p_{j,k}{q}_{j,k}n_k(T_{\nu(s,d)+1})\notag\\
\ge&\gamma \DLP_{\bQ}-\gamma\sum_{r\in[\nu(s,d)+1]}\frac{t_r-t_{r-1}}{T}\left(\DLP_{\bQ}-\mathsf{J}^{\mathsf{PES}}_r\right)-\left(16\pmax \sqrt{n(d+1)\log [(d+1)KT]}\log T\right)t_1\notag\\
\ge&\gamma \DLP_{\bQ}-\left(\frac{2\amax\DLP_{\bQ}}{\Bmin}+\pmax\right)\sqrt{n(d+1)K\log[(d+1)KT]}\sum_{r\in[\nu(s,d)+1]}\frac{t_r-t_{r-1}}{\sqrt{t_{r-1}}} \notag \\
& \qquad \qquad \qquad \qquad \qquad \qquad \qquad \qquad \qquad \qquad \qquad -\left(16\pmax \sqrt{n(d+1)\log [(d+1)KT]}\log T\right)t_1\notag\\
\ge&\gamma \DLP_{\bQ}-\left(\left(\frac{16\amax\DLP_{\bQ}}{\Bmin}+24\pmax\right) \sqrt{n(d+1)\log [(d+1)KT]}\log T\right)t_1,
\end{align}
where the last inequality follows from Lemma \ref{lem:points}. 
Combining (\ref{eqn:BlindNRM-UB-5}) and (\ref{eqn:BlindNRM-UB-7}), we know that the total revenue collected by the policy $\phi$ is lower bounded by
\[
\gamma \DLP_{\bQ}-\left(\frac{16\amax\DLP_{\bQ}}{\Bmin}+25\pmax\right) \sqrt{n(d+1)\log [(d+1)KT]}\log T\cdot t_1.
\]
Thus both $R^\phi_s(T)$ and $R^\phi(T)$ are upper bounded by
\begin{align*}
    &(1-\gamma)\DLP_{\bQ}+\left(\frac{16\amax\DLP_{\bQ}}{\Bmin}+25\pmax\right) \sqrt{n(d+1)\log [(d+1)KT]}\log T\cdot t_1\\
    =&\left(\frac{33\DLP_{\bQ}}{\Bmin}\amax+25\pmax\right)\sqrt{n(d+1)\log[(d+1)KT]}\log T\cdot t_1\\
    \le&(33n\pmax T\amax/\Bmin+25\pmax)\sqrt{n(d+1)\log[(d+1)KT]}\log T\cdot t_1\\
    \le&(33\pmax\amax/\underline{b}+25\pmax)n\sqrt{n(d+1)\log[(d+1)KT]}\log T\cdot K^{1-\frac{1}{2-2^{-\nu(s,d)}}}T^{\frac{1}{2-2^{-\nu(s,d)}}}.
\end{align*}
\hfill\Halmos \endproof

\proof{Proof of Proposition~\ref{prop:BwKUB}.} 
Consider DLP-G as defined by \eqref{eqn:obj-G}, \eqref{eqn:constraint:inventory-G}, \eqref{eqn:constraint:time-G}, and \eqref{eqn:constraint:NonNeg-G}.
Let $\mathsf{DLP}_{\bmR, \bC}$ denote the DLP-G with the underlying distributions being $\bmR, \bC$. Let $\DLP_{\bmR, \bC}$ denote the optimal objective value of $\mathsf{DLP}_{\bmR, \bC}$.

Let $\tilde{l}$ be the last epoch in the execution of policy $\phi$, and let $\tau$ be the last period before the policy $\phi$ stops. We know that $\tau+1$ is a stopping time and we have $T_{\tilde{l}-1}< \tau\le t_{\tilde{l}}\le T$. Since $\phi$ makes at most $(K-1)(\tilde{l}-1)$ switches before $T_{\tilde{l}-1}$ and makes at most $d+1$ switches after $T_{\tilde{l}-1}$, its total number of switches is always upper bounded by $(K-1)\nu(s,d)+(d+1)\le s$.

We use a coupling argument for the regret analysis. Consider a virtual policy $\phi^{\mathsf{v}}$ that runs under exactly the same demand realization process and acts exactly the same as $\phi$ until period $\tau$, but keeps running until the end of epoch $\nu(s,d)+1$ regardless of the resource constraints. 
Without conflicts to the previously defined notation in Algorithm \ref{alg:bsse2}, for each sample path of the action and demand realization process under the execution of policy $\phi^{\mathsf{v}}$, let $T_l$ denote the last period of epoch $l$ under policy $\phi^{\mathsf{v}}$ ($l\in[\nu(s,d)+1]$), let $n_{k}(t)$ be the total number of periods that arm $k$ is chosen by $\phi^\mathsf{v}$ during period 1 to $t$ ($k\in[K], t\in[T_{\nu(s,d)+1}]$); on this sample path, let $\bar{c}_{i,k}(t)$ be the average realized consumption of resource $i$ when arm $k$ is pulled, during period 1 to $t$, and let $\bar{r}_k$ be the average realized rewards generated when arm $k$ is pulled, during period 1 to $t$ ($k\in[K], t\in[T_{\nu(s,d)+1}]$). 
For all $t\in[T_{\nu(s,d)+1}]$, define the confidence radius as
\[
\radius_k(t)=\sqrt{\frac{\log [(d+1)KT]}{n_k(t)}},~~\forall k\in[K].
\]
For notational simplicity, define
\[
\begin{cases}U^{\mathsf{rew}}_{k}(T_{l})=\min\left\{\bar{r}_{k}(T_{l})+\rmax\radius_k(T_{l}),U^{\mathsf{rew}}_k(T_{l-1})\right\}
,\\L^{\mathsf{rew}}_{k}(T_{l})=\max\left\{\bar{r}_{k}(T_{l})-\rmax\radius_k(T_{l}),L_k^{\mathsf{rew}}(T_{l-1})\right\},\end{cases}~~\forall k\in[K],
\]
\[
\begin{cases}U^{\mathsf{cost}}_{i,k}(T_{l})=\min\left\{\bar{c}_{i,k}{(T_{l})}+\cmax\radius_k(T_{l}),U^{\mathsf{cost}}_{i,k}(T_{l-1})\right\}
,\\L^{\mathsf{cost}}_{i,k}(T_{l})=\max\left\{\bar{c}_{i,k}{(T_{l})}-\cmax\radius_k(T_{l}),L_{i,k}^{\mathsf{cost}}(T_{l-1})\right\},\end{cases}~~\forall i\in[d],\forall k\in[K].
\]
for $l=\nu(s,d)$.

Define the \textit{clean event} $\mathcal{E}$ as
$$\left\{\forall i\in[d], k\in[K],  t\in[T_{\nu(s,d)+1}],~\begin{cases}\left|\bar{r}_{k}(t)-r_k\right|\le\rmax\radius_k(t),\\\left|\bar{c}_{i,k}(t)-c_{i,k}\right|\le \cmax\radius_k(t).
\end{cases} 
 \right\}.$$
By the Hoeffding's inequality for general bounded random variables (see Theorem A.1 in \cite{slivkins2019introduction}) and a standard union bound argument (see Chapter 1.3.1 in \cite{slivkins2019introduction}), we have $\text{Pr}(\mathcal{E})\ge 1-\frac{2}{(d+1)KT}$ under distributions $\bQ$ and policy $\phi^{\mathsf{v}}$. Since the clean event happens with very high probability, we can just focus on a \textit{clean execution} of policy $\phi^{\mathsf{v}}$: an execution in which the clean event holds. 
Conditional on the clean event,  it holds that $\forall i\in[d], k\in[K],  l\in[\nu(s,d)]$,
\begin{align}\label{eqn:BlindNRM-UB-0-G}
\begin{cases}
    U_{k}^{\mathsf{rew}}(T_l)-2\rmax\radius_k(t)\le L^{\mathsf{rew}}_{k}(T_l)\le r_{k}\le U_{k}^{\mathsf{rew}}(T_l)\le L^{\mathsf{rew}}_{k}(T_l)+2\rmax\radius_k(t),\\
    U_{i,k}^{\mathsf{cost}}(T_l)-2\cmax\radius_k(t)\le L^{\mathsf{cost}}_{i,k}(T_l)\le c_{i,k}\le U_{i,k}^{\mathsf{cost}}(T_l)\le L^{\mathsf{cost}}_{i,k}(T_l)+2\cmax\radius_k(t).
\end{cases}
\end{align}
In the rest of the proof, we always assume that $\mathcal{E}$ holds.

For all $i\in[d]$ and  $l\in[\nu(s,d)+1]$, we have
\begin{align}\label{eqn:BlindNRM-UB-1-G}
&\sum_{k\in[K]} \bar{c}_{i,k}(T_{l})n_{k}(T_{l})\notag\\
\le &\sum_{k\in[K]} c_{i,k}n_k(T_{l})+\sum_{k\in[K]}\cmax\radius_k(T_l)n_k(T_{l})\notag\\
=&\sum_{k\in[K]} c_{i,k}n_k(T_{l})+\cmax\sqrt{\log[(d+1)KT]}\sum_{k\in[K]}\sqrt{n_k(T_{l})}\notag\\
\le&\sum_{k\in[K]} c_{i,k}n_k(T_{l})+\cmax\sqrt{\log[(d+1)KT]}\sqrt{K{T_{l}}},
\end{align}
where the last inequality is by Jensen's inequality. Similarly, for all $i\in[d]$ and  $l\in[\nu(s,d)+1]$, we also have
\begin{align*}
    \sum_{k\in[K]} \bar{c}_{i,k}(T_{l})n_{k}(T_{l})\ge\sum_{k\in[K]} c_{i,k}n_k(T_{l})-\cmax\sqrt{\log[(d+1)KT]}\sqrt{K{T_{l}}}.
\end{align*}

For notational convenience, define  $\bx^{\nu(s,d)+1,i}=\bx^*_{\bmR, \bC}$ for all $i\in[K]$. Then we have $n_{k}(T_{l})=\sum_{r=1}^{l} \gamma N_k^r=\gamma\sum_{r=1}^l\left(\frac{t_{r}-t_{r-1}}{T}\sum_{o=1}^{K}\frac{(\bx^{r,o})_k}{K}\right)$ for all $l\in[\nu(s,d)+1]$. Then for all $i\in[d]$ and  $l\in[\nu(s,d)+1]$, we have
\begin{align}\label{eqn:BlindNRM-UB-2-G}
&\sum_{k\in[K]} c_{i,k}n_{k}(T_l)\notag\\
=&\gamma\sum_{r\in[l]}\frac{t_r-t_{r-1}}{T}\sum_{k\in[K]} c_{i,k}\sum_{o\in[K]}\frac{(\bx^{r,o})_k}{K}\notag\\
\le&\gamma\sum_{r\in[l]}\frac{t_r-t_{r-1}}{T}\sum_{k\in[K]}\left(L_{i,k}^{\mathsf{cost}}(T_{l-1})+2\cmax\radius_k(T_{l-1})\right)\sum_{o\in[K]}\frac{(\bx^{r,o})_k}{K}\notag\\
=&\gamma\sum_{r\in[l]}\frac{t_r-t_{r-1}}{T}\sum_{o\in[K]}\frac{\sum_{k\in[K]}L_{i,k}^{\mathsf{cost}}(T_{l-1})(\bx^{r,o})_k}{K}+2\gamma\cmax\sum_{r\in[l]}\frac{t_r-t_{r-1}}{T}\sum_{k\in[K]}\radius_k(T_{r-1})\sum_{o\in[K]}\frac{(\bx^{r,o})_k}{K}\notag\\
\le&\gamma\sum_{r\in[l]}\frac{t_r-t_{r-1}}{T}\sum_{o\in[K]}\frac{B_i}{K}+2\gamma\cmax\sum_{r\in[l]}\frac{t_r-t_{r-1}}{T}\sum_{k\in[K]}\radius_k(T_{r-1})\sum_{o\in[K]}\frac{(\bx^{r,o})_k}{K}\notag\\
=&\gamma \frac{t_l}{T}B_i+2\gamma\cmax\sum_{r\in[l]}\frac{t_r-t_{r-1}}{T}\sum_{k\in[K]}\radius_k(T_{r-1})\sum_{o\in[K]}\frac{(\bx^{r,o})_k}{K},
\end{align}
where the first inequality is by (\ref{eqn:BlindNRM-UB-0-G}), and the second inequality is by the  feasibility of $\bx^{r,o}$ to the second-stage exploration LP of epoch $r$. We now bound $\sum_{k\in[K]}\radius_k(T_{r-1})(\bx^{r,o})_k$. Since $[L^{\mathsf{rew}}_{k}(T_{r-1}),U_{k}^{\mathsf{rew}}(T_{r-1})]\subset [L^{\mathsf{rew}}_{k}(T_{r-2}),U_{k}^{\mathsf{rew}}(T_{r-2})]\subset\cdots\subset [L^{\mathsf{rew}}_{k}(T_{0}),U_{k}^{\mathsf{rew}}(0)]$ and $[L^{\mathsf{cost}}_{i,k}(T_{r-1}),U_{i,k}^{\mathsf{cost}}(T_{r-1})]\subset [L^{\mathsf{cost}}_{i,k}(T_{r-2}),U_{i,k}^{\mathsf{cost}}(T_{r-2})]\subset\cdots\subset [L^{\mathsf{cost}}_{i,k}(T_{0}),U_{i,k}^{\mathsf{cost}}(0)]$, for the first-stage pessimistic LP in Algorithm \ref{alg:bsse2}, $\mathsf{J}_l^{\mathsf{PES}}$ is non-decreasing as $l$ grows, which in turns makes the constraints in the second-stage exploration LPs more stringent as $l$ grows, leading to non-increasing $\bx^{l,j}$ as $l$ grows $(\forall j\in[K])$. This ensures that $(\bx^{i,k})_k\ge(\bx^{r,k})_k\ge(\bx^{r,o})_k$ for all $i< r,o\in[K],k\in[K]$. Hence
\[
n_k(T_{r-1})=\sum_{i=1}^{r-1}\gamma N_k^i\ge\gamma\sum_{i=1}^{r-1}\frac{t_i-t_{i-1}}{T}\frac{(\bx^{i,k})_k}{K}
\ge\gamma\sum_{i=1}^{r-1}\frac{t_i-t_{i-1}}{T}\frac{(\bx^{r,o})_k}{K}=\gamma\frac{t_{r-1}}{T}\frac{(\bx^{r,o})_k}{K}.
\]
Thus
\begin{align}\label{eqn:BlindNRM-UB-3-G}
&\sum_{k\in[K]}\radius_k(T_{r-1})\sum_{o\in[K]}\frac{(\bx^{r,o})_k}{K}\notag\\
=&\sum_{o\in[K]}\sum_{k\in[K]}\sqrt{\frac{\log [(d+1)KT]}{n_k(T_{r-1})}}\frac{(\bx^{r,o})_k}{K}\notag\\
\le&\sum_{o\in[K]}\sqrt{\frac{T\log [(d+1)KT]}{\gamma K  t_{r-1}}}\sum_{k\in[K]}\sqrt{{(\bx^{r,o})_k}}\notag\\
\le&\sum_{o\in[K]}\sqrt{\frac{T\log [(d+1)KT]}{\gamma Kt_{r-1}}}\sqrt{(d+1)T}\notag\\
=& T\sqrt{\frac{(d+1)K\log [(d+1)KT]}{\gamma t_{r-1}}},
\end{align}
where the last inequality is by the fact that at most $(d+1)$ components of $\bx^{r,o}$ are non-zero and by Jensen's inequality.
Combining (\ref{eqn:BlindNRM-UB-2-G}) and (\ref{eqn:BlindNRM-UB-3-G}), we have
\begin{align}\label{eqn:BlindNRM-UB-4-G}
&\sum_{k\in[K]} c_{i,k}n_{k}(T_l)\notag\\\le& \gamma\frac{t_l}{T}B_i+2\cmax\sqrt{\gamma(d+1)K\log[(d+1)KT]}\sum_{r\in[l]}\frac{t_r-t_{r-1}}{\sqrt{t_{r-1}}}\notag\\
\le& \gamma\frac{t_l}{T}B_i+\left(16\cmax \sqrt{(d+1)\log [(d+1)KT]}\log T\right)t_1,
\end{align}
where the last inequality follows from Lemma \ref{lem:points}.
Combining (\ref{eqn:BlindNRM-UB-1-G}) and (\ref{eqn:BlindNRM-UB-4-G}), we know that for all $i\in[d]$ and $l=\nu(s,d)+1$,
\[
\sum_{k\in[K]} \bar{c}_{i,k}(T_{\nu(s,d)+1})n_{k}(T_{\nu(s,d)+1})\le  \gamma\frac{B_i}{T}t_{\nu(s,d)+1}+\left(17\cmax \sqrt{(d+1)\log [(d+1)KT]}\log T\right)t_1\le B_i.
\]

The above inequality indicates that conditional on the clean event, policy $\phi^{\mathsf{v}}$ will not violate any resource constraint. By the coupling relationship between $\phi$ and $\phi^\mathsf{v}$, we know that $\tau=T_{\nu(s,d)+1}$ conditional on the clean execution of policy $\phi^{\mathsf{v}}$. Thus conditional on the clean event, the total revenue collected by policy $\phi$ is
\begin{align}\label{eqn:BlindNRM-UB-5-G}
    &\sum_{k\in[K]}\bar{r}_{k}(T_{\nu(s,d)+1})n_k(T_{\nu(s,d)+1})\notag\\
    \ge&\sum_{k\in[K]}r_kn_k(T_{\nu(s,d)+1})-\sum_{k\in[K]}\rmax\radius_k(T_{\nu(s,d)+1})n_k(T_{\nu(s,d)+1})\notag\\
    \ge&\sum_{k\in[K]}r_kn_k(T_{\nu(s,d)+1})-\rmax\sqrt{\log[(d+1)KT]}\sqrt{KT_{\nu(s,d)+1}}.
\end{align}
We now bound $\sum_{k\in[K]}r_kn_k(T_{\nu(s,d)+1})$ conditional on the clean event. We have
\begin{align}\label{eqn:BlindNRM-UB-6-G}
&\sum_{k\in[K]}r_kn_k(T_{\nu(s,d)+1})\notag\\
=&\gamma\sum_{r\in[\nu(s,d)+1]}\frac{t_r-t_{r-1}}{T}\sum_{k\in[K]} r_{k}\sum_{o\in[K]}\frac{(\bx^{r,o})_k}{K}\notag\\
\ge&\gamma\sum_{r\in[\nu(s,d)+1]}\frac{t_r-t_{r-1}}{T}\sum_{k\in[K]}\left(U_{k}^{\mathsf{rew}}(T_{r-1})-2\rmax\radius_k(T_{r-1})\right)\sum_{o\in[K]}\frac{(\bx^{r,o})_k}{K}\notag\\
=&\gamma\sum_{r\in[\nu(s,d)+1]}\frac{t_r-t_{r-1}}{T}\sum_{o\in[K]}\frac{\sum_{k\in[K]}U_{k}^{\mathsf{rew}}(T_{r-1})(\bx^{r,o})_k}{K} \notag \\
& \qquad \qquad -2\gamma\rmax \sum_{r\in[\nu(s,d)+1]}\frac{t_r-t_{r-1}}{T}\sum_{k\in[K]}\radius_k(T_{r-1})\sum_{o\in[K]}\frac{(\bx^{r,o})_k}{K}\notag\\
\ge&\gamma\sum_{r\in[\nu(s,d)+1]}\frac{t_r-t_{r-1}}{T}\sum_{o\in[K]}\frac{\mathsf{J}^{\mathsf{PES}}_r}{K}-2\gamma\rmax \sum_{r\in[\nu(s,d)+1]}\frac{t_r-t_{r-1}}{T}\sum_{k\in[K]}\radius_k(T_{r-1})\sum_{o\in[K]}\frac{(\bx^{r,o})_k}{K}\notag\\
\ge&\gamma \DLP_{\bmR, \bC}-\gamma\sum_{r\in[\nu(s,d)+1]}\frac{t_r-t_{r-1}}{T}\left(\DLP_{\bmR, \bC}-\mathsf{J}^{\mathsf{PES}}_r\right)-\left(16 \rmax \sqrt{(d+1)\log [(d+1)KT]}\log T\right)t_1,
\end{align}
where the last inequality follows from (\ref{eqn:BlindNRM-UB-3-G}) and Lemma \ref{lem:points}, which imply
\begin{align*}
&2\gamma\rmax\sum_{r\in[l]}\frac{t_r-t_{r-1}}{T}\sum_{k\in[K]}\radius_k(T_{r-1})\sum_{o\in[K]}\frac{(\bx^{r,o})_k}{K}\\\le&2\rmax\sqrt{\gamma(d+1)K\log [(d+1)KT]}\sum_{r\in[l]}\frac{t_r-t_{r-1}}{ \sqrt{t_{r-1}}}\\
\le&\left(16\rmax\sqrt{\gamma(d+1)\log[(d+1)KT]}\log T\right)t_1.
\end{align*}
Let $\bx^*_{\bmR, \bC}$ denote an optimal solution to $\mathsf{DLP}_{\bmR, \bC}$, then \[\min_{i\in[d]}\frac{B_i}{\sum_{k\in[K]}U_{i,k}^{\mathsf{cost}}(T_{r-1})(\bx^*_{\bmR, \bC})_k}\cdot\bx^*_{\bmR, \bC}\] is a feasible solution to the first-stage pessimistic LP at epoch $r\in[\nu(s,d)+1]$. Thus
\begin{align}\label{eq:LP-bound-G}
\mathsf{J}^{\mathsf{PES}}_r\ge&\left(\min_{i\in[d]}\frac{B_i}{\sum_{k\in[K]}U_{i,k}^{\mathsf{cost}}(T_{r-1})(\bx^*_{\bmR, \bC})_k}\right)\sum_{k\in[K]}L^{\mathsf{rew}}_k(T_{l-1})  (\bx^*_{\bmR, \bC})_k\notag\\
\ge&\left(\min_{i\in[d]}\frac{B_i}{\sum_{k\in[K]}U_{i,k}^{\mathsf{cost}}(T_{r-1})(\bx^*_{\bmR, \bC})_k}\right)\sum_{k\in[K]}\left( r_{k}-\rmax\radius_k(T_{l-1})\right)  (\bx^*_{\bmR, \bC})_k\notag\\
\ge&\left(\min_{i\in[d]}\frac{B_i}{\sum_{k\in[K]}U_{i,k}^{\mathsf{cost}}(T_{r-1})(\bx^*_{\bmR, \bC})_k}\right)\DLP_{\bmR, \bC}-\rmax \sum_{k\in[K]}\radius_k(T_{r-1})(\bx^*_{\bmR, \bC})_k.
\end{align}
For all $r\in[\nu(s,d)+1]$, $i\in[d]$,
\begin{align}\label{eq:LP-bound-sp1-G}
\sum_{k\in[K]}U_{i,k}^{\mathsf{cost}}(T_{r-1})(\bx^*_{\bmR, \bC})_k\le&\sum_{k\in[K]}\left( c_{i,k}+2\cmax\radius_k(T_{r-1})\right)(\bx^*_{\bmR, \bC})_k\notag\\
\le& B_i+2\cmax\sum_{k\in[K]}\radius_k(T_{r-1})(\bx^*_{\bmR, \bC})_k
\end{align}
By (\ref{eqn:BlindNRM-UB-0-G}), we know that $\forall i\in[r-1]$, $\DLP_{\bmR, \bC}\ge\mathsf{J}^{\mathsf{PES}}_i$ and $\bx^*_{\bmR, \bC}$ is feasible to the second-stage exploration LP of epoch $i$, thus $(\bx^{i,k})_k\ge(\bx^*_{\bmR, \bC})_k$ for all $k\in[K]$.
Therefore, we have
\[
n_k(T_{r-1})=\sum_{i=1}^{r-1}\gamma N_k^i\ge\gamma\sum_{i=1}^{r-1}\frac{t_i-t_{i-1}}{T}\frac{(\bx^{i,k})_k}{K}
\ge\gamma\sum_{i=1}^{r-1}\frac{t_i-t_{i-1}}{T}\frac{(\bx^*_{\bmR, \bC})_k}{K}=\gamma\frac{t_{r-1}}{T}\frac{(\bx^{*}_{\bmR, \bC})_k}{K},
\]
Thus
\begin{align}\label{eq:LP-bound-sp2-G}
&\sum_{k\in[K]}\radius_k(T_{r-1}){(\bx^{*}_{\bmR, \bC})_k}\notag\\
=&\sum_{k\in[K]}\sqrt{\frac{\log [(d+1)KT]}{n_k(T_{r-1})}}{(\bx^{*}_{\bmR, \bC})_k}\notag\\
\le&\sqrt{\frac{KT\log [(d+1)KT]}{\gamma  t_{r-1}}}\sum_{k\in[K]}\sqrt{{{(\bx^{*}_{\bmR, \bC})_k}}}\notag\\
\le&\sqrt{\frac{KT\log [(d+1)KT]}{\gamma t_{r-1}}}\sqrt{(d+1)T}\notag\\
=& T\sqrt{\frac{(d+1)K\log [(d+1)KT]}{\gamma t_{r-1}}},
\end{align}
{where the last inequality follows from the fact that there are at most $(d+1)$ non-zero components of $\bx^*_{\bmR, \bC}$ and Jensen's inequality.}
Combining (\ref{eq:LP-bound-G}), (\ref{eq:LP-bound-sp1-G}) and (\ref{eq:LP-bound-sp2-G}), we have
\begin{align}\label{eq:LP-bound-f-G}
&\DLP_{\bmR, \bC}-\mathsf{J}_r^{\mathsf{PES}}\notag\\\le&
\frac{2\cmax\sqrt{(d+1)K\log[(d+1)KT]}\frac{T}{\sqrt{\gamma t_{r-1}}}}{\Bmin}\DLP_{\bmR, \bC}+\rmax\sqrt{(d+1)K\log[(d+1)KT]}\frac{T}{\sqrt{\gamma t_{r-1}}}\notag\\
\le&\sqrt{(d+1)K\log[(d+1)KT]}\left(2\cmax\frac{\DLP_{\bmR, \bC}}{\Bmin}+\rmax\right)\frac{T}{\sqrt{\gamma t_{r-1}}}
\end{align}
for all $r\in[\nu(s,d)+1]$. Combining (\ref{eq:LP-bound-f-G}) and (\ref{eqn:BlindNRM-UB-6-G}), we have
\begin{align}\label{eqn:BlindNRM-UB-7-G}
&\sum_{k\in[K]}r_kn_k(T_{\nu(s,d)+1})\notag\\
\ge&\gamma \DLP_{\bmR, \bC}-\gamma\sum_{r\in[\nu(s,d)+1]}\frac{t_r-t_{r-1}}{T}\left(\DLP_{\bmR, \bC}-\mathsf{J}^{\mathsf{PES}}_r\right)-\left(16 \rmax \sqrt{(d+1)\log [(d+1)KT]}\log T\right)t_1\notag\\
\ge&\gamma \DLP_{\bmR, \bC}-\left(\frac{2\cmax\DLP_{\bmR, \bC}}{\Bmin}+\rmax\right)\sqrt{(d+1)K\log[(d+1)KT]}\sum_{r\in[\nu(s,d)+1]}\frac{t_r-t_{r-1}}{\sqrt{t_{r-1}}} \notag \\
& \qquad \qquad \qquad \qquad \qquad \qquad \qquad \qquad \qquad \qquad  - \left(16 \rmax \sqrt{(d+1)\log [(d+1)KT]}\log T\right)t_1\notag\\
\ge&\gamma \DLP_{\bmR, \bC}-\left(\left(\frac{16\cmax\DLP_{\bmR, \bC}}{\Bmin}+24\rmax\right) \sqrt{(d+1)\log [(d+1)KT]}\log T\right)t_1,
\end{align}
where the last inequality follows from Lemma \ref{lem:points}. 
Combining (\ref{eqn:BlindNRM-UB-5-G}) and (\ref{eqn:BlindNRM-UB-7-G}), we know that the total revenue collected by the policy $\phi$ is lower bounded by
\[
\gamma \DLP_{\bmR, \bC}-\left(\frac{16\cmax\DLP_{\bmR, \bC}}{\Bmin}+25\rmax\right) \sqrt{(d+1)\log [(d+1)KT]}\log T\cdot t_1.
\]
Thus both $R^\phi_s(T)$ and $R^\phi(T)$ are upper bounded by
\begin{align*}
    &(1-\gamma)\DLP_{\bmR, \bC}+\left(\frac{16\cmax\DLP_{\bmR, \bC}}{\Bmin}+25\rmax\right) \sqrt{(d+1)\log [(d+1)KT]}\log T\cdot t_1\\
    =&\left(\frac{33\DLP_{\bmR, \bC}}{\Bmin}\cmax+25\rmax\right)\sqrt{(d+1)\log[(d+1)KT]}\log T\cdot t_1\\
    \le&(33\rmax T\cmax /\Bmin+25\rmax)\sqrt{(d+1)\log[(d+1)KT]}\log T\cdot t_1\\
    \le&(33\rmax\cmax /\underline{b}+25\rmax)\sqrt{(d+1)\log[(d+1)KT]}\log T\cdot K^{1-\frac{1}{2-2^{-\nu(s,d)}}}T^{\frac{1}{2-2^{-\nu(s,d)}}}.
\end{align*}


\hfill\Halmos \endproof

\section{Proof of Theorem~\ref{thm:BlindNRMLB}}\label{app:BlindNRMLB}
\subsection{Preliminaries}

Without loss of generality, we assume that $T\ge 2K$. 
For any $n_1,n_2\in[T]$, let $[n_1:n_2]$ denote the set $\{n_1,n_1+1,\dots,n_2\}$. For any random variable $X$, let $\mathbb{P}_X$ denote the probability measure induced by the random variable $X$.

We construct a \bnrm problem $\mathcal{P}$ as follows. 
For any problem input $\bm{B} \in [\underline{b}T,T]^d$, define $b_1=B_1/T,\dots,b_d=B_d/T$. We have $b_i\in[\underline{b},1],\,\forall i\in[d]$. 
Let there be $n\ge d+1$ products. 
Let the $d\times n$ consumption matrix be
\begin{align*}
A= 2 \cdot
\bigl[
\bm{0}_{d\times1} \quad \text{diag}(b_1,\dots,b_d) \quad \bm{0}_{d\times (n-(d+1))}
\bigr],
\end{align*}
where $\text{diag}(b_1,\dots,b_d)$ stands for the $d\times d$ diagonal matrix whose diagonal entries are $b_1,\dots,b_d$. For any $j\in[n]$, $k\in[K]$, let the price be
\begin{align*}
p_{j,k}=\begin{cases}1,&\text{if } j=1,\\
0, & \text{otherwise.}
\end{cases}
\end{align*}
Based on the above \bnrm problem $\mathcal{P}$, we will construct different \bnrm instances by specifying different demand distributions $\bQ$.

We prove Theorem~\ref{thm:BlindNRMLB} even when we restrict $\bQ$ to Bernoulli demand distributions. 
Recall that we use $q_{j,k} = \bE[Q_{j,k}]$ to stand for the mean value of the distribution $Q_{j,k}$. 
When restricted to Bernoulli distributions, such a $q_{j,k}$ uniquely describes the distribution of $Q_{j,k}$. 
Thus every  $\bq \in[0,1]^{n\times K}$ uniquely determines a \bnrm instance  $\cI_{\bq}:=(T,\bm{B},K,d,n,P,A,s;\bq)$. 
Specifically, we parameterize $\bq$ by two vectors $\bmu_1\in[-\frac{1}{2},\frac{1}{2}]^{K}$ and $\bmu_2\in[-\frac{1}{2},\frac{1}{2}]^{K}$, such that for all $k\in[K]$ and $j\in[n]$,
\begin{align*}
q_{j,k;\bmu}=\begin{cases}\frac{1}{2}+\mu_{1,k},&\text{if }j=1,\\
\frac{1}{2}-\mu_{2,k},&\text{else if }j=(k-1)\%(d+1)+1,\\
\frac{1}{2}+\mu_{2,k},&\text{else if }j=k\%(d+1)+1,\\
\frac{1}{2}, &\text{else if }j\in[2,d+1],\\
0, &\text{else},
\end{cases}
\end{align*}
where $\%$ stands for the modulo operation. 
In the lower bound proof, we will assign different values to $\bmu$ to construct different \bnrm instances. 
Below we will use $\cI_{\bmu}:=(T,\bm{B},K,d,n,P,A,s;\bmu)$ to stand for a \bnrm instance, which highlights the dependence on $\bmu$. 
Let $\mathsf{DLP}_{\bmu}$ denote the DLP as defined by \eqref{eqn:obj}, \eqref{eqn:constraint:inventory}, \eqref{eqn:constraint:time}, \eqref{eqn:constraint:NonNeg}, on the  instance $\cI_{\bmu}$:
\begin{align*}
\mathsf{J}^{\mathsf{DLP}_{\bmu}} =\max_{\bx}  & ~\sum_{k\in[K]} \left(\frac{1}{2}+\mu_{1,k}\right)x_k \\
\text{s.t.} ~ & ~{b_i}\left(\sum_{k\in[K]} \frac{1}{2} x_k + \sum_{k': k'\%(d+1)=i}\mu_{2,k'}x_{k'} - \sum_{k'': k''\%(d+1)=i+1}\mu_{2,k''}x_{k''}\right)\leq \frac{b_i T}{2}, ~~~~ \forall\ i \in [d], \\
&~\sum_{k\in[K]}x_k\le T,\\
&~x_k  \geq 0,~~~~ \forall k\in[K].
\end{align*}

For any instance $\mathcal{I}_{\bmu}$, let $Q_{j,k;\bmu}^{t} \sim \mathsf{Ber}(q_{j,k;\bmu})$ denote the i.i.d. random demand of product $j$ under price  $p_{j,k}$ at round $t$ ($j\in[n], k\in[K],t\in[T]$). 
For any $k\in[K], t\in[T]$, let $X_{\bmu}^t(k)=(Q_{j,k;\bmu}^t)_{j\in[n]}$ denote the $n$-dimensional random vector whose $j^{\rm th}$ component is the random demand of product $j\in[n]$ under action $k$ at round $t$. For any $k\in[K]$ and $n_1,n_2\in[T]$, let $(X_{\bmu}^t(k))_{t\in[n_1:n_2]}$ denote the $n\times(n_2-n_1+1)$-dimensional random matrix which consists of random  vectors $X_{\bmu}^t(k)$ from round $n_1$ to round $n_2$. 

\subsection{Inevitable Revenue Loss of Any Policy Due to Imbalanced Use of Actions}
We first study the properties of $\mathcal{I}_{\bmu}$ in the distributiuonally-known setup. 
Specifically, we focus on the case when $K \geq d+1$. 
For any $\bmu$, we consider a stochastic packing (\textsf{SP}) problem instance $\mathcal{I}=(T,\bm{B}, K,d;\bm{R},\bC)$ which is equivalent to $\mathcal{I}_{\bmu}$,
specified as follows.
\begin{itemize}
\item $B_i=\frac{b_i}{2} T,~\forall i=1,\dots,n$
\item $\bC$ and $\bm{R}$ consist of Bernoulli distributions
\item $\forall k\in[K]$, $r_k=\frac{1}{2}+\mu_{1,k}$
\item $\forall i\in[d],\forall k\in[K]$, $c_{i,k}=\begin{cases}
    \frac{b_i}{2}+b_i\mu_{2,k}, &\text{if }k \% (d+1) = i, \\
    \frac{b_i}{2}-b_i\mu_{2,k}, &\text{if }k \% (d+1) = i+1, \\
    \frac{b_i}{2}, &\text{otherwise}.
\end{cases}$
\vspace{2mm}
\end{itemize}
On this problem instance, we will show that even when
\begin{itemize}
    \item $(\mu_{i,k})_{i\in[2],k\in[K]}$ are completely known,
    \item $(\mu_{i,k})_{i\in[2],k\in[K]}$ can be very small and shrink as $T$ increases, 
\end{itemize}
any algorithm that makes less than enough switches must incur a large revenue loss compared with DLP-G as defined by $\big\{\max\eqref{eqn:obj-G} \ | \  \eqref{eqn:constraint:inventory-G}, \eqref{eqn:constraint:time-G}, \eqref{eqn:constraint:NonNeg-G} \ hold\big\}$.
More concretely, let there be $(d+1)$ families of actions, which we define as
\begin{align}
\mathcal{A}_1 = \Big\{k\big\vert k \% (d+1) = 1\Big\}, && \mathcal{A}_2 = \Big\{k\big\vert k \% (d+1) = 2\Big\}, && ... && \mathcal{A}_{d+1} = \Big\{k\big\vert k \% (d+1) = 0\Big\}. \label{eqn:FamilyOfActions}
\end{align}
We will show that, for any algorithm, as long as there exists a family of actions, such that this algorithm almost surely selects action(s) from this family for no more than $\zeta \cdot T / (d+1)$ times for some constant $\zeta \in [0,1)$, then this algorithm must incur a large revenue loss.

\begin{lemma}
\label{lem:WorstCaseLP}
Given arbitrary $\bm{b} \in (0,1)^d$, $T\ge 1$, $d\ge 1$, $K\ge d+1$, and  $\zeta \in [0,1)$. For any $\eta\in[0,1/2]$ (which could depend on $T$), consider the following $\bmu$:
\[
\mu_{i,k}=\begin{cases}
\frac{\gap}{d+1}, &\text{if }i=1, k\%(d+1)=1,\\
0, &\text{if }i=1, k\%(d+1)\in [2:d],\\
-\frac{\gap}{d+1}, &\text{if }i=1, k\%(d+1)=0,\\
    \gap, &\text{if }i=2,k\in[K].
\end{cases}\] Then for the \textsf{SP} problem instance $\mathcal{I}=(T,\bm{B}, K,d;\bm{R},\bC)$ specified above, and for any policy $\pi$ that selects some family of actions \eqref{eqn:FamilyOfActions} for no more than $\zeta \cdot T / (d+1)$ times almost surely, this policy $\pi$ must satisfy
\[
\Rev(\pi) \leq \DLPG -\DLPG\left(\frac{ (1-\zeta)\eta}{d(d+1)^2}-{\frac{1}{\min_{i\in[d]} b_i} \sqrt{3K\log T}}T^{-\frac{1}{2}}\right)+  \sqrt{3KT\log T}.
\]
As a corollary, if $\eta=O(T^{-\alpha})$ where $\alpha\in[0,1/2)$, then $\Rev(\pi) = \DLPG-\Omega(T^{1-\alpha})$.
\end{lemma}


Furthermore, under the instance specified by Lemma \ref{lem:WorstCaseLP}, for any policy $\pi$ that selects some action for no more than $\zeta \cdot T / (d+1)$ times with some constant probability $\xi\in(0,1]$ (instead of almost surely), it must also incur a large revenue loss. This is because conditional on a high probability event of the demand realization process, the upper bound in Lemma \ref{lem:WorstCaseLP} essentially holds for the empirical total revenue collected by any realized action sequence, as long as the sequence fails to select every action for more than $\zeta \cdot T / (d+1)$ times. As a result, as long as we can provide a ``per-action-sequence'' proof for Lemma \ref{lem:WorstCaseLP}, we can easily extend Lemma \ref{lem:WorstCaseLP} to Lemma \ref{lem:WorstCaseLPProb}.


\begin{lemma}
\label{lem:WorstCaseLPProb}
Given arbitrary $\bm{b}\in (0,1)^d$, $T\ge 1$, $d\ge 1$, $K\ge d+1$, and $\zeta\in [0,1),\xi\in(0,1]$. For any $\eta\in[0,1/2]$ (which could depend on $T$), consider the following $\bmu$:
\[
\mu_{i,k}=\begin{cases}
\frac{\gap}{d+1}, &\text{if }i=1, k\%(d+1)=1,\\
0, &\text{if }i=1, k\%(d+1)\in [2:d],\\
-\frac{\gap}{d+1}, &\text{if }i=1, k\%(d+1)=0,\\
    \gap, &\text{if }i=2,k\in[K].
\end{cases}\] Then for the \textsf{SP} problem instance $\mathcal{I}=(T,\bm{B}, K,d;\bm{R},\bC)$ specified above, and for any policy $\pi$ that selects some family of actions \eqref{eqn:FamilyOfActions} for no more than $\zeta \cdot T / (d+1)$ times with probability $\xi$, this policy $\pi$ must satisfy
\[
\Rev(\pi) \leq \DLPG -\xi\DLPG\left(\frac{ (1-\zeta)\eta}{d(d+1)^2}-{\frac{1}{\min_{i\in[d]} b_i} \sqrt{3K\log T}}T^{-\frac{1}{2}}\right)+ \xi \sqrt{3KT\log T}.
\]
\end{lemma}


Lemma~\ref{lem:WorstCaseLPProb} is a generalization of Lemma~\ref{lem:WorstCaseLP}, whose proof is omitted, as we will indeed provide a ``per-action-sequence'' proof for Lemma \ref{lem:WorstCaseLP}. Below we prove Lemma~\ref{lem:WorstCaseLP} using a combination of (i) the clean event analysis technique as we have used in the proof of Theorem~\ref{thm:NRMLB} (see Step 2), and (ii) a careful comparison and analysis of the objective values of several LP's (see Step 3).

\proof{Proof of Lemma~\ref{lem:WorstCaseLP}.} Since $\eta\in[0,1/2]$, the specified $\bmu$ ensures that $r_k,c_{i,k}\in[0,1]$ for all $k\in[K],i\in[d]$, and hence the components of $\bm{R},\bC$ are indeed valid Bernoulli random variables. Since the realization of a Bernoulli random variable always belongs to $\{0,1\}$, in this proof only, we define $\cmax=\rmax=1$.

\noindent \textbf{Step 1}: 
We argue that DLP-G under the above specification is feasible, and that its optimal solution has at least $(d+1)$ non-zero variables.
To see this, write the DLP-G under the above specification.
For any $k \in [d+1]$, recall the definition of $\mathcal{A}_k$ from \eqref{eqn:FamilyOfActions}.
We then write DLP-G as follows.
%
\begin{align}
\DLPG = \max_{(x_1,\dots,x_{K})} \frac{1}{2} \sum_{k\in[K]} x_k + \frac{\eta}{d+1} \sum_{k\in\mathcal{A}_1} x_k - \frac{\eta}{d+1} \sum_{k\in\mathcal{A}_{d+1}} x_k &&&  \label{eq:ec51} \\
\text{s.t.} \quad\quad\quad\quad\quad\quad \frac{b_i}{2} \sum_{k\in[K]} x_k \ + \ b_i\eta\bigg(\sum_{k\in\mathcal{A}_i} x_k - \sum_{k\in\mathcal{A}_{i+1}} x_k\bigg) & \leq \frac{b_i}{2} T && \forall\ i \in [d]\label{eq:ec52}\\
\sum_{k\in [K]}x_k &\leq T && \nonumber \\
x_k & \geq 0 && \forall\ k\in[K] \nonumber
\end{align}
Note that the constraint \eqref{eq:ec52} can be written as $\frac{1}{2} \sum_{k\in[K]} x_k + \eta\sum_{k\in\mathcal{A}_i} x_k - \eta\sum_{k\in\mathcal{A}_{i+1}} x_k \leq \frac{T}{2}, \forall\ i \in [d]$.
One class of feasible solutions to DLP-G is $$\sum_{k\in\mathcal{A}_{1}} x_k = \sum_{k\in\mathcal{A}_{2}} x_k = ... = \sum_{k\in\mathcal{A}_{d+1}} x_k = T/(d+1).$$

Next we show that this class of feasible solutions is optimal.
To see this, for any $k \in [d+1]$, let $\iota_k = \sum_{k\in\mathcal{A}_k} x_k$ denote the number of times that an action from family $\mathcal{A}_k$ is selected.
Then we introduce an equivalent LP using $\iota_1, \iota_2, ..., \iota_{d+1}$ as the decision variables, while also introducing slack variables $\rho_1, \rho_2, ..., \rho_{d+1}$.
We then re-write the above LP into the canonical form:
\begin{align*}
\min_{(x_1,\dots,x_{d+1}), (y_1,\dots,y_{d+1})} \sum_{k\in[d+1]} \ \left( - \frac{1}{2} \right) \ \iota_k + \left( -\frac{\eta}{d+1}\right) \iota_1 - \left(-\frac{\eta}{d+1}\right) \iota_{d+1} &&& \\
\text{s.t.} \ \sum_{k\in[d+1]} \ \frac{1}{2} \iota_k \ + \ \eta(\iota_i-\iota_{i+1}) + \rho_i & = \frac{T}{2} && \forall\ i \in [d] \\
\sum_{k\in [d+1]}\iota_k + \rho_{d+1} & = T  && \nonumber \\
\iota_k, \rho_k & \geq 0 && \forall\ k\in[d+1]. \nonumber
\end{align*}

To analyze this LP, denote $\tilde{A}$ to be the matrix of the linear program in the canonical form, and $\tilde{\bm{c}}$ to be the objective vector in the canonical form.
Denote for all $i,j\in[d+1]$,
$$\tilde{B}_{ij} = \begin{cases}
    c_{i,k}, &\text{if } i \leq d,\\
    1, &\text{if } i = d+1
\end{cases},$$
which is the basis matrix at this feasible solution.

We make the following observations: (i) all the slack variables $\rho_1, ..., \rho_{d+1}$ are zero, and all the $\iota_1, ..., \iota_{d+1}$ are non-negative.
(ii) we check the reduced costs by first denoting $\tilde{\bm{c}}'_B = - \frac{1}{d+1} \bm{1}' \tilde{B}$, and so we have $\bar{\bm{c}}' = \tilde{\bm{c}}' - \tilde{\bm{c}}'_{\tilde{B}} \tilde{B}^{-1} \tilde{A} = (0,0, ..., 0, 1/(d+1), 1/(d+1), ..., 1/(d+1))' \geq \bm{0}'$, where the inequality for the vector is component-wise.
Due to Definition 3.3 from \citet{bertsimas1997introduction}, we know that this feasible solution is optimal.
This optimal solution has $(d+1)$ non-zero values and is non-degenerate. Also, we know that $\DLPG=\frac{T}{2}$.

\noindent \textbf{Step 2}: Now we identify a clean event, conditional on which the realized inventory consumption and the realized rewards are close to the expected inventory consumption and the expected rewards, respectively. So we can use LP to approximate the empirical reward collected by any realized action sequence.

For any $k \in [K]$, let $y_k$ be the total number of periods that arm $k$ is pulled, under policy $\pi$.
Notice that $y_k$ may be random, i.e., the number of periods that arm $k$ is pulled is determined by the random trajectory of reward and cost realization and action selection.
Notice that $(y_1,\dots,y_K)$ always respects the inventory and time constraints.

Let $C_{i,k}^s(t)$ denote the random amount of resource $i$ consumed, during the first $t$ periods that the arm $k$ is pulled; let $R^s_{k}{(t)}$ denote the random amount of rewards generated, during the first $t$ periods that arm $k$ is pulled.
By Hoeffding's inequality: 
\begin{align*}
\forall k, \forall i, \forall t, & \Pr\left( \left| C^s_{i,k}(t) - t c_{i,k} \right| \leq \cmax \sqrt{3 t \log{T}} \right) \geq 1 - \frac{2}{T^6}; \\
\forall k, \forall t, & \Pr\left( \left| R^s_{k}(t) - t r_{k} \right| \leq \rmax \sqrt{3 t \log{T}} \right) \geq 1 - \frac{2}{T^6}.
\end{align*}
Denote the following event $E$:
\begin{align*}
E = \left\{ 
\begin{aligned}
\forall k, \forall i, \forall t, & \left| C^s_{i,k}(t) - t c_{i,k} \right| \leq \cmax \sqrt{3 t \log{T}}, \\
\forall k, \forall t, & \left| R^s_{k}(t) - t r_{k} \right| \leq \rmax \sqrt{3 t \log{T}}.
\end{aligned}
\right\}.
\end{align*}
Using a union bound we have:
$$\Pr\left( E \right) \geq 1 - \frac{4}{T^3}$$
because $K,d$ are both less than $T$, and each arm cannot be pulled longer than $T$ periods.
The happening of such event implies that 
\begin{align*}
\forall k, \forall i,  & \left| C^s_{i,k} - y_k c_{i,k} \right| \leq \cmax \sqrt{3 y_k \log{T}}, \\
\forall k,& \left| R^s_{k} - y_k r_{k} \right| \leq \rmax \sqrt{3 y_k \log{T}},
\end{align*}
i.e., 
the realized rewards and costs are close to the expected values, suggesting that we can use LP to approximately bound the total reward collected by any (realized) action sequence $(y_1,\dots,y_K)$.
To do so, we use the following arguments. 
Conditional on $E$, for any realization  of $(y_1,\dots,y_K)$, the total collected reward can be upper bounded by
\begin{align*}
\sum_{k \in [K]} R^s_{k} & \leq \sum_{k \in [K]} ( r_{k} y_{k} + \rmax \sqrt{3y_k\log{T}} ) \\
& \leq ( \sum_{k \in [K]} r_{k} y_{k} ) + \rmax \sqrt{3KT\log{T}}.
\end{align*}
On the other hand, the consumption of each resource $i$ must not violate the resource constraints.
$$\sum_{k \in [K]} C^s_{i,k} \leq B_i.$$
Lower bounding $C^s_{i,k}$ by $( y_{k} c_{i, k} - \cmax \sqrt{3y_k\log{T}} )$ we have
$$\sum_{k \in [K]} y_{k} c_{i,k} \leq B_i + \sum_{k \in [K]} \cmax \sqrt{3y_k\log{T}} \leq B_i + \cmax \sqrt{3KT\log{T}}.$$
These suggest that conditional on $E$, $(y_1,\dots,y_K)$ almost surely satisfies the following constraints:
\begin{align*} \sum_{k\in[K]}c_{i,k} \ y_k &\leq B_i + \cmax \sqrt{3KT\log{T}} && \forall\ i \in [d] \\
\sum_{k\in [K]}y_k &\leq T && \\
y_k & \geq 0 && \forall\ k\in[K]\\
\sum_{l\in[d+1]}\bI\bigg\{y_l > \zeta\frac{T}{d+1}\bigg\} & \leq d &&,
\end{align*}
with its total collected reward upper bounded by 
\[
 \sum_{k \in [K]} r_{i,k} y_{k}   +\rmax \sqrt{3KT\log{T}}.
\]

Recall that the optimal solution to the DLP-G evenly selects $T/(d+1)$ many actions from each family of actions $\mathcal{A}_k, \forall k \in [d+1]$.
For any $l \in [d+1]$, define the following linear program parameterized by $l$,
\begin{align*}
\bm{\mathsf{(DLP_l-G)}} \quad \DLPG_{l} = \max_{(x_k)_{k \in [K]}} \sum_{k\in[K]} r_k \ x_k &&& \\
\text{s.t.} \ \sum_{k\in[K]} c_{i,k} \ x_k &\leq B_i && \forall\ i \in [d] \\
\sum_{k\in [K]}x_k &\leq T && \\
\sum_{k \in \mathcal{A}_l} x_k & \leq \zeta \frac{T}{d+1} && \\
x_k & \geq 0 && \forall\ k\in[K].
\end{align*}
such that this linear program selects no more than $\zeta T / (d+1)$ times of an action from family $\mathcal{A}_l$.
Now for any $l \in [d+1]$, construct the following linear program parameterized by $l$, which we denote as a ``perturbed linear program'':
\begin{align*}
\bm{\mathsf{(DLP_l-G \ Perturbed)}} \quad \mathsf{J}^\mathsf{Perturbed-G}_l = \max_{(x_k)_{k \in [K]}} \sum_{k\in[K]} r_{k} \ x_k &&& \\
\text{s.t.} \ \sum_{k\in[K]} c_{i,k} \ x_k &\leq B_i + \cmax \sqrt{3KT\log{T}} && \forall\ i \in [d] \\
\sum_{k\in [K]}x_k &\leq T && \\
\sum_{k \in \mathcal{A}_l} x_k & \leq \zeta \frac{T}{d+1} && \\
x_k & \geq 0 && \forall\ k\in[K].
\end{align*}
Since from each solution $\bm{x}^*$ of the $\mathsf{Perturbed \ DLP_l-G}$, we can find a corresponding discounted solution $\bm{x}^* / (1+\frac{ \cmax \sqrt{3KT\log{T}}}{\Bmin})$ that is feasible to the $\mathsf{DLP_l-G}$.
This suggests that $\mathsf{J}^\mathsf{Perturbed}_l \leq \DLPG_l \cdot (1+\frac{\cmax \sqrt{3KT\log{T}}}{\Bmin})$, because DLP-G is a maximization problem.

Next we define an instance-dependent gap between the maximum objective value of $\mathsf{DLP_l-G}$, and the objective value of DLP-G.
Let $\Delta = (\DLPG - \max_{l \in [K]} \DLPG_l) / \DLPG$ be such an instance-dependent gap normalized by $\DLPG$.

Putting everything together, we obtain the following result: conditional on event $E$ that happens with probability at least $1 - \frac{4}{T^3}$, for any policy $\pi$ satisfying the requirements of Lemma \ref{lem:WorstCaseLP} (that selects some family of actions for no more than $\zeta T / (d+1)$ times) and any possible realization  of $(y_1,\dots,y_K)$,  the total collected reward is  upper bounded by
\begin{align*}
&~~~ \max_{l \in [d+1]} \mathsf{J}^\mathsf{Perturbed-G}_l + \rmax \sqrt{3KT\log{T}} \\
& \leq \max_{l \in [d+1]} \DLPG_l \cdot (1+\frac{\cmax \sqrt{3KT\log{T}}}{\Bmin}) + \rmax \sqrt{3KT\log{T}} \\
& \leq (\DLPG - \Delta \DLPG) \cdot (1+\frac{\cmax \sqrt{3KT\log{T}}}{\Bmin}) +\rmax \sqrt{3KT\log{T}}.
\end{align*}


\noindent \textbf{Step 3}: Now we examine the instance-dependent gap $\Delta$ on this problem instance as specified at the beginning of this proof.
Recall that $\Delta$ is defined as $\Delta = (\DLPG - \max_{l \in \mathcal{Z}(\bm{x}^*)} \DLPG_l) / \DLPG$.
Next we check each $\DLPG_l$ and find the one with the largest objective value.
First notice that $\DLPG = T/2$, i.e., the optimal objective value is exactly $T/2$.

Second, we show that $\DLPG_l$ can be decomposed as follows:
\begin{align}
\label{eqn:DecomposeLP}
\DLPG_l = \zeta \DLPG + (1-\zeta) \DLPGO_l
\end{align}
where $\DLPGO_l$ is given by the following LP:
\begin{align*}
\bm{\mathsf{(DLP_l-G-0)}} \quad \DLPGO_{l} = \max_{(x_k)_{k \in [K]}} \sum_{k\in[K]} r_k \ x_k &&& \\
\text{s.t.} \ \sum_{k\in[K]} c_{i,k} \ x_k &\leq B_i && \forall\ i \in [d] \\
\sum_{k\in [K]}x_k &\leq T && \\
\sum_{k \in \mathcal{A}_l} x_k & = 0 && \\
x_k & \geq 0 && \forall\ k\in[K].
\end{align*}
This decomposition holds due to the following two arguments.
(i) Denote the optimal solution to $\DLPG$ as $\bm{x}^{\mathsf{DLP-G}}$, which is given explicitly by $x^{\mathsf{DLP-G}}_1 = x^{\mathsf{DLP-G}}_2 = ... = x^{\mathsf{DLP-G}}_{d+1} = T / (d+1)$. 
Denote, for each $l \in [d+1]$, the optimal solution to each $\mathsf{DLP_l-G-0}$ problem as $\bm{x}^{\mathsf{DLP_l-G-0}}$. 
Then, their convex combination $(\zeta \bm{x}^{\mathsf{DLP-G}} + (1-\zeta) \bm{x}^{\mathsf{DLP_l-G-0}})$ satisfy the constraints of $\mathsf{DLP_l-G}$. 
To see this, note that both $\bm{x}^{\mathsf{DLP-G}}$ and $\bm{x}^{\mathsf{DLP_l-G-0}}$ satisfy the resource constraints for any $i$ as well as the time constraints.
Moreover, $$\sum_{k \in \mathcal{A}_l} \Big(\zeta x_k^{\mathsf{DLP-G}} + (1-\zeta) x_k^{\mathsf{DLP_l-G-0}}\Big) = \zeta \frac{T}{d+1}$$ which satisfies the additional constraint from $\mathsf{DLP_l-G}$, too.
Due to the optimality of $\DLPG_l$, we know that $\DLPG_l \geq \zeta \DLP + (1-\zeta) \DLPGO_l$. 
(ii) The optimal solution to $\DLPG_l$, denoted as $\bm{x}^\mathsf{DLP-G}_l$ must respect the constraints. In particular, the constraint $x_l \leq \zeta \frac{T}{d+1}$ must be binding. So that from $\bm{x}^\mathsf{DLP-G}_l$ we can construct $\bm{x}^\mathsf{DLP-G}_l - T/(d+1) \cdot \bm{1}$, which is feasible to $\mathsf{DLP_l-G-0}$. Then due  to the optimality of $\DLPGO_l$, we know that $\DLPG_l \leq \zeta \DLP + (1-\zeta) \DLPGO_l$. Combining both sides we prove \eqref{eqn:DecomposeLP}.

Third, to bound each $\DLPG_l$, we focus on each $\DLPGO_l$.

(i) When $l=1$, then the $1^\text{st}$ constraint becomes $\sum_{k \in \{2,3,...,d+1\}} 1/2 \ x_k - \eta \ x_2 \leq T/2$, which is dominated by the $(d+1)^\text{th}$ constraint (the time constraint).
Then we solve a linear program with $d$ variables and $d$ constraints.
\begin{align*}
\DLPGO_l = \max_{\bm{x}} && \left[\frac{1}{2} \hphantom{+ \frac{\eta}{d+1}}, \frac{1}{2} \hphantom{+ \frac{\eta}{d+1}}, ..., \ \frac{1}{2} \hphantom{+ \frac{\eta}{d+1}}, \frac{1}{2} - \frac{\eta}{d+1}\right] \cdot \bm{x} & \\
\text{s.t.} && \left[ 
\begin{aligned}
& 1+2\eta \hspace{8pt}  && 1-2\eta               && ... && 1                    && 1                     \\
& 1                     && 1+2\eta  \hspace{8pt} && ... && 1                    && 1                     \\
&                       &&                       &&     &&                      &&                       \\
& 1                     &&                       && ... && 1+2\eta \hspace{8pt} && 1-2\eta \hspace{11pt} \\
& 1                     &&                       && ... && 1                    && 1                     \\
\end{aligned}
\right] \hphantom{'} \cdot \bm{x} & \leq T \cdot \bm{1}\\
&& \bm{x} & \geq \bm{0}
\end{align*}
The optimal solution is be $x_2 = x_3 = ... = x_{d+1} = T/d$, and the objective value is $\frac{T}{2} \cdot (1 - \frac{2 \eta}{d (d+1)})$.

(ii) When $2\leq l \leq d$, first observe that both the $l^\text{th}$ constraint and the $(d+1)^\text{th}$ constraint are dominated by the $(l-1)^\text{th}$ constraint. Then we solve a linear program with $d$ variables and $(d-1)$ constraints.
\begin{align*}
\DLPGO_l = \max_{\bm{x}} && \left[\frac{1}{2} + \frac{\eta}{d+1}, \frac{1}{2} \hphantom{+ \frac{\eta}{d+1}}, ..., \frac{1}{2} \hphantom{+ \frac{\eta}{d+1}}, \frac{1}{2} \hphantom{+ \frac{\eta}{d+1}}, \frac{1}{2} \hphantom{+ \frac{\eta}{d+1}}, ..., \ \frac{1}{2} \hphantom{+ \frac{\eta}{d+1}}, \frac{1}{2} - \frac{\eta}{d+1}\right] \cdot \bm{x} & \\
\text{s.t.} && \left[ 
\begin{aligned}
& 1+2\eta \hspace{8pt} && 1-2\eta              && ... && 1                    && 1                    && 1                    && ... && 1                    && 1                     \\
& 1                    && 1+2\eta \hspace{8pt} && ... && 1                    && 1                    && 1                    && ... && 1                    && 1                     \\
&                      &&                      &&     &&                      &&                      &&                      &&     &&                      &&                       \\
& 1                    && 1                    && ... && 1-2\eta              && 1                    && 1                    && ... && 1                    && 1                     \\
& 1                    && 1                    && ... && 1+2\eta \hspace{8pt} && 1                    && 1                    && ... && 1                    && 1                     \\
& 1                    && 1                    && ... && 1                    && 1+2\eta \hspace{8pt} && 1-2\eta              && ... && 1                    && 1                     \\
& 1                    && 1                    && ... && 1                    && 1                    && 1+2\eta \hspace{8pt} && ... && 1                    && 1                     \\
&                      &&                      &&     &&                      &&                      &&                      &&     &&                      &&                       \\
& 1                    && 1                    && ... && 1                    && 1                    && 1                    && ... && 1-2\eta \hspace{8pt} && 1                     \\
& 1                    && 1                    && ... && 1                    && 1                    && 1                    && ... && 1+2\eta              && 1-2\eta \hspace{11pt} \\
\end{aligned}
\right] \hphantom{'} \cdot \bm{x} & \leq T \cdot \bm{1}\\
&& \bm{x} & \geq \bm{0}
\end{align*}
The optimal solution is be $x_{l+1} = x_{l+2} = ... = x_{d+1} = T/(d-l+1)$, and the objective value is $\frac{T}{2} \cdot (1 - \frac{2 \eta}{(d-l+1) \cdot (d+1)})$.

(iii) When $l = d+1$, then the $(d+1)^\text{th}$ constraint is dominated by the $d^\text{th}$ constraint. Then we solve a linear program with $d$ variables and $d$ constraints.
\begin{align*}
\DLPGO_l = \max_{\bm{x}} && \left[\frac{1}{2} + \frac{\eta}{d+1}, \frac{1}{2} \hphantom{+ \frac{\eta}{d+1}}, ..., \ \frac{1}{2} \hphantom{+ \frac{\eta}{d+1}}, \frac{1}{2} \hphantom{+\frac{\eta}{d+1}} \right] \cdot \bm{x} & \\
\text{s.t.} && \left[ 
\begin{aligned}
& 1+2\eta \hspace{8pt}  && 1-2\eta               && ... && 1                    && 1                    \\
& 1                     && 1+2\eta  \hspace{8pt} && ... && 1                    && 1                    \\
&                       &&                       &&     &&                      &&                      \\
& 1                     &&                       && ... && 1+2\eta \hspace{8pt} && 1-2\eta              \\
& 1                     &&                       && ... && 1                    && 1+2\eta \hspace{11pt} \\
\end{aligned}
\right] \cdot \bm{x} & \leq T \cdot \bm{1}\\
&& \bm{x} & \geq \bm{0}
\end{align*}
The optimal solution requires a little computation. It is $x_i = \frac{T (d+1-i)}{d(d+1)+4\eta}, \forall i \in [d]$, and the objective value is $\frac{T}{2} \cdot (1 - \frac{4 \eta}{d(d+1)^2 + 4 \eta (d+1)})$.

After comparing the three cases, we know that
$$\Delta = \frac{4 (1-\zeta) \eta}{d(d+1)^2 + 4 \eta (d+1)} \geq \frac{2 (1-\zeta) \eta}{d(d+1)^2}.$$

\noindent \textbf{Step 4}: Now we combine Steps~2 and~3. From Step~2, we know that conditional on event $E$ which happens with probability at least $1-\frac{4}{T^3}$, for any possible realization  of $(y_1,\dots,y_K)$,  the total collected reward is  upper bounded by
\begin{align*}
\Rev(\pi)  \leq (\DLPG - \Delta \DLPG) \cdot (1+\frac{\cmax \sqrt{3KT\log{T}}}{\Bmin}) + \rmax \sqrt{3KT\log{T}}.
\end{align*}
So unconditionally,
\begin{align*}
\Rev(\pi) \leq & (1 - \frac{4}{T^3}) \cdot \left\{ (\DLPG - \Delta \DLPG) \cdot (1+\frac{\cmax \sqrt{3KT\log{T}}}{\Bmin}) + \rmax \sqrt{3KT\log{T}} \right\} + \frac{4}{T^3} \cdot \DLPG \\
\leq & \DLPG \left\{ (1-\frac{4}{T^3})(1-\Delta)(1+\frac{\cmax \sqrt{3KT\log(T)}}{\Bmin}) + \frac{4}{T^3} \right\} + \rmax \sqrt{3KT\log T} \\
\leq & \DLPG (1 - \frac{1}{2} \Delta)(1+\frac{\cmax \sqrt{3KT\log T}}{\Bmin}) +\rmax \sqrt{3KT\log T} \\
\leq & \DLPG (1 - \frac{(1-\zeta) \eta}{d(d+1)^2})(1+{\frac{1}{\min_{i\in[d]} b_i}\cmax \sqrt{3K\log T}}T^{-\frac{1}{2}}) + \rmax \sqrt{3KT\log T}\\
\leq & \DLPG -\DLPG\left(\frac{ (1-\zeta)\eta}{d(d+1)^2}-{\frac{1}{\min_{i\in[d]} b_i}\cmax \sqrt{3K\log T}}T^{-\frac{1}{2}}\right)+ \rmax \sqrt{3KT\log T}.
\end{align*}
Noting that in this proof $\cmax=\rmax=1$, we finish the proof. 
\Halmos
\endproof

\subsection{Key Definitions}
We then turn to the distributionally-unkwown setup.
Our proof builds on the ``tracking the cover time'' argument of \cite{simchi2019phase,simchi2023phase} and extends it to the resource-constrained setting.  

For any $t < t' \in [T]$, let $[t:t'] = \{t, t+1, ..., t'\}$ denote the set of integers between $t$ and $t'$. For notational convenience in this lower bound proof, we use $[K]$ (instead of $P$) to represent the action set, where each action $k\in[K]$ corresponds to the price vector $\bp_k\in P$.

For any admissible policy $\phi\in\Phi[\infty]$, for any instance $\mathcal{I}_{\bmu}$, for any $t\in[T]$, we use $z_t$ to denote the random action selected by policy $\phi$ under instance $\Imu$. Note that $z_t$ depends on both $\phi$ and $\bmu$. Whenever $z_t$ appears within an expectation or probability operator, the corresponding policy and instance (which may be denoted by variants such as $\phi^\dagger$ and $\bmu'$) will be explicitly specified in the operator’s superscript and subscript. We use $\mathcal{F}_t=\sigma\left(\left(z_1,(Q^1_{j,k;\bmu}\right)_{j\in[n],k\in[K]}),\dots,\left(z_t,(Q^t_{j,k;\bmu}\right)_{j\in[n],k\in[K]})\right)$ to denote the sigma algebra generated by all of the randomness up to round $t$.

For any admissible policy $\phi\in\Phi[\infty]$, for any instance $\mathcal{I}_{\bmu}$, we use $\Prob^{\phi}_{\bmu}$ to denote the probability measure induced by policy $\phi$'s random actions $(z_t)_{t\in[T]}$ and nature's random demand variables $(Q^t_{j,k;\bmu})_{j\in[n],k\in[K],t\in[T]}$. Let $\mathbb{E}^{\phi}_{\bmu}$ be the expectation operator associated with $\Prob^{\phi}_{\bmu}$.

For any limited-switch policy $\phi\in\Phi[s]$, for any instance $\mathcal{I}_{\bmu}$, we make some key definitions below.

1. We first define a series of ordered stopping times $\tau_1\le\tau_2\dots\le\tau_{\nu(s,d)}\le\tau_{\nu(s,d)+1}$.
\begin{itemize}
    \item $\tau_1=\min\{1\le t\le T: \text{all the actions in $[K]$ have been chosen in $[1:\tau_1]$}\}$ if the set is non-empty and $\tau_1=\infty$ otherwise.
    \item $\tau_2=\min\{1\le t\le T: \text{all the actions in $[K]$ have been chosen in $[\tau_1:\tau_2]$}\}$ if the set is non-empty and $\tau_2=\infty$ otherwise.
    \item Generally, $\tau_j=\min\{1\le t\le T: \text{all the actions in $[K]$ have been chosen in $[\tau_{j-1}:\tau_j]$}\}$ if the set is non-empty and $\tau_j=\infty$ otherwise, for all $i=2,\dots,\nu(s,d)+1$.
\end{itemize}
It can be verified that $\tau_1,\dots,\tau_{\nu(s,d)+1}$ are stopping times with respect to the filtration $\mathbb{F}=(\mathcal{F}_t)_{t\in [T]}$. We also define $\tau_{\rm stop}$ to be the last round of the (effective) selling process: if $\tau_{\rm stop}<T$, then it means that some resource is exhausted before round $T$ and the selling process stops immediately. For notational convenience, for any realization of $\tau_{\rm stop}$, we say that policy $\phi$ keeps choosing the action chosen in round $\tau_{\rm stop}$ for all $t\in[\tau_{\rm stop}+1:T]$ (though the selling process has already stopped).

2. We then define a series of random variables (which depend on the stopping times).
\begin{itemize}
    \item $S(1,\tau_1)$ is the number of switches occurs in $[1:\tau_1]$ (note that if there is a switch happening between $\tau_1$ and $\tau_1+1$, we do not count it in $S(1,\tau_1)$).
    \item For all $j=2,\dots,\nu(s,d)$, $S(\tau_{j-1},\tau_{j})$ is the number of switches occurs in $[\tau_{j-1}:\tau_{j}]$ (note that if there is a switch happening between $\tau_{j-1}-1$ and $\tau_{j-1}$, or between $\tau_{j}$ and $\tau_{j}+1$, we do not count it in $S(\tau_{j-1},\tau_{j})$).
    \item  $S(\tau_{\nu(s,d)},T)$ is the number of switches occurs in $[\tau_{\nu(s,d)}:T]$ (note that if there is a switch happening between $\tau_{\nu(s,d)-1}$ and $\tau_{\nu(s,d)}$, we do not count it in $S(\tau_{\nu(s,d)},T)$).
\end{itemize}
We remark that if the selling horizon stops at round $\tau_{\rm stop}<T$ due to exhausted inventory, then there is no switch happening in $[\tau_{\rm stop}:T]$.

3. Next we define a series of events. Recall that $t_1,\dots,t_{\nu(s,d)}\in[T]$ are some fixed values specified in Algorithm \ref{alg:bsse}. If $\nu(s,d)\ne0$, we define
\begin{itemize}
    \item $E_1=\{\tau_1>t_1\}$.
    \item For all $j=2,\dots,\nu(s,d)$, $E_j=\{\tau_{j-1}\le t_{j-1}, \tau_{j}> t_j\}$.
    \item $E_{\nu(s,d)+1}=\{\tau_{\nu(s,d)}\le t_{\nu(s,d)}\}$.
\end{itemize}
 If $\nu(s,d)=0$, we define $E_1=\{\tau_1>t_1\}=\{\tau_1>T\}$ and $\overline{E_1}=\{\tau_1\le t_1\}=\{\tau_1\le T\}$.

4. And we define a series of shrinking errors. If $\nu(s,d)\ne0$, we define
\begin{itemize}
    \item $\Delta_1=1/2$.
    \item For $j=2,\dots,\nu(s,d)$, $\Delta_j=\frac{K^{-1/2}\left(K/T\right)^{(1-2^{1-j})/(2-2^{-\nu(s,d)})}}{2K(\nu(s,d)+1)}\in (0,\frac{1}{2K(\nu(s,d)+1)}]$. (That is, $\Delta_j\approx \frac{1}{2K(\nu(s,d)+1)}\frac{1}{\sqrt{t_{j-1}}}$.)
    \item $\Delta_{\nu(s,d)+1}=\frac{K^{-1/2}\left(K/T\right)^{(1-2^{-\nu(s,d)})/(2-2^{-\nu(s,d)})}}{4K(\nu(s,d)+1)}\in (0,\frac{1}{4K(\nu(s,d)+1)}]$. (That is, $\Delta_{\nu(s,d)+1}\approx \frac{1}{4K(\nu(s,d)+1)}\frac{1}{\sqrt{t_{\nu(s,d)}}}$.)
\end{itemize}
If $\nu(s,d)=0$, we define $\Delta_1=1/4$.

5. For notational convenience, define $\tau_0=1$ and $\tau_{\max}=\sup_{j\in[0:\nu(s,d)+1]}\{\tau_j\mid\tau_j<\infty\}$, and let $S_{\times}$ denote the (random) set of actions that are not chosen in $[\tau_{\max}:T]$.\footnote{Since $\phi\in\Phi[s]$ and $K>d+1$, $S_{\times}$ is always non-empty.} Define $z_{\infty}$ (which is a unconditioned random variable) such that
conditional on any realization of $S_{\times}$, $z_{\infty}=i$ with probability $1/|S_{\times}|$ for all $i\in S_{\times}$.

\subsection{Technical Lemmas}

 For any two probability measures $\mathbb{P}$ and $\mathbb{Q}$ defined on the same measurable space, let $D_{\rm{TV}}(\mathbb{P}\|\mathbb{Q})$ denote the total variation distance between $\mathbb{P}$ and $\mathbb{Q}$, and $D_{\rm{KL}}(\mathbb{P}\|\mathbb{Q})$ denote the Kullback-Leibler (KL) divergence between $\mathbb{P}$ and $\mathbb{Q}$, see detailed definitions in Chapter 15 of \cite{wainwright2019high}.

\begin{lemma}\label{lm:kl}
For any $p,q\in(0,1)$, the KL divergence between $\mathsf{Ber}(p)$ and $\mathsf{Ber}(q)$ is
\[
D_{\rm KL}(\mathsf{Ber}(p)\|\mathsf{Ber}(q))=q\log\frac{q}{p}+(1-q)\log\frac{1-q}{1-p}\le\frac{1}{\min\{p,q,1-p,1-q\}}|p-q|^2.
\]
\end{lemma}

The above lemma is a standard result in information theory, and we omit its proof here.

We also have the following lemma on the number of switches. 
\begin{lemma}
\label{lm:switch}
For any limited-switch policy $\phi\in\Phi[s]$, for any instance $\mathcal{I}_{\bmu}$, the occurrence of $E_{\nu(s,d)+1}$ implies the occurrence of the event $\{\text{the number of switches occurs in }[\tau_{\nu(s,d)}:T]\text{ is no more than }K+d-1\}$ almost surely.
\end{lemma}

\proof{Proof of Lemma \ref{lm:switch}.}
When $E_{\nu(s,d)+1}$ happens, $\tau_{\nu(s,d)}\le t_{\nu(s,d)}\le T$, thus all $\tau_1,\dots, \tau_{\nu(s,d)}\le T$. Since in each of $[1:\tau_1],[\tau_1,\tau_2],\dots,[\tau_{\nu(s,d)-1}:\tau_{\nu(s,d)}]$, all $K$ actions were visited, we know that $S(1,\tau_1)\ge K-1$, $S(\tau_{1},\tau_2)\ge K-1$, $\dots$, $S(\tau_{\nu(s,d)-1},\tau_{\nu(s,d)})\ge K-1$. Thus we have
$$
S(1,\tau_1)+S(\tau_{1},\tau_2)+\cdots+S(\tau_{\nu(s,d)-1},\tau_{\nu(s,d)})\ge \nu(s,d)(K-1).
$$
Since $\phi\in\Phi[s]$, we further know that
$$
S(\tau_{\nu(s,d)},T)\le s-[S(1,\tau_1)+S(\tau_{1},\tau_2)+\cdots+S(\tau_{\nu(s,d)-1},\tau_{\nu(s,d)})]\le s-\nu(s,d)(K-1)\le K+d-1
$$
happens almost surely. As a result, the occurrence of $E_{\nu(s,d)+1}$ implies the occurrence of the event $\{\text{the number of switches occurs in }[\tau_{\nu(s,d)}:T]\text{ is no more than }K+d-1\}$ almost surely.
\Halmos
\endproof

\begin{lemma}\label{lem:reg-dec}
    For any instance $\mathcal{I}_{\bmu}$, for any admissible policy $\phi\in\Phi[\infty]$, we have
   \[ \left(\frac{1}{2}+\mu_{1,k^*}\right)T-\mathsf{Rev}_{\bQ_{\bmu}}(\phi)\ge\sum_{k\in[K]}\mathbb{E}^\phi_{\bmu}\left[\sum_{t\in[T]}\bI(z_t=k)\right]\left(\mu_{1,k^*}-\mu_{1,k}\right),\]
   where $\mu_{1,k^*}=\max_{k\in[K]}\{\frac{1}{2}+\mu_{1,k}\}$.
\end{lemma}
\proof{Proof of Lemma \ref{lem:reg-dec}.}
Consider a virtual policy $\phi^{\mathsf{v}}$
that runs under
exactly the same demand realization process and acts exactly the same as $\phi$ until period $\tau_{\text{stop}}$, but
keeps running and (importantly) \emph{collecting revenue} until the end of period $T$ as if there were no resource constraint. Since the action selection and reward collection process of policy $\phi^{\mathsf{v}}$ is not restricted by any resource constraint, this process is equivalent to running an equivalent  $K$-armed bandit learning policy on a $K$-armed bandit instance where the reward of arm $k\in[K]$ is $\mathsf{Ber}\left(\frac{1}{2}+\mu_{1,k}\right)$. Let $\mathsf{Rev}^{\mathsf{v}}_{\bQ_{\bmu}}(\phi^{\mathsf{v}})$ denote the total expected reward  collected in such an equivalent $K$-armed bandit instance. By the standard regret decomposition lemma for $K$-armed bandit (c.f. Lemma 4.5 of \cite{lattimore2020bandit}), we have
\[ \left(\frac{1}{2}+\mu_{1,k^*}\right)T-\mathsf{Rev}^{\mathsf{v}}_{\bQ_{\bmu}}(\phi^{\mathsf{v}})=\sum_{k\in[K]}\mathbb{E}^{\phi^{\mathsf{v}}}_{\bmu}\left[\sum_{t\in[T]}\bI(z_t=k)\right]\left(\mu_{1,k^*}-\mu_{1,k}\right).\]
Since $\mathsf{Rev}_{\bQ_{\bmu}}(\phi)\le\mathsf{Rev}^{\mathsf{v}}_{\bQ_{\bmu}}(\phi^{\mathsf{v}})$ (i.e., resource constraints cannot make the expected revenue higher), we have
 \begin{align*} \left(\frac{1}{2}+\mu_{1,k^*}\right)T-\mathsf{Rev}_{\bQ_{\bmu}}(\phi)&\ge \left(\frac{1}{2}+\mu_{1,k^*}\right)T-\mathsf{Rev}^{\mathsf{v}}_{\bQ_{\bmu}}(\phi^{\mathsf{v}})\\&=
\sum_{k\in[K]}\mathbb{E}^{\phi^{\mathsf{v}}}_{\bmu}\left[\sum_{t\in[T]}\bI(z_t=k)\right]\left(\mu_{1,k^*}-\mu_{1,k}\right)\\
 &\ge\sum_{k\in[K]}\mathbb{E}^\phi_{\bmu}\left[\sum_{t\in[T]}\bI(z_t=k)\right]\left(\mu_{1,k^*}-\mu_{1,k}\right),\end{align*}
 where the last inequality utilizes the specific coupling relationship between $\phi$ and $\phi^{\mathsf{v}}$ as well as the non-negativity of $\left(\mu_{1,k^*}-\mu_{1,k}\right)$.\Halmos
\endproof

\begin{lemma}\label{lem:reg-dec-dlp}
    Let $k' \in [K]$ be an action and $\Delta > 0$. Consider two instances, $\cI_{\bmu}$ and $\cI_{\bmu'}$, such that $\mu'_{1,k'} = \mu_{1,k'} - \Delta$, $\mu'_{2,k'} = \mu_{2,k'}$, and $\mu'_{2,k} = \mu_{2,k}$ for all $k \in [K]$. For any admissible policy $\phi \in \Phi[\infty]$, we have
    \[
    \mathsf{J}^{\mathsf{DLP}_{\bmu}} - \mathsf{Rev}_{\bQ_{\bmu'}}(\phi) \geq \mathbb{E}_{\bmu'}^\phi\left[\sum_{t \in [T]} \bI(z_t = k')\right] \Delta.
    \]
\end{lemma}

\proof{Proof of Lemma~\ref{lem:reg-dec-dlp}.} 
Given any $\phi = (\phi^1, \dots, \phi^T) \in \Phi[\infty]$ and ${\bmu'}$, we construct a comparator policy $\phi^{\dagger} \in \Phi[\infty]$ as follows. For any \bnrm instance, at each period $t = 1, 2, \dots$, as long as $z_t \in [K]$ (i.e., the algorithm has not stopped):
\begin{itemize}
    \item If $z_t \neq k'$, then add the action $z_t$ and the corresponding demand observations $(Q^t_{j,z_t})_{j \in [n]}$ to the history $\mathcal{H}_t$.
    \item If $z_t = k'$, replace the demand observation $Q_{1,k'}^t$ with an independent realization of $\Ber(q_{1,k';\bmu'})$, denoted as $Q_{1,k';\bmu'}^t$. Then, add the action $k'$ along with the \emph{modified} demand observations $\left(Q_{1,k';\bmu'}^t, (Q^t_{j,k})_{j \in [2:n]}\right)$ to the history $\mathcal{H}_t$. Note that only the demand observation of product 1 is modified; demand observations of other products remain unmodified.
\end{itemize}
The policy $\phi^\dagger$ then selects the action in period $t+1$ according to $\phi^{t+1}(\mathcal{H}_t)$, treating $\mathcal{H}_t$ as the observed history. Importantly, the modification is solely used for constructing $\phi^\dagger$ and does not alter the underlying instance. That is, when evaluating $\phi^\dagger$, its total reward is calculated using the actual demand realizations, not the modified ones. 

Clearly, the action sequence generated by $\phi$ under $\cI_{\bmu'}$ is distributionally identical to the action sequence generated by $\phi^\dagger$ under $\cI_{\bmu}$. This implies that
$
\mathbb{E}_{\bmu'}^\phi\left[\sum_{t \in [T]} \bI(z_t = k)\right] = \mathbb{E}_{\bmu}^{\phi^\dagger}\left[\sum_{t \in [T]} \bI(z_t = k)\right], \forall k \in [K].
$

Moreover, we have
\begin{align*}
    \mathsf{Rev}_{\bQ_{\bmu'}}(\phi)
    &= \mathbb{E}_{\bmu'}^\phi\left[\sum_{t \in [T]} \sum_{k \in [K]} Q^t_{1,k;\bmu'} \bI(z_t = k) \right] = \sum_{t \in [T]} \sum_{k \in [K]} \mathbb{E}_{\bmu'}^\phi\left[Q^t_{1,k;\bmu'} \bI(z_t = k)\right] \\
    &= \sum_{t \in [T]} \sum_{k \in [K]} \mathbb{E}_{\bmu'}^\phi\left[q_{1,k;\bmu'} \bI(z_t = k)\right] = \sum_{k=1}^{K} q_{1,k;\bmu'} \mathbb{E}_{\bmu'}^\phi\left[\sum_{t \in [T]} \bI(z_t = k)\right].
\end{align*}
Similarly, we obtain
\[
\mathsf{Rev}_{\bQ_{\bmu}}(\phi^\dagger) = \sum_{k=1}^{K} q_{1,k;\bmu} \mathbb{E}_{\bmu}^{\phi^\dagger}\left[\sum_{t \in [T]} \bI(z_t = k)\right].
\]
Using the identity 
$
\mathbb{E}_{\bmu'}^\phi\left[\sum_{t \in [T]} \bI(z_t = k)\right] = \mathbb{E}_{\bmu}^{\phi^\dagger}\left[\sum_{t \in [T]} \bI(z_t = k)\right], \forall k \in [K],
$
we deduce that
\[
\mathsf{Rev}_{\bQ_{\bmu}}(\phi^\dagger) - \mathsf{Rev}_{\bQ_{\bmu'}}(\phi) = \mathbb{E}_{\bmu'}^\phi\left[\sum_{t \in [T]} \bI(z_t = k')\right] \Delta.
\]
Since $\phi^\dagger \in \Phi[\infty]$, we have $\mathsf{Rev}_{\bQ_{\bmu}}(\phi^\dagger) \leq \mathsf{J}^{\mathsf{DLP}_{\bmu}}$, which completes the proof.
\Halmos
\endproof

\subsection{Proof of Lower Bound}
Based on all of the materials developed before, we start to prove Theorem~\ref{thm:BlindNRMLB}. For better exposition, we first prove our results for the case of $\nu(s,d)\ne0$, then consider the corner case of $\nu(s,d)=0$.

\proof{Proof of Theorem~\ref{thm:BlindNRMLB} when $\nu(s,d)\ne0$.}
Let  $\gap=\Delta_{\nu(s,d)+1}/4$. Consider the following $\bmu$:
\[
\mu_{i,k}=\begin{cases}
\frac{\gap}{d+1}, &\text{if }i=1, k\%(d+1)=1,\\
0, &\text{if }i=1, k\%(d+1)\in [2:d],\\
-\frac{\gap}{d+1}, &\text{if }i=1, k\%(d+1)=0,\\
    \gap, &\text{if }i=2,k\in[K],
\end{cases}
\]
For any $s$-switch learning policy $\phi\in\Phi[s]$, by the union bound, we have
$$
\sum_{j=1}^{\nu(s,d)+1}\mathbb{P}_{{\bmu}}^{\phi}(E_j)\ge \mathbb{P}_{{\bmu}}^{\phi}(\cup_{j=1}^{\nu(s,d)+1}E_j)=1.
$$
Therefore, there exists $j^*\in[\nu(s,d)+1]$ such that $\mathbb{P}_{{\bmu}}^{\phi}(E_{j^*})\ge 1/(\nu(s,d)+1)$.

\textbf{Case 1: $j^*=1$.}

Since $\mathbb{P}_{{\bmu}}^{\phi}(E_1)=\mathbb{P}_{\bmu}^{\phi}(\tau_1>t_1)\ge  1/(\nu(s,d)+1)$ and
$$
\mathbb{P}_{\bmu}^{\phi}(\tau_1>t_1)=\sum_{k=1}^{K}\mathbb{P}_{\bmu}^{\phi}(\tau_1>t_1,z_{\tau_1}=k),
$$
we know that there exists $k'\in[K]$ such that
$$
\mathbb{P}_{\bmu}^{\phi}(\tau_1>t_1,z_{\tau_1}=k')\ge \frac{\mathbb{P}_{{\bmu}}^{\phi}(E_1)}{K}\ge \frac{1}{K(\nu(s,d)+1)}.
$$
When $\tau_1<\infty$, since $\tau_1$ is the first time that all actions in $[K]$ has been chosen in $[1:\tau_1]$, the event $\{z_{\tau_1}=k'\}$ must imply the event $\{k'\text{ was not chosen in }[1:\tau_1-1]\}$. When $\tau_1=\infty$, since $z_{\tau_{1}}=z_\infty$ only takes values in $S_{\times}$ (which is the set of actions that were not chosen in $[1:T]$), the event $\{z_{\tau_1}=k'\}$ must imply the event $\{k'\text{ was not chosen in }[1:T]\}$. Thus, no matter whether $\tau_1<\infty$ or not, the event $\{\tau_1>t_1,z_{\tau_1}=k'\}$ must imply the event $\mathcal{E}_{k'}[1:t_1-1]:=\{k'\text{ was not chosen in }[1:t_1-1]\}$. Therefore, we have
\begin{equation*}
    \Prob_{\bmu}^{\phi}(\mathcal{E}_{k'}[1:t_1-1])\ge \mathbb{P}_{\bmu}^{\phi}(\tau_1>t_1,z_{\tau_1}=k')\ge \frac{1}{K(\nu(s,d)+1)}.
\end{equation*}
Meanwhile, the occurrence of the event $\mathcal{E}_{k'}[1:t_1-1]$ is independent of random matrix $(X_{\bmu}^{t}(k'))_{t\in[1:t_1-1]}$ and random matrices $(X_{\bmu}^{t}(k))_{t\in[t_1:T]}$ for all $k\in[K]$, i.e., the occurrence of the event $\mathcal{E}_{k'}[1:t_1-1]$ only depends on random matrices $(X_{\bmu}^{t}(k))_{t\in[1:t_1-1]}$ for $k\ne k'$ and policy $\phi$. Let $\mathbb{Q}_{\bmu}^{\phi}$ be the probability measure induced by the following random variables:
\begin{itemize}
\item random matrices $(X_{\bmu}^{t}(k))_{t\in[1:t_1-1]}$ for $k\ne k'$,
    \item policy $\phi$'s random actions $(z_t)_{t\in[1:t_1-1]}$ under instance $\mathcal{I}_{\bmu}$.
\end{itemize}
We have
\begin{equation}\label{eq:app1}
    \mathbb{Q}_{\bmu}^{\phi} (\mathcal{E}_{k'}[1:t_1-1])=\Prob_{\bmu}^{\phi}(\mathcal{E}_{k'}[1:t_1-1])\ge \frac{1}{K(\nu(s,d)+1)}.
\end{equation}

We now consider a new instance $\mathcal{I}_{\bmu'}$ with $\mu'_{1,k'}=\Delta_1$, $\mu'_{2,k'}=0$, and $\mu'_{i,k}=\mu_{i,k}$ for all $(i,k)\in[2]\times([K]\setminus\{k'\})$. Again, the occurrence of the event $\mathcal{E}_{k'}[1:t_1-1]$ is independent of random matrix $(X_{\bmu'}^{t}(k'))_{t\in[1:t_1-1]}$ and random matrices $(X_{\bmu'}^t(k))_{[t_1:T]}$ for all  $k\in[K]$. Let $\mathbb{Q}_{\bmu'}^\pi$ be the probability measure induced by the following random variables:
\begin{itemize}
\item random matrices $(X_{\bmu'}^{t}(k))_{t\in[1:t_1-1]}$ for $k\ne k'$,
    \item policy $\phi$'s random actions $(z_t)_{t\in[1:t_1-1]}$ under instance $\mathcal{I}_{\bmu'}$.
\end{itemize}
We have
\begin{equation}\label{eq:app2}
    \mathbb{Q}_{\bmu'}^\phi(\mathcal{E}_{k'}[1:t_1-1]) =\Prob_{\bmu'}^{\phi}(\mathcal{E}_{k'}[1:t_1-1]).
\end{equation}

But note that $(X_{\bmu'}^{t}(k))_{[1:t_1-1]}$ and $(X_{\bmu}^{t}(k))_{[1:t_1-1]}$ have exactly the same distribution for all $k\ne k'$. Thus from (\ref{eq:app1}) and (\ref{eq:app2}) we have
\begin{align*}
\Prob_{\bmu'}^{\phi}(\mathcal{E}_{k'}[1:t_1-1])=\Prob_{\bmu}^{\phi}(\mathcal{E}_{k'}[1:t_1-1])\ge \frac{1}{K(\nu(s,d)+1)}.
\end{align*}
However, under instance $\mathcal{I}_{\bmu'}$, the optimal solution to $\mathsf{DLP}_{\bmu'}$ satisfies $x_{k'}=T$, and $\mathsf{J}^{\mathsf{DLP}_{\bmu'}}=(\frac{1}{2}+\Delta_1)T$. 
By Lemma \ref{lem:reg-dec}, 
\[\mathsf{J}^{\mathsf{DLP}_{\bmu'}}-\Rev_{\bQ_{\bmu'}}(\phi)=(\frac{1}{2}+\Delta_1)T-\Rev_{\bQ_{\bmu'}}(\phi)\ge\mathbb{E}^\phi_{\bmu'}\left[\sum_{t\in[T]}\bI(z_t\ne k')\right]\left(\Delta_1-\max_{k\ne k'}\{\mu_{1,k}\}\right).\]
This means choosing any action other than $k'$ for one period will incur at least \[\Delta_1-\max_{k\ne k'}\{\mu_{1,k}\}\ge\Delta_1-\frac{\gap}{d+1}\ge\frac{1}{4}\] expected revenue loss compared with $\mathsf{J}^{\mathsf{DLP}_{\bmu'}}$. Since $\mathcal{E}_{k'}[1:t_1-1]$ indicates that the policy does not choose $k'$ for at least $t_1-1$ rounds, by the previous three inequalities in this paragraph, we have
\begin{align*}
R^{\phi}(T)\ge R_s^{\phi}(T)&= \sup_{\bQ}\left\{\Rev_{\bQ}(\pi_\bQ^*[s])-\Rev_{\bQ}(\phi)\right\}\\
&\ge \Rev_{\bQ_{\bmu'}}(\pi_{\bQ_{\bmu'}}^*[s])-\Rev_{\bQ_{\bmu'}}(\phi)\\
&=\left(\mathsf{J}^{\mathsf{DLP}_{\bmu'}}-\Rev_{\bQ_{\bmu'}}(\phi)\right)-\left(\mathsf{J}^{\mathsf{DLP}_{\bmu'}}-\Rev_{\bQ_{\bmu'}}(\pi_{\bQ_{\bmu'}}^*[s])\right)\\
&\ge \frac{1}{4}\mathbb{E}_{\bmu'}^\phi\left[\sum_{t\in[T]}\bI(z_t\ne k')\right]-\left(\mathsf{J}^{\mathsf{DLP}_{\bmu'}}-\Rev_{\bQ_{\bmu'}}(\pi_{\bQ_{\bmu'}}^*[s])\right)\\
&\ge 
\frac{1}{4}\Prob_{\bmu'}^\phi (\mathcal{E}_{k'}[1:t_1-1])\left[(t_1-1)\right]-\left(\mathsf{J}^{\mathsf{DLP}_{\bmu'}}-\Rev_{\bQ_{\bmu'}}(\pi_{\bQ_{\bmu'}}^*[s])\right)\\
&\ge\frac{t_1-1}{4K(\nu(s,d)+1)}-\left(\mathsf{J}^{\mathsf{DLP}_{\bmu'}}-\Rev_{\bQ_{\bmu'}}(\pi_{\bQ_{\bmu'}}^*[s])\right)\\
&\ge\frac{K^{-1/(2-2^{-\nu(s,d)})}}{8(\nu(s,d)+1)}T^{1/(2-2^{-\nu(s,d)})}-\left(\mathsf{J}^{\mathsf{DLP}_{\bmu'}}-\Rev_{\bQ_{\bmu'}}(\pi_{\bQ_{\bmu'}}^*[s])\right).
\end{align*}

Now we bound the $\left(\mathsf{J}^{\mathsf{DLP}_{\bmu'}}-\Rev_{\bQ_{\bmu'}}(\pi_{\bQ_{\bmu'}}^*[s])\right)$ term. Let $\pi$ be the Tweaked LP policy suggested by Algorithm \ref{alg:NRMUB}. By viewing $\mathcal{I}_{\bmu'}$ as an \stp instance and applying Proposition~\ref{prop:stpUB} and (\ref{eq:known-ubg}), we have $\pi\in\Pi[s]$ and 
\[
\mathsf{J}^{\mathsf{DLP}_{\bmu'}}-\Rev_{\bQ_{\bmu'}}(\pi)\le   \left(\frac{2}{T/2}\sqrt{T\log T}+\frac{d}{T^2}\right) \mathsf{J}^{\mathsf{DLP}_{\bmu'}}\le4\sqrt{T\log T}+ d/T,
\]
which indicates
\begin{equation}\label{eq:knowngap}
\mathsf{J}^{\mathsf{DLP}_{\bmu'}}-\Rev_{\bQ_{\bmu'}}(\pi_{\bQ_{\bmu'}}^*[s])\le\mathsf{J}^{\mathsf{DLP}_{\bmu'}}-\Rev_{\bQ_{\bmu'}}(\pi)\le4\sqrt{T\log T}+ 1.
\end{equation}
Therefore, we have
\[R^{\phi}(T)\ge R_s^{\phi}(T)\ge\frac{K^{-1/(2-2^{-\nu(s,d)})}}{8(\nu(s,d)+1)}T^{1/(2-2^{-\nu(s,d)})}-4\sqrt{T\log T}-1.\]

\textbf{Case 2: $ j^*\in [2:\nu(s,d)]$.}

Since $\mathbb{P}_{{\bmu}}^{\phi}(E_{j^*})=\mathbb{P}_{\bmu}^{\phi}(\tau_{j^*-1}\le t_{j^*-1},\tau_{j^*}>t_{j^*})\ge  1/(\nu(s,d)+1)$ and
$$
\mathbb{P}_{\bmu}^{\phi}(\tau_{j^*-1}\le t_{j^*-1},\tau_{j^*}>t_{j^*})=\sum_{k=1}^{K}\mathbb{P}_{\bmu}^{\phi}(\tau_{j^*-1}\le t_{j^*-1},\tau_{j^*}>t_{j^*},z_{\tau_j^*}=k),
$$
we know that there exists $k'\in[K]$ such that
$$
\mathbb{P}_{\bmu}^{\phi}(\tau_{j^*-1}\le t_{j^*-1},\tau_{j^*}>t_{j^*},z_{\tau_j^*}=k')\ge \frac{\mathbb{P}_{{\bmu}}^{\phi}(E_{j^*})}{K}\ge \frac{1}{K(\nu(s,d)+1)}.
$$
When $\tau_{j^*}<\infty$, since $\tau_{j^*}$ is the first time that all actions in $[K]$ has been chosen in $[\tau_{j^*-1}:\tau_{j^*}]$, the event $\{z_{\tau_{j^*}}=k'\}$ must imply the event $\{k'\text{ was not chosen in }[\tau_{j^*-1}:\tau_{j^*}-1]\}$. When $\tau_{j^*}=\infty$, since $z_{\tau_{j^*}}=z_\infty$ only takes values in $S_{\times}$ (which is the set of actions that were not chosen in $[\tau_{j^*-1}:T]$), the event $\{z_{\tau_{j^*}}=k'\}$ must imply the event $\{k'\text{ was not chosen in }[\tau_{j^*-1}:T]\}$. Thus, no matter whether $\tau_{j^*}<\infty$ or not, the event $\{\tau_{j^*-1}\le t_{j^*-1},\tau_{j^*}>t_{j^*},z_{\tau_j^*}=k'\}$ must imply the event $\mathcal{E}_{k'}[t_{j^*-1}:t_{j^*}]:=\{k'\text{ was not chosen in }[t_{j^*-1}:t_{j^*}]\}$. Therefore, we have
\begin{equation*}
    \Prob_{\bmu}^{\phi}(\mathcal{E}_{k'}[t_{j^*-1}:t_{j^*}])\ge \mathbb{P}_{\bmu}^{\phi}(\tau_{j^*-1}\le t_{j^*-1},\tau_{j^*}>t_{j^*},z_{\tau_j^*}=k')\ge \frac{1}{K(\nu(s,d)+1)}.
\end{equation*}
Meanwhile, the occurrence of the event $\mathcal{E}_{k'}[t_{j^*-1}:t_{j^*}]$ is independent of random matrix $(X_{\bmu}^{t}(k'))_{t\in[t_{j^*-1}:t_{j^*}]}$ and random matrices $(X_{\bmu}^{t}(k))_{t\in[t_{j^*+1}:T]}$ for all $k\in[K]$, i.e., the occurrence of the event $\mathcal{E}_{k'}[t_{j^*-1}:t_{j^*}]$ only depends on  random matrix $(X_{\bmu}^{t}(k'))_{t\in[1:t_{j^*-1}-1]}$, random matrices $(X_{\bmu}^{t}(k))_{t\in[1:t_{j^*}]}$ for $k\ne k'$, and policy $\phi$. Let $\mathbb{Q}_{\bmu}^\phi$ be the probability measure induced by the following random variables:
\begin{itemize}
    \item random matrix $(X_{\bmu}^{t}(k'))_{t\in[1:t_{j^*-1}-1]}$ and random matrices $(X_{\bmu}^{t}(k))_{t\in[1:t_{j^*}]}$ for $k\ne k'$,
    \item  policy $\phi$'s random actions $(z_t)_{t\in[1:t_1-1]}$ under instance $\mathcal{I}_{\bmu}$.
\end{itemize}
We have
\begin{equation}\label{eq:app3}
    \mathbb{Q}_{\bmu}^\phi (\mathcal{E}_{k'}[t_{j^*-1}:t_{j^*}])=\Prob_{\bmu}^{\phi}(\mathcal{E}_{k'}[t_{j^*-1}:t_{j^*}])\ge \frac{1}{K(\nu(s,d)+1)}.
\end{equation}

We now consider a new instance ${\mathcal{I}_{\bmu'}}$ with $\mu'_{1,k'}=\mu_{1,k'}+\Delta_{j^*}$, $\mu'_{2,k}=0$, and $\mu'_{i,k}=\mu_{i,k}$ for all $(i,k)\in[2]\times([K]\setminus\{k'\})$. Again, the occurrence of the event $\mathcal{E}_{k'}[t_{j^*-1}:t_{j^*}]$ is independent of random matrix  $(X_{\bmu'}^{t}(k'))_{t\in[t_{j^*-1}:t_{j^*}]}$ and random matrices $(X_{\bmu'}^{t}(k))_{t\in[t_{j^*+1}:T]}$ for all $k\in[K]$. Let $\mathbb{Q}_{\bmu'}^\phi$ be the probability measure induced by the following random variables:
\begin{itemize}
    \item random matrix $(X_{\bmu'}^{t}(k'))_{t\in[1:t_{j^*-1}-1]}$ and random matrices $(X_{\bmu'}^{t}(k))_{t\in[1:t_{j^*}]}$ for $k\ne k'$,
    \item  policy $\phi$'s random actions $(z_t)_{t\in[1:t_1-1]}$ under instance $\mathcal{I}_{\bmu}$.
\end{itemize}
We have
\begin{equation}\label{eq:app4}
   \mathbb{Q}_{\bmu'}^\phi (\mathcal{E}_{k'}[t_{j^*-1}:t_{j^*}])=\Prob_{\bmu'}^{\phi}(\mathcal{E}_{k'}[t_{j^*-1}:t_{j^*}]).
\end{equation}

We now try to bound the difference between the $\mathbb{Q}_{\bmu}^\phi (\mathcal{E}_{k'}[t_{j^*-1}:t_{j^*}])$ and the $\mathbb{Q}_{\bmu'}^\phi (\mathcal{E}_{k'}[t_{j^*-1}:t_{j^*}])$. Let $\mathbb{Q}_{\bmu}$ be the (policy-independent) probability measure induced by
\begin{itemize}
    \item random matrix $(X_{\bmu}^{t}(k'))_{t\in[1:t_{j^*-1}-1]}$ and random matrices $(X_{\bmu}^{t}(k))_{t\in[1:t_{j^*}]}$ for $k\ne k'$,
\end{itemize}
and $\mathbb{Q}_{\bmu'}$ be the (policy-independent) probability measure induced by
\begin{itemize}
    \item random matrix $(X_{\bmu'}^{t}(k'))_{t\in[1:t_{j^*-1}-1]}$ and random matrices $(X_{\bmu'}^{t}(k))_{t\in[1:t_{j^*}]}$ for $k\ne k'$.
\end{itemize}
Since the KL divergence is additive to for independent distributions, we have
\begin{align}\label{eq:klbound}
    &~~~D_{\rm{KL}}\left(\mathbb{Q}_{\bmu}~{\huge{\parallel}}~\mathbb{Q}_{\bmu'}\right)\notag\\
    &=\sum_{t\in[1:t_{j^*-1}-1]}D_{\rm{KL}}\left(\mathbb{P}_{X_{\bmu}^t(k')}~{\huge{\parallel}}~\mathbb{P}_{X_{\bmu'}^t(k')}\right)+\sum_{k\in[K]\setminus\{k'\}}\sum_{t\in[1:t_{t_{j^*}}]}D_{\rm{KL}}\left(\mathbb{P}_{X_{\bmu}^t(k)}~{\huge{\parallel}}~\mathbb{P}_{X_{\bmu'}^t(k)}\right)\notag\\
    &=\sum_{t\in[1:t_{j^*-1}-1]}D_{\rm{KL}}\left(\mathbb{P}_{X_{\bmu}^t(k')}~{\huge{\parallel}}~\mathbb{P}_{X_{\bmu'}^t(k')}\right)+0\notag\\
    &=\sum_{t\in[1:t_{j^*-1}-1]}\sum_{j\in[n]}D_{\rm{KL}}\left(\mathbb{P}_{Q_{j,k';\bmu}^t}~{\huge{\parallel}}~\mathbb{P}_{Q_{j,k';\bmu'}^t}\right)\notag\\
    &=\sum_{t\in[1:t_{j^*-1}-1]}\sum_{j\in[n]}D_{\rm{KL}}\left({\mathsf{Ber}(q_{j,k';\bmu})}~{\huge{\parallel}}~\mathsf{Ber}(q_{j,k';\bmu'})\right)\notag\\
    &=\sum_{t\in[1:t_{j^*-1}-1]}\sum_{j:q_{j,k';\bmu}\ne q_{j,k';\bmu'}}D_{\rm{KL}}\left({\mathsf{Ber}(q_{j,k';\bmu})}~{\huge{\parallel}}~\mathsf{Ber}(q_{j,k';\bmu'})\right)\notag\\
    &=(t_{j^*-1}-1)\left[{D_{\rm{KL}}\left({\mathsf{Ber}(\frac{1}{2}+\mu_{1,k'})}{\huge{\parallel}}\mathsf{Ber}(\frac{1}{2}+\mu'_{1,k'})\right)+D_{\rm{KL}}\left({\mathsf{Ber}(\frac{1}{2}+\gap)}{\huge{\parallel}}\mathsf{Ber}(\frac{1}{2})\right)+D_{\rm{KL}}\left({\mathsf{Ber}(\frac{1}{2}-\gap)}{\huge{\parallel}}\mathsf{Ber}(\frac{1}{2})\right)}\right]\notag\\
    &\le(t_{j^*-1}-1)\left[\frac{1}{1/2-\Delta_{j^*}}\left(\Delta_{j^*}\right)^2+\frac{2}{1/2-\gap}\gap^2\right]\notag\\
    &\le (t_{j^*-1}-1)\left[\frac{12}{5}\left(\Delta_{j^*}\right)^2+\frac{6}{5}\left(\Delta_{j^*}\right)^2\right]\notag\\
    &\le \frac{18}{5}(t_{j^*-1}-1)\left(\Delta_{j^*}\right)^2
\end{align}
where the first inequality follows from Lemma \ref{lm:kl}, and the second inequality follows from $2\gap\le\Delta_{j^*}\le\frac{1}{2K(\nu(s,d)+1)}\le\frac{1}{2\times2\times3}$ (note that Case 2 is meaningful only when $\nu(s,d)\ge2$.)
We further have
\begin{align*}
&\mathbb{Q}_{\bmu}^\phi (\mathcal{E}_{k'}[t_{j^*-1}:t_{j^*}]) - \mathbb{Q}_{\bmu'}^\phi (\mathcal{E}_{k'}[t_{j^*-1}:t_{j^*}])|\\
\le&{D_{\rm{TV}}}\left(\mathbb{Q}_{\bmu}^\phi~{\huge{\parallel}}~\mathbb{Q}_{\bmu'}^\phi\right)\\
\le&\sqrt{\frac{1}{2}D_{\rm{KL}}\left(\mathbb{Q}_{\bmu}^\phi~{\huge{\parallel}}~\mathbb{Q}_{\bmu'}^\phi\right)}\\
\le&\sqrt{\frac{1}{2}D_{\rm{KL}}\left(\mathbb{Q}_{\bmu}~{\huge{\parallel}}~\mathbb{Q}_{\bmu'}\right)}
\\
\le&\sqrt{\frac{1}{2}\left[\frac{18}{5}(t_{j^*-1}-1){\left(\Delta_{j^*}\right)^2}\right]}\\
\le&\frac{3\sqrt{t_{j^*-1}}\Delta_{j^*}}{2}\le\frac{3}{4K(\nu(s,d)+1)},
\end{align*}
where the first inequality is by the definition of total variation distance of two probability measures, the second inequality is by Pinsker's inequality in information theory, the third inequality is by the data-processing inequality in information theory, and the fourth inequality is by (\ref{eq:klbound}).

Combining the above inequality with (\ref{eq:app3}) and (\ref{eq:app4}), we have
$$
\Prob_{\bmu'}^{\phi}(\mathcal{E}_{k'}[t_{j^*-1}:t_{j^*}])\ge\Prob_{\bmu}^{\phi}(\mathcal{E}_{k'}[t_{j^*-1}:t_{j^*}])-\frac{3}{4K(\nu(s,d)+1)}\ge \frac{1}{4K(\nu(s,d)+1)}.
$$
However, under instance $\mathcal{I}_{\bmu'}$, the optimal solution to $\mathsf{DLP}_{\bmu'}$ satisfies $x_{k'}=T$, and $\mathsf{J}^{\mathsf{DLP}_{\bmu'}}=(1/2+\mu_{1,k'}+\Delta_{j^*})T$. 
By Lemma \ref{lem:reg-dec}, 
\begin{align*}\mathsf{J}^{\mathsf{DLP}_{\bmu'}}-\Rev_{\bQ_{\bmu'}}(\phi)&=(1/2+\mu_{1,k'}+\Delta_{j^*})T-\Rev_{\bQ_{\bmu'}}(\phi)\\
&\ge\mathbb{E}^\phi_{\bmu'}\left[\sum_{t\in[T]}\bI(z_t\ne k')\right]\left(\mu_{1,k'}+\Delta_{j^*}-\max_{k\ne k'}\{\mu_{1,k}\}\right).\end{align*}
This means choosing any action other than $k'$ for one period will incur at least 
\[
(\mu_{1,k'}+\Delta_{j^*})-\max_{k\ne k'}\{\mu_{1,k}\}\ge \Delta_{j^*}-\frac{2}{d+1}\gap\ge\frac{\Delta_{j^*}}{2}
\]
expected revenue loss compared with $\mathsf{J}^{\mathsf{DLP}_{\bmu'}}$. Since $\mathcal{E}_{k'}[t_{j^*-1}:t_{j^*}]$ indicates that the policy does not choose $k'$ for at least $t_{j^*}-t_{j^*-1}+1$ rounds, by the previous three inequalities in this paragraph, we have
\begin{align*}
    &~~~\mathsf{J}^{\mathsf{DLP}_{\bmu'}}-\Rev_{\bQ_{\bmu'}}(\phi)\\
&\ge
\Prob_{\bmu'}^\phi (\mathcal{E}_{k'}[t_{j^*-1}:t_{j^*}])(t_{j^*}-t_{j^*-1}+1){\Delta_{j^*}}/2\\
&\ge\frac{1}{4K(\nu(s,d)+1)}\left(K(T/K)^{\frac{2-2^{1-j^*}}{2-2^{-\nu(s,d)}}}-K(T/K)^{\frac{2-2^{2-j^*}}{2-2^{-\nu(s,d)}}}\right)\frac{K^{-\frac{1}{2}}\left(K/T\right)^{\frac{1-2^{1-j^*}}{2-2^{-\nu(s,d)}}}}{4K(\nu(s,d)+1)}\\
&\ge\frac{K^{-\frac{3}{2}}}{16(\nu(s,d)+1)^2}\left((T/K)^{\frac{1}{2-2^{-\nu(s,d)}}}-(T/K)^{\frac{1-2^{1-j^*}}{2-2^{-\nu(s,d)}}}\right)\\
&\ge\frac{K^{-\frac{3}{2}-\frac{1}{2-2^{-\nu(s,d)}}}T^{\frac{1}{2-2^{-\nu(s,d)}}}}{16(\nu(s,d)+1)^2}\left(1-(T/K)^{\frac{-2^{1-j^*}}{2-2^{-\nu(s,d)}}}\right)\\
&\ge\frac{K^{-\frac{3}{2}-\frac{1}{2-2^{-\nu(s,d)}}}T^{\frac{1}{2-2^{-\nu(s,d)}}}}{16(\nu(s,d)+1)^2}\left(1-(T/K)^{\frac{-2^{1-\nu(s,d)}}{2-2^{-\nu(s,d)}}}\right)\\
&\ge\frac{K^{-\frac{3}{2}-\frac{1}{2-2^{-\nu(s,d)}}}T^{\frac{1}{2-2^{-\nu(s,d)}}}}{16(\nu(s,d)+1)^2}\left(1-(T/K)^{-2^{-\nu(s,d)}}\right).
\end{align*}
When $\nu(s,d)\le\log_2\log_2(T/K)$, we have
$$
(T/K)^{-2^{-\nu(s,d)}}\le (T/K)^{-\frac{1}{\log_2(T/k)}}=\frac{1}{(T/K)^{\log_{T/K}(2)}}=\frac{1}{2}.
$$
Thus we know that
$$
\mathsf{J}^{\mathsf{DLP}_{\bmu'}}-\Rev_{\bQ_{\bmu'}}(\phi)\ge \frac{K^{-\frac{3}{2}-\frac{1}{2-2^{-\nu(s,d)}}}T^{\frac{1}{2-2^{-\nu(s,d)}}}}{16(\nu(s,d)+1)^2}\left(1-(T/K)^{-2^{-\nu(s,d)}}\right)\ge \frac{K^{-\frac{3}{2}-\frac{1}{2-2^{-\nu(s,d)}}}}{32(\nu(s,d)+1)^2}T^{\frac{1}{2-2^{-\nu(s,d)}}}
$$
when $\nu(s,d)\le\log_2\log_2(T/k)$. 

By analysis identical to the derivation of (\ref{eq:knowngap}) in Case 1, we know that
\[\mathsf{J}^{\mathsf{DLP}_{\bmu'}}-\Rev_{\bQ_{\bmu'}}(\pi_{\bQ_{\bmu'}}^*[s])\le4\sqrt{T\log T}+ 1.\] Thus we have
\begin{align*}
R^{\phi}(T)\ge R_s^{\phi}(T)&= \sup_{\bQ}\left\{\Rev_{\bQ}(\pi_\bQ^*[s])-\Rev_{\bQ}(\phi)\right\}\\
&\ge \Rev_{\bQ_{\bmu'}}(\pi_{\bQ_{\bmu'}}^*[s])-\Rev_{\bQ_{\bmu'}}(\phi)\\
&=\left(\mathsf{J}^{\mathsf{DLP}_{\bmu'}}-\Rev_{\bQ_{\bmu'}}(\phi)\right)-\left(\mathsf{J}^{\mathsf{DLP}_{\bmu'}}-\Rev_{\bQ_{\bmu'}}(\pi_{\bQ_{\bmu'}}^*[s])\right)\\
&\ge
\left(\mathsf{J}^{\mathsf{DLP}_{\bmu'}}-\Rev_{\bQ_{\bmu'}}(\phi)\right)-4\sqrt{T\log T}-1\\
&\ge\frac{K^{-\frac{3}{2}-\frac{1}{2-2^{-\nu(s,d)}}}}{32(\nu(s,d)+1)^2}T^{\frac{1}{2-2^{-\nu(s,d)}}}-4\sqrt{T\log T}-1
\end{align*}
when $\nu(s,d)\le\log_2\log_2(T/k)$.

\textbf{Case 3: $j^*=\nu(s,d)+1.$}

Since $\mathbb{P}_{{\bmu}}^{\phi}(E_{\nu(s,d)+1})=\mathbb{P}_{\bmu}^{\phi}(\tau_{\nu(s,d)}\le t_{\nu(s,d)})\ge  1/(\nu(s,d)+1)$ and
$$
\mathbb{P}_{\bmu}^{\phi}(\tau_{\nu(s,d)}\le t_{\nu(s,d)})=\sum_{k=1}^{K}\mathbb{P}_{\bmu}^{\phi}(\tau_{\nu(s,d)}\le t_{\nu(s,d)},z_{\tau_{\nu(s,d)+1}}=k),
$$
we know that there exists $k'\in[K]$ such that
$$
\mathbb{P}_{\bmu}^{\phi}(\tau_{\nu(s,d)}\le t_{\nu(s,d)},z_{\tau_{\nu(s,d)+1}}=k')\ge \frac{\mathbb{P}_{{\bmu}}^{\phi}(E_{\nu(s,d)+1})}{K}\ge \frac{1}{K(\nu(s,d)+1)}.
$$
Define $\hat{t}=\left\lfloor \frac{T-t_{\nu(s,d)}}{8(d+1)}\right\rfloor$. 
Then either
\begin{equation}\label{eq:app5}
    \mathbb{P}_{\bmu}^{\phi}\left(\tau_{\nu(s,d)}\le t_{\nu(s,d)},\tau_{\nu(s,d)+1}>t_{\nu(s,d)}+\hat{t},z_{\tau_{\nu(s,d)+1}}=k'\right)\ge \frac{1}{2K(\nu(s,d)+1)},
\end{equation}
or
\begin{equation}\label{eq:app6}
    \mathbb{P}_{\bmu}^{\phi}\left(\tau_{\nu(s,d)}\le t_{\nu(s,d)},\tau_{\nu(s,d)+1}\le t_{\nu(s,d)}+\hat{t},z_{\tau_{\nu(s,d)+1}}=k'\right)\ge \frac{1}{2K(\nu(s,d)+1)}.
\end{equation}

If (\ref{eq:app5}) holds true, then we consider a new instance $\mathcal{I}_{\bmu'}$ with $\mu'_{1,k'}=\mu_{1,k'}+\Delta_{\nu(s,d)+1}$, $\mu'_{2,k'}=0$, and $\mu'_{i,k}=\mu_{i,k}$ for all $(i,k)\in[2]\times([K]\setminus\{k'\})$. Define the event $\mathcal{E}_{k'}[t_{\nu(s,d)}:t_{\nu(s,d)}+\hat{t}]:=\{k'\text{ was not chosen in }[t_{\nu(s,d)}:t_{\nu(s,d)}+\hat{t}]\}$. From (\ref{eq:app5}) we know that $\mathbb{P}_{\bmu}^{\phi}(\mathcal{E}_{k'}[t_{\nu(s,d)}:t_{\nu(s,d)}+\hat{t}])\ge 1/(2K(\nu(s,d)+1))$. Using arguments analogous to Case 2, we can derive that
$$
\mathbb{P}_{\bmu'}^{\phi}(\mathcal{E}_{k'}[t_{\nu(s,d)}:t_{\nu(s,d)}+\hat{t}])\ge\mathbb{P}_{\bmu}^{\phi}(\mathcal{E}_{k'}[t_{\nu(s,d)}:t_{\nu(s,d)}+\hat{t}])-\frac{3}{8K(\nu(s,d)+1)}\ge\frac{1}{8K(\nu(s,d)+1)}
$$
and
\begin{align*}
R^{\phi}(T)\ge R_s^\phi(T)&\ge \Prob_{\bmu'}^\phi (\mathcal{E}_{k'}[t_{\nu(s,d)}:t_{\nu(s,d)}+\hat{t}])(\hat{t}+1){\Delta_{\nu(s,d)+1}}/2-4\sqrt{T\log T}-1\\
&\ge \frac{K^{-\frac{3}{2}-\frac{1}{2-2^{-\nu(s,d)}}}T^{\frac{1}{2-2^{-\nu(s,d)}}}}{512(d+1)(\nu(s,d)+1)^2}\left(1-(T/K)^{-2^{-\nu(s,d)-1}}\right)-4\sqrt{T\log T}-1\\
&\ge\frac{K^{-\frac{3}{2}-\frac{1}{2-2^{-\nu(s,d)}}}}{768(d+1)(\nu(s,d)+1)^2}T^{\frac{1}{2-2^{-\nu(s,d)}}}-4\sqrt{T\log T}-1
\end{align*}
when $\nu(s,d)+1\le\log_2\log_3(T/K)$.

Now we consider the case that (\ref{eq:app6}) holds true.  Let $\mathcal{E}_{k'}$ denote the event $\{\tau_{\nu(s,d)}\le t_{\nu(s,d)},\tau_{\nu(s,d)+1}\le t_{\nu(s,d)}+\hat{t},z_{\tau_{\nu(s,d)+1}}=k'\}$. According to Lemma \ref{lm:switch}, the event $\{\tau_{\nu(s,d)}\le t_{\nu(s,d)}\}$ implies that the number of switches occurs in $[\tau_{\nu(s,d)}:T]$ is no more than $K+d-1$. Meanwhile, the event $\{\tau_{\nu(s,d)+1}\le t_{\nu(s,d)}+\hat{t}<\infty\}$ implies that the number of switches occurs in $[\tau_{\nu(s,d)}:\tau_{\nu(s,d)+1}]$ is at least $K-1$. As a result, the event $\{\tau_{\nu(s,d)}\le t_{\nu(s,d)},\tau_{\nu(s,d)+1}\le t_{\nu(s,d)}+\hat{t}\}$ implies that there are no more than $d$ switches occurring in $[\tau_{\nu(s,d)+1}:T]$.  

Meanwhile, since the event $\mathcal{E}_{k'}$ implies that action $k'$ is not chosen in $[t_{\nu(s,d)}:\tau_{\nu(s,d)+1}-1]$, the occurrence of the event $\mathcal{E}_{k'}$ only depends on  random matrix $(X_{\bmu}^{t}(k'))_{t\in[1:t_{\nu(s,d)}]}$, random matrices $(X_{\bmu}^{t}(k))_{t\in[1:t_{\nu(s,d)}+\hat{t}]}$ for $k\ne k'$, and policy $\phi$. Consider a new instance $\mathcal{I}_{\bmu'}$ with $\mu'_{1,k'}=\mu_{1,k'}-\Delta_{\nu(s,d)+1}$, $\mu'_{2,k'}=\mu_{2,k'}$, and $\mu'_{i,k}=\mu_{i,k}$ for all $(i,k)\in[2]\times([K]\setminus\{k'\})$. Using arguments analogous to Case 2, we can derive that
$$
\mathbb{P}_{\bmu'}^{\phi}(\mathcal{E}_{k'})\ge\mathbb{P}_{\bmu}^{\phi}(\mathcal{E}_{k'})-\frac{3}{10K(\nu(s,d)+1)}\ge\frac{1}{5K(\nu(s,d)+1)}.
$$

Let $\mathcal{E}'$ denote the event $\{\text{no switch occurs between period }\tau_{\nu(s,d)+1}\text{ and period }t_{\nu(s,d)}+2\hat{t}\}$, and let $\overline{\mathcal{E}'}$ denote its complement. We know that either 
\begin{equation}\label{eq:app7}
    \mathbb{P}_{\bmu'}^{\phi}\left(\mathcal{E}_{k'}\cap \mathcal{E'}\right)\ge \frac{1}{10K(\nu(s,d)+1)},
\end{equation}
or
\begin{equation}\label{eq:app8}
    \mathbb{P}_{\bmu'}^{\phi}\left(\mathcal{E}_{k'}\cap \overline{\mathcal{E'}}\right)\ge \frac{1}{10K(\nu(s,d)+1)}.
\end{equation}


Under instance $\mathcal{I}_{\bmu'}$, for any $l\in[d+1]$, we can partition the action set $[K]$ into $(d+1)$ disjoint subsets:
\[
A_l=\{k\in[K]\mid k\%(d+1)=l\%(d+1)\}.
\]
Let $l':=k'\%(d+1)$. We know that for any $l\in[d+1]\setminus\{l'\}$, all the actions in $A_l$ are the same (i.e., they have the same reward and cost distributions), except for $l'$, whose associated action set $A_{l'}$ includes an action $k'$ which is different from the rest of actions in $A_{l'}\setminus\{k'\}$ (note that $A_{l'}\setminus\{k'\}$ is guaranteed to be non-empty because of $K\ge 2(d+1)$). Since action $k'$ is strictly dominated by all other actions in $A_{l'}\setminus\{k'\}$ (i.e., compared with other actions in $A_{l'}\setminus\{k'\}$, action $k'$ has the same cost distributions but has a strictly worse reward distribution), we know that any optimal solution to $\mathsf{DLP}_{\bmu'}$ must have its $k'$-th coordinate being 0 (i.e., action $k'$ should always be avoided in the distributionally-known case). By step 1 of the proof of Lemma \ref{lem:WorstCaseLP}, we know that there exists an optimal solution to $\mathsf{DLP}_{\bmu'}$, denoted as $\bx^*$, which satisfies the following property:
\[
\forall l\in[d+1],~~\sum_{k\in A_l} \bx^*_k=\frac{T}{d+1}.
\]
Moreover, $x^*_{k'}=0$. 
We know that  $\mathsf{J}^{\mathsf{DLP}_{\bmu'}}=\frac{T}{2}=\mathsf{J}^{\mathsf{DLP}_{\bmu}}$. 

If (\ref{eq:app7}) holds, since $\mathcal{E}_{k'}\cap \mathcal{E'}$ implies the event \[\{\text{action }k'\text{ is continuously chosen in every period from period } t_{\nu(s,d)}+\hat{t}\text{ to period }t_{\nu(s,d)}+2\hat{t}\},\] we know that action $k'$ is continuously chosen for at least $\hat{t}+1$ periods with probability at least $1/(10K(\nu(s,d)+1))$. However, since the expected reward of action $k'$ is only $\frac{1}{2}+\mu'_{1,k'}=\frac{1}{2}+\mu_{1,k'}-\Delta_{\nu(s,d)+1}$, applying Lemma~\ref{lem:reg-dec-dlp}, 
we know that choosing action $k'$ for one period will incur at least
$\Delta_{\nu(s,d)+1}$ 
expected revenue loss compared with $\mathsf{J}^{\mathsf{DLP}_{\bmu'}}=\mathsf{J}^{\mathsf{DLP}_{\bmu}}=\frac{T}{2}$. Using arguments analogous to Case 2, we have \begin{align*}
R^{\phi}(T)\ge R_s^\phi(T)&\ge \Prob_{\bmu'}^\phi (\mathcal{E}_{k'}\cap\mathcal{E}')(\hat{t}+1){\Delta_{\nu(s,d)+1}}-4\sqrt{T\log T}-1\\
&\ge \frac{K^{-\frac{3}{2}-\frac{1}{2-2^{-\nu(s,d)}}}T^{\frac{1}{2-2^{-\nu(s,d)}}}}{320(d+1)(\nu(s,d)+1)^2}\left(1-(T/K)^{-2^{-\nu(s,d)-1}}\right)-4\sqrt{T\log T}-1\\
&\ge\frac{K^{-\frac{3}{2}-\frac{1}{2-2^{-\nu(s,d)}}}}{480(d+1)(\nu(s,d)+1)^2}T^{\frac{1}{2-2^{-\nu(s,d)}}}-4\sqrt{T\log T}-1
\end{align*}
when $\nu(s,d)+1\le\log_2\log_3(T/K)$.

Now we consider the case that (\ref{eq:app8}) holds true. Since $\mathcal{E}_{k'}$ implies that there are no more than $d$ switches occurring in $[\tau_{\nu(s,d)+1}:T]$ and $\mathcal{E}'$ implies that there is at least one switch occurring in $[\tau_{\nu(s,d)+1}:t_{\nu(s,d)}+2\hat{t}]$, the event $\mathcal{E}_{k'}\cap\mathcal{E}'$ implies that there are no more than $d-1$ switches occurring in $[t_{\nu(s,d)}+2\hat{t}:T]$ (and this event can only happen when $d\ge 1$). When $\nu(s,d)+1\le\log_2\log_{4d+3}(T/K)$, we have
\[
\frac{t_{\nu(s,d)}}{T}=(T/K)^{-2^{-\nu(s,d)-1}}\le\frac{1}{4d+3}
\]
and
\[
t_{\nu(s,d)}+2\hat{t}\le t_{\nu(s,d)}+\frac{T-t_{\nu(s,d)}}{4(d+1)}=\frac{T+(4d+3)t_{\nu(s,d)}}{4(d+1)}\le\frac{T}{2(d+1)}.
\]
Thus when $\nu(s,d)+1\le\log_2\log_{4d+3}(T/K)$, the event $\mathcal{E}_{k'}\cap\mathcal{E}'$ implies that there are no more than $d-1$ switches occurring in $\left[\lfloor\frac{T}{2(d+1)}\rfloor:T\right]$, which further implies that there exists $l\in[d+1]$, such that no action in $A_l$ is chosen during $\left[\lfloor\frac{T}{2(d+1)}\rfloor:T\right]$ ---  consequently, the total number of periods that $\phi$ chooses actions in $A_l$ during $[1:T]$ is smaller than $\lfloor\frac{T}{2(d+1)}\rfloor$. Hence, (\ref{eq:app8}) implies that \[\Prob_{\bmu'}^\phi\left(\exists l\in[d+1]\text{ such that } \sum_{t\in[T]}\sum_{k\in A_l}\bI(z_t=k)< \left\lfloor\frac{T}{2(d+1)}\right\rfloor\right)\ge\frac{1}{10K(\nu(s,d)+1)}.\]
Let $\phi^\dagger$ be the comparator policy constructed based on $\phi$ and $\bmu'$ as described in the proof of Lemma~\ref{lem:reg-dec-dlp}. Since the action sequence generated by $\phi$ under $\cI_{\bmu'}$ is distributionally identical to the action sequence generated by $\phi^\dagger$ under $\cI_{\bmu}$, we have
\[\Prob_{\bmu}^{\phi^\dagger}\left(\exists l\in[d+1]\text{ such that } \sum_{t\in[T]}\sum_{k\in A_l}\bI(z_t=k)< \left\lfloor\frac{T}{2(d+1)}\right\rfloor\right)\ge\frac{1}{10K(\nu(s,d)+1)}.\]
This inequality immediately makes the condition described in Lemma \ref{lem:WorstCaseLPProb} holds, with $\zeta=\frac{1}{2}$,  $\xi=\frac{1}{10K(\nu(s,d)+1)}$, and $\bmu$ being precisely what is required in Lemma~\ref{lem:WorstCaseLPProb}. Applying Lemma \ref{lem:WorstCaseLPProb}, we know that\footnote{Note that we do not need to consider the case of $d=0$, as (\ref{eq:app8}) only holds when $d\ge1$.}
\begin{align*}
    \Rev_{\bQ_{\bmu}}(\phi^\dagger)&\le \mathsf{J}^{\mathsf{DLP}_{\bmu}} -\mathsf{J}^{\mathsf{DLP}_{\bmu}}\left(\frac{ \xi(1-\zeta)\gap}{d(d+1)^2}-{\frac{\xi }{\min_{i\in[d]}b_i} \sqrt{3K\log T}}T^{-\frac{1}{2}}\right)+\xi \sqrt{3KT\log T}
    \\
    &=\frac{T}{2}-\frac{T}{2}\frac{\gap }{20Kd(d+1)^2(\nu(s,d)+1)}+\frac{ \left(1/\underline{b}+2\right)\sqrt{3KT\log T}}{20K(\nu(s,d)+1)}\\
    &\le\frac{T}{2}-\frac{K^{-2/3-\frac{1}{2-2^{-\nu(s,d)}}}}{640d(d+1)^2(\nu(s,d)+1)^2}T^{\frac{1}{2-2^{-\nu(s,d)}}}+\frac{(1/\underline{b}+2)\sqrt{KT\log T}}{10K(\nu(s,d)+1)}.
\end{align*}
Furthermore, the last equality in the proof of Lemma~\ref{lem:reg-dec-dlp} implies that
\[
\Rev_{\bQ_{\bmu'}}(\phi)\le\Rev_{\bQ_{\bmu}}(\phi^\dagger)\le\frac{T}{2}-\frac{K^{-2/3-\frac{1}{2-2^{-\nu(s,d)}}}}{640d(d+1)^2(\nu(s,d)+1)^2}T^{\frac{1}{2-2^{-\nu(s,d)}}}+\frac{(1/\underline{b}+2)\sqrt{KT\log T}}{10K(\nu(s,d)+1)}.
\]
This implies that
\[
R^{\phi}(T)\ge R_s^\phi(T)\ge\frac{K^{-2/3-\frac{1}{2-2^{-\nu(s,d)}}}}{640(d+1)^3(\nu(s,d)+1)^2}T^{\frac{1}{2-2^{-\nu(s,d)}}}-\frac{(1/\underline{b}+2) \sqrt{KT\log T}}{10K(\nu(s,d)+1)}-4\sqrt{T\log T}-1
\]
when $\nu(s,d)+1\le\log_2\log_{4d+3}(T/K)$.

\textbf{Putting Everything Together.}


Combining Case 1, 2 and 3, we know that for all $d\ge 0, K\ge 2(d+1), T\ge 2K,s\ge0$ and $\phi\in\Phi[s]$, we have
\begin{align}\label{eq:lbcase1}
R^{\phi}(T)\ge R^{\phi}_s(T)&\ge \frac{1}{768}\frac{K^{-\frac{3}{2}-\frac{1}{2-2^{-\nu(s,d)}}}}{(d+1)^3(\nu(s,d)+1)^2}T^{\frac{1}{2-2^{-\nu(s,d)}}}-\left(\frac{1}{\underline{b}}+6\right)\sqrt{T\log T}\notag\\
&\ge \frac{1}{768}\frac{K^{-\frac{3}{2}-\frac{1}{2-2^{-\nu(s,d)}}}}{(d+1)^3(\log T)^2}T^{\frac{1}{2-2^{-\nu(s,d)}}}-\left(\frac{1}{\underline{b}}+6\right)\sqrt{T\log T}
\end{align}
when $\nu(s,d)+1\le\log_2\log_{4d+3}(T/K)$. 

Meanwhile, by considering the family of instances \[\{\mathcal{I}_{\bmu''}\mid \mu''_{1,k}\in[-1/2,1/2], \mu''_{2,k}=0, \forall k\in[K]\},\]
our \bnrm problem completely reduces to the classical $K$-armed (Bernoulli) bandit problem,  and the standard $\Omega(\sqrt{KT})$ lower bound for \mab tells that there exists a numerical constant $C_{\mab}>0$ such that
$$
R^\phi(T)\ge R_s^{\phi}(T)\ge C_{\mab}\sqrt{KT}
$$
always holds. 
When $\nu(s,d)+1>\log_2\log_{4d+3}(T/K)$, we have
\begin{align*}
\frac{K^{-\frac{3}{2}-\frac{1}{2-2^{-\nu(s,d)}}}}{(d+1)^3}T^{\frac{1}{2-2^{-\nu(s,d)}}}&=\frac{K(T/K)^{\frac{1}{2-2^{-\nu(s,d)}}}}{K^{5/2}(d+1)^3}\\
&\le \frac{K{((4d+3)T/K)^{\frac{1}{2}}}}{K^{5/2}(d+1)^3}\\
&\le 2\frac{\sqrt{KT}}{K^{5/2}(d+1)^{5/2}},
\end{align*}
thus 
\begin{align}\label{eq:lbcase2}
    R^\phi(T)\ge R_s^{\phi}(T)&\ge C_{\mab}\sqrt{KT}\ge \frac{C_{\mab}}{2(d+1)^{1/2}}T^{\frac{1}{2-2^{-\nu(s,d)}}}.
\end{align}
Combining (\ref{eq:lbcase1}) and (\ref{eq:lbcase2}), we know that no matter $\nu(s,d)+1\le\log_2\log_{4d+3}(T/K)$ or not, it always holds that
\[
R^{\phi}(T)\ge R^{\phi}_s(T)
\ge \min\{\frac{1}{768},C_{\mab}\}\frac{K^{-\frac{3}{2}-\frac{1}{2-2^{-\nu(s,d)}}}}{(d+1)^3(\log T)^2}T^{\frac{1}{2-2^{-\nu(s,d)}}}-\left(\frac{1}{\underline{b}}+6\right)\sqrt{T\log T}.
\]
Moreover, since we always have $
R^\phi(T)\ge R_s^{\phi}(T)\ge C_{\mab}\sqrt{KT}
$, it holds that
    \begin{align*}
    R^{\phi}(T)&\ge R^{\phi}_s(T)\\&\ge \frac{C_{\mab}\left(\min\{\frac{1}{768},C_{\mab}\}\frac{K^{-\frac{3}{2}-\frac{1}{2-2^{-\nu(s,d)}}}}{(d+1)^3(\log T)^2}T^{\frac{1}{2-2^{-\nu(s,d)}}}-\frac{(1+6\underline{b})\sqrt{T\log T}}{\underline{b}}\right)+\frac{(1+6\underline{b})\sqrt{\log T}}{\underline{b}}C_{\mab}\sqrt{KT}}{C_{\mab}+\frac{(1+6\underline{b})\sqrt{\log T}}{\underline{b}}}\\
    &\ge \frac{C_{\mab}\min\{\frac{1}{768},C_{\mab}\}}{C_{\mab}+\frac{(1+6\underline{b})\sqrt{\log T}}{\underline{b}}}\frac{K^{-\frac{3}{2}-\frac{1}{2-2^{-\nu(s,d)}}}}{(d+1)^3(\log T)^2}T^{\frac{1}{2-2^{-\nu(s,d)}}}\\
    &\ge \left(\min\{c\underline{b},c'\}\cdot{(d+1)^{-3}K^{-\frac{3}{2}-\frac{1}{2-2^{-\nu(s,d)}}}}{(\log T)^{-\frac{5}{2}}}\right)T^{\frac{1}{2-2^{-\nu(s,d)}}}
    \end{align*}
where $c,c'>0$ are some numerical constants completely determined by the numerical constant $C_{\mab}$.
$\hfill\Box$
\endproof

\proof{Proof of Theorem~\ref{thm:BlindNRMLB} when $\nu(s,d)=0$.}  There is nothing really different in the proof for this case (in fact, we write this case separately only for notational considerations). 
When $\nu(s,d)=0$, we have $s\le K+d-2$. Define the same $\bmu$  as in the previous proof, with $\gap=\Delta_1/4$. 
If $s\le d-1\le K-2$, then $\tau_1=\infty$, and $E_1=\{\tau_1> t_1\}$ happens almost surely. Using arguments analogous to Case 1 of the previous proof, we obtain a $T/(8K)$ lower bound for both $R^{\phi}(T)$ and $R^{\phi}_s(T)$. When $s\ge d$, we have
$
\Prob_{\bmu}^\phi(E_1)+\Prob_{\bmu}^\phi(\overline{E_1})=1,
$
thus either $\Prob_{\bmu}^\phi(E_1)\ge 1/2$, or $\Prob_{\bmu}^\phi(\overline{E_1})\ge 1/2$. The former case corresponds to Case 1 in the previous proof, and the latter case corresponds to Case 3 in the previous proof. By almost the same arguments, we have
\[
 R^{\phi}(T)\ge R^{\phi}_s(T)\ge \left(\min\{c\underline{b},c'\}\cdot{(d+1)^{-3}K^{-\frac{5}{2}}}{(\log T)^{-\frac{5}{2}}}\right)T
\]
where $c,c'>0$ are some numerical constants.
\Halmos
\endproof

\section{Proof of Theorem~\ref{thm:linearBNRM}}\label{app:linearBNRM}

\subsection{Preliminary Lemmas}

For any parameters $(\tilde{\bm{\alpha}}, \tilde{\beta}) \in \Theta \subseteq [-C_\theta,C_\theta]^{n \times (n+1)}$ and any price vector $\bm{p} \in P_c$, define
\begin{align}
J(\tilde{\bm{\alpha}},\tilde{\beta},\bm{p}) = \bm{p}^\top(\tilde{\bm{\alpha}}+\tilde{\beta}\bm{p}). \label{eqn:defn:JQP}
\end{align}

Recall that we have assumed in Section \ref{sec:LinearDemands} that, there exists a price vector $\bar{\bm{p}} \in P_c$ such that the average demand consumption strictly respects the resource constraints (under the true unknown coefficients $(\bm{\alpha},\beta)$), that is, there exists $\bar{\bm{p}} \in P_c$ and a positive constant $C_\delta > 0$ such that $A(\bm{\alpha} + \beta \bar{\bm{p}}) \leq \bm{B}/T - C_\delta \bm{1}_n$, where $\bm{1}_n$ stands for a length-$n$ vector with all elements equal to one. 
For the true unknown coefficients $(\bm{\alpha},\beta) \in \Theta \subseteq [-C_\theta,C_\theta]^{n \times (n+1)}$ and any  parameters $(\tilde{\bm{\alpha}},\tilde{\beta}) \in \Theta \subseteq [-C_\theta,C_\theta]^{n \times (n+1)}$, we define the following quantity.
\begin{align}
\gamma_{\text{lin}}(\tilde{\bm{\alpha}},\tilde{\beta}) = \frac{1}{C_\delta} \big( \amax \sqrt{n} \| \bm{\alpha} - \tilde{\bm{\alpha}} \|_2 + \amax \pmax n \| \beta - \tilde{\beta} \|_F \big) > 0, \label{eqn:gamma_lin}
\end{align}
where for any vector $\bm{\alpha}$ we let $\|\bm{\alpha}\|_2$ be the $\ell_2$ norm of the vector, and for any matrix $\beta$ we let $\|\beta\|_F = \big(\sum_{i=1}^n \sum_{j=1}^n \beta_{ij}^2\big)^{\frac{1}{2}}$ be the Frobenius norm of the matrix.

\begin{lemma}
\label{lem:PurturbedQP}
For the true unknown coefficients $(\bm{\alpha},\beta) \in \Theta \subseteq [-C_\theta,C_\theta]^{n \times (n+1)}$ and any  parameters $(\tilde{\bm{\alpha}},\tilde{\beta}) \in \Theta \subseteq [-C_\theta,C_\theta]^{n \times (n+1)}$, define $\DLP_{\bm{\alpha},\beta}$ and $\DLP_{\tilde{\bm{\alpha}},\tilde{\beta}}$ to be the objective values of the following two quadratic programs.
\begin{align}
\DLP_{\bm{\alpha},\beta} = \max_{\bm{p}} \ \bm{p}^\top(\bm{\alpha}+\beta\bm{p}) & \label{eqn:MyQP1} \\
\text{s.t.} \ A (\bm{\alpha}+\beta \bm{p}) &\leq \bm{B}/T \nonumber \\
\bm{p} & \in P_c, \nonumber
\end{align}
and
\begin{align}
\DLP_{\tilde{\bm{\alpha}},\tilde{\beta}} = \max_{\bm{p}} \ \bm{p}^\top(\tilde{\bm{\alpha}}+\tilde{\beta}\bm{p}) & \label{eqn:MyQP2} \\
\text{s.t.} \ A (\tilde{\bm{\alpha}}+\tilde{\beta}\bm{p}) &\leq \bm{B}/T \nonumber \\
\bm{p} & \in P_c. \nonumber
\end{align}If $\gamma_\mathsf{lin}(\tilde{\bm{\alpha}},\tilde{\beta}) \leq 1$,
then we have
\begin{align*}
\DLP_{\bm{\alpha},\beta} - \DLP_{\tilde{\bm{\alpha}},\tilde{\beta}} 
\leq \sqrt{n} \Big(\|\bm{\alpha} - \tilde{\bm{\alpha}}\|_2 + \pmax \sqrt{n} \|\beta - \tilde{\beta}\|_F\Big) \bigg(\pmax + \frac{C_\theta \amax n (\pmax - \pmin)(1+2n\pmax)}{C_\delta} \bigg).
\end{align*}
\end{lemma}

\proof{Proof of Lemma~\ref{lem:PurturbedQP}.}
For the two quadratic programs \eqref{eqn:MyQP1} and \eqref{eqn:MyQP2} as defined in Lemma~\ref{lem:PurturbedQP}, let the optimal solutions be $\bm{p}^*$ and $\tilde{\bm{p}}^*$, respectively.
Recall that we assume the existence of $\bar{\bm{p}}$ under the unknown coefficients $(\bm{\alpha}, \beta)$.
Using $\gamma_{\text{lin}}:=\gamma_{\text{lin}}(\tilde{\bm{\alpha}},\tilde{\beta})$ as defined in \eqref{eqn:gamma_lin}, we further define 
\begin{align*}
\bm{p}_s = (1-\gamma_{\text{lin}}) \bm{p}^* + \gamma_{\text{lin}} \bar{\bm{p}}.
\end{align*}
Applying Cauchy-Schwarz inequality, we have $\|\bm{p}^* - \bm{p}_s\|_2 = \gamma_{\text{lin}} \|\bm{p}^* - \bar{\bm{p}}\|_2 \leq \gamma_{\text{lin}} (\pmax - \pmin) \sqrt{n}$.

The proof proceeds in two steps.
In step one, we show that $\bm{p}_s$ is a feasible solution for both quadratic programs \eqref{eqn:MyQP1} and \eqref{eqn:MyQP2}.
In step two, we upper bound the gap between $\DLP_{\bm{\alpha},\beta}$ and $\DLP_{\tilde{\bm{\alpha}},\tilde{\beta}}$, the two objective values of the two quadratic programs.

\noindent \textbf{Step one}: 
First, it is easy to see that when $\gamma_{\text{lin}} \le 1$, both $\bm{p}^*$ and $\bar{\bm{p}}$ satisfies the constraints of \eqref{eqn:MyQP1}.
So $\bm{p}_s$ is a feasible solution of \eqref{eqn:MyQP1}.

Second, for any price vector $\bm{p} \in P_c$, we upper bound the following difference.
\begin{align*}
A(\tilde{\bm{\alpha}} + \tilde{\beta} \bm{p}) - A(\bm{\alpha} + \beta \bm{p}) = A (\tilde{\bm{\alpha}} - \bm{\alpha}) + A(\tilde{\beta} - \beta) \bm{p}.
\end{align*}
For each dimension $i \in [n]$, let $A_i$ be the $i$-th row vector of $A$.
Then we have 
\begin{align*}
A_i (\tilde{\bm{\alpha}} - \bm{\alpha}) + A_i (\tilde{\beta} - \beta) \bm{p} \leq \amax \sqrt{n} \|\bm{\alpha} - \tilde{\bm{\alpha}}\|_2 + \amax \pmax n \|\beta - \tilde{\beta}\|_F.
\end{align*}
As a result,
\begin{align*}
A (\tilde{\bm{\alpha}} + \tilde{\beta} \bm{p}_s) & \leq A (\bm{\alpha} + \beta \bm{p}_s) + \Big( \amax \sqrt{n} \|\bm{\alpha} - \tilde{\bm{\alpha}}\|_2 + \amax \pmax n \|\beta - \tilde{\beta}\|_F \Big) \bm{1}_n \\
& \leq (1-\gamma_{\text{lin}}) A (\bm{\alpha} + \beta \bm{p}^*) + \gamma_{\text{lin}} A (\bm{\alpha} + \beta \bar{\bm{p}}) + \Big( \amax \sqrt{n} \|\bm{\alpha} - \tilde{\bm{\alpha}}\|_2 + \amax \pmax n \|\beta - \tilde{\beta}\|_F \Big) \bm{1}_n \\
& \leq (1-\gamma_{\text{lin}}) \frac{\bm{B}}{T} + \gamma_{\text{lin}} \Big(\frac{\bm{B}}{T} - C_\delta \bm{1}_n\Big) + \Big( \amax \sqrt{n} \|\bm{\alpha} - \tilde{\bm{\alpha}}\|_2 + \amax \pmax n \|\beta - \tilde{\beta}\|_F \Big) \bm{1}_n \\
& = \frac{\bm{B}}{T},
\end{align*}
which suggests that $\bm{p}_s$ is a feasible solution of \eqref{eqn:MyQP2}.

\noindent \textbf{Step two}:
Using expression \eqref{eqn:defn:JQP}, we have
\begin{multline}
\DLP_{\bm{\alpha},\beta} - \DLP_{\tilde{\bm{\alpha}},\tilde{\beta}} = J(\bm{\alpha},\beta,\bm{p}^*) - J(\tilde{\bm{\alpha}},\tilde{\beta},\tilde{\bm{p}}^*) \leq J(\bm{\alpha},\beta,\bm{p}^*) - J(\tilde{\bm{\alpha}},\tilde{\beta},\bm{p}_s) \\
= J(\bm{\alpha},\beta,\bm{p}^*) - J(\bm{\alpha},\beta,\bm{p}_s) + J(\bm{\alpha},\beta,\bm{p}_s) - J(\tilde{\bm{\alpha}},\tilde{\beta},\bm{p}_s), \label{eqn:2to4terms}
\end{multline}
where the inequality holds because $\tilde{\bm{p}}^*$ is the optimal solution to \eqref{eqn:MyQP2}.
We next upper bound $J(\bm{\alpha},\beta,\bm{p}^*) - J(\bm{\alpha},\beta,\bm{p}_s)$ and $J(\bm{\alpha},\beta,\bm{p}_s) - J(\tilde{\bm{\alpha}},\tilde{\beta},\bm{p}_s)$ respectively.

We first show that
\begin{align*}
J(\bm{\alpha},\beta,\bm{p}^*) - J(\bm{\alpha},\beta,\bm{p}_s) = \ & \bm{p}^{*\top} \bm{\alpha} + \bm{p}^{*\top} \beta \bm{p}^* - \big( \bm{p}_s^\top \bm{\alpha} + \bm{p}_s^\top \beta \bm{p}_s \big) \\
= \ & (\bm{p}^{*\top} - \bm{p}_s^\top) \bm{\alpha} + \big(\bm{p}^{*\top} \beta \bm{p}^* - \bm{p}^{*\top} \beta \bm{p}_s \big) + \big(\bm{p}^{*\top} \beta \bm{p}_s - \bm{p}_s^\top \beta \bm{p}_s \big) \\
\leq \ & \|\bm{p}^* - \bm{p}_s\|_2 \|\bm{\alpha}\|_2 + \|\bm{p}^{*\top} \beta\|_2 \|\bm{p}^* - \bm{p}_s\|_2 + \|\bm{p}^* - \bm{p}_s\|_2 \|\beta \bm{p}_s\|_2 \\
\leq \ & C_\theta n \gamma_{\text{lin}} (\pmax - \pmin) + 2 C_\theta \pmax n^2 \gamma_{\text{lin}} (\pmax - \pmin),
\end{align*}
where the first inequality is applying Cauchy-Schwarz inequality; and the second inequality is bounding each component by its maximum.

We then show that
\begin{align*}
J(\bm{\alpha},\beta,\bm{p}_s) - J(\tilde{\bm{\alpha}},\tilde{\beta},\bm{p}_s) = \ & (\bm{\alpha}^\top - \tilde{\bm{\alpha}}^\top) \bm{p}_s + \bm{p}_s^\top (\beta - \tilde{\beta}) \bm{p}_s \\
\leq \ & \pmax \sqrt{n} \|\bm{\alpha} - \tilde{\bm{\alpha}}\|_2 + \pmax^2 n \|\beta - \tilde{\beta}\|_F,
\end{align*}
where the inequality is applying Cauchy-Schwarz inequality.

Putting the above two inequalities back to \eqref{eqn:2to4terms} and using the expression of $\gamma_{\text{lin}}$ we have
\begin{align*}
\DLP_{\bm{\alpha},\beta} - \DLP_{\tilde{\bm{\alpha}},\tilde{\beta}} 
\leq \sqrt{n} \Big(\|\bm{\alpha} - \tilde{\bm{\alpha}}\|_2 + \pmax \sqrt{n} \|\beta - \tilde{\beta}\|_F\Big) \bigg(\pmax + \frac{C_\theta \amax n (\pmax - \pmin)(1+2n\pmax)}{C_\delta} \bigg),
\end{align*}
which finishes the proof.
\Halmos 
\endproof

\begin{lemma}
\label{lem:PurturbedQP2}
For the true unknown coefficients $(\bm{\alpha},\beta) \in \Theta \subseteq [-C_\theta,C_\theta]^{n \times (n+1)}$ and any parameters $(\tilde{\bm{\alpha}},\tilde{\beta}) \in \Theta \subseteq [-C_\theta,C_\theta]^{n \times (n+1)}$, let $\bm{p}^*$ be the optimal solution to the quadratic program \eqref{eqn:MyQP1} and  $\tilde{\bm{p}}^*$ be the optimal solution to the quadratic program \eqref{eqn:MyQP2}. If $\gamma_\mathsf{lin}(\tilde{\bm{\alpha}},\tilde{\beta}) \leq 1$, then we have
\begin{multline*}
J(\bm{\alpha},\beta,\bm{p}^*) - J(\bm{\alpha},\beta,\tilde{\bm{p}}^*) \\
\le 4\bigg(\pmax + \frac{C_\theta \amax (\pmax - \pmin)(1+\pmax)}{C_\delta} \bigg)\sqrt{1+\pmax^2 }n^3\sqrt{\|\bm{\alpha} - \tilde{\bm{\alpha}}\|_2^2 +  \|\beta - \tilde{\beta}\|_F^2}.
\end{multline*}
\end{lemma}

\proof{Proof of Lemma~\ref{lem:PurturbedQP2}.}
Note that,
\begin{align*}
J(\bm{\alpha},\beta,\bm{p}^*) - J(\bm{\alpha},\beta,\tilde{\bm{p}}^*) 
= J(\bm{\alpha},\beta,\bm{p}^*) - J(\tilde{\bm{\alpha}},\tilde{\beta},\tilde{\bm{p}}^*) + J(\tilde{\bm{\alpha}},\tilde{\beta},\tilde{\bm{p}}^*) - J(\bm{\alpha},\beta,\tilde{\bm{p}}^*).
\end{align*}
The first difference, $J(\bm{\alpha},\beta,\bm{p}^*) - J(\tilde{\bm{\alpha}},\tilde{\beta},\tilde{\bm{p}}^*)$, can be upper bounded using Lemma~\ref{lem:PurturbedQP}.
Now we upper bound the second difference $J(\tilde{\bm{\alpha}},\tilde{\beta},\tilde{\bm{p}}^*) - J(\bm{\alpha},\beta,\tilde{\bm{p}}^*)$.
\begin{align*}
J(\tilde{\bm{\alpha}},\tilde{\beta},\tilde{\bm{p}}^*) - J(\bm{\alpha},\beta,\tilde{\bm{p}}^*) = \ & (\bm{\alpha}^\top - \tilde{\bm{\alpha}}^\top) \tilde{\bm{p}}^* + \tilde{\bm{p}}^{*\top} (\beta - \tilde{\beta}) \tilde{\bm{p}}^* \\
\leq \ & \pmax \sqrt{n} \|\bm{\alpha} - \tilde{\bm{\alpha}}\|_2 + \pmax^2 n \|\beta - \tilde{\beta}\|_F,
\end{align*}
where the inequality is applying Cauchy-Schwarz inequality.

Combining both parts we have
\begin{align*}
&J(\bm{\alpha},\beta,\bm{p}^*) - J(\bm{\alpha},\beta,\tilde{\bm{p}}^*) \\
&= \sqrt{n} \Big(\|\bm{\alpha} - \tilde{\bm{\alpha}}\|_2 + \pmax \sqrt{n} \|\beta - \tilde{\beta}\|_F\Big) \bigg(2\pmax + \frac{C_\theta \amax n (\pmax - \pmin)(1+2n\pmax)}{C_\delta} \bigg)\\
&\le\bigg(2\pmax + \frac{C_\theta \amax n (\pmax - \pmin)(1+2n\pmax)}{C_\delta} \bigg)\sqrt{n}\sqrt{1+\pmax^2 n}\sqrt{\|\bm{\alpha} - \tilde{\bm{\alpha}}\|_2^2 +  \|\beta - \tilde{\beta}\|_F^2}\\
&\le 4\bigg(\pmax + \frac{C_\theta \amax (\pmax - \pmin)(1+\pmax)}{C_\delta} \bigg)\sqrt{1+\pmax ^2}n^3\sqrt{\|\bm{\alpha} - \tilde{\bm{\alpha}}\|_2^2 +  \|\beta - \tilde{\beta}\|_F^2}
\end{align*}
which finishes the proof.
\Halmos 
\endproof

\subsection{Proof of Theorem~\ref{thm:linearBNRM}}

In this section we prove Theorem~\ref{thm:linearBNRM}.
{
In the proof of Theorem~\ref{thm:linearBNRM}, we only consider the case when 
\begin{align*}
T > \frac{C_{\text{lin}} n^{3} \sqrt{\log[n(d+1)T]} {T}^{2/3}}{\underline{b}},
\end{align*}
which implies $\gamma > 0$. 
Otherwise, if 
\begin{align*}
T \leq \frac{C_{\text{lin}} n^{3} \sqrt{\log[n(d+1)T]} {T}^{2/3}}{\underline{b}},
\end{align*}
the proof of Theorem~\ref{thm:linearBNRM} becomes straightforward: $T$ being sufficiently small ensures that the regret upper bound in Theorem~\ref{thm:linearBNRM} exceeds a constant multiple of $n^2T$, thus holding trivially.
}

\proof{Proof of Theorem~\ref{thm:linearBNRM}.}
Without loss of generality, we assume $T\ge n+1$ and $s\ge n+1$; otherwise, the regret upper bound to be proved will be trivial. 
Let $\beta_i$ denote the $i$-th row of $\beta$ ($i=1,\dots,n$).

For any underlying demand parameters $[\bm{\alpha};\beta]$, let $\mathsf{DLP}_{\bm{\alpha},\beta}$ denote the DLP defined by \begin{align}
\DLP_{\bm{\alpha},\beta} = \max_{\bm{p}} \ \bm{p}^\top(\bm{\alpha}+{\beta}\bm{p}) & \label{eqn:obj-l} \\
\text{s.t.} \ A (\bm{\alpha}+{\beta}\bm{p}) &\leq \bm{B}/T  \label{eqn:constraint:inventory-l} \\
\bm{p} & \in P_c. \label{eqn:constraint:NonNeg-l}
\end{align}

Define $T_2=\gamma T$. Let $\tilde{l}$ be the last epoch in the execution of policy $\phi$, and let $\tau$ be the last period before the policy $\phi$ stops. We know that $\tau+1$ is a stopping time and we have $T_{\tilde{l}-1}< \tau\le T$. Clearly, $\phi$ is a $s$-switch learning policy by definition. Moreover, since $\phi$ makes at most $n(\tilde{l}-1)$ switches before $T_{\tilde{l}-1}$ and makes at most 1 switch after $T_{\tilde{l}-1}$, its total number of switches is always upper bounded by $n+1$; since we have assumed $s\ge n+1$, the switching constraint will never be violated.

We use a coupling argument for the regret analysis. Consider a virtual policy $\phi^{\mathsf{v}}$ that runs under exactly the same demand realization process and acts exactly the same as $\phi$ until period $\tau$, but keeps running until the end of round $T$ regardless of the resource constraints. Without conflicts to the previously defined notation in Algorithm \ref{alg:lsete}, let $T_l$ denote the last period of epoch $l$ under policy $\phi^{\mathsf{v}}$ ($l=1,2$), 

Let $(\hat{\bm{\alpha}},\hat{\beta})$ be the least squares estimates that policy $\phi^{\mathsf{v}}$ at the beginning of epoch 2. By Theorem 2.2 and Remark 2.3 of \cite{rigollet2023high} (i.e., the standard estimation guarantee of fixed-design linear regression), we know that
\[
\|\hat{\bm{\alpha}}-\bm{\alpha}\|_2^2+\|\hat{\beta}-\beta\|_F^2=\sum_{i=1}^n (\|\hat{{\alpha}_i}-{\alpha}_i\|_2^2+\|\hat{\beta}_i-\beta_i\|_2^2 )\le n\cdot\frac{(n+1)+\log(n/\delta)}{\lambda_{\min}(X^\top X)}
\]
with probability at least $1-\delta$, 
where $\|\cdot\|_F$ denotes the Frobenius norm, and
\[
X:=\left[\begin{aligned} 1 && \bm{z}_1^\top\\1&& \dots\\1&& \bm{z}_{T_1}^\top\end{aligned}\right].
\]
Given the specific fixed design of $\bm{z}_1,\dots,\bm{z}_{T_1}$ in epoch 
1 of Algorithm \ref{alg:lsete} (i.e., all of $\pmin\bm{1}$ and $\pmin\bm{1}+(\pmax-\pmin)\bm{e}_j$ ($j\in[n]$) are chosen for $T_1/(n+1)$ periods), it is easy to show that $\lambda_{\min}(X^\top X)/T_1$ is lower bounded by  $ \min\{(\pmax-\pmin)^2,1\}/(2(n+1)^2)>0$. Let $C_{\text{design}}:=\min\{(\pmax-\pmin)^2,1\}$. We know that
\[
\|\hat{\bm{\alpha}}-\bm{\alpha}\|_2^2+\|\hat{\beta}-\beta\|_F^2\le{2n(n+1)^2}\frac{n+1+\log(n/\delta)}{T_1C_{\text{design}}}
\]
happens with probability at least $1-\delta$.

Define the confidence radius as
\[
\radius(t)=\sqrt{\frac{6\log [(d+1)T]}{t}}.
\]
Define the \emph{clean event} $\mathcal{E}$ as
$
\mathcal{E}_{\rm conc}\cup\mathcal{E}_{\rm LS}
$,
where 
\[\mathcal{E}_{\rm conc}:=\left\{\forall i\in[d], t\in[T_2],~\left|\frac{1}{t}\sum_{t'=1}^t \sum_{j=1}^n(\epsilon_{t'})_j\right|\le\sqrt{n}\radius(t).
 \right\}\] and
\[\mathcal{E}_{\rm LS}:=
\left\{\|\hat{\bm{\alpha}}-\bm{\alpha}\|_2^2+\|\hat{\beta}-\beta\|_F^2\le2n(n+1)^2\frac{n+1+\log(n(d+1)T)}{T_1 C_{\rm design}}\right\}.\]
By the Hoeffding's inequality for standard sub-Gaussian random variables (see Corollary 1.7 in \cite{rigollet2023high}) and a standard union bound argument (see Chapter 1.3.1 in \cite{slivkins2019introduction}), we have $\text{Pr}(\mathcal{E}_{\rm conc})\ge 1-\frac{2}{(d+1)T}$. Combining it with $\Pr(\mathcal{E}_{\rm LS})\ge 1-\frac{1}{(d+1)T}$, we have $\Pr(\mathcal{E})\ge 1-\frac{3}{(d+1)T}$. Since the clean event $\mathcal{E}$ happens with very high probability, we can just focus on a \textit{clean execution} of policy $\phi^{\mathsf{v}}$: an execution in which the clean event holds. By the coupling relationship between $\phi$ and $\phi^{\mathsf{v}}$, we know that conditional on the clean event, policy $\phi$ behaves exactly the same as $\phi^{\mathsf{v}}$. 
In the rest of the proof, we always assume that $\mathcal{E}$ holds.

For all $i\in[d]$ and  $l\in\{1,2\}$, we have
\begin{align*}
& \sum_{t=1}^{T_l} A_i \bQ_{t}(\bm{z}_t)=\sum_{t=1}^{T_l} A_i (\bm{\alpha}+\beta \bm{z}_t) + \sum_{t=1}^{T_l} A_i\epsilon_t\notag\\
\le&\sum_{t=1}^{T_l} A_i (\bm{\alpha}+\beta \bm{z}_t) + T_l\cdot\amax\sqrt{n}\nabla(T_l)\notag\\
=&\sum_{t=1}^{T_l} A_i (\bm{\alpha}+\beta \bm{z}_t) + \amax\sqrt{6n\log[(d+1)T]}\sqrt{T_l}
\end{align*}
where the first inequality follows from the occurrence of $\mathcal{E}_{\rm conc}$. 
Furthermore, we have
\begin{align}\label{eqn:linearBNRM-UB-1}
& \sum_{t=1}^{T_l} A_i \bQ_{t}(\bm{z}_t)\notag\\
\le&\sum_{t=1}^{T_l} A_i (\bm{\alpha}+\beta \bm{z}_t) + \amax\sqrt{6n\log[(d+1)T]}\sqrt{T_l}\notag\\
\le&\sum_{t=1}^{T_1} A_i (\bm{\alpha}+\beta \bm{z}_t) + (T_l-T_1) A_i (\bm{\alpha}+\beta \bp^*_{\hat{\bm{\alpha}},\hat{\beta}}) +\amax\sqrt{6n\log[(d+1)T]}\sqrt{T_l}\notag\\
\le& n\amax(C_\theta+n C_\theta\pmax) T^{2/3}+ (T_l-T_1) A_i (\bm{\alpha}+\beta \bp^*_{\hat{\bm{\alpha}},\hat{\beta}}) +\amax\sqrt{6n\log[(d+1)T]}\sqrt{T}\notag\\
\le& \amax n^2(C_\theta(1+\pmax)+\sqrt{6\log[(d+1)T]})T^{2/3}+ (T_l-T_1) A_i (\bm{\alpha}+\beta \bp^*_{\hat{\bm{\alpha}},\hat{\beta}})\notag\\
\le& \amax n^2(2\max\{C_\theta,1\}(1+\pmax)\sqrt{6\log[(d+1)T]})T^{2/3}+ (T_l-T_1) A_i (\bm{\alpha}+\beta \bp^*_{\hat{\bm{\alpha}},\hat{\beta}})\notag\\
\le&(C_{\rm lin}/3) n^2\sqrt{\log[n(d+1)T]}T^{2/3}+(T_l-T_1) A_i (\bm{\alpha}+\beta \bp^*_{\hat{\bm{\alpha}},\hat{\beta}}).
\end{align}
As we have mentioned, we only need to consider the case when $\gamma>0$, which implies $C_{\rm lin}n^3\sqrt{\log[n(d+1)T]}T^{2/3}\le \Bmin$. 
{Therefore, when $l=1$, $\sum_{t=1}^{T_1} A_i \bQ_{t}(\bm{z}_t)\le B_i$ for all $i\in[d]$}; hence $\tilde{l}=2$ and $T_1=\gamma t_1$. When $l=2$, since
\begin{align}\label{eq:difficult}
    & (T_2-T_1)A_i(\bm{\alpha}+\beta \bp^*_{\hat{\bm{\alpha}},\hat{\beta}})\notag\\
    \le& (T_2-T_1)A_i(\hat{\bm{\alpha}}+\hat{\beta} \bp^*_{\hat{\bm{\alpha}},\hat{\beta}})+(T_2-T_1)\amax\sqrt{n}\left(\|\hat{\bm{\alpha}}-\bm{\alpha}\|_2+\pmax\sqrt{n}\|\hat{\beta}-\beta\|_F\right)\notag\\
    \le &(T_2-T_1)A_i(\hat{\bm{\alpha}}+\hat{\beta} \bp^*_{\hat{\bm{\alpha}},\hat{\beta}})+(T_2-T_1)\amax\sqrt{n}\sqrt{(1+\pmax^2n)2n(n+1)^2\frac{n+1+\log[n(d+1)T]}{T_1 C_{\rm design}}}\notag\\
    \le& \gamma B_i+\frac{T_2-T_1}{\sqrt{T_1}}\amax\sqrt{1+\pmax^2} n^{3/2}(n+1)^{3/2}{\sqrt{6\log[n(d+1)T]/C_{\text{design}}}}\notag\\
    \le& \gamma B_i+2\sqrt{3}{T^{2/3}}\amax\sqrt{1+\pmax^2} n^{3/2}(n+1)^{3/2}{\sqrt{\log[n(d+1)T]/C_{\text{design}}}}\notag\\
    \le& \gamma B_i+ {10T^{2/3}}\amax\sqrt{1+\pmax^2}n^{3}{\sqrt{\log[n(d+1)T]/C_{\text{design}}}}
\end{align}
under $\mathcal{E}_{\rm conc}$, 
combining \eqref{eqn:linearBNRM-UB-1} and \eqref{eq:difficult}, we know that the specification of $\gamma$ in Algorithm \ref{alg:lsete} ensures that $\sum_{t=1}^{T_2} A_i \bQ_{t}(\bm{z}_t)\le B_i$ for all $i\in[d]$.

The previous inequality indicates that conditional on the clean event, policy $\phi^{\mathsf{v}}$ will not violate any resource constraint up to $T_2$. By the coupling relationship between $\phi$ and $\phi^\mathsf{v}$, we know that $\tau\ge T_2$ conditional on the clean execution of policy $\phi^{\mathsf{v}}$. Thus conditional on the clean event, the total revenue collected by policy $\phi$ is at least
\begin{align}\label{eqn:linearBNRM-UB-5}
    &\sum_{t=1}^{T_2} \bm{z}_t^\top (\bm{\alpha}+\beta \bm{z}_t)+ \sum_{t=1}^{T_2} \bm{z}_t^\top\epsilon_t\notag\\
    \ge&\sum_{t=1}^{T_2} \bm{z}_t^\top (\bm{\alpha}+\beta \bm{z}_t)-\pmax\sqrt{6n\log[(d+1)T]}\sqrt{T_{2}}.
\end{align}
In what follows, we bound $\sum_{t=1}^{T_2} \bm{z}_t^\top (\bm{\alpha}+\beta \bm{z}_t)$ conditional on the clean event. 
Let $\bp^*$ be an optimal solution to $\DLP_{\bm{\alpha},\beta}$; we have $A({\bm{\alpha}}+{\beta}\bp^*)\le\bm{B}/T$. 

{

We only focus on the case when $T$ is sufficiently large such that 
\begin{align}
\lfloor T^{\frac{2}{3}}\rfloor \geq \frac{2 n^2 (n+1)^2 \bigg( n+1 + \log{[ n(d+1)T ]} \bigg) \amax^2 (n \pmax^2 + 1)}{C_{\text{design}} C_\delta^2}. \label{eqn:Tlargeenough}
\end{align}
Otherwise, if 
$T$ is small such that
\begin{align*}
\lfloor T^{\frac{2}{3}}\rfloor < \frac{2 n^2 (n+1)^2 \bigg( n+1 + \log{[ n(d+1)T ]} \bigg) \amax^2 (n \pmax^2 + 1)}{C_{\text{design}} C_\delta^2},
\end{align*}
this implies 
\begin{align*}
T^{\frac{1}{3}} < \frac{2 n (n+1) \big(n+1+\log{[n(d+1)T]}\big)^{\frac{1}{2}} \amax (n \pmax^2 + 1)^{\frac{1}{2}}}{C_{\text{design}}^{\frac{1}{2}} C_\delta}.
\end{align*}
Note that, in each time period, there is at most a loss of revenue on $n$ products.
As a result, the regret can be upper bounded by $n\pmax T = n\pmax T^{\frac{2}{3}}  \cdot T^{\frac{1}{3}} < c T^{\frac{2}{3}} n^{3} \big(\log{[n(d+1)T]}\big)^{\frac{1}{2}},$ where $c$ is a constant that does not depend on $T, \bm{B}, d, n$, and $\underline{b}$. 

Now, when $T$ is sufficiently large such that \eqref{eqn:Tlargeenough} holds, conditioning on $\mathcal{E}$ we have
\begin{multline*}
\amax \sqrt{n} \|\hat{\bm{\alpha}}-\bm{\alpha}\|_2 + \amax \pmax n \|\hat{\beta}-\beta\|_F \leq \sqrt{\amax^2 n + \amax^2 \pmax^2 n^2} \sqrt{\|\hat{\bm{\alpha}}-\bm{\alpha}\|_2^2+\|\hat{\beta}-\beta\|_F^2} \\
\leq \sqrt{\amax^2 n + \amax^2 \pmax^2 n^2} \sqrt{2n(n+1)^2\frac{n+1+\log[n(d+1)T]}{T_1 C_{\rm design}}} 
\leq C_\delta.
\end{multline*}
the first inequality is Cauchy-Schwarz, the second inequality is conditioning on $\mathcal{E}$, the third inequality is because $T$ is large such that \eqref{eqn:Tlargeenough} holds.
The above inequality means that $\gamma_\mathsf{lin} \leq 1$. So we can then apply Lemma \ref{lem:PurturbedQP2}.
}

Define $C_{\ref{lem:PurturbedQP2}}:=4\bigg(\pmax + \frac{C_\theta \amax (\pmax - \pmin)(1+\pmax)}{C_\delta} \bigg)\sqrt{1+\pmax^2}$, by Lemma \ref{lem:PurturbedQP2}, we have
\begin{align*}
&\sum_{t=T_1+1}^{T_2} \bm{z}_t^\top (\bm{\alpha}+\beta \bm{z}_t)\notag\\
=&\sum_{t=T_1+1}^{T_2} {\bp^*}^\top (\bm{\alpha}+\beta \bp^*)-(T_2-T_1)\left({\bp^*}^\top (\bm{\alpha}+\beta \bp^*)-{\bp^*_{\hat{\bm{\alpha}},\hat{\beta}}}^\top (\bm{\alpha}+\beta \bp^*_{\hat{\bm{\alpha}},\hat{\beta}})\right)\notag\\
\ge&\sum_{t=T_1+1}^{T_2} {\bp^*}^\top (\bm{\alpha}+\beta \bp^*)-(T_2-T_1)C_{\ref{lem:PurturbedQP2}}n^3\sqrt{\|\bm{\alpha}_1 - \bm{\alpha}_2\|_2^2 +  \|\beta_1 - \beta_2\|_F^2}\notag\\
\ge& \sum_{t=T_1+1}^{T_2} {\bp^*}^\top (\bm{\alpha}+\beta \bp^*)-(T_2-T_1)C_{\ref{lem:PurturbedQP2}}n^3\sqrt{2n(n+1)^2\frac{(n+1)+\log[n(d+1)T]}{T_1 C_{\rm design}}}\\
\ge& \sum_{t=T_1+1}^{T_2} {\bp^*}^\top (\bm{\alpha}+\beta \bp^*)-\frac{T_2-T_1}{\sqrt{T_1}}C_{\ref{lem:PurturbedQP2}}n^{7/2}(n+1)^{3/2}\sqrt{6{\log[n(d+1)T]}/{C_{\text{design}}}}\\
\ge&\sum_{t=T_1+1}^{T_2} {\bp^*}^\top (\bm{\alpha}+\beta \bp^*)-2\sqrt{3}{T^{2/3}}C_{\ref{lem:PurturbedQP2}}n^{7/2}(n+1)^{3/2}\sqrt{{\log[n(d+1)T]}/{C_{\text{design}}}}\\
\ge&\sum_{t=T_1+1}^{T_2} {\bp^*}^\top (\bm{\alpha}+\beta \bp^*)-10C_{\ref{lem:PurturbedQP2}}n^5\sqrt{{\log[n(d+1)T]}/{C_{\text{design}}}}T^{2/3}
\end{align*}
Hence,
\begin{align*}
&\sum_{t=1}^{T} {\bp^*}^\top (\bm{\alpha}+\beta \bp^*)-\sum_{t=1}^T \bm{z}_t^\top (\bm{\alpha}+\beta \bm{z}_t)\\
\le& \sum_{t=1}^{T}{\bp^*}^\top (\bm{\alpha}+\beta \bp^*) -\sum_{t=T_1+1}^{T_2} {\bp^*}^\top (\bm{\alpha}+\beta \bp^*)+ 10C_{\ref{lem:PurturbedQP2}}n^5\sqrt{{\log[n(d+1)T]}/{C_{\text{design}}}}T^{2/3}\\
\le& (T_1+(T-T_2)){\bp^*}^\top (\bm{\alpha}+\beta \bp^*)+10C_{\ref{lem:PurturbedQP2}}n^5\sqrt{{\log[n(d+1)T]}/{C_{\text{design}}}}T^{2/3}\\
\le& (\gamma T^{2/3}+(1-\gamma)T)n\pmax(C_\theta+ n C_\theta \pmax)+10C_{\ref{lem:PurturbedQP2}}n^5\sqrt{{\log[n(d+1)T]}/{C_{\text{design}}}}T^{2/3}\\
\le& {\max\{c/\underline{b},c'\}n^5\sqrt{\log[n(d+1)T]}}\cdot T^{2/3}
\end{align*}
where $c,c'$ are some absolute constants completely determined by $C_{\text{lin}}$ (defined in Algorithm \ref{alg:lsete}), $C_{\ref{lem:PurturbedQP2}},C_{\text{design}}$ (defined in this proof) and $C_{\theta},\pmax,\amax$ (given by the problem). Hence $c,c'$ are completely determined by $\pmin,\pmax,\amax,C_\theta,C_\delta$.
\Halmos 
\endproof


\section{Additional Algorithms} \label{sec:AdditionalAlgos}
\begin{algorithm}[htbp]
\caption{An inventory-updating version of Algorithm \ref{alg:bsse}}
\label{alg:update}
\leftline{{\bf Input:} Problem parameters $(T,\bm{B},K,d,n,P,A)$; switching budget $s$.} 
\leftline{{\bf Initialization:} Calculate $\nu(s,d)=\left\lfloor\frac{s-d-1}{K-1}\right\rfloor$. Define $t_0=0$ and}
\[
t_l=\left\lfloor K^{1-\frac{2-2^{-(l-1)}}{2-2^{-\nu(s,d)}}}T^{\frac{2-2^{-(l-1)}}{2-2^{-\nu(s,d)}}}\right\rfloor,~~\forall l=1,\dots,\nu(s,d)+1.
\]
{Set $\gamma=\max\big\{1- 17\frac{\amax\sqrt{n(d+1)\log[(d+1)KT]}\log T}{\Bmin}{t_1},0\big\}$.} \\
\leftline{{\bf Notation:} {Let $T_l$ denote the ending period of epoch $l$ (which will be determined by the algorithm).}} Let $B_i^{l}$ denote the consumed inventory of resource $i$ in the first $l$ epochs. 
{Let $\bm{z}_t \in \{\bp_1,\dots,\bp_K\}$ denote the algorithm's selected price vector at period $t$. Let $z_0$ be a random price vector in $\{\bp_1,\dots,\bp_K\}$.}\\
{{\bf Policy:}}
\begin{algorithmic}[1]
\FOR{epoch $l=1,\dots,\nu(s,d)$}
\IF{$l=1$}
\STATE{Set $T_0=L^{\mathsf{rew}}_k(0)=L^{\mathsf{cost}}_{i,k}(0)=0$ and $U^{\mathsf{rew}}_k(0)=U^{\mathsf{cost}}_{i,k}(0)=\infty$ for all $i\in[d],k\in[K]$.}
\ELSE
\STATE{Let $n_{k}(T_{l-1})$ be the total number of periods that price vector $\bp_k$ is chosen up to period $T_{l-1}$, and $\bar{q}_{j,k}(T_{l-1})$ be the empirical mean demand of product $j$ sold at price vector $\bp_k$ up to period $T_{l-1}$. Calculate $\radius_k(T_{l-1})=\sqrt{\frac{\log\left[(d+1)KT\right]}{n_k(T_{l-1})}}$ and 
\[
\begin{cases}U^{\mathsf{rew}}_{k}(T_{l-1})=\min\left\{\sum_{j\in[n]}p_{j,k}\bar{q}_{j,k}(T_{l-1})+||\bp_k||_2\radius_k(T_{l-1}),U^{\mathsf{rew}}_k(T_{l-2})\right\}
,\\L^{\mathsf{rew}}_{k}(T_{l-1})=\max\left\{\sum_{j\in[n]}p_{j,k}\bar{q}_{j,k}(T_{l-1})-||\bp_k||_2\radius_k(T_{l-1}),L_k^{\mathsf{rew}}(T_{l-2})\right\},\end{cases}~~\forall k\in[K],
\]
\[
\begin{cases}U^{\mathsf{cost}}_{i,k}(T_{l-1})=\min\left\{\sum_{j\in[n]}a_{ij}\bar{q}_{j,k}{(T_{l-1})}+||A_i||_2\radius_k(T_{l-1}),U^{\mathsf{cost}}_{i,k}(T_{l-2})\right\}
,\\L^{\mathsf{cost}}_{i,k}(T_{l-1})=\max\left\{\sum_{j\in[n]}a_{ij}\bar{q}_{j,k}{(T_{l-1})}-||A_i||_2\radius_k(T_{l-1}),L_{i,k}^{\mathsf{cost}}(T_{l-2})\right\},\end{cases}~~\forall i\in[d],\forall k\in[K].
\]}
\ENDIF
\STATE{Solve the first-stage pessimistic LP:
\begin{align*}
\mathsf{J}^{\mathsf{PES}}_{l} = \max_{(x_1,\dots,x_K)} \sum_{k\in[K]}L^{\mathsf{rew}}_k(T_{l-1})  x_k &&  \\
\text{s.t.} \ \sum_{k\in[K]}U_{i,k}^{\mathsf{cost}}(T_{l-1})  x_k &\leq B_i- B_i^{l-1} & \forall i \in [d] \\
\sum_{k\in [K]}x_k &\leq T- T_{l-1} &  \\
x_k & \geq 0 & \forall k\in[K] 
\end{align*}}
\algstore{myalg}
\end{algorithmic}
\end{algorithm}

\begin{algorithm}                     
\begin{algorithmic}[1]
\algrestore{myalg}
\STATE{For each $j\in[K]$, solve the second-stage exploration LP:
\begin{align*}
\bx^{l,j} = \arg\max_{(x_1,\dots,x_K)} \ x_j &&  \\
\text{s.t.} \ \sum_{k\in[K]}U^{\mathsf{rew}}_{k}(T_{l-1})x_k&\ge \mathsf{J}_{l}^{\mathsf{PES}} &\\
\sum_{k\in[K]}L^{\mathsf{cost}}_{i,k}(T_{l-1}) x_k &\leq B_i-B_i^{l-1} & \forall i \in [d]\\
\sum_{k\in [K]}x_k &\leq T- T_{l-1}&  \\
x_k & \geq 0 & \forall\ k\in[K] 
\end{align*}
}
\STATE{For all $k\in[K]$, let $N_k^l=\frac{(t_{l}-t_{l-1})}{T-T_{l-1}}\sum_{j=1}^{K}\frac{1}{K}(\bx^{l,j})_k$. Let $\bm{z}_{T_{l-1}+1}=\bm{z}_{T_{l-1}}$. Starting from this action, choose each price vector $\bp_k$ for $\gamma N_k^l$ consecutive periods, $k\in[K]$ (we overlook the rounding issues here, which are easy to fix in regret analysis). Stop the algorithm once time horizon is met or one of the resources is exhausted.} 
\STATE{End epoch $l$.  Mark the last period in epoch $l$ as $T_l$.}
\ENDFOR
\STATE{For epoch $\nu(s,d)+1$, let $\bar{q}_{j,k}(T_{\nu(s,d)})$ be the empirical mean demand of product $j$ sold at price vector $\bp_k$ up to period $T_{\nu(s,d)}$, and calculate $\tilde{\bm{q}}=\left(\bar{q}_{j,k}(T_{\nu(s,d)})\right)_{j\in[n],k\in[K]}$. Find an optimal solution to 
\begin{align*}
 \max_{(x_1,\dots,x_K)} \sum_{k\in[K]}\sum_{j\in[n]}p_{j,k} \ \bar{q}_{j,k}(T_{\nu(s,d)-1}) \ x_k &&&  \\
\text{s.t.} \ \sum_{k\in[K]}\sum_{j\in[n]} a_{ij} \ \bar{q}_{j,k}(T_{\nu(s,d)-1}) \ x_k &\leq B_i-B_i^{\nu(s,d)} && \forall\ i \in [d]  \\
\sum_{k\in [K]}x_k &\leq T-T_{\nu(s,d)} &&  \\
x_k & \geq 0 && \forall\ k\in[K] 
\end{align*}
with the least number of non-zero variables, denoted by $\bx^*$. Let $N_k^{\nu(s,d)+1}={(\bx^*)_k}$ for all $k\in[K]$. Let $\bm{z}_{T_{\nu(s,d)}+1}=\bm{z}_{T_{\nu(s,d)}}$. Starting from this action, choose each price vector $\bp_k$ for $\gamma N_k^{\nu(s,d)+1}$ consecutive periods, $k\in[K]$ (we overlook the rounding issues here, which are easy to fix in regret analysis). Stop the algorithm once time horizon is met or one of the resources is exhausted. End epoch $\nu(s,d)+1$.}
\end{algorithmic}
\end{algorithm}

\clearpage

\section{Additional Simulation Results} \label{sec:AdditionalSimulations}

\subsection{Performance Relative to the Deterministic Linear Program}

We present in this section the simulation results for all algorithms, including the updating inventory version of all the algorithms we have presented in Section~\ref{sec:simulation}.

\begin{figure}[!htb]
\centering
\includegraphics[width=0.7\textwidth]{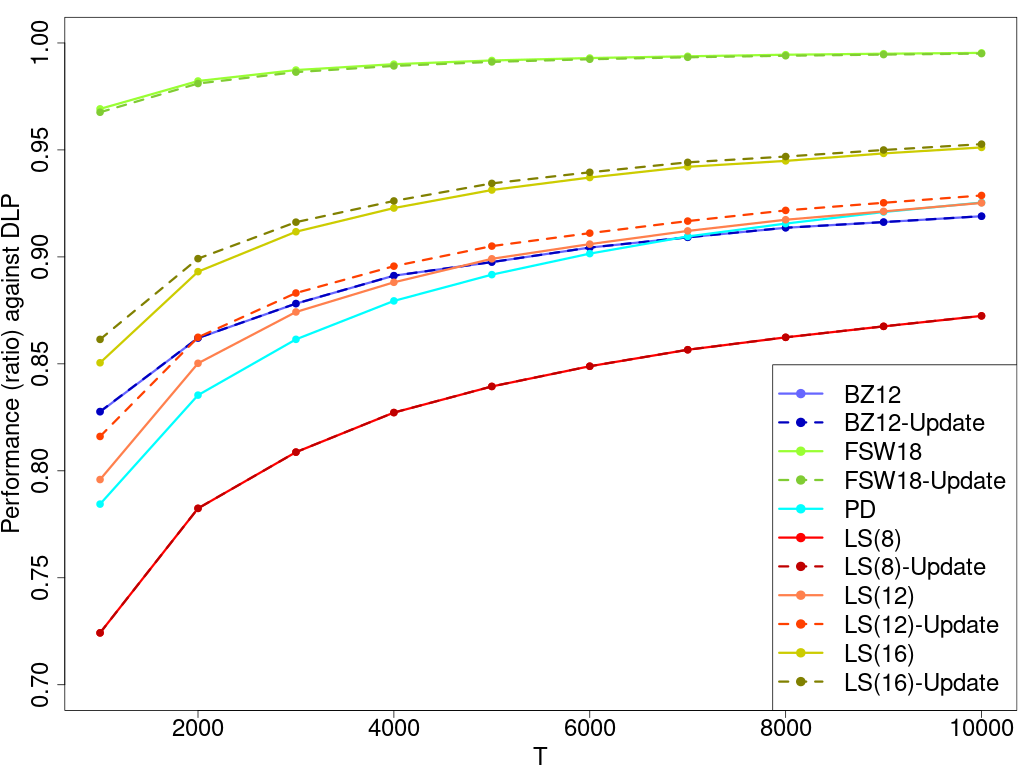}
\caption{Numerical results under linear demand and small inventory. The solid lines are the original method without updating inventory; the dashed lines are updating inventory.}
\label{fig:additional:linear:small}
{\footnotesize {\it Note:} The solid lines are the original method without updating inventory; the dashed lines are updating inventory. The performance of algorithms are normalized relative to the DLP upper bound, thus between 0 and 1.}
\end{figure}

\begin{figure}[!htb]
\centering
\includegraphics[width=0.7\textwidth]{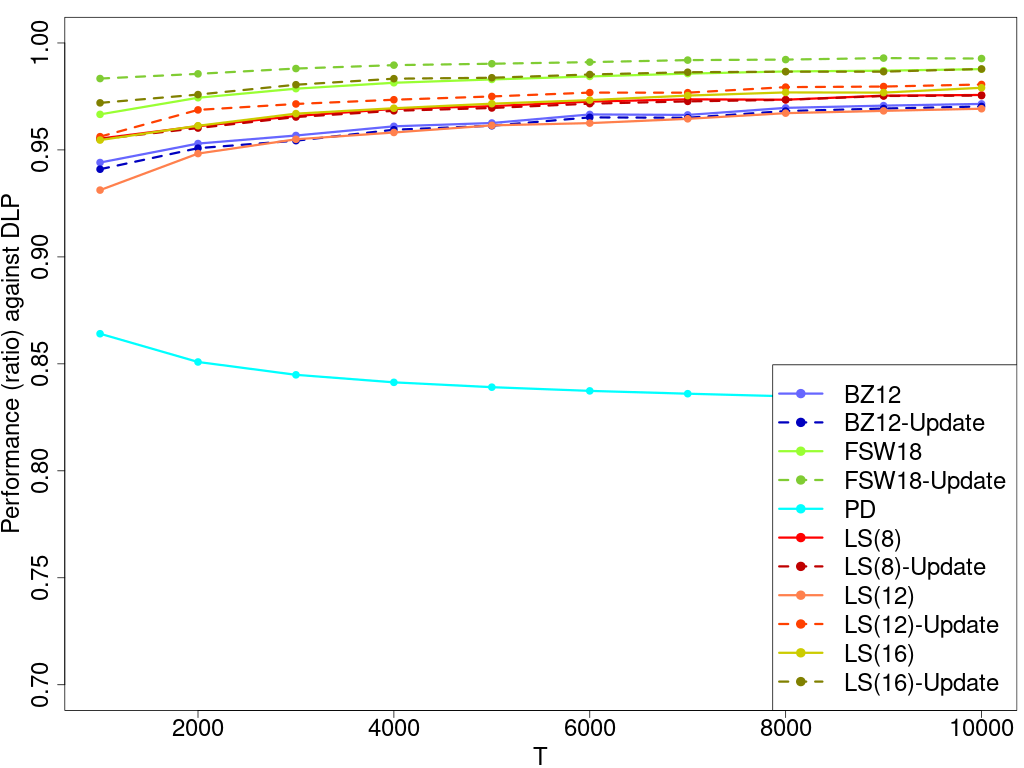}
\caption{Numerical results under linear demand and large inventory when there are $K=5$ price vectors.}
\label{fig:additional:linear:large}
{\footnotesize {\it Note:} The solid lines are the original method without updating inventory; the dashed lines are updating inventory. The performance of algorithms are normalized relative to the DLP upper bound, thus between 0 and 1.}
\end{figure}

\begin{figure}[!htb]
\centering
\includegraphics[width=0.7\textwidth]{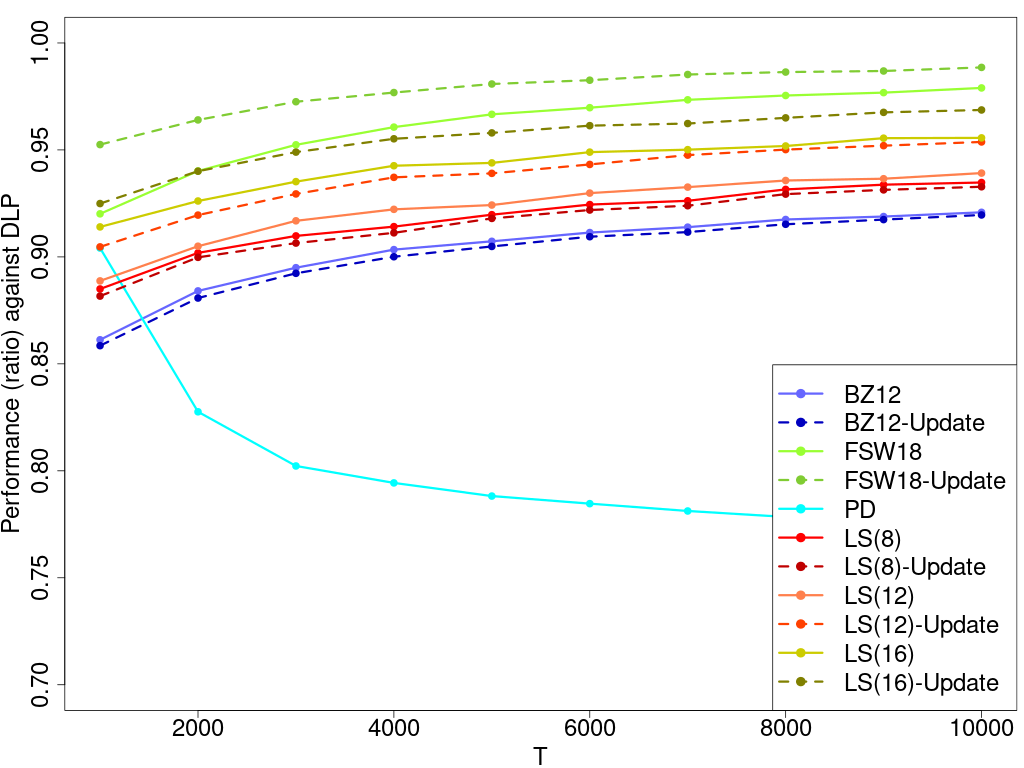}
\caption{Numerical results under exponential demand and small inventory when there are $K=5$ price vectors.}
\label{fig:additional:exponential:small}
{\footnotesize {\it Note:} The solid lines are the original method without updating inventory; the dashed lines are updating inventory. The performance of algorithms are normalized relative to the DLP upper bound, thus between 0 and 1.}
\end{figure}

\begin{figure}[!htb]
\centering
\includegraphics[width=0.7\textwidth]{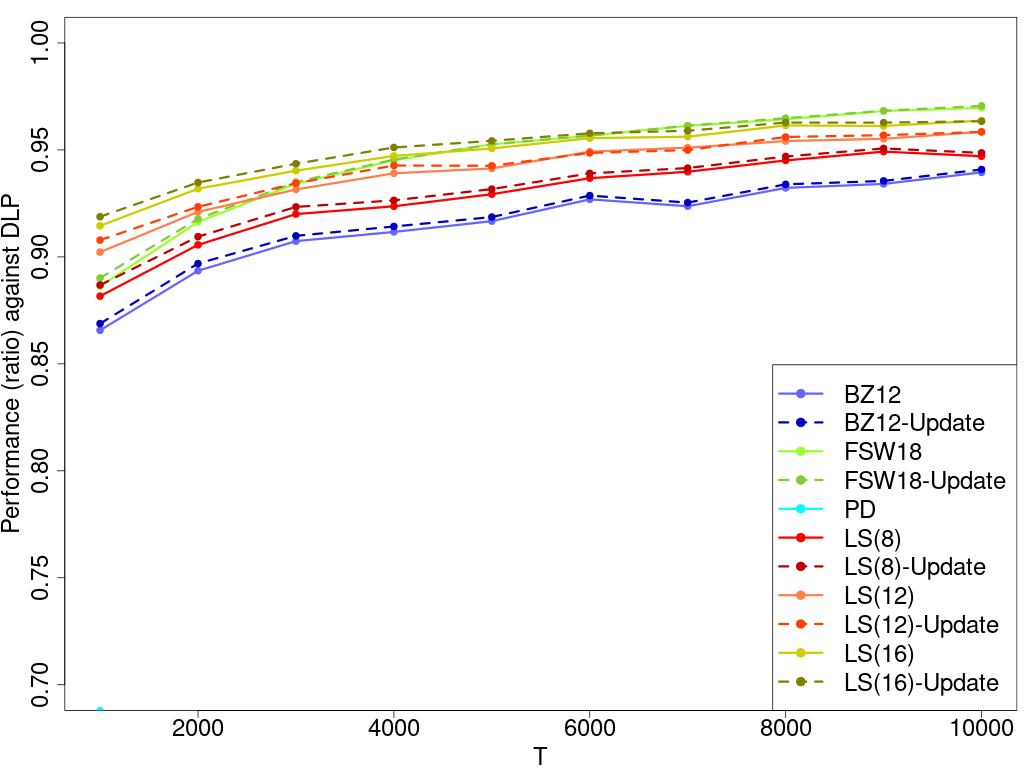}
\caption{Numerical results under exponential demand and large inventory when there are $K=5$ price vectors.}
\label{fig:additional:exponential:large}
{\footnotesize {\it Note:} The solid lines are the original method without updating inventory; the dashed lines are updating inventory. The performance of algorithms are normalized relative to the DLP upper bound, thus between 0 and 1.}
\end{figure}

\begin{figure}[!htb]
\centering
\includegraphics[width=0.7\textwidth]{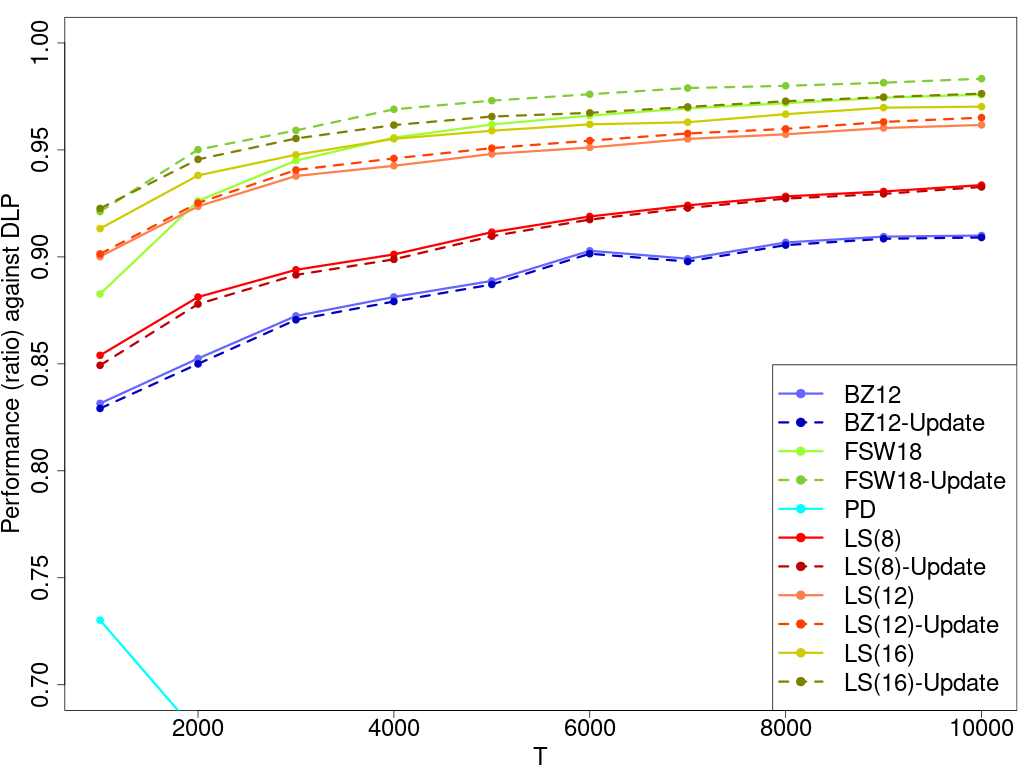}
\caption{Numerical results under logit demand and small inventory when there are $K=5$ price vectors.}
\label{fig:additional:logit:small}
{\footnotesize {\it Note:} The solid lines are the original method without updating inventory; the dashed lines are updating inventory. The performance of algorithms are normalized relative to the DLP upper bound, thus between 0 and 1.}
\end{figure}

\begin{figure}[!htb]
\centering
\includegraphics[width=0.7\textwidth]{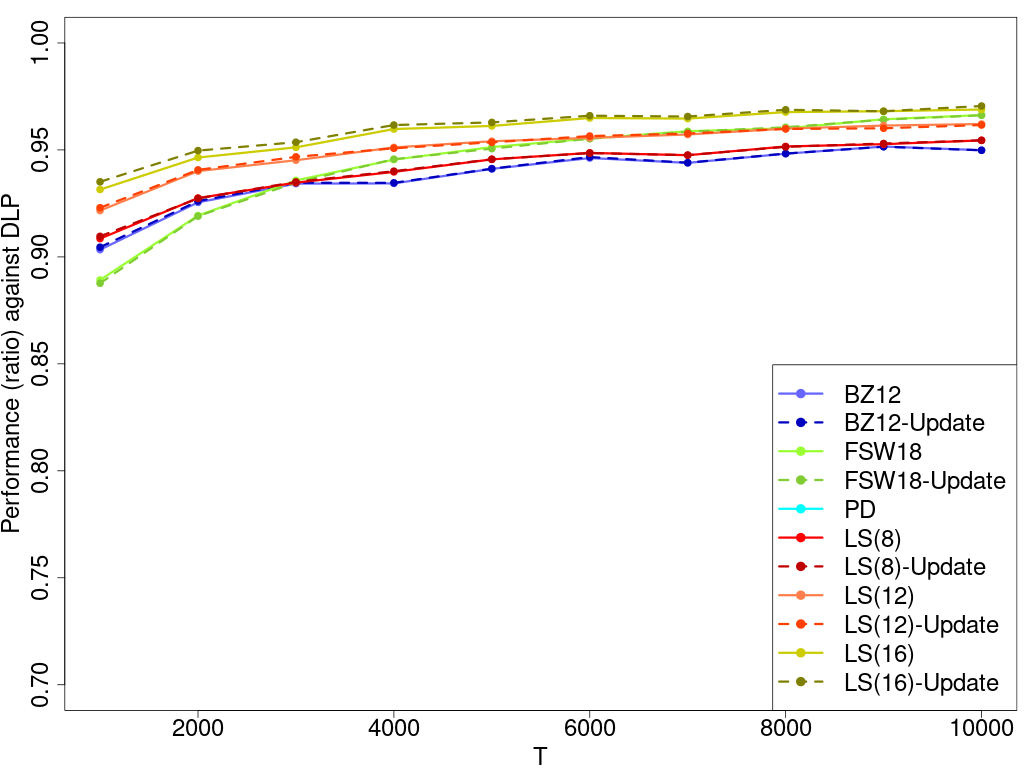}
\caption{Numerical results under logit demand and large inventory when there are $K=5$ price vectors.}
\label{fig:additional:logit:large}
{\footnotesize {\it Note:} The solid lines are the original method without updating inventory; the dashed lines are updating inventory. The performance of algorithms are normalized relative to the DLP upper bound, thus between 0 and 1.}
\end{figure}

We first present the simulation results in the first simulation setup where there are 4 price $(p_1, p_2) \in \{(1, 1.5),(1, 2),(2, 3),(4, 4),(4, 6.5)\}$.
The blue dashed line is \textbf{BZ12} with updating inventory.
The green dashed line is \textbf{FSW18} with updating inventory.
The \textbf{PD} algorithm does not have the updating inventory version.
The red, orange, and olive dashed lines are our algorithms under switching budgets of $s=8, 12, 16$, respectively, with updating inventory.
For most cases, the inventory updating version of all the algorithms are slightly better than the no inventory updating version.
See Figures~\ref{fig:additional:linear:small}--~\ref{fig:additional:logit:large}.

\begin{figure}[!htb]
\centering
\includegraphics[width=0.7\textwidth]{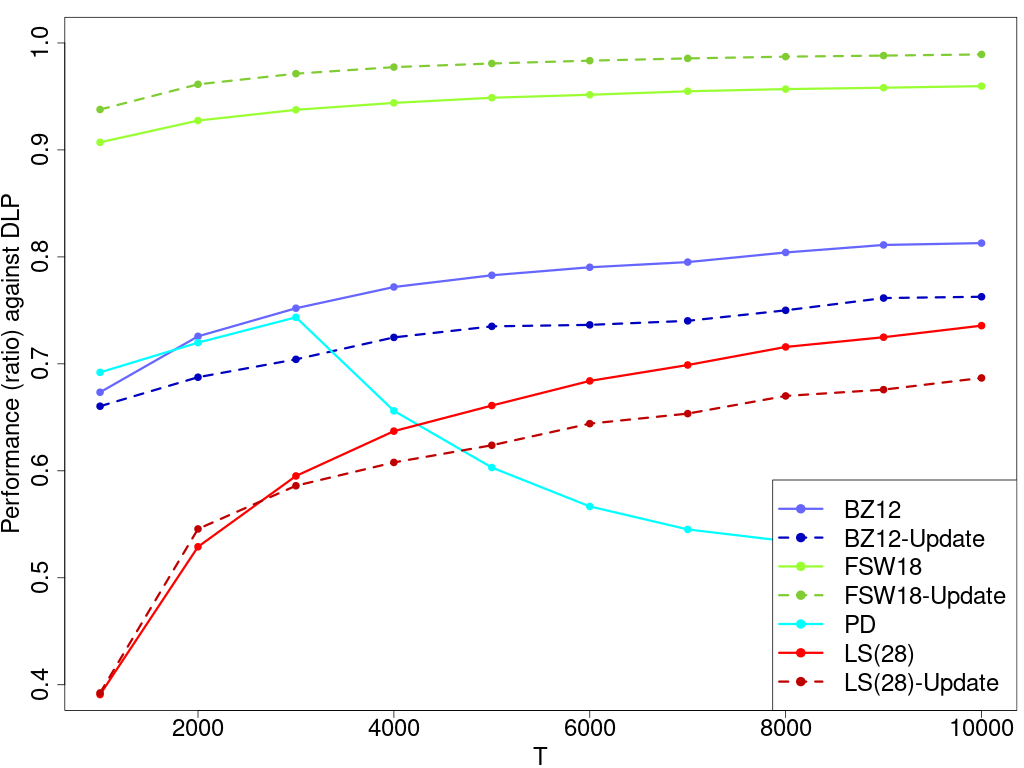}
\caption{Numerical results under linear demand and small inventory when there are $K=15$ price vectors.}
\label{fig:additionalK=15:linear:small}
{\footnotesize {\it Note:} The solid lines are the original method without updating inventory; the dashed lines are updating inventory. The performance of algorithms are normalized relative to the DLP upper bound, thus between 0 and 1.}
\end{figure}

\begin{figure}[!htb]
\centering
\includegraphics[width=0.7\textwidth]{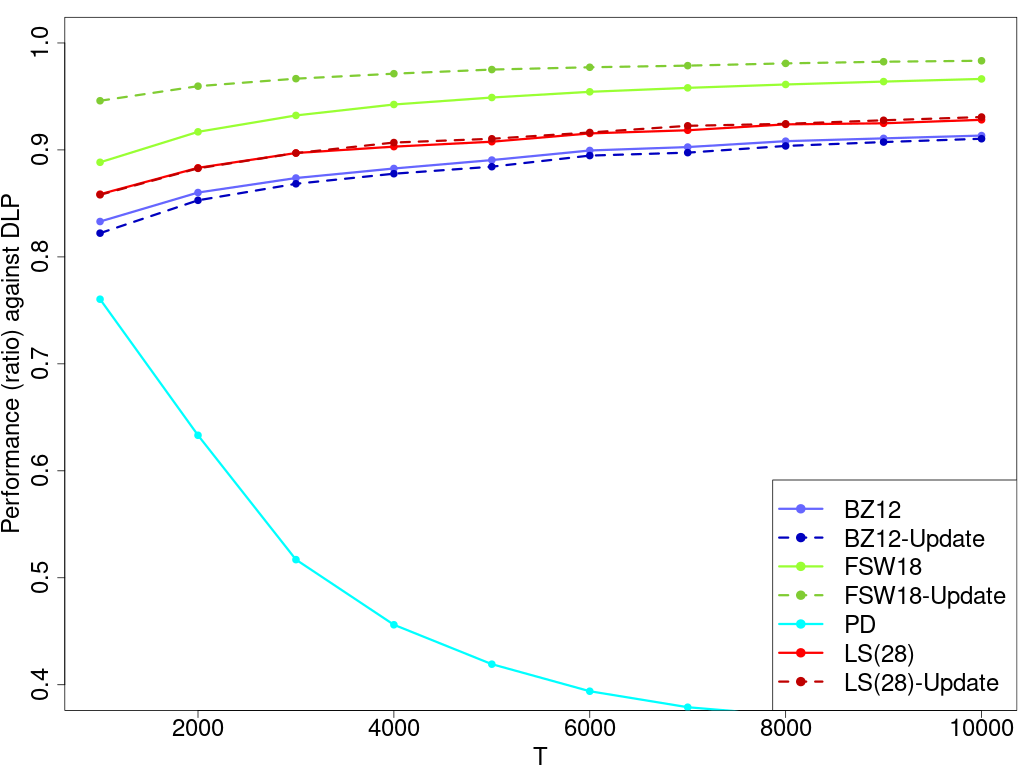}
\caption{Numerical results under linear demand and large inventory when there are $K=15$ price vectors.}
\label{fig:additionalK=15:linear:large}
{\footnotesize {\it Note:} The solid lines are the original method without updating inventory; the dashed lines are updating inventory. The performance of algorithms are normalized relative to the DLP upper bound, thus between 0 and 1.}
\end{figure}

\begin{figure}[!htb]
\centering
\includegraphics[width=0.7\textwidth]{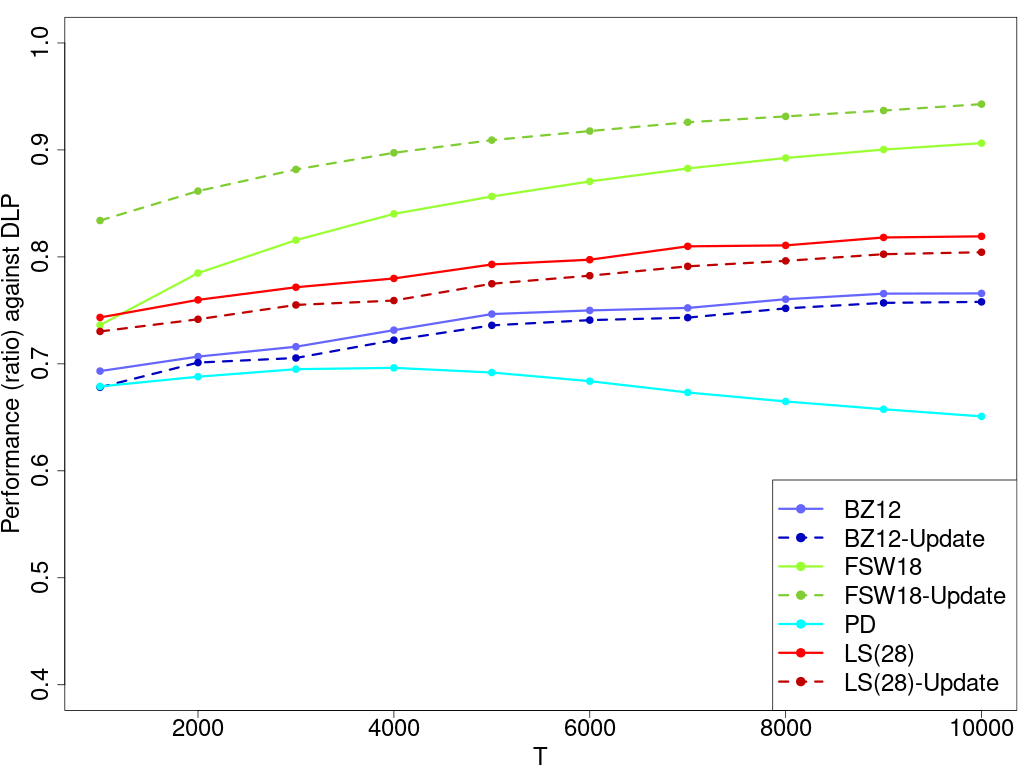}
\caption{Numerical results under exponential demand and small inventory when there are $K=15$ price vectors.}
\label{fig:additionalK=15:exponential:small}
{\footnotesize {\it Note:} The solid lines are the original method without updating inventory; the dashed lines are updating inventory. The performance of algorithms are normalized relative to the DLP upper bound, thus between 0 and 1.}
\end{figure}

\begin{figure}[!htb]
\centering
\includegraphics[width=0.7\textwidth]{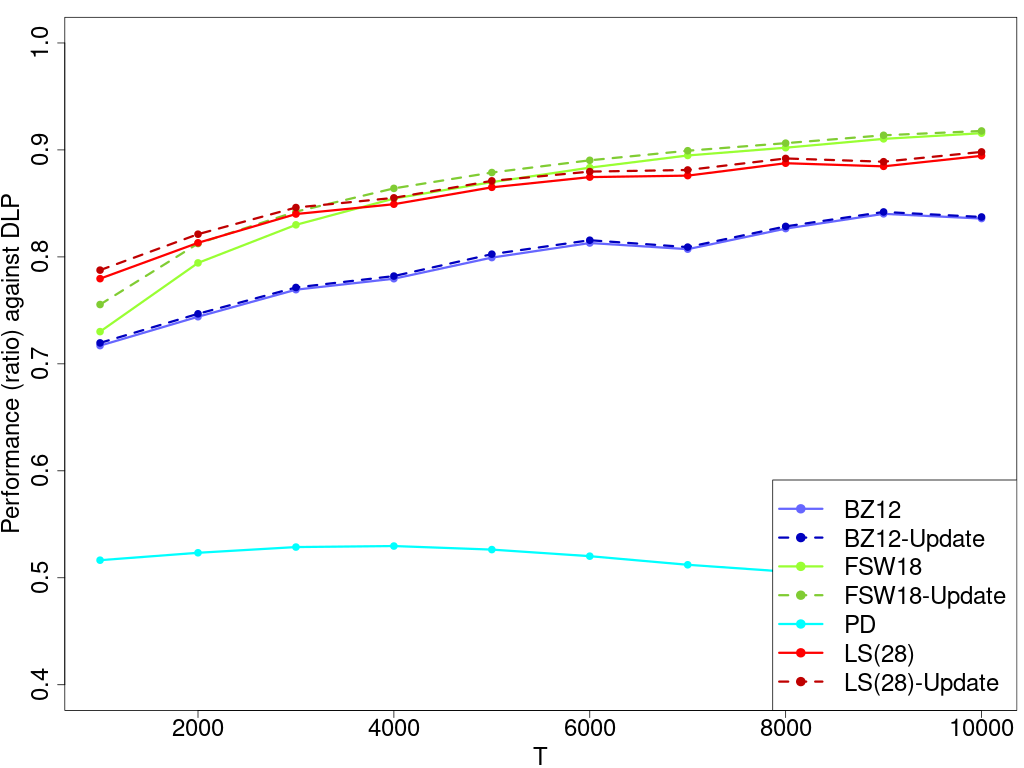}
\caption{Numerical results under exponential demand and large inventory when there are $K=15$ price vectors.}
\label{fig:additionalK=15:exponential:large}
{\footnotesize {\it Note:} The solid lines are the original method without updating inventory; the dashed lines are updating inventory. The performance of algorithms are normalized relative to the DLP upper bound, thus between 0 and 1.}
\end{figure}

\begin{figure}[!htb]
\centering
\includegraphics[width=0.7\textwidth]{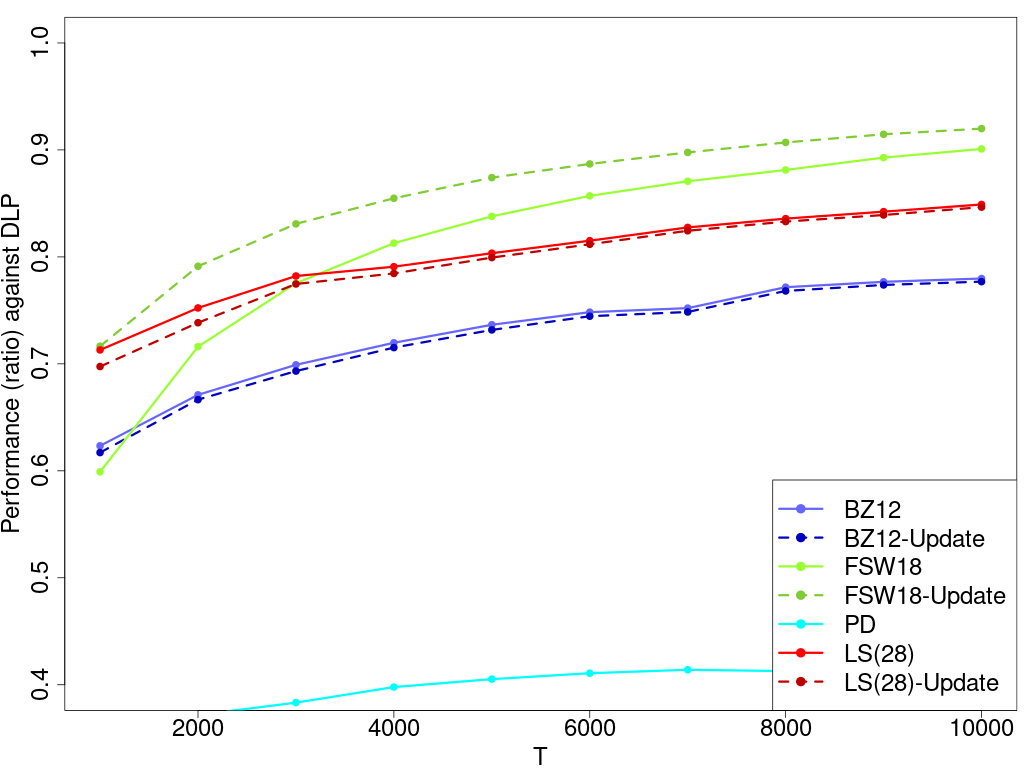}
\caption{Numerical results under logit demand and small inventory when there are $K=15$ price vectors.}
\label{fig:additionalK=15:logit:small}
{\footnotesize {\it Note:} The solid lines are the original method without updating inventory; the dashed lines are updating inventory. The performance of algorithms are normalized relative to the DLP upper bound, thus between 0 and 1.}
\end{figure}

\begin{figure}[!htb]
\centering
\includegraphics[width=0.7\textwidth]{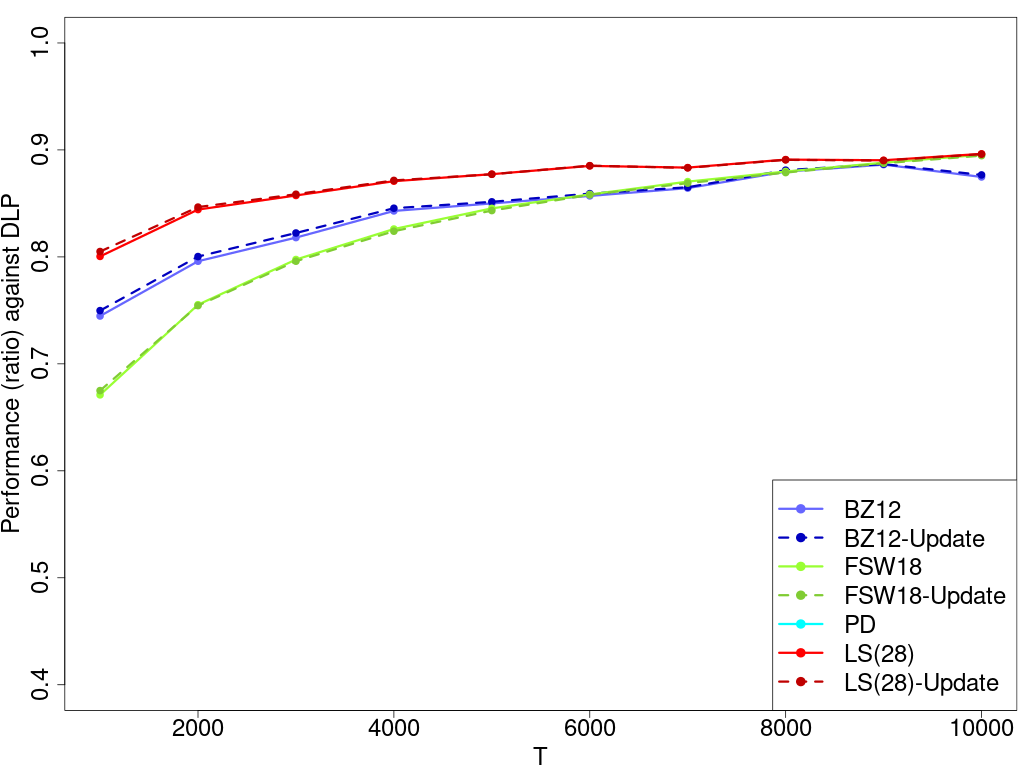}
\caption{Numerical results under logit demand and large inventory when there are $K=15$ price vectors.}
\label{fig:additionalK=15:logit:large}
{\footnotesize {\it Note:} The solid lines are the original method without updating inventory; the dashed lines are updating inventory. The performance of algorithms are normalized relative to the DLP upper bound, thus between 0 and 1.}
\end{figure}

We next present the simulation results in the second simulation setup where there are 15 price $(p_1,p_2) \in \{(0.5,0.5)$, $(0.5,0.8)$, $(0.5,1)$, $(0.5,1.5)$, $(0.8,0.8)$, $(0.8,1)$, $(0.8,1.5)$, $(1,1.5)$, $(1,2)$, $(2,3)$, $(2,4)$, $(4,4)$, $(4,6.5)$, $(4,8), (5,8)\}$.
The blue dashed line is \textbf{BZ12} with updating inventory.
The green dashed line is \textbf{FSW18} with updating inventory.
The \textbf{PD} algorithm does not have the updating inventory version.
The red dashed line is our algorithm under switching budgets of $s=28$ with updating inventory.
For most cases, the inventory updating version of all the algorithms are slightly better than the no inventory updating version.
See Figures~\ref{fig:additionalK=15:linear:small}--~\ref{fig:additionalK=15:logit:large}.

\clearpage

\subsection{The Number of Switches}

\begin{table}[htbp]
\TABLE
{Summary of the number of switches made in all six scenarios.
\label{tbl:NSwitchesOneExample}}
{\tiny
\begin{tabular}{|l|r|r|r|r|r|r|r|r|r|r|}
\multicolumn{11}{c}{(a) Linear demand, small inventory} \\ \hline
$T$           & 1000    & 2000    & 3000     & 4000     & 5000     & 6000     & 7000     & 8000     & 9000     & 10000    \\ \hline
BZ12          & 3.42   & 3.28   & 3.24   & 3.26   & 3.27   & 3.21   & 3.23   & 3.18   & 3.20   & 3.18   \\ \hline
BZ12-Update   & 3.21   & 3.16   & 3.14   & 3.10   & 3.11   & 3.10   & 3.08   & 3.07   & 3.08   & 3.09   \\ \hline
FSW18         & 32.25  & 36.98  & 39.71  & 40.90  & 41.95  & 43.32  & 44.94  & 44.64  & 46.54  & 47.70  \\ \hline
FSW18-Update  & 33.03  & 38.53  & 41.72  & 43.34  & 44.68  & 46.14  & 47.67  & 47.85  & 49.55  & 50.30  \\ \hline
PD            & 128.39 & 186.27 & 230.57 & 264.88 & 295.50 & 322.43 & 342.63 & 365.27 & 385.45 & 404.26 \\ \hline
LS(8)         & 3.28   & 3.26   & 3.21   & 3.18   & 3.20   & 3.14   & 3.16   & 3.13   & 3.14   & 3.13   \\ \hline
LS(8)-Update  & 3.03   & 3.04   & 3.02   & 3.02   & 3.03   & 3.02   & 3.02   & 3.01   & 3.02   & 3.01   \\ \hline
LS(12)        & 7.04   & 7.01   & 7.00   & 7.00   & 7.00   & 7.00   & 7.00   & 7.00   & 7.00   & 7.00   \\ \hline
LS(12)-Update & 6.84   & 7.00   & 7.00   & 7.00   & 7.00   & 7.00   & 7.00   & 7.00   & 7.00   & 7.00   \\ \hline
LS(16)        & 6.91   & 7.12   & 7.10   & 7.14   & 7.11   & 7.08   & 7.08   & 7.10   & 7.11   & 7.09   \\ \hline
LS(16)-Update & 6.43   & 6.99   & 7.05   & 7.11   & 7.07   & 7.07   & 7.08   & 7.09   & 7.10   & 7.08   \\ \hline
\multicolumn{11}{c}{(b) Linear demand, large inventory} \\ \hline
$T$           & 1000    & 2000    & 3000     & 4000     & 5000     & 6000     & 7000     & 8000     & 9000     & 10000    \\ \hline
BZ12          & 3.95   & 3.97    & 3.99    & 3.99    & 3.99    & 4.00    & 4.00    & 4.00    & 4.00    & 4.00    \\ \hline
BZ12-Update   & 3.95   & 3.97    & 3.99    & 3.99    & 3.99    & 4.00    & 4.00    & 4.00    & 4.00    & 4.00    \\ \hline
FSW18         & 546.52 & 1075.95 & 1597.35 & 2103.45 & 2617.72 & 3110.36 & 3604.39 & 4097.72 & 4587.40 & 5070.29 \\ \hline
FSW18-Update  & 567.26 & 1110.47 & 1641.40 & 2151.35 & 2659.97 & 3165.47 & 3658.28 & 4155.13 & 4640.22 & 5129.77 \\ \hline
PD            & 159.25 & 221.84  & 267.58  & 303.46  & 334.31  & 361.32  & 382.31  & 403.65  & 424.27  & 441.35  \\ \hline
LS(8)         & 3.97   & 3.99    & 4.00    & 4.00    & 4.00    & 4.00    & 4.00    & 4.00    & 4.00    & 4.00    \\ \hline
LS(8)-Update  & 3.97   & 3.99    & 4.00    & 4.00    & 4.00    & 4.00    & 4.00    & 4.00    & 4.00    & 4.00    \\ \hline
LS(12)        & 7.82   & 7.89    & 7.92    & 7.93    & 7.93    & 7.95    & 7.95    & 7.96    & 7.97    & 7.98    \\ \hline
LS(12)-Update & 7.81   & 7.88    & 7.91    & 7.93    & 7.94    & 7.96    & 7.94    & 7.96    & 7.97    & 7.98    \\ \hline
LS(16)        & 10.20  & 10.25   & 10.30   & 10.24   & 10.30   & 10.32   & 10.27   & 10.30   & 10.28   & 10.27   \\ \hline
LS(16)-Update & 10.17  & 10.25   & 10.23   & 10.29   & 10.29   & 10.29   & 10.28   & 10.24   & 10.33   & 10.26   \\ \hline
\multicolumn{11}{c}{(c) Exponential demand, small inventory} \\ \hline
$T$           & 1000    & 2000    & 3000     & 4000     & 5000     & 6000     & 7000     & 8000     & 9000     & 10000    \\ \hline
BZ12          & 4.05   & 4.05    & 4.07    & 4.06    & 4.07    & 4.09    & 4.05    & 4.06    & 4.07    & 4.08    \\ \hline
BZ12-Update   & 4.01   & 4.04    & 4.04    & 4.04    & 4.06    & 4.07    & 4.04    & 4.05    & 4.06    & 4.06    \\ \hline
FSW18         & 518.10 & 1022.15 & 1507.10 & 1941.80 & 2376.27 & 2823.99 & 3230.40 & 3640.19 & 4025.02 & 4437.37 \\ \hline
FSW18-Update  & 526.07 & 1007.10 & 1444.13 & 1871.25 & 2280.57 & 2686.55 & 3084.54 & 3509.88 & 3884.50 & 4249.87 \\ \hline
PD            & 437.04 & 490.38  & 518.71  & 567.02  & 605.95  & 639.08  & 670.57  & 701.88  & 727.85  & 752.42  \\ \hline
LS(8)         & 4.04   & 4.05    & 4.08    & 4.05    & 4.07    & 4.06    & 4.07    & 4.04    & 4.05    & 4.05    \\ \hline
LS(8)-Update  & 4.00   & 4.02    & 4.05    & 4.03    & 4.05    & 4.03    & 4.04    & 4.03    & 4.03    & 4.03    \\ \hline
LS(12)        & 7.32   & 7.24    & 7.21    & 7.18    & 7.16    & 7.13    & 7.11    & 7.09    & 7.08    & 7.08    \\ \hline
LS(12)-Update & 7.32   & 7.26    & 7.22    & 7.18    & 7.16    & 7.14    & 7.11    & 7.10    & 7.09    & 7.08    \\ \hline
LS(16)        & 9.82   & 9.88    & 9.86    & 9.84    & 9.82    & 9.78    & 9.81    & 9.79    & 9.69    & 9.67    \\ \hline
LS(16)-Update & 9.84   & 9.83    & 9.85    & 9.87    & 9.84    & 9.82    & 9.83    & 9.78    & 9.72    & 9.69    \\ \hline
\multicolumn{11}{c}{(d) Exponential demand, large inventory} \\ \hline
$T$           & 1000    & 2000    & 3000     & 4000     & 5000     & 6000     & 7000     & 8000     & 9000     & 10000    \\ \hline
BZ12          & 4.12   & 4.15   & 4.19    & 4.21    & 4.20    & 4.22    & 4.20    & 4.19    & 4.20    & 4.20    \\ \hline
BZ12-Update   & 4.11   & 4.13   & 4.15    & 4.17    & 4.17    & 4.19    & 4.16    & 4.16    & 4.15    & 4.16    \\ \hline
FSW18         & 544.75 & 934.05 & 1218.74 & 1429.95 & 1637.41 & 1800.63 & 1970.92 & 2078.62 & 2145.22 & 2278.42 \\ \hline
FSW18-Update  & 534.53 & 930.88 & 1226.41 & 1445.83 & 1648.23 & 1801.71 & 1961.06 & 2051.71 & 2167.49 & 2255.26 \\ \hline
PD            & 437.06 & 490.38 & 518.64  & 566.99  & 606.43  & 640.52  & 671.52  & 703.15  & 729.55  & 754.22  \\ \hline
LS(8)         & 4.15   & 4.22   & 4.20    & 4.17    & 4.16    & 4.19    & 4.14    & 4.18    & 4.17    & 4.17    \\ \hline
LS(8)-Update  & 4.10   & 4.15   & 4.12    & 4.10    & 4.10    & 4.12    & 4.10    & 4.10    & 4.09    & 4.10    \\ \hline
LS(12)        & 6.97   & 6.96   & 6.95    & 6.99    & 6.96    & 6.96    & 6.97    & 6.96    & 6.97    & 6.97    \\ \hline
LS(12)-Update & 6.88   & 6.90   & 6.92    & 6.95    & 6.92    & 6.93    & 6.94    & 6.95    & 6.96    & 6.96    \\ \hline
LS(16)        & 7.90   & 7.74   & 7.66    & 7.58    & 7.57    & 7.54    & 7.46    & 7.46    & 7.40    & 7.41    \\ \hline
LS(16)-Update & 7.52   & 7.54   & 7.53    & 7.48    & 7.39    & 7.42    & 7.40    & 7.44    & 7.33    & 7.34    \\ \hline
\multicolumn{11}{c}{(e) Logit demand, small inventory} \\ \hline
$T$           & 1000    & 2000    & 3000     & 4000     & 5000     & 6000     & 7000     & 8000     & 9000     & 10000    \\ \hline
BZ12          & 4.45   & 4.49    & 4.56    & 4.58    & 4.60    & 4.67    & 4.65    & 4.69    & 4.71    & 4.72    \\ \hline
BZ12-Update   & 4.43   & 4.49    & 4.56    & 4.56    & 4.60    & 4.66    & 4.65    & 4.69    & 4.70    & 4.72    \\ \hline
FSW18         & 496.27 & 928.45  & 1346.83 & 1739.66 & 2156.84 & 2560.80 & 2961.68 & 3371.62 & 3778.39 & 4176.79 \\ \hline
FSW18-Update  & 561.68 & 1047.78 & 1525.00 & 1941.58 & 2379.56 & 2794.50 & 3205.90 & 3629.77 & 4047.03 & 4455.53 \\ \hline
PD            & 593.12 & 947.99  & 1011.55 & 1039.39 & 1050.22 & 1079.23 & 1087.37 & 1105.07 & 1109.79 & 1125.06 \\ \hline
LS(8)         & 4.50   & 4.58    & 4.65    & 4.68    & 4.75    & 4.76    & 4.79    & 4.81    & 4.82    & 4.83    \\ \hline
LS(8)-Update  & 4.48   & 4.56    & 4.63    & 4.67    & 4.75    & 4.75    & 4.79    & 4.80    & 4.82    & 4.83    \\ \hline
LS(12)        & 7.07   & 7.01    & 7.00    & 7.01    & 7.00    & 7.00    & 7.00    & 7.00    & 7.00    & 7.00    \\ \hline
LS(12)-Update & 7.06   & 7.03    & 7.01    & 7.00    & 7.00    & 7.01    & 7.00    & 7.00    & 7.00    & 7.00    \\ \hline
LS(16)        & 9.74   & 9.73    & 9.74    & 9.82    & 9.79    & 9.91    & 9.82    & 9.77    & 9.84    & 9.84    \\ \hline
LS(16)-Update & 9.72   & 9.82    & 9.67    & 9.81    & 9.86    & 9.88    & 9.87    & 9.83    & 9.79    & 9.92    \\ \hline
\multicolumn{11}{c}{(f) Logit demand, large inventory} \\ \hline
$T$           & 1000    & 2000    & 3000     & 4000     & 5000     & 6000     & 7000     & 8000     & 9000     & 10000    \\ \hline
BZ12          & 4.08   & 4.04   & 4.05    & 4.03    & 4.01    & 4.01    & 4.00    & 4.01    & 4.01    & 4.00    \\ \hline
BZ12-Update   & 4.06   & 4.02   & 4.03    & 4.01    & 4.01    & 4.01    & 4.00    & 4.00    & 4.00    & 4.00    \\ \hline
FSW18         & 541.18 & 976.71 & 1344.60 & 1682.86 & 1983.57 & 2347.32 & 2637.03 & 2957.84 & 3185.34 & 3374.18 \\ \hline
FSW18-Update  & 541.50 & 976.33 & 1356.69 & 1698.51 & 2012.54 & 2351.86 & 2625.33 & 2942.12 & 3205.04 & 3360.45 \\ \hline
PD            & 593.12 & 947.99 & 1011.55 & 1039.39 & 1050.22 & 1079.23 & 1087.37 & 1105.07 & 1109.79 & 1125.06 \\ \hline
LS(8)         & 4.03   & 4.03   & 4.00    & 4.01    & 4.00    & 4.00    & 4.00    & 4.00    & 4.00    & 4.00    \\ \hline
LS(8)-Update  & 4.01   & 4.01   & 4.00    & 4.00    & 4.00    & 4.00    & 4.00    & 4.00    & 4.00    & 4.00    \\ \hline
LS(12)        & 6.89   & 6.92   & 6.90    & 6.93    & 6.94    & 6.95    & 6.95    & 6.95    & 6.96    & 6.96    \\ \hline
LS(12)-Update & 6.88   & 6.91   & 6.92    & 6.92    & 6.93    & 6.95    & 6.95    & 6.94    & 6.95    & 6.96    \\ \hline
LS(16)        & 7.49   & 7.48   & 7.48    & 7.47    & 7.47    & 7.44    & 7.51    & 7.43    & 7.43    & 7.46    \\ \hline
LS(16)-Update & 7.40   & 7.44   & 7.52    & 7.47    & 7.49    & 7.48    & 7.48    & 7.44    & 7.46    & 7.48    \\ \hline
\end{tabular}
}
{}
\end{table}

 

\end{document}